# A GPT-4 Reticular Chemist for Guiding MOF Discovery

Zhiling Zheng,+[b] Zichao Rong,+[b] Nakul Rampal,[b] Christian Borgs,[c] Jennifer T. Chayes,[d] and Omar M. Yaghi*[a]

[a] Prof. Dr. O. M. Yaghi
Department of Chemistry, Kavli Energy Nanoscience Institute, and Bakar Institute of Digital Materials for the Planet, College of Computing, Data Science, and Society, University of California, Berkeley
Berkeley, CA 94720 (United States)
and
KACST—UC Berkeley Center of Excellence for Nanomaterials for Clean Energy Applications, King Abdulaziz City for Science and Technology
Riyadh 11442 (Saudi Arabia)
E-mail: yaghi@berkeley.edu

[b] Z. Zheng, Z. Rong, Dr. N. Rampal
Department of Chemistry, Kavli Energy Nanoscience Institute, and Bakar Institute of Digital Materials for the Planet, College of Computing, Data Science, and Society, University of California, Berkeley
Berkeley, CA 94720 (United States)

[c] Prof. Dr. C. Borgs
Department of Electrical Engineering and Computer Sciences and Bakar Institute of Digital Materials for the Planet, College of Computing, Data Science, and Society, University of California, Berkeley
Berkeley, CA 94720 (United States)

[d] Prof. Dr. J. T. Chayes
Department of Electrical Engineering and Computer Sciences, Department of Statistics, Department of Mathematics, School of Information, and Bakar Institute of Digital Materials for the Planet, College of Computing, Data Science, and Society, University of California, Berkeley
Berkeley, CA 94720 (United States)

[+] Z.Z. and Z.R. contributed equally to this work.

**Abstract:** We present a new framework integrating the AI model GPT-4 into the iterative process of reticular chemistry experimentation, leveraging a cooperative workflow of interaction between AI and a human researcher. This GPT-4 Reticular Chemist is an integrated system composed of three phases. Each of these utilizes GPT-4 in various capacities, wherein GPT-4 provides detailed instructions for chemical experimentation and the human provides feedback on the experimental outcomes, including both success and failures, for the in-context learning of AI in the next iteration. This iterative human-AI interaction enabled GPT-4 to learn from the outcomes, much like an experienced chemist, by a prompt-learning strategy. Importantly, the system is based on natural language for both development and operation, eliminating the need for coding skills, and thus, make it accessible to all chemists. Our collaboration with GPT-4 Reticular Chemist guided the discovery of an isoreticular series of MOFs, with each synthesis fine-tuned through iterative feedback and expert suggestions. This workflow presents a potential for broader applications in scientific research by harnessing the capability of large language models like GPT-4 to enhance the feasibility and efficiency of research activities.

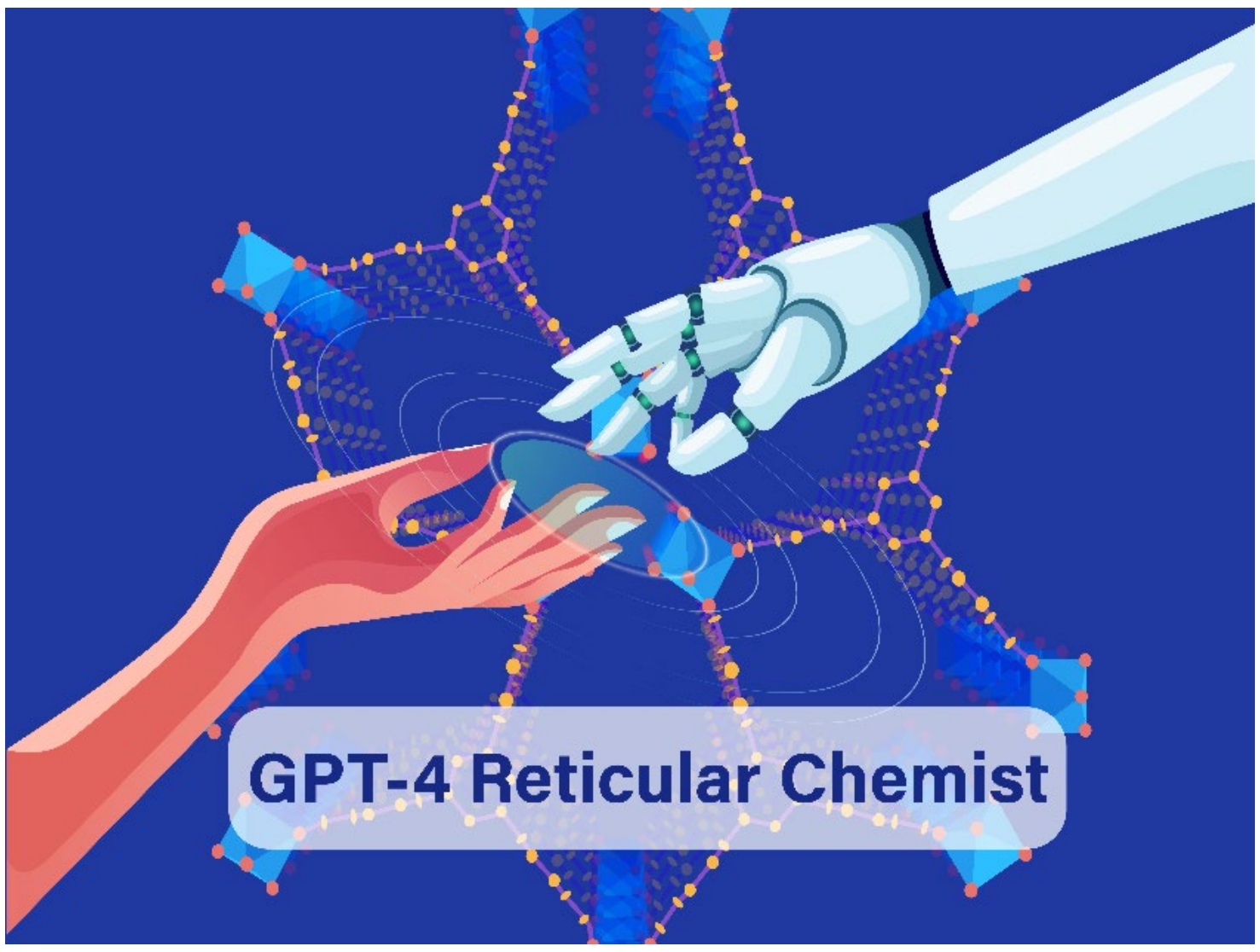

## Introduction

The advancement of reticular chemistry relies on our ability to expand the repertoire of materials and devise innovative strategies to exploit their properties.[1] The mastery of this chemistry requires working knowledge in various disciplines of chemistry (organic, inorganic, physical, analytical, and computational), physics, biology, as well as material science and engineering.[1c, 2] However, training a reticular chemist to such a degree of proficiency is time-intensive, necessitating a comprehensive understanding of project design and workflow, as well as expertise beyond the traditional confines of chemistry. Indeed, the development of this depth and breadth of knowledge is typically acquired over years of education and training. We believe that the emerging field of artificial intelligence (AI) and more specifically large language models (LLMs) like GPT-4 are poised to help bridge the knowledge gap between these disciplines and do so with efficiency and speed.[3]

Implementation of this vision requires the seamless integration of AI models into the day-to-day activities and workflow of a reticular chemist.[4] In the present report, we use GPT-4 as a demonstration of how this could be done. GPT-4, being proficient in language comprehension,[3b, 5] can facilitate experiment design,[6] review and summarize research articles,[7] provide educational introductions to various concepts and techniques,[8] and assist in data interpretation[9] when guided by carefully designed prompts with explicit instructions.[3f, 5c, 10] Yet, as it stands, GPT-4 alone lacks the capability to perform a core component of chemistry – experimentation, which is essentially a series of iterative trial-and-error cycles driven by observations and hypotheses. Here, we show initial steps for integrating GPT-4 into the iterative process of chemical experimentation and propose a novel framework wherein GPT-4 works in tandem with a human researcher at any expertise level. Our strategy features having GPT-4 provide detailed step-by-step instructions for experiments, and the human researcher carries out these procedures and provides detailed feedback on the observations and results. This iterative in-context learning process allows the GPT-4 Reticular Chemist to have access to the experimental outcomes, thereby enabling it to learn from both successes and failures, just like an experienced chemist. This human-AI interaction loop continues until the project goal is achieved, with GPT-4 dynamically adjusting its strategy based on the outcomes of previous iterations. This innovative approach reaps mutual benefits – GPT-4 Reticular Chemist gains real-world experimental validation of its plans and learns from the outcomes, and the human benefits from accelerated research progress under the professional guidance of GPT-4, akin to having a personalized mentor.

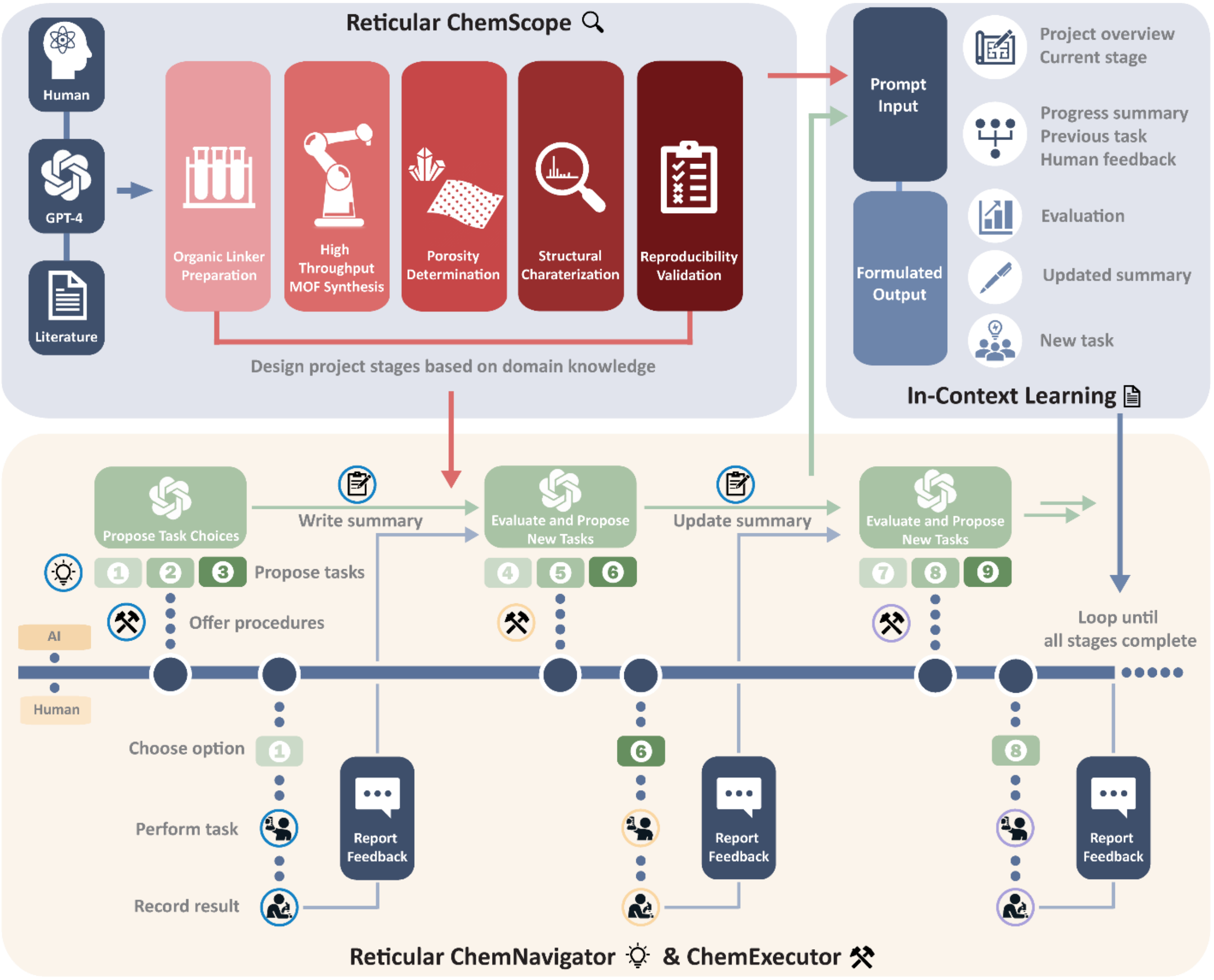

**Figure 1.** Schematic representation of the GPT-4 based framework functioning as an advanced reticular chemist through strategic prompt engineering and in-context learning informed by human feedback. This GPT-4 Reticular Chemist workflow encapsulates three states: the "Reticular ChemScope", responsible for devising the foundational research project blueprint by dividing it into major domains of activity; the "Reticular ChemNavigator", which assesses the latest activities and presents three potential task options for the human researcher; and the "Reticular ChemExecutor" that furnishes comprehensive procedural guidelines for the selected task, facilitating step-by-step human engagement. All three states are powered by GPT-4 but are given three different types of prompts. The GPT-4's in-context learning capacity is nurtured through a combination of a pre-engineered prompt system and continual human feedback, linked to each task performed. The development and execution of this workflow only require natural language, with no coding involved.

## Results and Discussion

We designed a novel prompt learning strategy framework that integrates GPT-4 into reticular chemistry, utilizing human feedback to expedite the discovery of new MOFs. Impressively, this symbiotic human-AI collaboration based on two-way learning process led to the sequential discovery of four isoreticular MOFs bearing the same generic chemical formula of $[Al_3(\mu\text{-}OH)_3(HCOO)_3(BTB\text{-}X)]$ and further defined and elaborated below. Notably, the synthesis of these MOFs and their linkers and the subsequent optimization and characterization were all designed by GPT-4 and executed by the human researcher.

Our developed framework allows the fast adaptation of Large Language Models (LLMs) such as GPT-4 to the realm of reticular chemistry via prompt engineering and in-context learning, driven by human feedback. The process comprises three interlinked phases, each employing GPT-4 as a reticular chemist with varying degrees of interaction with human counterparts through tailored prompts, we termed Reticular ChemScope, Reticular ChemNavigator, and Reticular ChemExecutor, which are illustrated in Figure 1.

The first phase Reticular ChemScope had GPT-4 create a project blueprint based on given inputs, followed by the second phase Reticular ChemNavigator where the GPT-4 suggested tasks and guided progress based on human feedback. Lastly, in the third phase Reticular ChemExecutor, GPT-4 provided detailed task steps and a feedback template, enabling effective task execution and improvement. The detailed design and description of each phase can be found in the Materials and Methods Section (Supporting Information, Section S1). In addition, an interactive prompt refinement strategy was used to enable GPT-4 to self-compose prompts by providing role descriptions and duties, which were then refined through an iterative process of suggesting, testing, and improvement. The final versions of the prompts for all three phases are depicted in Figures S36–S41.

It should be emphasized that the ChemNavigator serves as the heart of the three-phase framework. It is where most of the proposed experiment – human wet lab experiments – feedback loop takes place and serves as the cornerstone of our human-AI interaction research design. Inspired by previous studies on prompt engineering of GPT models,[11] our prompt was designed to include specific elements such as role definition, goal instructions, memory of previous trials, evaluation process, and structured output as three task choices. These elements were included based on our reasoning to ensure that the input information is provided in an explicit way, to store a summarized memory of the previous information, and to make GPT-4's logic in suggesting tasks more interpretable. Together, they contribute to GPT-4's reticular chemist behavior, manageability of development process, accumulation of past learning, interpretability and transparency in GPT-4's decision-making process, and variety of task choices based on available resources or schedules, respectively. Following the proposed task by ChemNavigator, the ChemExecutor then makes a detailed experimental plan for the human.

In particular, each iteration involves a single instance of human-AI conversation. It begins with the human sharing a summary of prior experiments, successes, and failures, which GPT-4 processes as “memory”. GPT-4 also considers the most recent task it had recommended in the previous iteration, along with the human's feedback on the outcomes. The central question posed to GPT-4 is how to advance, given the current status of the project, to achieve the goal of discovering and studying a new MOF. Hence, even if every time a new dialogue session is initiated, GPT-4 capitalizes on in-context learning from prior interactions, allowing it to pick up from the latest point of progress. Using this information, GPT-4 evaluates the current state of the project, updates its memory to include a description of the last trial and errors, and suggests three potential actions. These actions are all aimed at progressing towards the completion of the current stage. Since the GPT-4 was asked to come up with reasoning behind its suggestions, the human researcher can critically evaluate them. This evaluation aids in selecting a course of action, based not only on its merit but also on the availability of resources, laboratory instruments, and the expertise level of the human researcher. Subsequent to the task's execution, as guided by GPT-4's instructions, the human documents the observations and summarizes the outcomes. This summary, presented to GPT-4 in textual format, reinforces its in-context learning, assisting it in making more informed decisions in subsequent steps. This iterative process continues until GPT-4 identifies progress indicators that align with the objectives indicated in the current stage's description and provides an explanation (Supporting Information, Section S8). Guided by GPT-4's suggestions, the human researcher then decides whether to proceed to the next stage or continue refining the current one.

The complete journey of MOF discovery involved the use of GPT-4 Reticular Chemist workflow outlined in Figure 1 and encompasses the creation of four new MOFs (Figure 2a: MOF-521-H, -*o*F, -*m*F, -$CH_3$) as detailed in Figures S47–S165, where the input into and the output from GPT-4 are shown. Notably, one of these compounds, MOF-521-H, was previously discovered accidentally as a byproduct during the synthesis of MOF-520,[12] another BTB-based aluminum MOF, and remained unpublished. GPT-4 Reticular Chemist, however, unaware of and thus unbiased by the synthesis details from this previous study, independently designed a strategy for these four new MOFs, including MOF-521-H. Interestingly, the optimized synthesis condition established by GPT-4 was subtly different from what humans had previously discovered alone (Supporting Information, Section S2 and Section S9).

Succinctly, upon initiation, the first stage saw Reticular ChemNavigator guiding the human researcher to seek synthesis routes to obtain the linkers, primarily through a thorough literature review (Figures S1–S6). Following a preliminary plan for synthesis, the dialogue directed the human to conduct the organic synthesis reaction, overcome the challenges, and finally confirm the linker by proton NMR (Figures S7–S10). With the linkers in hand, and guiding the human in summarizing the literature search,[13] the second research phase phase focused on identifying the best conditions for MOFs formation, encompassing modifications to metal-linker ratio, temperature, reaction time, modulator(s), and their proportion (Table 1 and Supporting Information, Section S9). Alongside powder X-ray diffraction (PXRD), microscopic observations were used to report the progress. After several rounds of optimization guided by the suggestions of Reticular ChemNavigator, single crystals of all compounds were obtained. Except for MOF-521-$CH_3$, whose rod-shaped single crystals were too small for single crystal X-ray diffraction (SXRD), the remaining three compounds were sufficiently large to have their structures characterized by SXRD.[14]

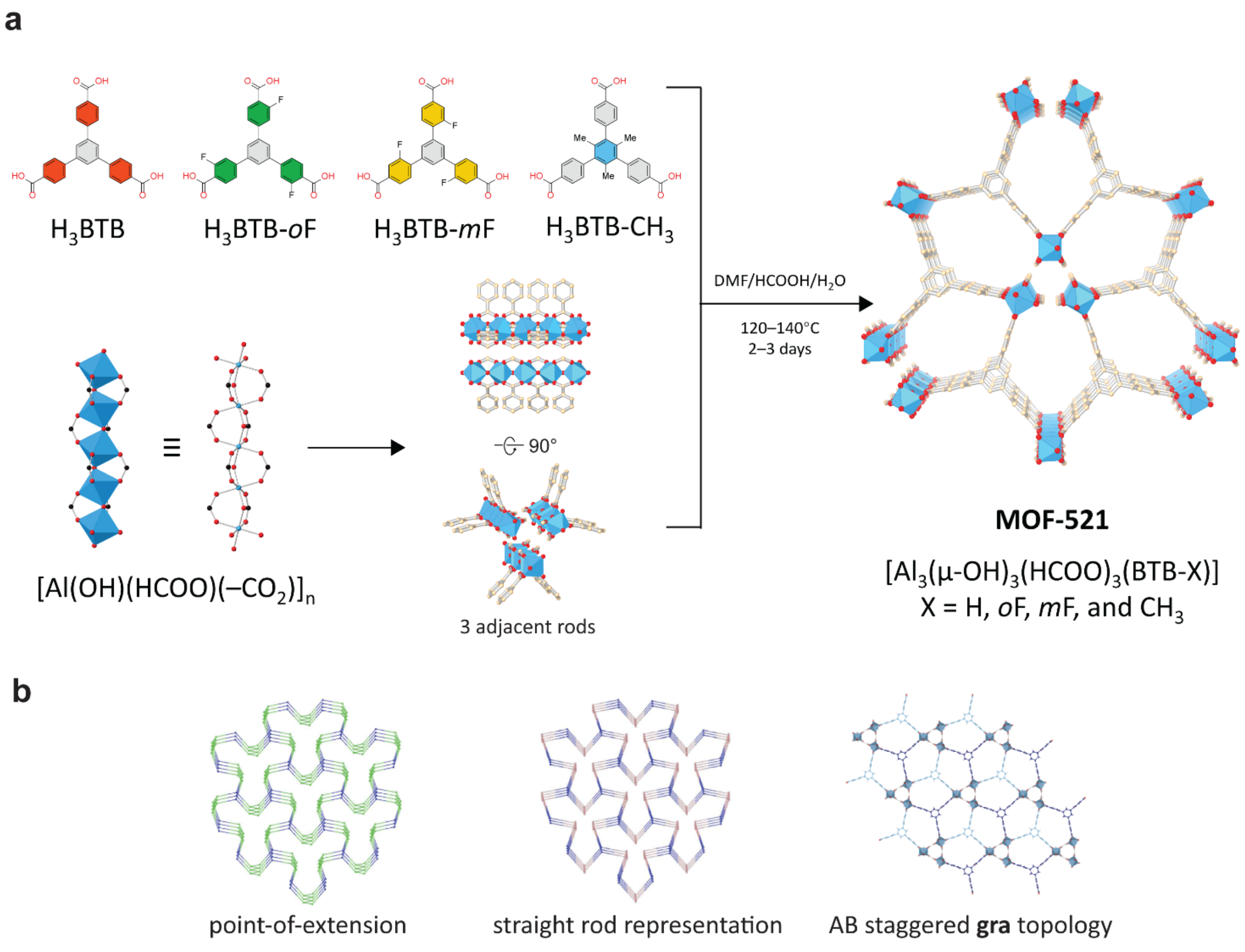

**Figure 2.** (a) Synthesis and structures of MOF-521. (b) Topological analysis of MOF-521. Color code: Al, blue; O, red; F, green; C, yellow. Hydrogen atoms are omitted for clarity.

Throughout single crystal X-ray diffraction analysis, it was discovered that the MOF-521 compounds crystallize in the P-62c (No. 190) space group with almost the same unit cell parameters (a=b=21.9 Å, c=6.6 Å). All MOF-521 compounds are built from rod secondary building units (SBUs): two of the six corners of each $AlO_6$ octahedron are shared with another two octahedra via $\mu_2$-OH respectively, two corners are bridged by carboxylate from the BTB linker, and the last two corners are capped by formates, thus forming a 1-dimensional straight rod (Figure 2a). There is one $\mu_2$-OH, one carboxylate from the BTB linker and one formate between the two adjacent $AlO_6$ octahedra. The rods are parallelly connected to each other via BTB linkers, while the $\mu_2$-OH groups point into the large pores and form hydrogen bonding with DMF or water. Positional disorder is found at the peripheral phenyl rings of BTB and can be refined to two symmetric positions. The dihedral angle between the peripheral phenyl ring and the central phenyl ring in MOF-521 (MOF-521-H: 62.9(4)°, MOF-521-*o*F: 51.9(4)°, and MOF-521-*m*F: 68.0(5)°) is larger than the one in MOF-520[15] (36.4(7)°, 36.7(8)° and 40.8(7)°). Adding functionalities onto the phenyl rings would introduce steric hindrance and therefore larger dihedral angles are favored. In addition, the carboxylate of the BTB is almost coplanar with the peripheral phenyl ring to which it connected so that the large dihedral angle determines the plane of the carboxylate to be perpendicular to the plane of the central phenyl ring and also the formation of rod SBUs.

**Table 1.** Screening conditions for MOF-521 synthesis guided by the Reticular ChemNavigator through iterative learning and inference processes.

| Exp.[a] | Linker | Modulator[b] | L:M Ratio[c] | Temp. (°C) | Time (h) |
|---|---|---|---|---|---|
| 1 | BTB-H | FA | 1:1 | 100 | 48 |
| 17 | BTB-H | FA | 2:3 | 140 | 48 |
| 21 | BTB-H | FA | 4:1 | 140 | 48 |
| 29 | BTB-H | FA | 3:4 | 120 | 72 |
| 32 | BTB-H | TFA | 3:4 | 120 | 72 |
| 33 | BTB-H | TFA/AA (1:1) | 3:4 | 120 | 72 |
| 35 | BTB-H | FA/HCl (1:1) | 3:4 | 120 | 72 |
| 41 | BTB-H | FA/$H_2O$ (1:2) | 3:4 | 120 | 72 |
| 46 | BTB-H | FA/$H_2O$ (4:1) | 3:4 | 120 | 84 |
| 52 | BTB-H | FA/$H_2O$ (4:1) | 3:4 | 140 | 72 |
| 54 | BTB-*o*F | FA | 2:1 | 140 | 48 |
| 56 | BTB-*o*F | FA/$H_2O$ (1:1) | 2:1 | 140 | 48 |
| 60 | BTB-*o*F | FA/$H_2O$ (4:1) | 3:4 | 140 | 48 |
| 63 | BTB-*o*F | FA/$H_2O$ (4:1) | 3:4 | 120 | 48 |
| 69 | BTB-*o*F | FA/$H_2O$ (4:1) | 3:4 | 120 | 96 |
| 71 | BTB-*m*F | FA/$H_2O$ (4:1) | 3:4 | 120 | 72 |
| 73 | BTB-*m*F | FA/$H_2O$ (4:1) | 3:4 | 110 | 72 |
| 78 | BTB-*m*F | FA/$H_2O$ (4:1) | 3:5 | 120 | 72 |
| 79 | BTB-*m*F | FA/$H_2O$ (5:1) | 3:4 | 120 | 72 |
| 82 | BTB-$CH_3$ | FA/$H_2O$ (4:1) | 3:4 | 120 | 48 |
| 84 | BTB-$CH_3$ | FA/$H_2O$ (4:1) | 3:4 | 130 | 72 |
| 86 | BTB-$CH_3$ | FA/$H_2O$ (4:1) | 1:2 | 130 | 72 |
| 91 | BTB-$CH_3$ | FA/$H_2O$ (4:1) | 3:4 | 150 | 72 |

[a] Experiment ID corresponds to each human-conducted experiment following the guidelines provided by the Reticular ChemNavigator, powered by GPT-4. Initial conditions were derived from a comprehensive literature review. For each iteration, GPT-4 suggests a modification of one parameter (e.g., linker to metal ratio, modulator, temperature, or time). For the sake of conciseness, representative conditions were documented in this table. Please refer to Supporting Information, Table S1 and Section S9, for comprehensive details. It is important to note that GPT-4 does not provide explicit predictions, but instead guides human to analyze PXRD patterns, considers previous success and failures reported by human, infers potential avenues for subsequent steps in MOF screening, and provides suggestions to human. [b] Modulator abbreviations: FA = formic acid; TFA = trifluoroacetic acid; AA = acetic acid. [c] This denotes the molar ratio of linker to metal ions.

Given the unique rod packing presented in four MOF-521 compounds, we sought to further examine its topology[16] to determine whether they were unprecedented MOFs. However, the inherently mathematical and analytical nature of this process surpassed GPT-4's capabilities,[3b, 3f] so this analysis was handled by the human side. Several topological analyses were applied to the MOF-521 structure, aiming to better describe its underlying topology, which include standard method, point-of-extension (PE), point-of-extension & metal (PE&M) and straight rod representation (STR).[17] Among them, we believe PE and STR best extract the topological features as they both simplify the rod shape just based on its connection direction to the BTB linkers, which is a conventional analysis for MOFs with rod SBUs (Figure 2b). Both methods result in a (3-c)$_3$(3-c) 2-nodal net, which is also termed as **etc** according to the Reticular Chemistry Structure Resource (RCSR).[18] We found that this (3-c)$_3$(3-c) 2-nodal net only has 3 entries recorded in the TOPOS database (GEKCAD: $In_{12}S_{34}Sn_4$; OMABEL: $C_{66}H_{78}B_3Cu_6N_{27}S_9W_3$; XEKJUT: $C_{24}H_{15}Cu_3N_{12}O_3$), which included one inorganic compound and two coordination polymers, suggesting MOF-521 is the first MOF to have this topology.

In addition, we note that even though the underlying topology is the same, the MOF-521 structure still differs strongly from the three entries in TOPOS database in that the 3 rod SBUs in MOF-521 are packed so closely to each other as evidenced by the small distance between the central axis of the rods (6.54 Å). The three rod SBUs are not covalently linked because the formates block the coordination sites of the aluminum metal center. This close parallel packing of rod SBUs has not been observed in other MOFs with rod SBUs. Furthermore, given the uniqueness of its SBU, another way to describe the MOF-521 structure is to view the three closely packed rod SBUs as one pseudo rod SBU since the internal space could not be accessed by any guest molecules, and then the structure has a **gra** topology (Figure 2b).

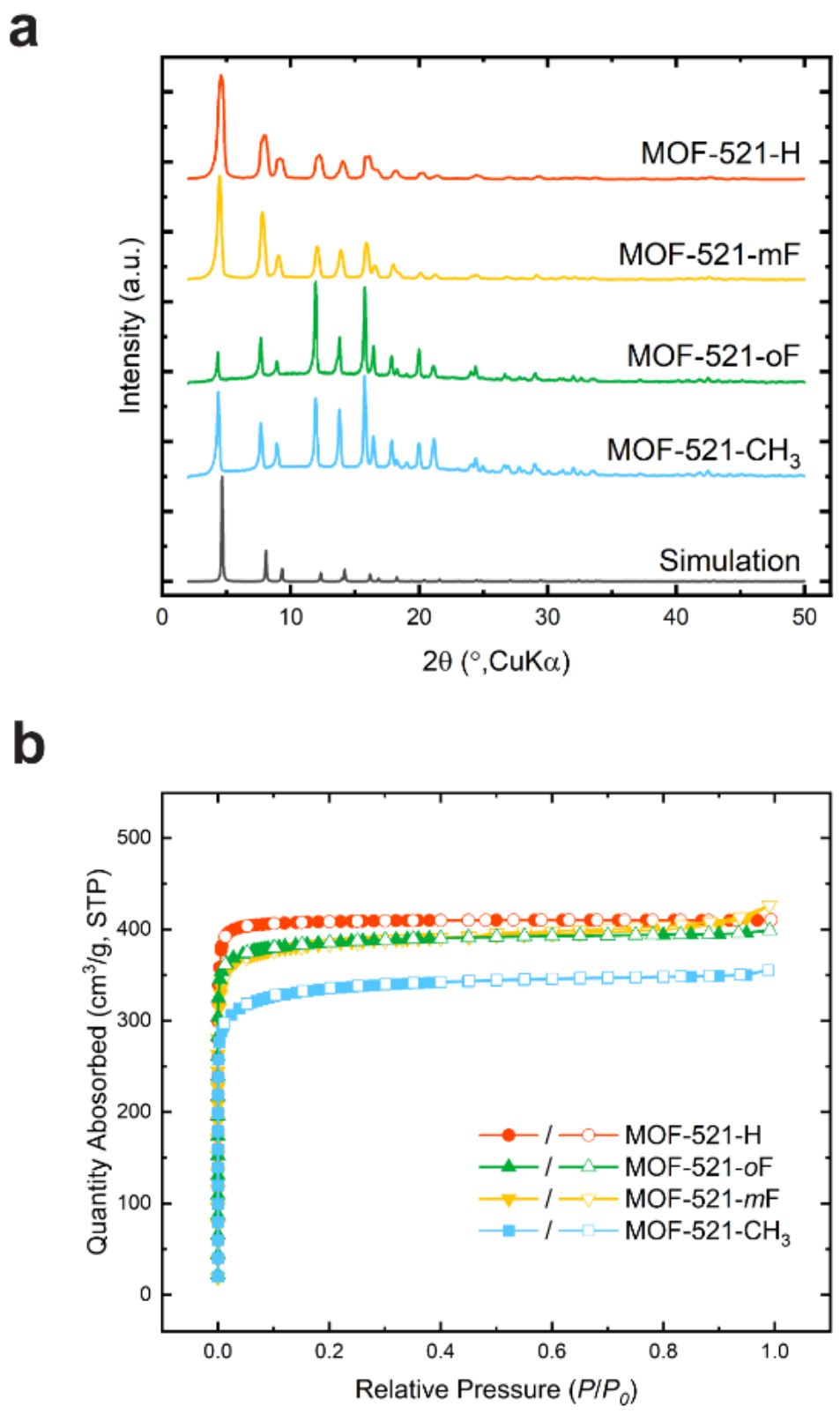

**Figure 3.** (a) PXRD patterns of MOF-521 compounds. The simulated pattern at the bottom was generated using single crystal structure of MOF-521-H. (b) Adsorption-desorption (filled circles-open circles) isotherms of nitrogen on MOF-521 compounds at a measurement temperature of 77 K.

As we moved forward, GPT-4 and the human found a good agreement between the simulated and experimental PXRD patterns, indicating that all four MOF-521 compounds were constructed under the same topology (Figure 3a). A crucial stage was reached in stage 3 when Reticular ChemNavigator initiated investigations into the permanent porosity of the obtained MOF-521 compounds. It gave useful suggestions on setting up the measurement and interpreting the analysis results. The permanent porosity of the MOF-521 family was confirmed by nitrogen sorption measurements, and the experimentally measured surface area of the four MOFs was consistent with the predictions made by Material Studio (Table 2). All the compounds show the Type I isotherms with no significant accompanied hysteresis, and there is a rough trend based on the relationship between the porosity and the size of the functional groups in the framework for both calculated and experimental data. The least substituted MOF-521-H demonstrates the highest BET surface area of 1696 $m^2$ $g^{-1}$, and MOF-521-*o*F and MOF-521-*m*F shows a comparable BET surface area of 1535 $m^2$ $g^{-1}$ and 1562 $m^2$ $g^{-1}$, respectively (Figures S21–S32). On the other hand, the surface area of MOF-521-$CH_3$ is 1311 $m^2$ $g^{-1}$ and turns to be much lower than other ones, suggesting that the substitution of center benzene has a greater influence on the pore environment and can be potentially used to tune the porosity.

**Table 2.** Porosity data for MOF-521 compounds.

| Compound | Formula | Crystal Density (g/cm$^3$) | Pore Width (Å) | $V_p$[a] (cm$^3$/g) | $A_{exp}$[b] (m$^2$/g) | $A_{calc}$[b] (m$^2$/g) |
|---|---|---|---|---|---|---|
| MOF-521-H | $Al_3(\mu\text{-}OH)_3(HCOO)_3(C_{27}H_{15}O_6)$ | 0.851 | 10.8 | 0.588 | 1696 | 1670 |
| MOF-521-*o*F | $Al_3(\mu\text{-}OH)_3(HCOO)_3(C_{27}H_{12}F_3O_6)$ | 0.916 | 10.7 | 0.564 | 1562 | 1540 |
| MOF-521-*m*F | $Al_3(\mu\text{-}OH)_3(HCOO)_3(C_{27}H_{12}F_3O_6)$ | 0.912 | 10.9 | 0.562 | 1535 | 1553 |
| MOF-521-$CH_3$ | $Al_3(\mu\text{-}OH)_3(HCOO)_3(C_{30}H_{21}O_6)$ | nd[c] | 9.3 | 0.509 | 1311 | 1550 |

[a] $V_p$ is the measured pore volume. [b] $A_{exp}$ and $A_{calc}$ are the experimental measured BET surface area and theoretical geometric surface area, respectively. The calculation was performed on *Materials Studios 8.0* with the $N_2$ as probe (diameter = 3.681 Å) according to published procedure.[19] [c] nd, no data.

In stage 4, to verify their compositions and chemical formulae, all four MOF-521 compounds underwent proton nuclear magnetic resonance (NMR) spectroscopy and elemental analysis as suggested by the Reticular ChemNavigator. The proton signals from four linkers were assigned based on their chemical shifts and splitting patterns (Figures S11–S14). The DMF solvent peak cannot be identified in the digestion NMR, suggesting that the compounds were fully activated. The appearance of the formate peaks at 8.5 ppm and the linker peak between 6.2 ppm and 7.8 ppm shows good agreement with the chemical formulae of MOF-521 compounds determined by SXRD and elemental analysis. In addition, it further confirms that, instead of moving freely in the pore of the MOF, formates exist as a building ligand on the framework and were not subjected to removal during the activation under vacuum. It's noteworthy that the data analysis was primarily conducted by human intervention, guided by recommendations from GPT-4. The findings were subsequently reported to GPT-4 for further analysis and interpretation, a strategic approach designed to reduce potential hallucination effects from large language model. We note that an experienced chemist may be able to do this without consultation, but the GPT-4 was able to offer three different directions by evaluating the circumstances and help human researcher, especially those who are not so familiar with some of the standard characterization techniques in reticular chemistry, to move on. Next, the thermogravimetric experiment was performed under nitrogen with a heating rate of 5 °C/min. The activated sample demonstrated a two-step weight loss and indicated that almost no solvent or water molecules were present in the pore (Figures S15–S18). We notice that the observed first weight loss between 250 – 350 °C matches with the weight of the formate ligand (calc. 19.2%; obs. 20.1%), and the second step (600 – 700°C) can be attributed to the decomposition of the framework.

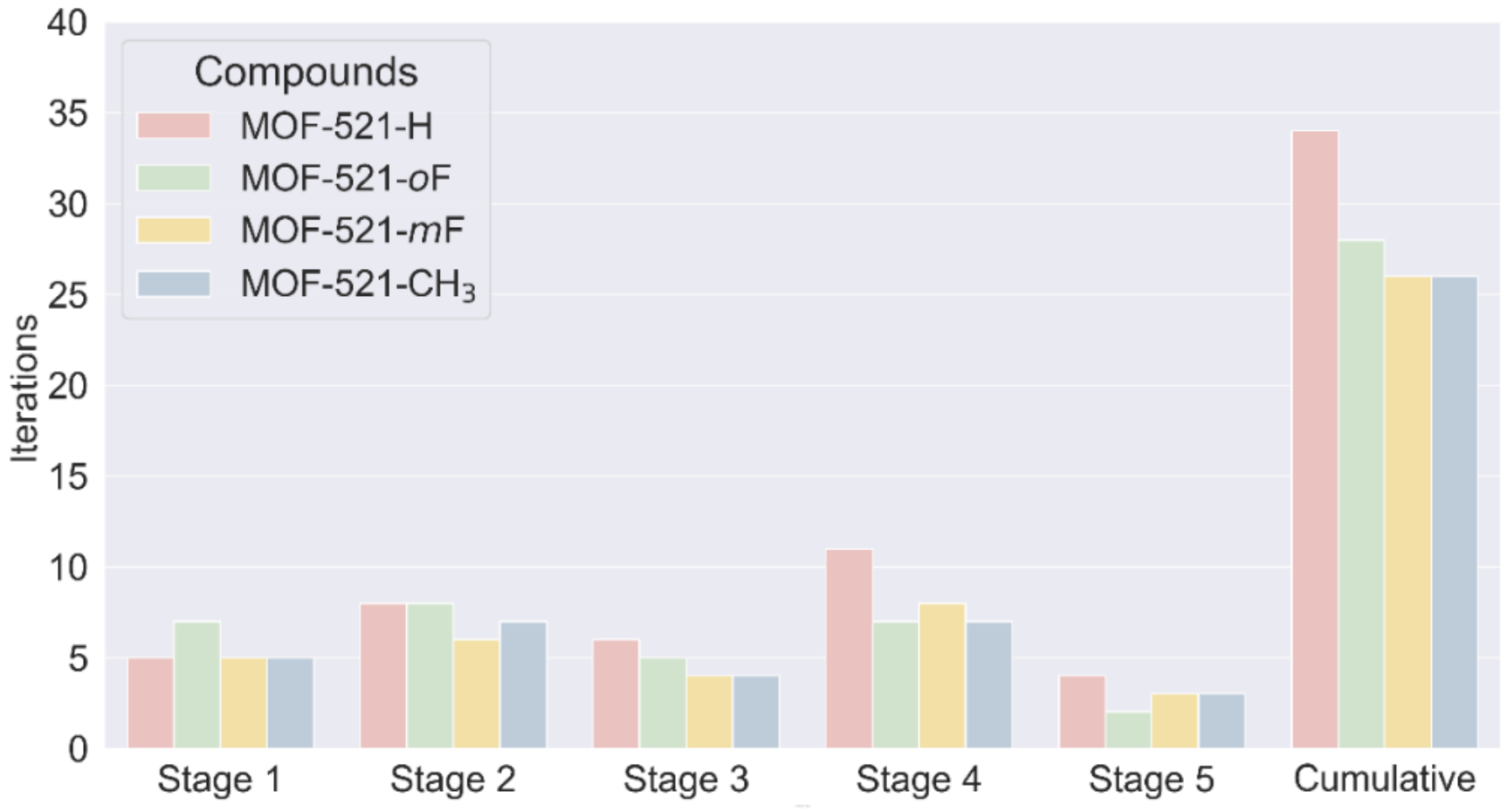

**Figure 4.** Number of prompting iteration took to complete each stage for the MOF-521 compounds based on BTB-H, BTB-*o*F, BTB-*m*F and BTB-$CH_3$ linkers. The number of iterations is a function of (i) the human input provided to AI, and (ii) the task picked by the human at each iteration of the approach.

As shown in Figure 4, the total numbers of prompt iterations are similar across the four MOF-521 compounds. The number of prompt iterations serves as a valuable measure of the efficiency and complexity of the synthesis, characterization and analysis of each compound. This observation is noteworthy as it underscores the reproducibility and robustness of GPT-4's effectiveness in acting as a reticular chemist to uncover new MOFs. The similarity in iterations across compounds is particularly noteworthy considering that each compound's discovery route differs in aspects such as organic linker preparation, synthesis condition optimization, and property characterization. In this light, the four linkers can be conceptualized as distinct discovery narratives, with each narrative standing as a testament to GPT-4's reproducible efficacy in acting as a reticular chemist for MOF exploration. Moreover, even within identical stages, GPT-4 encountered varied routes. For instance, in the synthesis condition screening phase, one route might prioritize temperature while another emphasizes the metal-to-linker ratio. This flexible approach by GPT-4, adapting and providing tailored suggestions to divergent scenarios, showcases its adeptness at addressing distinct challenges. The consistent number of iterations for each compound suggests that, irrespective of the distinct routes and decisions made, the steps and workload to achieve optimal stage goals remain similar. This consistency underscores GPT-4's ability to avoid situations where it becomes entangled in indecision or ineffective pathways.

Such observations validate not only the method's robustness but also its broader potential applicability for LLM-assisted scientific research. The iterative consistency further ratifies our division into five stages, emphasizing both its logical nature and efficiency. This segmentation, by breaking down the broader MOF discovery into digestible milestones, mirrors a typical human approach to complex tasks. Such structuring ensures evenly distributed efforts across stages, preventing any one stage from being disproportionately challenging or simple. In the meantime, it is essential to acknowledge that the role of GPT-4 in this research was not simply about discovering the most efficient path, but rather acting as a discerning consultant, providing expert suggestions tailored to researchers at various levels of expertise. As an illustrative point, when juxtaposed against human expert perspectives, over 80% of GPT-4's task recommendations were congruent with the immediate challenges faced by researchers and had evidently been taken account into prior project trajectory to ensure the task suggestions are helpful and practical (Table S5). This approach emphasizes a nuanced human-AI collaboration where the AI model serves as a valuable tool to guide and support rather than replace human judgment, reflecting the essence of a symbiotic relationship.

The in-context learning by GPT-4 Reticular Chemist within the system is further evidenced by the observation that the total iteration count for the latter three compounds was less than that for MOF-521-H. This is attributable to the fact that MOF-521-H, being the initial compound, lacked a precedent for guidance. However, once MOF-521-H successfully navigated through all stages, its compiled summary functioned as a directive for subsequent compounds, thereby expediting their processes. In essence, generating a summary based on past successes and failures is analogous to creating a repository of useful tools and guidance that can self-instruct other instances of GPT-4 Reticular Chemist to propose more efficient tasks for the discovery and development of similar MOFs. This reflects the system's inherent capability for progressive learning, emphasizing not only the feasibility and efficiency of this approach but also the potential for extending such methods to newcomers and cross-disciplinary researchers.

## Conclusion

This research signifies a noteworthy innovation in the traditional conduct of chemical experimentation, bridging the gap between AI and reticular chemistry by integrating the large language model, GPT-4, into the iterative chemical experimentation process. The developed framework emphasizes a symbiotic human-AI collaboration, facilitating research on the synthesis and characterization of new MOFs, as demonstrated by the four isoreticular MOFs. Our findings also underline the value of GPT-4's guidance across various stages of discovery, including areas that may appear similar in synthesis strategy or optimal conditions. The nuanced guidance provided by the model helps in interpreting the results more effectively and fine-tuning the specific conditions, even where similarities exist. This approach showcases the model's potential for scalable application in various scientific fields, adding to its unique contributions.

Moreover, the segmentation of this framework into three interactive phases underpins a manageable, systematic approach to material discovery, laying a foundation towards a visionary goal: the full automation and robotization of the experimental steps via automatic data mining[20] and high-throughput experiments[21]. Moreover, the progressive learning capability embedded within the system showcases the model's potential for scalable application in various scientific fields. Ultimately, this work serves as a pioneering step in the convergence of AI and reticular chemistry, opening new possibilities for future scientific discovery. It also serves as a testament to the constructive role of AI in driving scientific research by providing valuable insights and directions, akin to what a human mentor or collaborator might offer in a research setting.

## Acknowledgements

Z.Z. would like to extend special gratitude to Jiayi Weng from OpenAI for inspiring discussions on harnessing the potential of GPT-4. We appreciate the guidance from Drs. Xiaokun Pei, Ha Lac Nguyen, and Chuanshuai Li of the Yaghi Lab, Prof. Davide Proserpio, and Prof. Michael O'Keeffe, who provided insightful discussions on the topology of the new MOFs. We also acknowledge the financial support from the Defense Advanced Research Projects Agency (DARPA) under contract HR0011-21-C-0020 and the Bakar Institute of Digital Materials for the Planet (BIDMaP). We also thank the NIH (Grant S10-RR027172) for financial support of the X-ray crystallographic facility at UC Berkeley, and Dr. Nicholas Settineri for the support on using the facility. This research used resources of beamline 12.2.1 at the Advanced Light Source, which is a DOE Office of Science User Facility under contract no. DE-AC02-4105CH11231. In addition, Z.Z. is grateful for the financial support received through a Kavli ENSI Graduate Student Fellowship.

**Keywords:** metal–organic frameworks • synthesis • crystals • large language model • artificial intelligence

# Supporting Information

## A GPT-4 Reticular Chemist for Guiding MOF Discovery

Zhiling Zheng[a,b,c], Zichao Rong[a,b,c], Nakul Rampal[a,b,c], Christian Borgs[c,d],

Jennifer T. Chayes[c,d,e,f,g], and Omar M. Yaghi*[a,b,c,h]

a Department of Chemistry, University of California, Berkeley, California 94720, United States
b Kavli Energy Nanoscience Institute, University of California, Berkeley, California 94720, United States
c Bakar Institute of Digital Materials for the Planet, College of Computing, Data Science, and Society, University of California, Berkeley, California 94720, United States
d Department of Electrical Engineering and Computer Sciences, University of California, Berkeley, California 94720, United States
e Department of Mathematics, University of California, Berkeley, California 94720, United States
f Department of Statistics, University of California, Berkeley, California 94720, United States
g School of Information, University of California, Berkeley, California 94720, United States
h KACST–UC Berkeley Center of Excellence for Nanomaterials for Clean Energy Applications, King Abdulaziz City for Science and Technology, Riyadh 11442, Saudi Arabia
* Corresponding Author Email: yaghi@berkeley.edu

**Table of Contents**

## Section S1. Materials and Methods

**Large Language Model.** The Large Language Model (LLM) involved in this study is GPT-4, which was developed and is maintained by OpenAI. GPT-4 is an autoregressive language model that employs the transformer architecture.[1] It is crucial to point out that there are two primary modes for interfacing with GPT-4: firstly, via the web-based interaction of ChatGPT, operating on the GPT-4 model and accessible directly from the OpenAI website; secondly, through the application programming interface (API) of the GPT-4 model. While both are suitable for supporting our architecture, for the objectives outlined in this study, a more widely accessible variant of the GPT-4-based chatbot, denoted as ChatGPT or GPT-4, was utilized via the OpenAI official web portal at chat.openai.com. We chose the web-based chatbot interface also because it is easier to use and does not require any coding experience from scientists. All tasks involved in this study could be executed under the Default mode using GPT-4 as backbone model. Specifically, the chat model on which this study is based was the ChatGPT March 14, 2023 version[2], which is supported by *gpt-4-0314* model. Additionally, the choice of GPT-4 as the base model over GPT-3.5 was made considering GPT-4's larger model size and better inferencing capabilities,[2-3] along with its broader domain knowledge, making it more suitable for the diverse and detailed tasks required in this research. Subsequently, we present three distinct phases where the GPT-4 model was instructed by prompt. These phases are named "Reticular ChemScope," "Reticular ChemNavigator," and "Reticular ChemExecutor." Each phase has unique roles as determined by its respective prompt. It's crucial to highlight that all three phases are supported by GPT-4, with the different names signifying the various aims induced by their predefined prompts with designated roles.

**Reticular ChemScope.** In the first phase, GPT-4 was tasked with reading and understanding the standard practices of reticular chemistry, the project's goals and objectives, and current instruments. It then needed to apply this knowledge to develop a broader vision for the project. The term 'Scope' here refers to GPT-4's ability to 'zoom out' and perceive the big picture of the project, breaking it down into a series of stages and major goals. In particular, the GPT-4 model was instructed to behave like a professional reticular chemist and was tasked with comprehending and processing the standard practices of reticular chemistry, the objectives, and goals of the project, as well as the current instruments available and some other necessary information available. Using this information, GPT-4 constructed an overarching scheme or blueprint of the project, segmented into distinct stages with their corresponding goals. The human was then able to chat with GPT-4 to refine the blueprint of the project and provide more expectations if necessary. All the interaction was made using natural language and no code is needed. The given text from a peer-reviewed and well-cited reticular chemistry standard practice paper will allow for minimizing the hallucinations from GPT-4 model, leading to an effective means to visualize the big picture and enabling a systematic workflow for the project (Figure S37).

**Reticular ChemNavigator.** GPT-4 was mainly responsible for suggesting tasks and experiments, evaluating progress, and guiding the human counterpart through the stages of the project. The name of this phase emphasizes the element of navigation in the iterative process, as the GPT-4 navigates through trial-and-error summaries, human feedback, and current situations, all in pursuit of reaching each stage's completion. Therefore, the second phase involved an interactive dialogue between the human researcher with GPT-4. Here, based on the big picture designed in the previous phase, GPT-4 provided three task suggestions for the human participants to undertake, all aimed at achieving the objectives of the current project stage. Feedback from human participants, as well as previous trial-and-error summaries, were incorporated in this phase. These components were utilized by GPT-4 to evaluate the current situation and devise subsequent tasks, guiding the project toward the successful completion of each stage. In each iteration GPT-4 acting as Reticular ChemNavigator is reinitiated for every new prompt with the context on research summary, previous task suggested and the human feedback (Figure S39).

**Reticular ChemExecutor.** GPT-4 functioned in giving detailed steps for tasks and providing a feedback template. 'Executor' indicates GPT-4's role in enabling the execution of specific tasks, detailing the steps, and facilitating a structured feedback mechanism. Consequently, the final phase had GPT-4 detailed steps for a task chosen by the human participants from the three suggestions given in the previous phase. GPT-4 also provided a template for structured feedback, which allowed for effective communication and improvements in the implementation of tasks. In each iteration GPT-4 acting as Reticular ChemExecutor is reinitiated for every new prompt with the context on the task suggestion given by Reticular ChemNavigator as well the summary of previous activities (Figure S41).

**Prompt Components.** The prompt components to prompt GPT-4 can be classified into the following five categories: role definition and goal, memory, evaluation, and structured output. Our goal for the incorporation of these elements into the prompt was to enable GPT-4's functionalities to align with the requirements of human-GPT-4 interaction in the complex reticular chemistry research process. The design incorporated various elements, each playing a critical role, and the reasons for the choice of these components are described as follows. First, in the role definition, the inclusion of specific role descriptions within the prompt establishes a clear context for GPT-4's suggestions, allowing GPT-4 to assume a role akin to a collaborative partner in the research process, working in the field of reticular chemistry. This ensures that the recommendations provided are in accordance with the overall objectives and constraints of the project. In the meantime, regarding goal instructions, by embedding the main goals and specific targets within the prompt, GPT-4 can align its suggestions with the overarching aims of the research. This alignment is crucial for meaningful human-GPT-4 collaboration, directing the focus of GPT-4's outputs toward the successful completion of the project. Also, this will ensure that the suggestions made by GPT-4 do not go beyond the scope of the activities within reticular chemistry. Next, concerning the memory of previous trials, a summary of prior experiments, successes, and failures allows GPT-4 to process and recall important historical information. This aspect aids in the accumulation of past learning and facilitates intelligent suggestions that build upon previous experiences. It promotes a more nuanced, adaptive approach that can reflect on past actions to guide future decisions. Due to token limitations of the model, it is impossible to include all human-GPT-4 interaction history throughout the research process. As more and more information is added, GPT-4 can lose track of previous history, thus a short memory serves as a good tool for it to know the previous history. Following this, the evaluation process is critical as an embedded evaluation process enables GPT-4 to make assessments on the current state of the project, including successes or failures in previous trials. This not only adds depth to GPT-4's understanding but also enhances the transparency of its decision-making process, enabling humans to follow and critically evaluate the logic behind its recommendations. Finally, with structured output as three task choices, providing structured output in the form of three task choices adds flexibility and adaptability to GPT-4's responses. It takes into consideration practical constraints such as the availability of resources and time schedules, offering a variety of options that can suit different scenarios.

**Interactive Prompt Refinement.** The main aim was to utilize GPT-4's capability to self-instruct prompts based on the provided role descriptions and duties, and then refine these prompts through an iterative process of suggesting, testing, and improvement. Initially, raw prompts were composed by human users, which contained role descriptions, duties, and necessary supplementary information tailored to each phase. For instance, in the ChemScope phase, texts from reticular chemistry literature were used, outlining the standard process of reticular chemistry research. The ChemNavigator phase included a predefined project with a certain number of stages and their corresponding outcome goals, along with definitions of summaries. In the ChemExecutor phase, detailed descriptions of tasks chosen by human participants were included to guide the execution process. Next, GPT-4 was prompted with these raw inputs and the objectives of the prompts, encouraging the GPT-4 to ask questions for clarification and further refining the prompt. This active dialogue allowed the prompts to be revised and improved according to the AI's queries and the human users' feedback. After this initial loop, the first drafts of the prompts generated by GPT-4 were tested on a separate instance of GPT-4, devoid of any prior conversational context. The effectiveness of the prompts was assessed based on the generated outputs. Based on the test results, feedback was relayed back to the original GPT-4 instance (the prompt writer), highlighting the areas that needed addressing or improvement. The prompts were then updated, and the cycle of testing and refinement continued until the performance of the prompts met the satisfaction of the human users. The full prompt of Reticular ChemScope, Reticular ChemNavigator, and Reticular ChemExecutor obtained from this refinement strategy and demonstration on this process can be found in Section S8.

**Starting Materials.** Aluminum nitrate nonahydrate [$Al(NO_3)_3 \cdot 9H_2O$], *N,N*-dimethylformamide (DMF) (≥99.8%, HPLC), dichloromethane ($CH_2Cl_2$) (anhydrous, ≥99.8%, HPLC), and ethyl acetate (≥99.5%, HPLC) were purchased from Sigma-Aldrich Company. Bis(pinacolato)diboron, 1,3,5-tribromobenzene, 4-methoxycarbonylbenzeneboronic acid, 1,3,5-tribromo-2,4,6-trimethylbenzene, methyl 4-bromo-2-fluorobenzoate, and methyl 4-bromo-3-fluorobenzoate were purchased from Chemscene LLC. Cesium fluoride (CsF) was purchased from TCI America. Anhydrous acetone was purchased from Acros Organics. Formic acid (99 %) was obtained from EMD Millipore. [1,1′-Bis(diphenylphosphino)ferrocene] dichloropalladium(II) (Pd(dppf)$Cl_2$) and anhydrous magnesium sulfate ($MgSO_4$) were purchased from Fisher Scientific.

**Analytical Techniques and Instruments.** Elemental analysis was performed in the Microanalytical Laboratory of the College of Chemistry at UC Berkeley, using a PerkinElmer 2400 Series II CHNS elemental analyzer. PXRD patterns were recorded using a Bruker D8 Advance diffractometer with Cu Kα radiation (λ = 1.54056 Å). $^1$H NMR spectra were acquired on a Bruker AVB-400 NMR spectrometer or a Bruker AVQ-500 NMR spectrometer. Thermogravimetric analysis (TGA) curves were taken using a TA Q500 thermal analysis system with the heating rate of 5 °C/min under $N_2$ flow. $N_2$ adsorption isotherms were recorded on the ASAP 2420 system at 77 K with liquid nitrogen bath. Ultra-high-purity grade nitrogen gas (Praxair, 99.999% purity) was used for the adsorption measurements.

**X-Ray Diffraction Collection.** SXRD measurements on MOF-521-H and MOF-521-mF were performed on a Rigaku XtaLAB P200 equipped with a MicroMax 007HF rotating anode and a Pilatus 200K hybrid pixel array detector. Crystal was mounted on a Kapton®MiTeGen MicroMount™ in a minimal amount of Paratone® N oil and kept at 100 K throughout the collection. Data were collected using Cu Kα radiation (λ = 1.5406 Å). The diffraction data were processed in the CrysAlisPro software with a multi-scan absorption correction based on SCALE3 ABSPACK.[4]

SXRD measurement on MOF-521-oF was performed at the beamline 12.2.1 at the Advanced Light Source of Lawrence Berkeley National Laboratory (λ = 0.7288 Å) equipped with a PHOTON-II CMOS detector operating in a shutterless mode and a Si(111) monochromator. Crystal was mounted on a Kapton®MiTeGen MicroMount™ in a minimal amount of Paratone® N oil and kept at 100 K throughout the collection. The diffraction data were processed in the Bruker APEX4 software.[5] The data were integrated by SAINT program[6] followed by a multi-scan absorption correction carried out by SADABS program[7].

## Section S2. Experimental Section

### S2.1. Synthetic Procedures of Organic Linkers

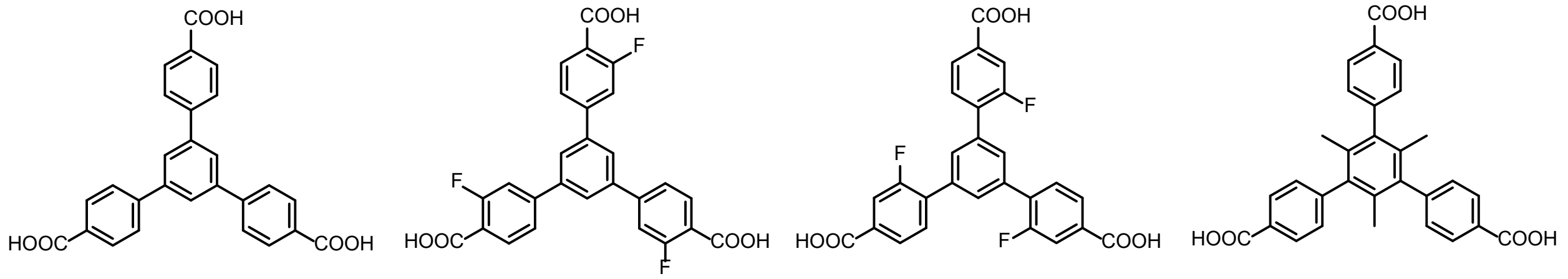

$H_3BTB$ / $H_3BTB$-H  $H_3BTB$-*o*F  $H_3BTB$-*m*F  $H_3BTB$-$CH_3$

**Figure S1.** Illustration of structures of organic linkers and their abbreviations.

**Synthesis of 1,3,5-Tris(4-carboxyphenyl)benzene ($H_3BTB$-H).** The synthesis of the linker was carried out with slight modifications to the process detailed in previous literature.[8] Initially, 1,3,5-tribromobenzene (1.00 g, 3.1 mmol), 4-Methoxycarbonylphenylboronic acid (2.05 g, 11.4 mmol) and potassium acetate (2.35 g, 24 mmol) were combined in 1,4-dioxane (80 mL). The reaction mixture was then degassed for 10 minutes under an argon atmosphere. Subsequently, Pd(dppf)$Cl_2$ (0.03 g, 0.04 mmol) was added to the reaction mixture under stirring conditions. The mixture was heated to 85°C for 4 days under argon. Post-heating, the solution was evaporated until dry, and the residue was subsequently extracted using $CH_2Cl_2$. The final product, BTB-H, was procured by hydrolyzing the crude product using 2M aqueous NaOH, followed by acidification using concentrated HCl (71% yield). $^1$H NMR (500 MHz, DMSO-$d_6$) δ8.09 (s, 6H), 8.06-8.07 (m, 12H).

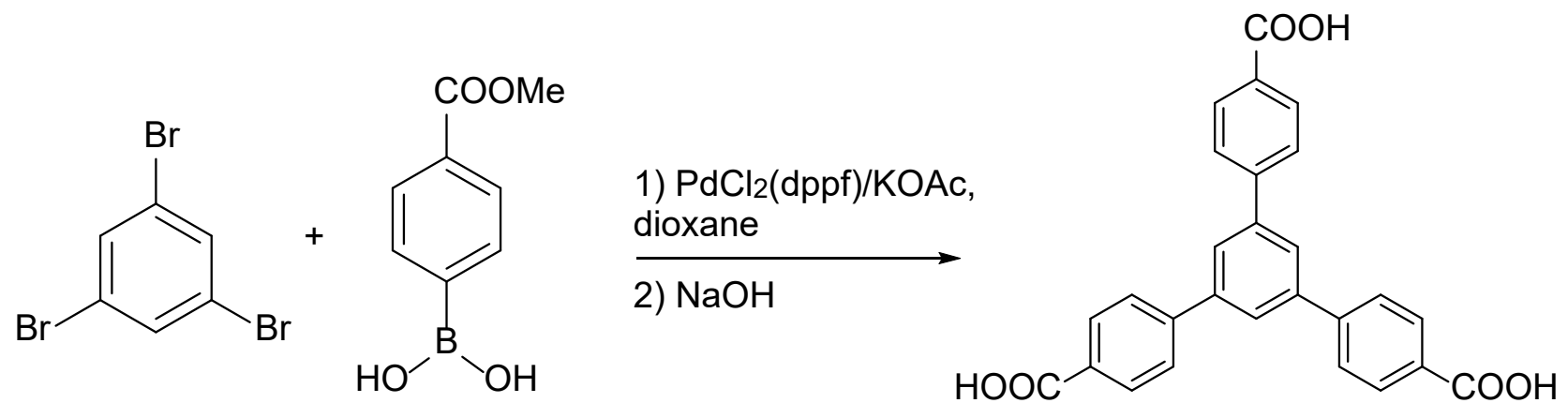

**Figure S2.** Synthesis of linker BTB-H.

**Synthesis of 1,3,5-Phenyltriboronic Acid Tris(pinacol)ester.** The precursor was synthesized according to the literature with some modification.[9] In a three-neck round bottom flask, anhydrous DMF (10 mL) was purged with nitrogen gas for 30 minutes. 1,3,5-tribromobenzene (10.00 g, 31.7 mmol), bis(pinacolato)diboron (25.4 g, 95.3 mmol), potassium acetate (18.7 g, 0.19 mol) and Pd(dppf)Cl2 (0.87 g, 1.2 mmol) were then quickly added into the flask. The resulting mixture was stirred vigorously and heated at 90 °C for 48 hours. Deionized water (500 mL) was added after the mixture cooled down to room temperature. Black precipitate was collected by filtration, and washed with deionized water three times, which was dried under vacuum (91% yield). $^1$H NMR (500 MHz, DMSO-$d_6$) δ 8.11 (s, 3H), 1.30 (s, 36H).

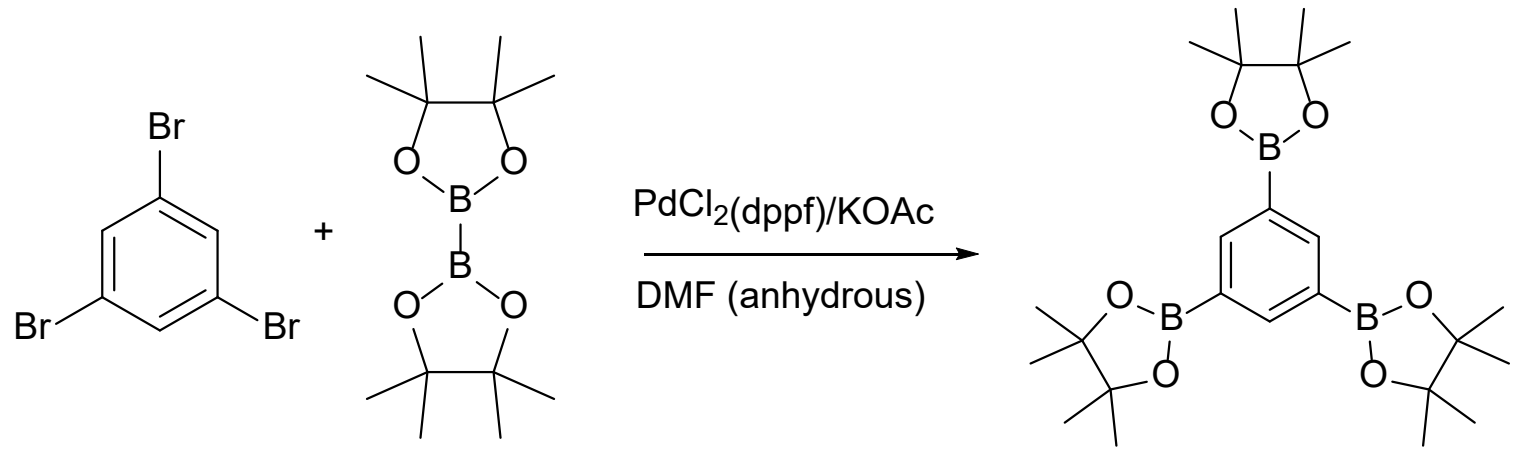

**Figure S3.** Synthesis of precursor 1,3,5-phenyltriboronic acid tris(pinacol)ester.

**Synthesis of 4-[3,5-Bis(4-carboxy-3-fluorophenyl)phenyl]-2-fluorobenzoic Acid ($H_3$BTB-*o*F).** The linker was synthesized according to the literature with some modification.[9] A mixture of dried 1,3,5-phenyltriboronic acid tris(pinacol)ester (0.91g, 2.0 mmol) and methyl 4-bromo-3-fluorobenzoate (1.6 g, 6.5 mmol) was dissolved in 48 mL mixed solvent of p-dioxane/H2O (1:1 v/v) in a three-neck round bottom flask, which was purged and protected under $N_2$ atmosphere After quickly adding of CsF (2.7 g, 18 mmol) and Pd(dppf)$Cl_2$ (0.11 g, 0.15 mmol), the suspension was heated and stirred vigorously at 90 °C for 24 hours. After cooling down to room temperature, 500 mL of 20% NaCl solution was added to the resulting suspension, and the mixture was extracted three times with 70 mL EtOAc using a 500-mL separatory funnel. The organic layers were combined, washed with saturated brine, dried with anhydrous $MgSO_4$ and filtered. A crude product was obtained after removing all the solvent by rotary evaporation, and further purified by quick chromatography using $CH_2Cl_2$/Hexane (8:1 v/v) as eluent. The final product BTB-*o*F was obtained by hydrolyzing the crude product in 9 mL THF and 0.5 M NaOH aqueous solution (9.0 mL, 4.5 mmol). The mixture was stirred vigorously at 50 °C for 24 hours. After removing the THF by rotary evaporation, the aqueous solution was acidified with concentrated HCl to pH = 2. The white precipitate was collected by filtration, washed with deionized water, and dried under vacuum (52% yield). $^1$H NMR (500 MHz, DMSO-$d_6$) δ7.89-7.87 (m, 6H), 7.86-7.80 (m, 6H).

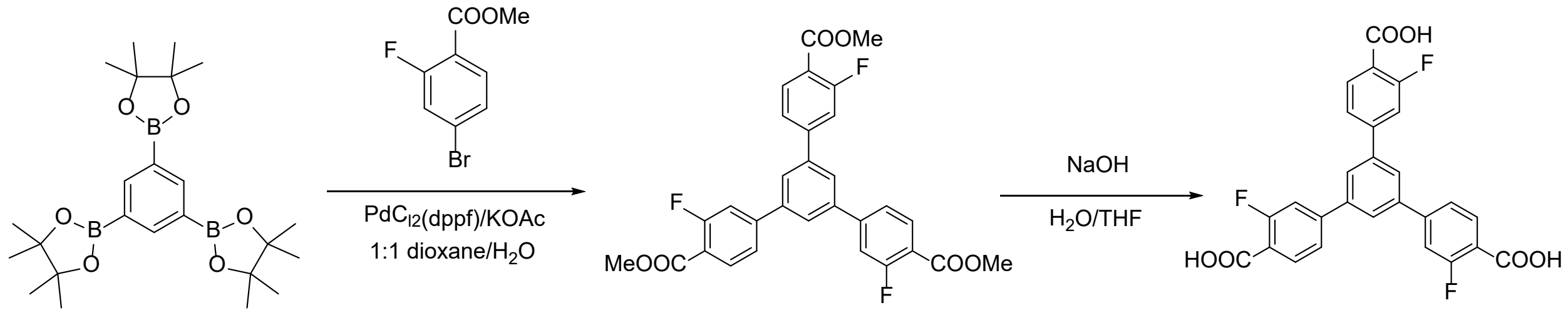

**Figure S4.** Synthesis of linker BTB-*o*F.

**Synthesis of 5'-(4-Carboxy-2-fluorophenyl)-2,2''-difluoro-[1,1':3',1''-terphenyl]-4,4''-dicarboxylic Acid ($H_3$BTB-*m*F).** The linker was also known as 4,4',4''-(2,4,6-trimethylbenzene-1,3,5-triyl)tribenzoic acid and was synthesized according to the literature with some modification.[9] In a three-neck round bottom flask, a mixture of dried 1,3,5-phenyltriboronic acid tris(pinacol)ester (0.91g, 2.0 mmol) and methyl 4-bromo-3-fluorobenzoate (1.6 g, 6.5 mmol) was dissolved in 48 mL mixed solvent of p-dioxane/$H_2O$ (1:1 v/v) under a purged and protected $N_2$ atmosphere. Quickly adding CsF (2.7 g, 18 mmol) and Pd(dppf)$Cl_2$ (0.11 g, 0.15 mmol) into the flask, the mixture was then heated and stirred vigorously at 90 °C for 24 hours. The resulting suspension was cooled to room temperature, mixed with 500 mL of 20% NaCl solution, and extracted three times with 70 mL EtOAc using a 500-mL separatory funnel. After combining and washing the organic layers with saturated brine, it was dried with anhydrous $MgSO_4$ and filtered. Solvent was removed by rotary evaporation to obtain a crude product, which was further purified by quick chromatography using $CH_2Cl_2$/Hexane (5:1 v/v) as eluent. The crude product was then hydrolyzed in 9 mL THF and 0.5 M NaOH aqueous solution (9.0 mL, 4.5 mmol) to generate the final product BTB-*m*F. This was stirred vigorously at 50 °C for 24 hours, then after the removal of THF by rotary evaporation, the aqueous solution was acidified to pH = 2 with concentrated HCl. The white precipitate was collected by filtration, washed with deionized water, and dried under vacuum (42% yield). $^1$H NMR (500 MHz, DMSO-$d_6$) δ8.19 (s, 3H), 8.03-7.91 (m, 9H).

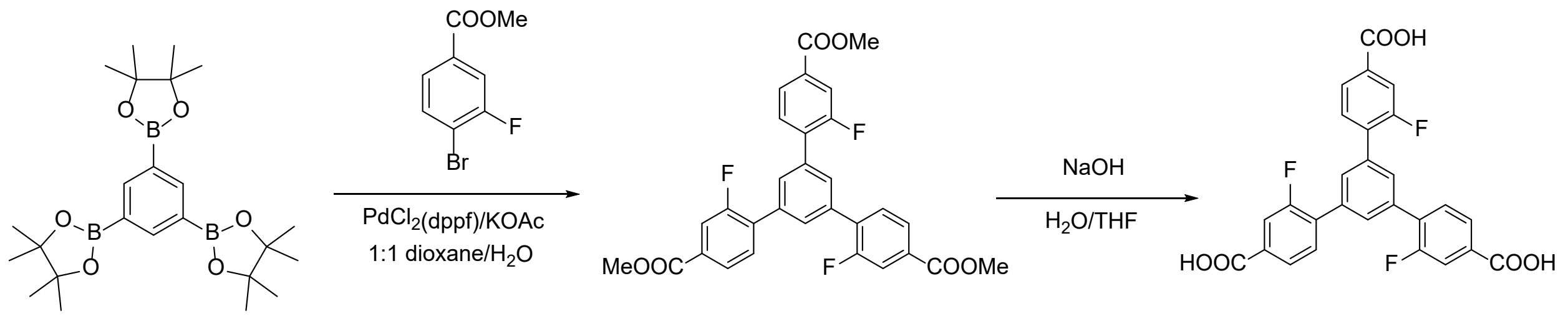

**Figure S5.** Synthesis of linker BTB-*m*F.

**Synthesis of 4,4′,4″-(2,4,6-Trimethylbenzene-1,3,5-triyl)tribenzoic Acid ($H_3$BTB-$CH_3$).** The linker was synthesized according to the literature with some modification.[10] In a three-neck round bottom flask, anhydrous dioxane (500 mL) and water (50 mL) was purged with nitrogen gas for 30 minutes. 1,3,5-tribromo-2,4,6-trimethylbenzene (5.3 g, 15 mmol), 4-methoxycarbonylbenzeneboronic acid (21 g, 75 mmol), potassium acetate (14 g, 0.15 mol) and Pd(dppf)$Cl_2$ (0.65 g, 0.9 mmol) were then quickly added into the flask. The resulting mixture was stirred vigorously and heated at 100 °C for 72 hours. Deionized water (500 mL) was added after the mixture cooled down to room temperature. and the mixture was extracted three times with 200 mL EtOAc. The organic layers were combined, washed with saturated brine, dried with anhydrous $MgSO_4$ and filtered. A crude product was obtained after removing all the solvent by rotary evaporation, and further purified by quick chromatography using EtOAc/Hexane (1:20 v/v) as eluent. The final product BTB-$CH_3$ was obtained by hydrolyzing the crude product in 200 mL THF and 0.5 M NaOH aqueous solution (150 mL, 0.75 mol). The mixture was stirred vigorously at 50 °C for 24 hours. After removing the THF by rotary evaporation, the aqueous solution was acidified with concentrated HCl to pH = 2. The white precipitation was collected by filtration, washed with deionized water, and dried under vacuum (68% yield). $^1$H NMR (500 MHz, DMSO-d6) δ8.03 (d, J = 8.5 Hz, 6H), 7.37 (d, J = 8.5 Hz, 6H), 1.63 (s, 9H).

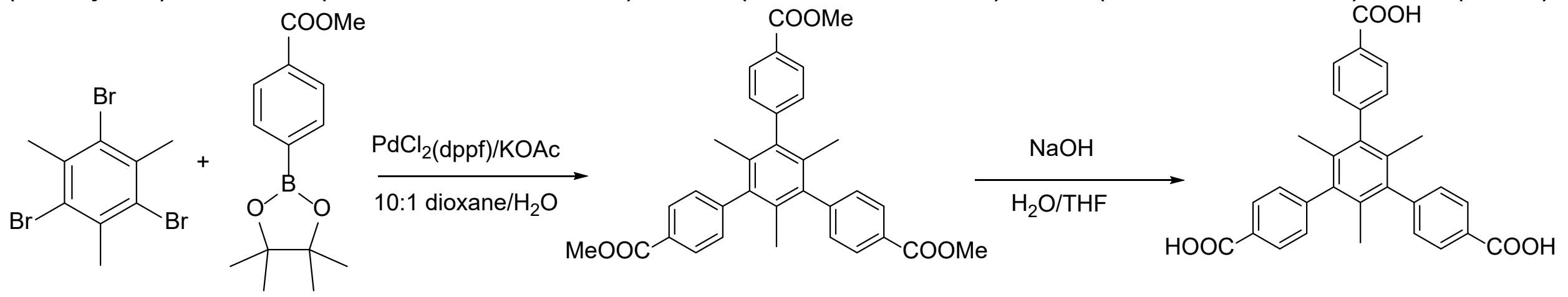

**Figure S6.** Synthesis of linker BTB-$CH_3$.

## S2.2. Synthetic Procedures of Metal-Organic Frameworks

**Optimized Synthesis of MOF-521-H Single Crystals.** In a 4 mL scintillation vial, 3.0 mL DMF was added to dissolve $Al(NO_3)_3·9H_2O$ (18.0 mg, 0.048 mmol) and $H_3BTB$ (15.8 mg, 0.036 mmol). After adding 0.24 mL formic acid and 30 μL deionized water to the solution, the vial was capped and placed in the preheated 140 °C for 2 days, and colorless needle-shape crystals were obtained. The crystals were washed with MeOH (3 × 20 mL) for 3 days before activation and characterization. Full activation of the MOF was conducted under dynamic vacuum (~$10^{-3}$ mbar) at 100°C for 12 hours, yielding pure and desolvated product.

**Optimized Synthesis of MOF-521-*o*F Single Crystals.** In a 4 mL scintillation vial, 3.0 mL DMF was added to dissolve $Al(NO_3)_3·9H_2O$ (18.0 mg, 0.048 mmol) and $H_3BTB$-*o*F (17.7 mg, 0.036 mmol). After adding 0.24 mL formic acid and 30 μL deionized water to the solution, the vial was capped and placed in the preheated oven at 120 °C for 2 days, and colorless needle-shape crystals were obtained. The crystals were washed with MeOH (3 × 20 mL) every 12 hours before activation and characterization. Full activation of the MOF was conducted under dynamic vacuum (~$10^{-3}$ mbar) at 120°C for 24 hours, yielding pure and desolvated product.

**Optimized Synthesis of MOF-521-*m*F Single Crystals.** In a 4 mL scintillation vial, 3.0 mL DMF was added to dissolve $Al(NO_3)_3·9H_2O$ (18.0 mg, 0.048 mmol) and linker $H_3BTB$-*m*F (17.7 mg, 0.036 mmol). After adding 0.24 mL formic acid and 30 μL deionized water to the solution, the vial was capped and placed in the preheated oven at 120 °C for 3 days, and colorless needle-shape crystals were obtained. The crystals were washed with MeOH (6 × 20 mL) every 8 hours before activation and characterization. Full activation of the MOF was conducted under dynamic vacuum (~$10^{-3}$ mbar) at 120°C for 12 hours, yielding pure and desolvated product.

**Optimized Synthesis of MOF-521-$CH_3$ Small Single Crystals.** In a 4 mL scintillation vial, 3.0 mL DMF was added to dissolve $Al(NO_3)_3·9H_2O$ (18.0 mg, 0.048 mmol) and $H_3BTB$-$CH_3$ (17.3 mg, 0.036 mmol). After adding 0.4 mL formic acid and 30 μL deionized water to the solution, the vial was capped and placed in the preheated oven at 130 °C for 3 days, and colorless needle-shape crystals were obtained. The crystals were washed with MeOH (2 × 20 mL) for 2 days before activation and characterization. Full activation of the MOF was conducted under dynamic vacuum (~$10^{-3}$ mbar) at 100°C for 12 hours, yielding pure and desolvated product.

**Synthesis of Mixed MOF-520 and MOF-521-H.** We observed in our prior studies that manual synthesis of MOFs, per established literature procedures[11], resulted in a simultaneous production of MOF-521-H alongside MOF-520. To elaborate, 90 mg (0.24 mmol) of $Al(NO3)_3·9H_2O$ was solubilized in 2 mL of anhydrous DMF, concurrently 75 mg (0.17 mmol) of $H_3BTB$ was also dissolved in a separate 2 mL solution of anhydrous DMF. These two solutions were combined, to which an additional 13 mL of anhydrous DMF was added. This was succeeded by the addition of 1 mL of formic acid and 0.3 mL of deionized water. The resultant solution was sealed in a 20 mL scintillation vial and subsequently subjected to heating at 140 °C for a duration of 3 to 5 days. The process culminated in the formation of a mixture of colorless block and needle crystals. The synthesized crystals were then washed with DMF (9 × 20 mL) over the span of 3 days and immersed in $CH_2Cl_2$ prior to activation.

**Elemental Analysis Data for the Activated MOFs**. *MOF-521-H*: Calcd for $Al_3C_{30}H_{21}O_{15}$ = $Al_3$(μ-OH)$(HCOO)_3$(BTB): Al, 11.52; C, 51.30; H, 3.01; O, 34.17%. Found: C, 51.02; H, 3.60%.

*MOF-521-oF*: Calcd for $Al_3C_{30}H_{18}O_{15}F_3$ = $Al_3$(μ-OH)$(HCOO)_3$(BTB-*o*F): Al, 10.70; C, 47.64; H, 2.40; O, 31.73; F, 7.54%. Found: C, 49.65; H, 2.56%.

*MOF-521-m*F: Calcd for $Al_3C_{30}H_{18}O_{15}F_3$ = $Al_3$(μ-OH)$(HCOO)_3$(BTB-*m*F): Al, 10.70; C, 47.64; H, 2.40; O, 31.73; F, 7.54%. Found: C, 47.38; H, 2.63%.

*MOF-521-$CH_3$*: Calcd for $Al_3C_{33}H_{27}O_{15}$ = $Al_3$(μ-OH)$(HCOO)_3$(BTB-$CH_3$): Al, 10.87; C, 53.24; H, 3.66; O, 32.24%. Found: C, 51.79; H, 3.66%.

**Table S1.** Screening conditions for MOF-521-H, MOF-521-*o*F, MOF-521-*m*F, and MOF-521-$CH_3$ synthesis.

| Exp.[a] | Linker | Modulator[b] | L:M Ratio[c] | Temp. (°C) | Time (h) |
|---|---|---|---|---|---|
| 1 | BTB-H | FA | 1:1 | 100 | 48 |
| 2 | BTB-H | FA | 3:4 | 100 | 48 |
| 3 | BTB-H | FA | 2:3 | 100 | 48 |
| 4 | BTB-H | FA | 1:2 | 100 | 48 |
| 5 | BTB-H | FA | 1:4 | 100 | 48 |
| 6 | BTB-H | FA | 2:1 | 100 | 48 |
| 7 | BTB-H | FA | 4:1 | 100 | 48 |
| 8 | BTB-H | FA | 1:1 | 120 | 48 |
| 9 | BTB-H | FA | 3:4 | 120 | 48 |
| 10 | BTB-H | FA | 2:3 | 120 | 48 |
| 11 | BTB-H | FA | 1:2 | 120 | 48 |
| 12 | BTB-H | FA | 1:4 | 120 | 48 |
| 13 | BTB-H | FA | 2:1 | 120 | 48 |
| 14 | BTB-H | FA | 4:1 | 120 | 48 |
| 15 | BTB-H | FA | 1:1 | 140 | 48 |
| 16 | BTB-H | FA | 3:4 | 140 | 48 |
| 17 | BTB-H | FA | 2:3 | 140 | 48 |
| 18 | BTB-H | FA | 1:2 | 140 | 48 |
| 19 | BTB-H | FA | 1:4 | 140 | 48 |
| 20 | BTB-H | FA | 2:1 | 140 | 48 |
| 21 | BTB-H | FA | 4:1 | 140 | 48 |
| 22 | BTB-H | FA | 1:1 | 160 | 48 |
| 23 | BTB-H | FA | 3:4 | 160 | 48 |
| 24 | BTB-H | FA | 2:3 | 160 | 48 |
| 25 | BTB-H | FA | 1:2 | 160 | 48 |
| 26 | BTB-H | FA | 1:4 | 160 | 48 |
| 27 | BTB-H | FA | 2:1 | 160 | 48 |
| 28 | BTB-H | FA | 4:1 | 160 | 48 |
| 29 | BTB-H | FA | 3:4 | 120 | 72 |
| 30 | BTB-H | FA | 3:4 | 140 | 72 |
| 31 | BTB-H | AA | 3:4 | 120 | 72 |
| 32 | BTB-H | TFA | 3:4 | 120 | 72 |
| 33 | BTB-H | TFA/AA (1:1) | 3:4 | 120 | 72 |
| 34 | BTB-H | FA/BA (1:1) | 3:4 | 120 | 72 |
| 35 | BTB-H | FA/HCl (1:1) | 3:4 | 120 | 72 |
| 36 | BTB-H | FA/$H_2O$ (1:1) | 3:4 | 120 | 72 |
| 37 | BTB-H | FA/TFA (1:1) | 3:4 | 120 | 72 |
| 38 | BTB-H | FA/$H_2O$ (4:1) | 3:4 | 120 | 72 |
| 39 | BTB-H | FA/$H_2O$ (2:1) | 3:4 | 120 | 72 |
| 40 | BTB-H | FA/$H_2O$ (1:1) | 3:4 | 120 | 72 |
| 41 | BTB-H | FA/$H_2O$ (1:2) | 3:4 | 120 | 72 |
| 42 | BTB-H | FA/$H_2O$ (1:2) | 3:4 | 120 | 72 |
| 43 | BTB-H | FA/$H_2O$ (1:4) | 3:4 | 120 | 48 |
| 44 | BTB-H | FA/$H_2O$ (4:1) | 3:4 | 120 | 60 |
| 45 | BTB-H | FA/$H_2O$ (4:1) | 3:4 | 120 | 72 |
| 46 | BTB-H | FA/$H_2O$ (4:1) | 3:4 | 120 | 84 |
| 47 | BTB-H | FA/$H_2O$ (4:1) | 3:4 | 120 | 96 |
| 48 | BTB-H | FA/$H_2O$ (4:1) | 3:4 | 100 | 72 |
| 49 | BTB-H | FA/$H_2O$ (4:1) | 3:4 | 110 | 72 |
| 50 | BTB-H | FA/$H_2O$ (4:1) | 3:4 | 120 | 72 |
| 51 | BTB-H | FA/$H_2O$ (4:1) | 3:4 | 130 | 72 |
| 52 | BTB-H | FA/$H_2O$ (4:1) | 3:4 | 140 | 72 |
| 53 | BTB-*o*F | FA | 1:4 | 140 | 48 |
| 54 | BTB-*o*F | FA | 2:1 | 140 | 48 |
| 55 | BTB-*o*F | FA | 4:1 | 140 | 48 |
| 56 | BTB-*o*F | FA/$H_2O$ (1:1) | 2:1 | 140 | 48 |
| 57 | BTB-*o*F | FA/$H_2O$ (2:1) | 2:1 | 140 | 48 |
| 58 | BTB-*o*F | FA/$H_2O$ (4:1) | 2:1 | 140 | 48 |
| 59 | BTB-*o*F | FA/$H_2O$ (4:1) | 1:2 | 140 | 48 |
| 60 | BTB-*o*F | FA/$H_2O$ (4:1) | 3:4 | 140 | 48 |
| 61 | BTB-*o*F | FA/$H_2O$ (4:1) | 1:1 | 140 | 48 |
| 62 | BTB-*o*F | FA/$H_2O$ (4:1) | 3:2 | 110 | 48 |
| 63 | BTB-*o*F | FA/$H_2O$ (4:1) | 3:4 | 120 | 48 |
| 64 | BTB-*o*F | FA/$H_2O$ (4:1) | 3:4 | 130 | 48 |
| 65 | BTB-*o*F | FA/$H_2O$ (4:1) | 3:4 | 140 | 48 |
| 66 | BTB-*o*F | FA/$H_2O$ (4:1) | 3:4 | 120 | 24 |
| 67 | BTB-*o*F | FA/$H_2O$ (4:1) | 3:4 | 120 | 48 |
| 68 | BTB-*o*F | FA/$H_2O$ (4:1) | 3:4 | 120 | 72 |
| 69 | BTB-*o*F | FA/$H_2O$ (4:1) | 3:4 | 120 | 96 |
| 70 | BTB-*m*F | FA/$H_2O$ (4:1) | 3:4 | 120 | 60 |
| 71 | BTB-*m*F | FA/$H_2O$ (4:1) | 3:4 | 120 | 72 |
| 72 | BTB-*m*F | FA/$H_2O$ (4:1) | 3:4 | 120 | 84 |
| 73 | BTB-*m*F | FA/$H_2O$ (4:1) | 3:4 | 110 | 72 |
| 74 | BTB-*m*F | FA/$H_2O$ (4:1) | 3:4 | 120 | 72 |
| 75 | BTB-*m*F | FA/$H_2O$ (4:1) | 3:4 | 130 | 72 |
| 76 | BTB-*m*F | FA/$H_2O$ (4:1) | 1:1 | 120 | 72 |
| 77 | BTB-*m*F | FA/$H_2O$ (4:1) | 3:4 | 120 | 72 |
| 78 | BTB-*m*F | FA/$H_2O$ (4:1) | 3:5 | 120 | 72 |
| 79 | BTB-*m*F | FA/$H_2O$ (5:1) | 3:4 | 120 | 72 |
| 80 | BTB-*m*F | FA/$H_2O$ (3:1) | 3:4 | 120 | 72 |
| 81 | BTB-*m*F | FA/$H_2O$ (4:1) | 3:4 | 120 | 72 |
| 82 | BTB-$CH_3$ | FA/$H_2O$ (4:1) | 3:4 | 120 | 48 |
| 83 | BTB-$CH_3$ | FA/$H_2O$ (4:1) | 3:4 | 120 | 72 |
| 84 | BTB-$CH_3$ | FA/$H_2O$ (4:1) | 3:4 | 130 | 72 |
| 85 | BTB-$CH_3$ | FA/$H_2O$ (4:1) | 3:4 | 130 | 72 |
| 86 | BTB-$CH_3$ | FA/$H_2O$ (4:1) | 1:2 | 130 | 72 |
| 87 | BTB-$CH_3$ | FA/$H_2O$ (4:1) | 3:4 | 130 | 72 |
| 88 | BTB-$CH_3$ | FA/$H_2O$ (4:1) | 1:1 | 130 | 72 |
| 89 | BTB-$CH_3$ | FA/$H_2O$ (4:1) | 3:4 | 130 | 72 |
| 90 | BTB-$CH_3$ | FA/$H_2O$ (4:1) | 3:4 | 140 | 72 |
| 91 | BTB-$CH_3$ | FA/$H_2O$ (4:1) | 3:4 | 150 | 72 |

[a] Experiment ID. [b] Modulator abbreviations: FA = formic acid; TFA = trifluoroacetic acid; AA = acetic acid; BA = benzoic acid. [c] This denotes the molar ratio of linker to metal ions.

## Section S3. $^1$H NMR Spectroscopy

Obtained BTB-X linkers (ca. 10 mg for each) were dissolved in 600 µl DMSO-*d*$_6$ with sonication. $^1$H NMR spectra of the solutions were collected on a Bruker NEO-500 NMR spectrometer.

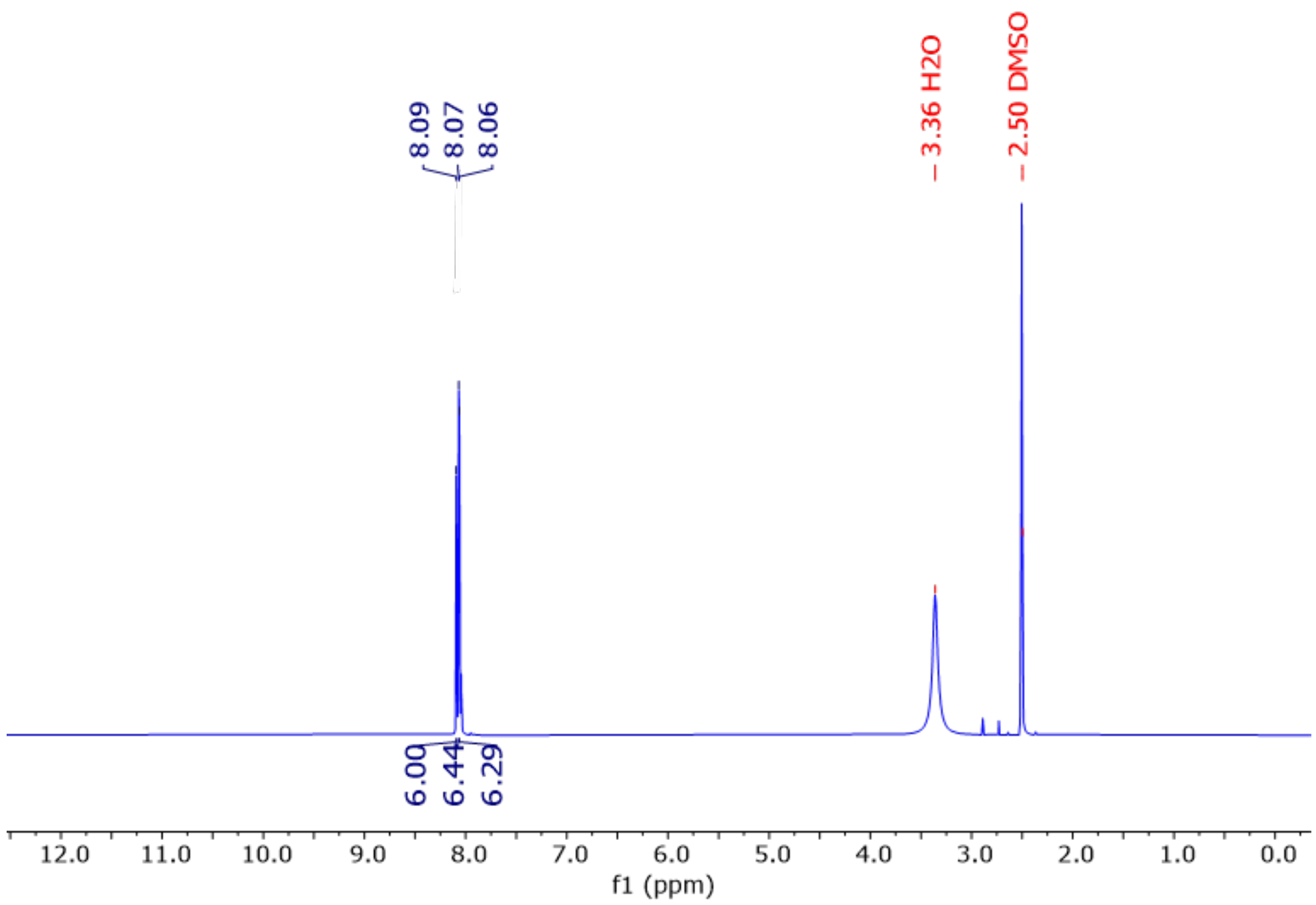

**Figure S7.** $^1$H-NMR spectrum of $H_3$BTB-H linker dissolved in DMSO-*d*$_6$.

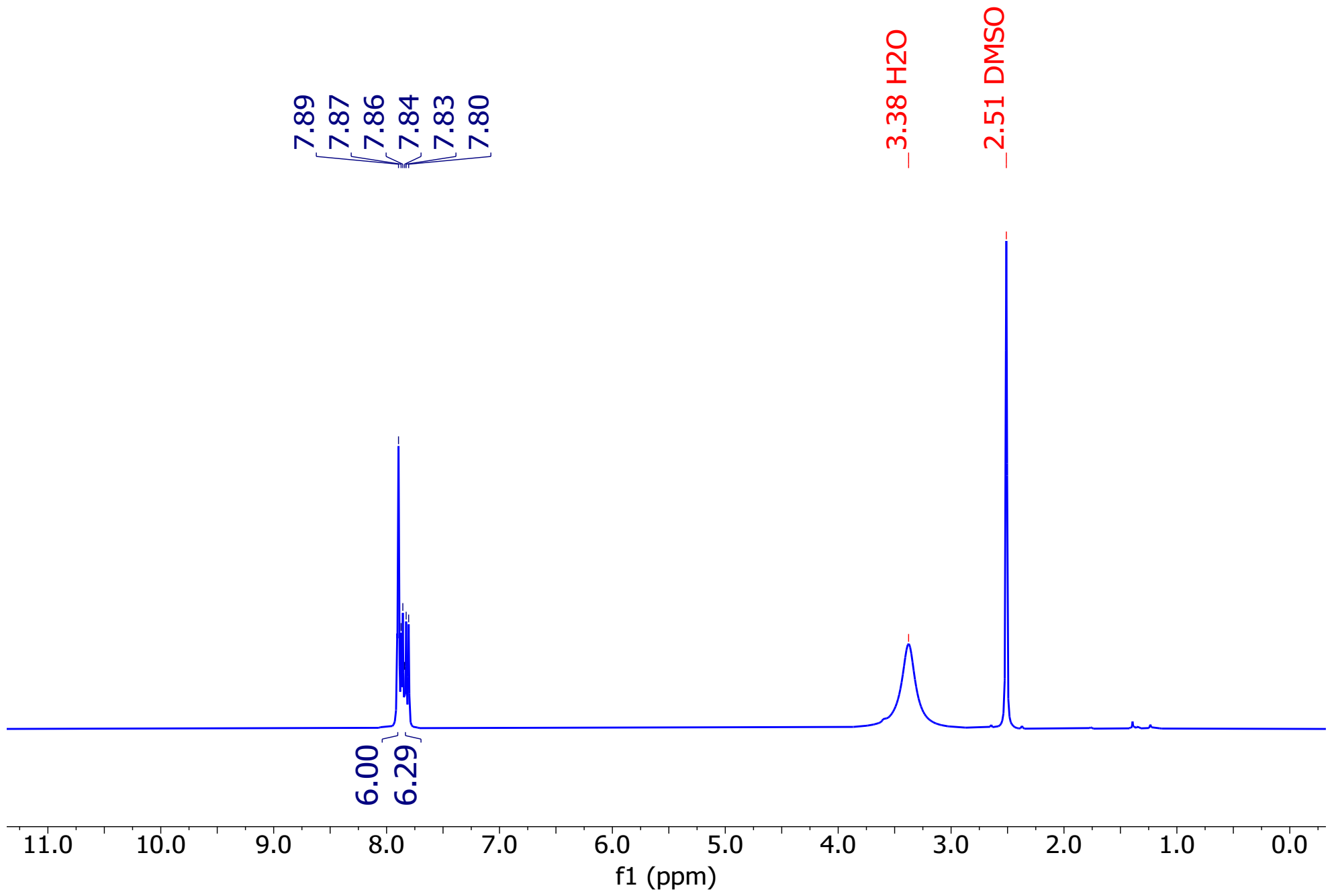

**Figure S8.** $^1$H-NMR spectrum of $H_3$BTB-*o*F linker dissolved in DMSO-*d*$_6$.

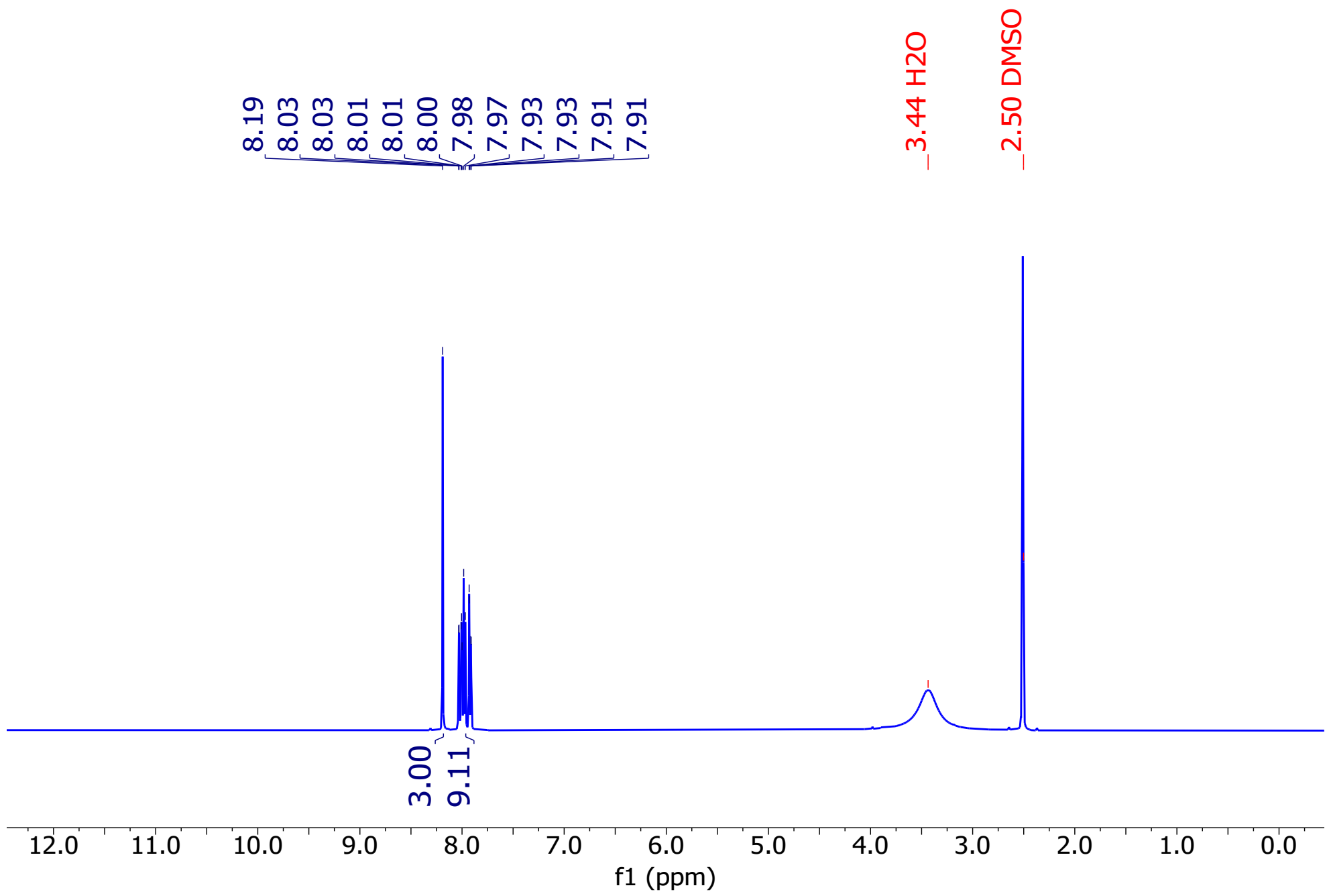

**Figure S9.** [1]H-NMR spectrum of $H_3$BTB-*m*F linker dissolved in DMSO-$d_6$.

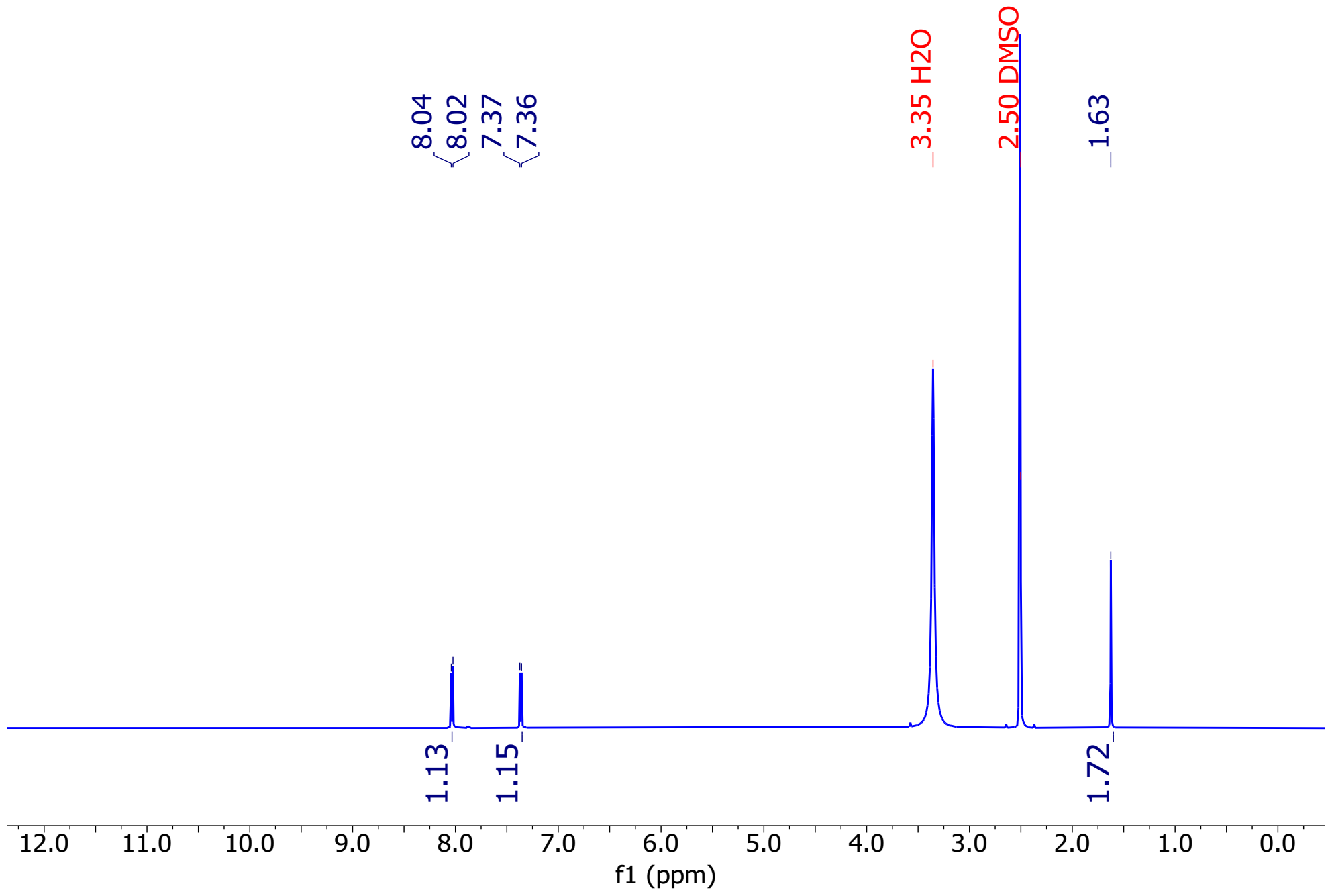

**Figure S10.** [1]H-NMR spectrum of $H_3$BTB-$CH_3$ linker dissolved in DMSO-$d_6$.

As-synthesized MOF-521-X samples were fully exchanged with anhydrous DMF and followed with anhydrous acetone. The samples were then activated by evacuation to remove the unreacted linker and guest molecules. The activated samples (ca. 10 mg for each) were dissolved in 600 µl 5% NaOD solution in $D_2O$ with sonication. $^1H$ NMR spectra of the digested solutions were collected on a Bruker AVB-400 NMR spectrometer.

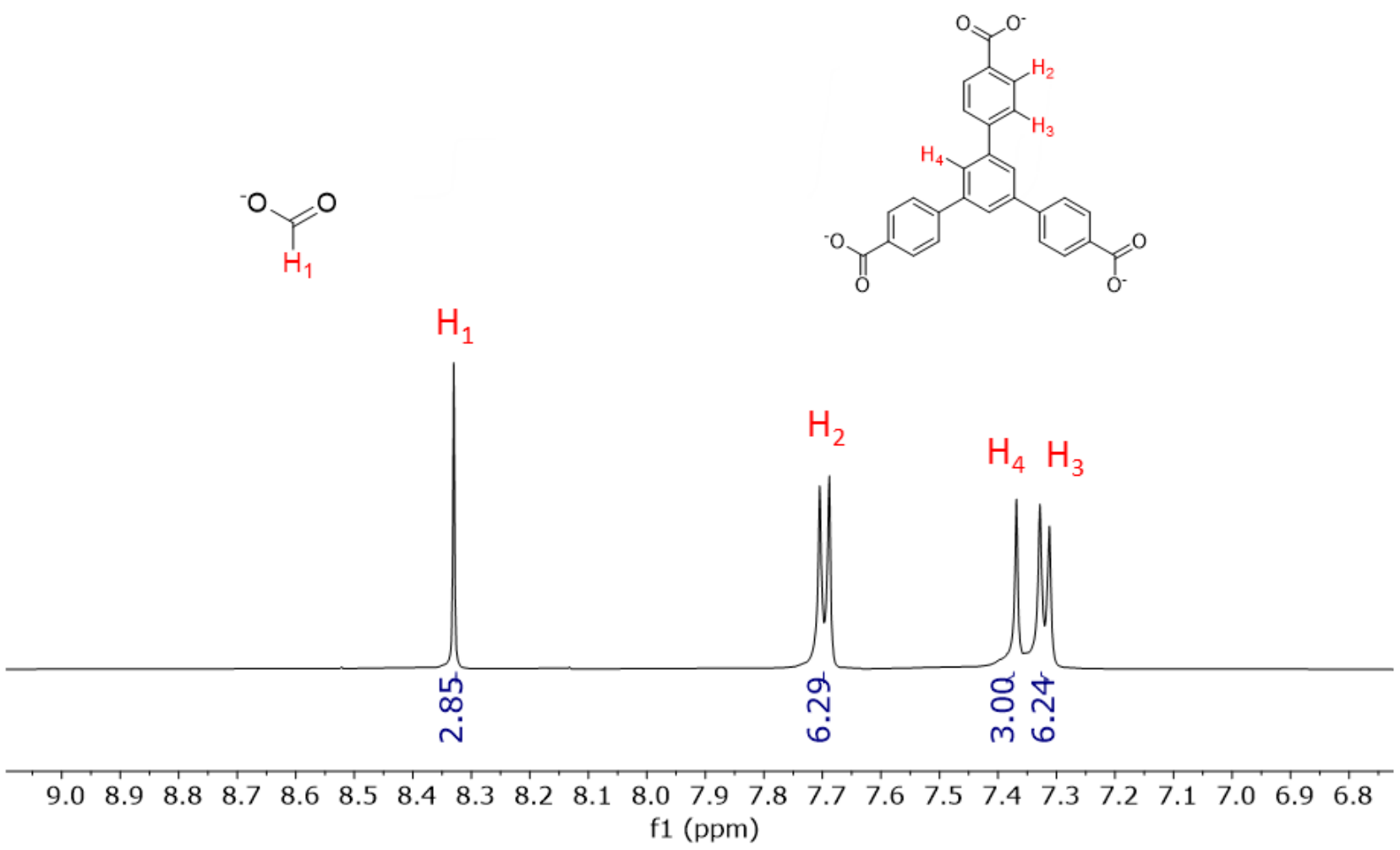

**Figure S11.** $^1H$-NMR spectrum of MOF-521-H after being thoroughly washed and digested in 5% NaOD in $D_2O$.

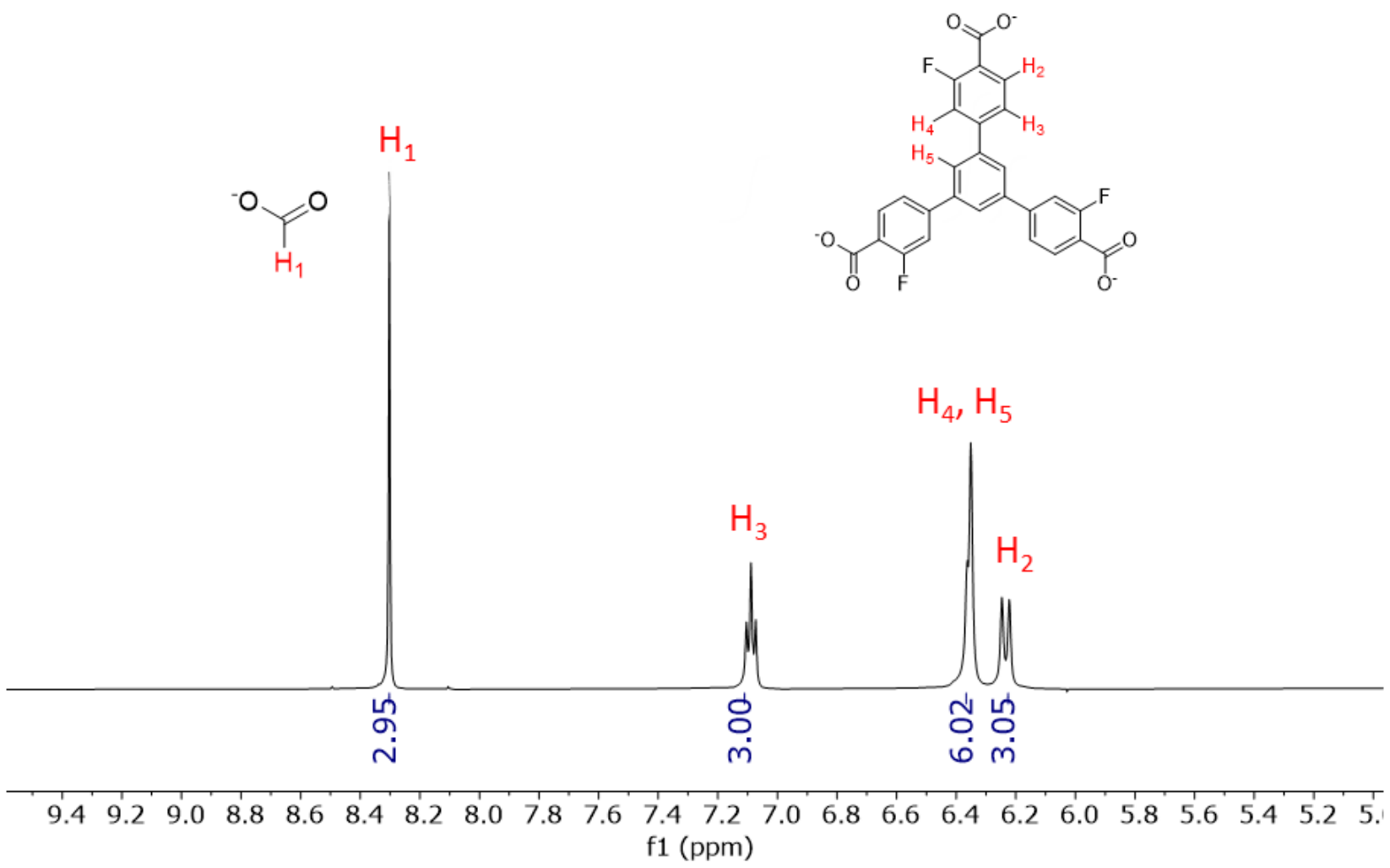

**Figure S12.** $^1H$-NMR spectrum of MOF-521-*o*F after being thoroughly washed and digested in 5% NaOD in $D_2O$.

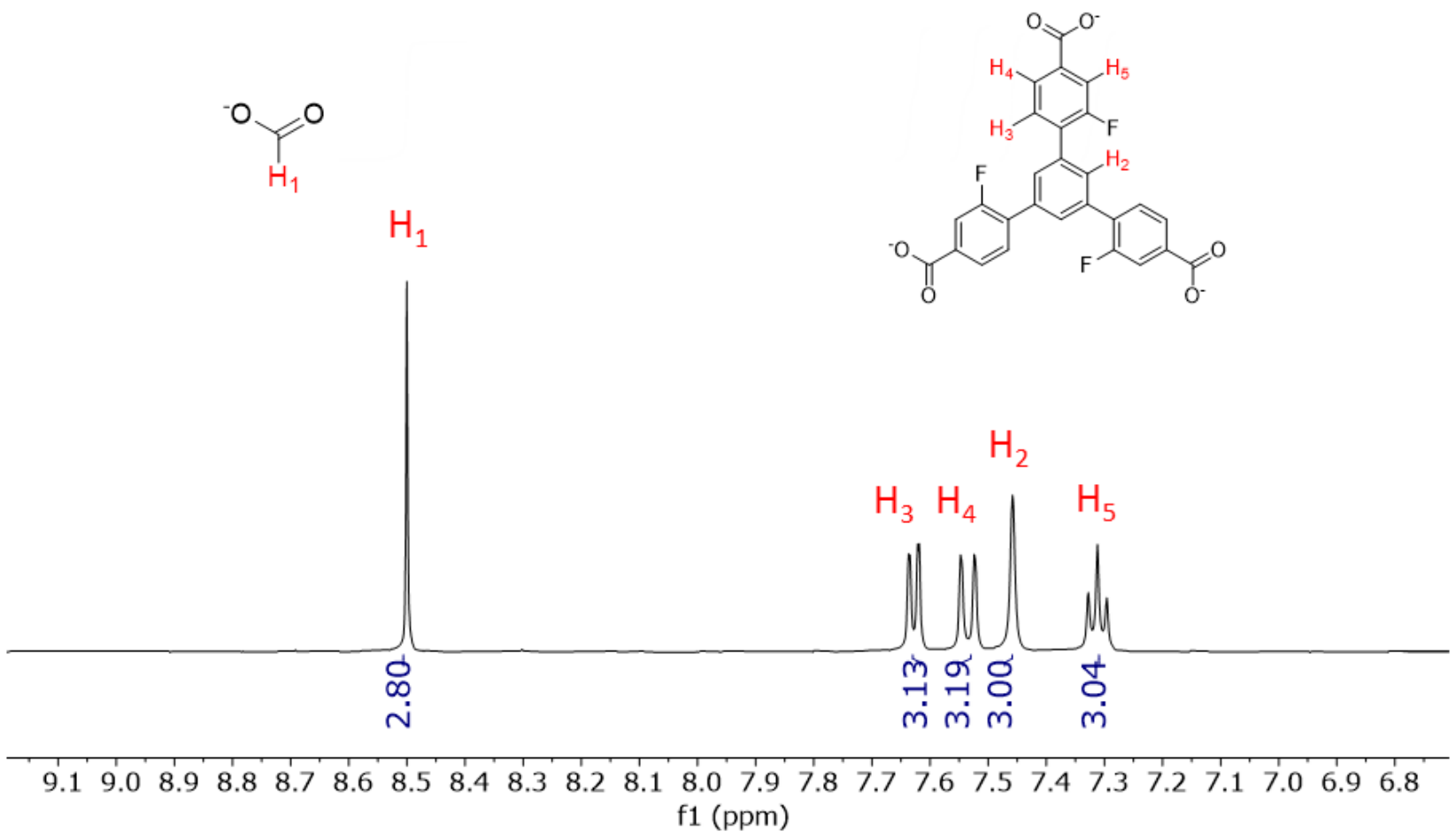

**Figure S13.** [1]H-NMR spectrum of MOF-521-*m*F after being thoroughly washed and digested in 5% NaOD in $D_2O$.

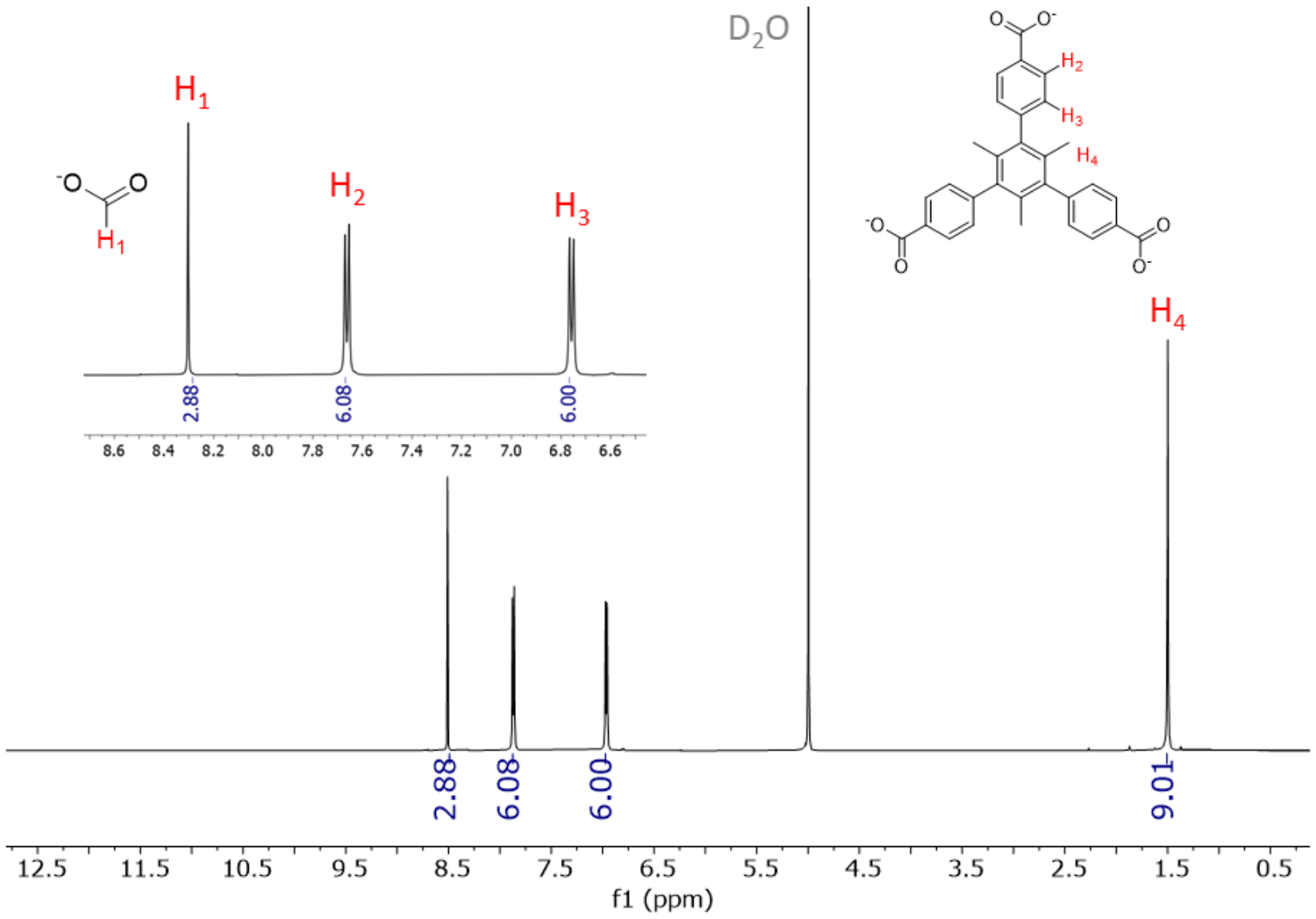

**Figure S14.** [1]H-NMR spectrum of MOF-521-$CH_3$ after being thoroughly washed and digested in 5% NaOD in $D_2O$.

## Section S4. Thermogravimetric Analysis

Thermal gravimetric analysis (TGA) curves were recorded on a TA Q500 thermal analysis system under nitrogen flow, ramping at 5 ˚C per min from room temperature to 800 ˚C. The thermal stability of all MOF-521 compounds was confirmed by no significant weight loss prior to 260 °C.

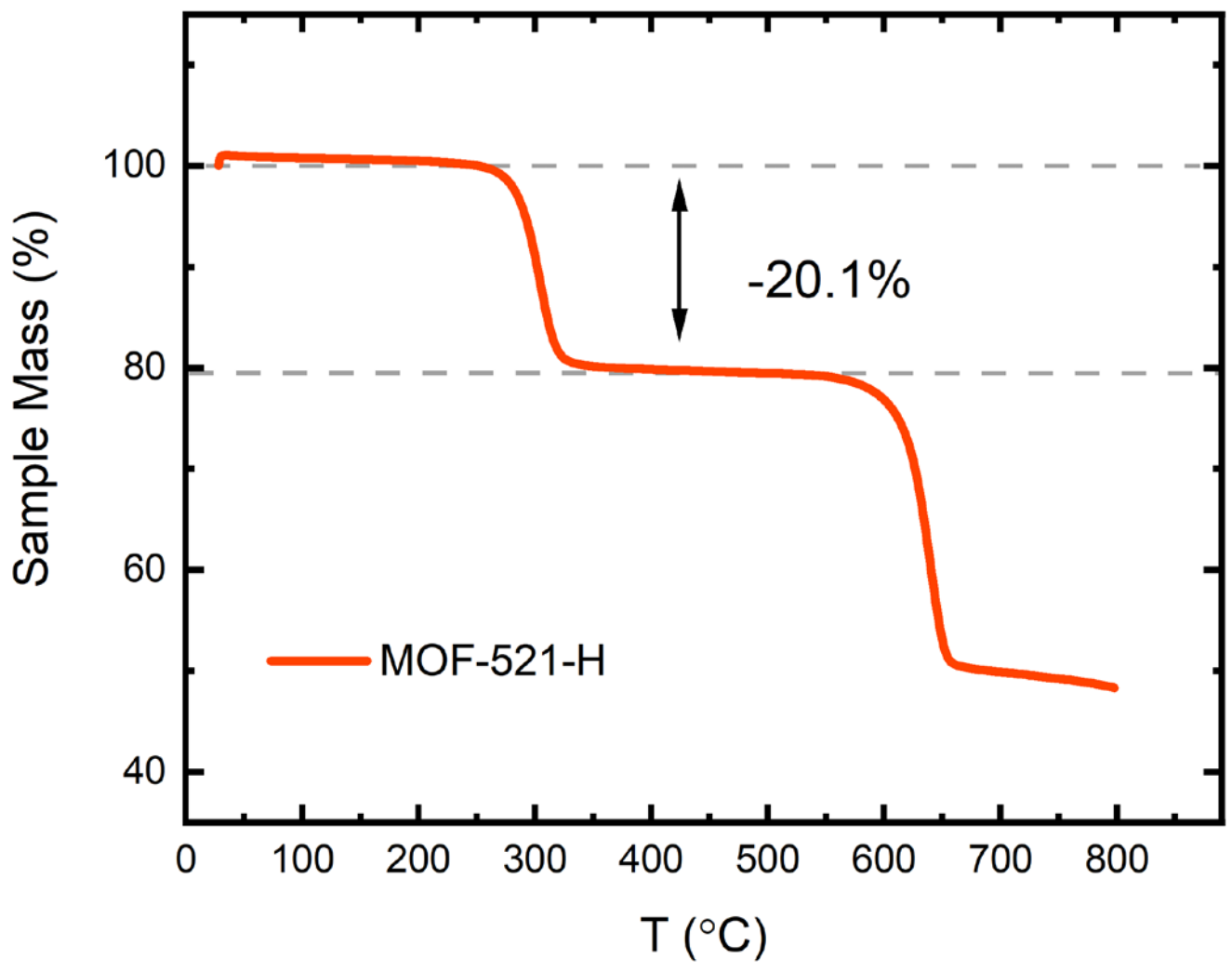

**Figure S15.** TGA trace of activated MOF-521-H under nitrogen flow.

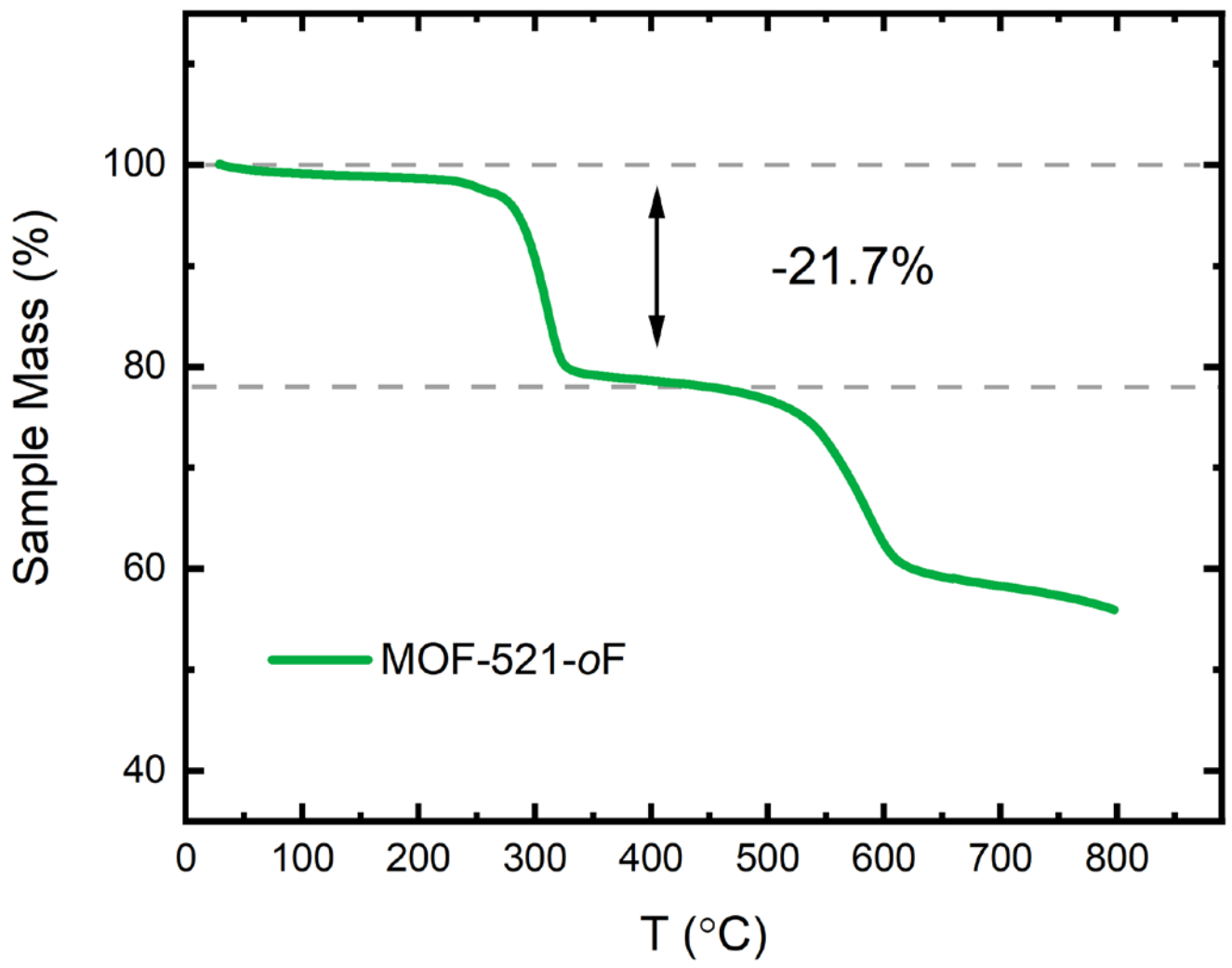

**Figure S16.** TGA trace of activated MOF-521-*o*F under nitrogen flow.

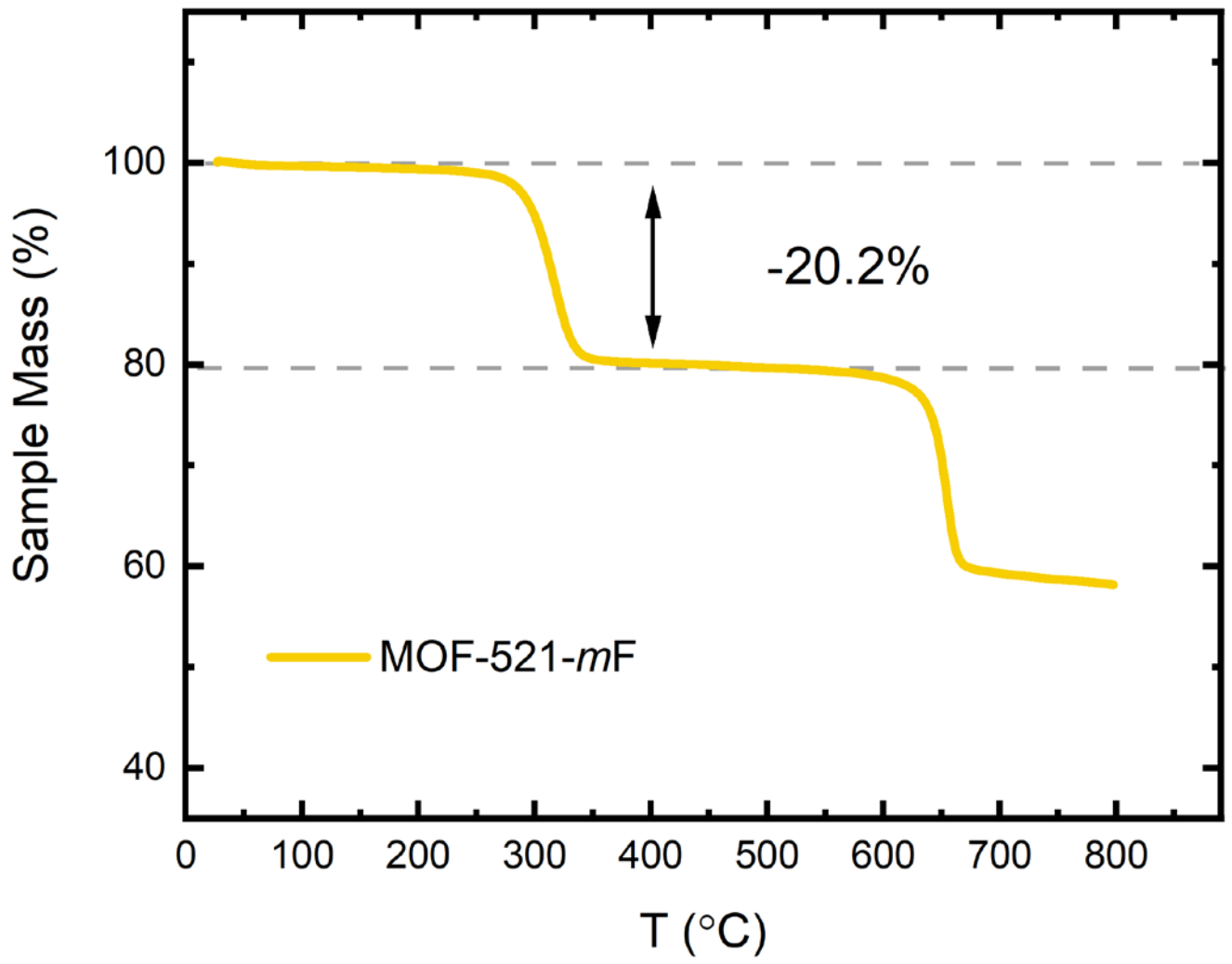

**Figure S17.** TGA trace of activated MOF-521-*m*F under nitrogen flow.

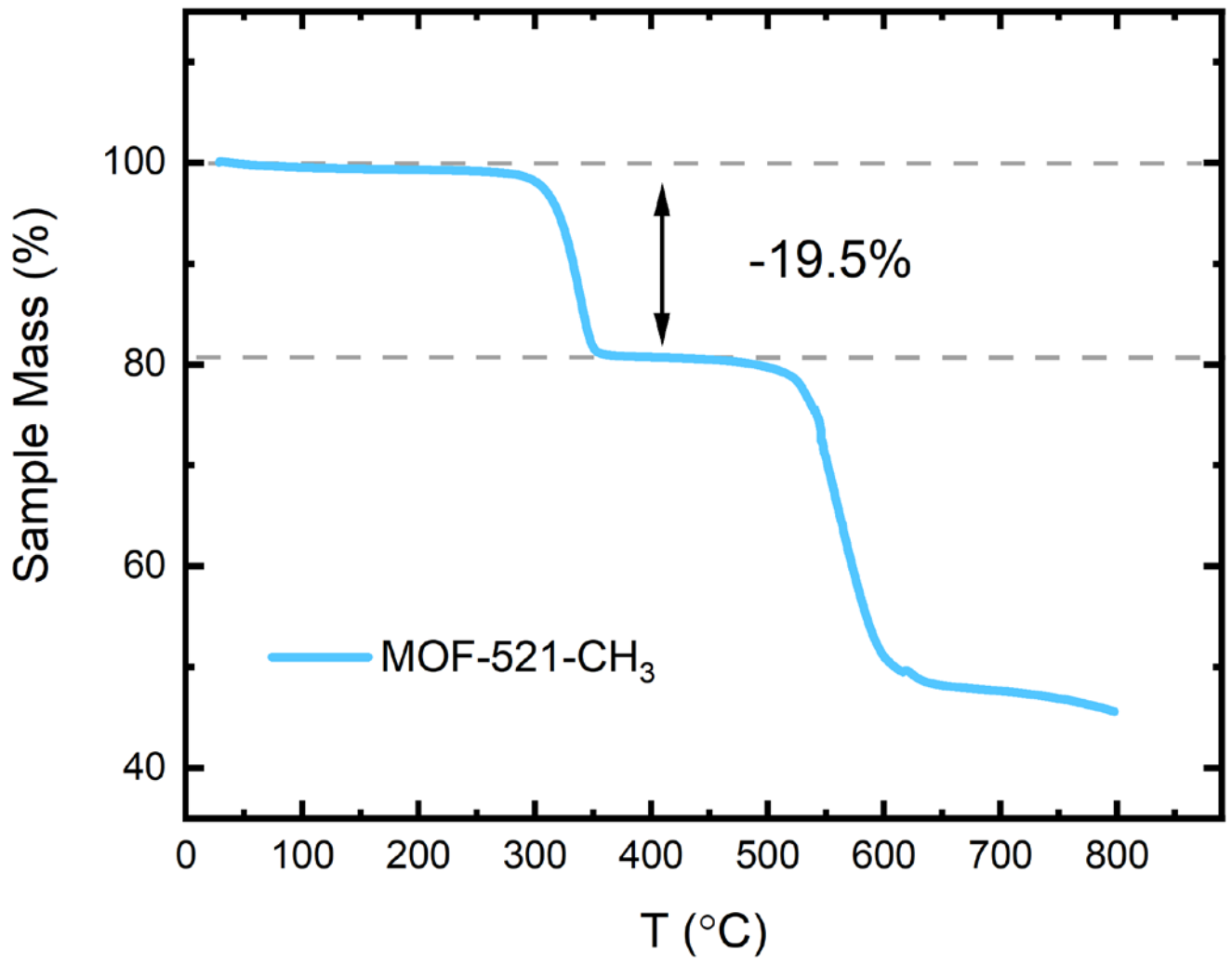

**Figure S18.** TGA trace of activated MOF-521-$CH_3$ under nitrogen flow.

## Section S5. Additional Powder X-Ray Diffraction Measurements

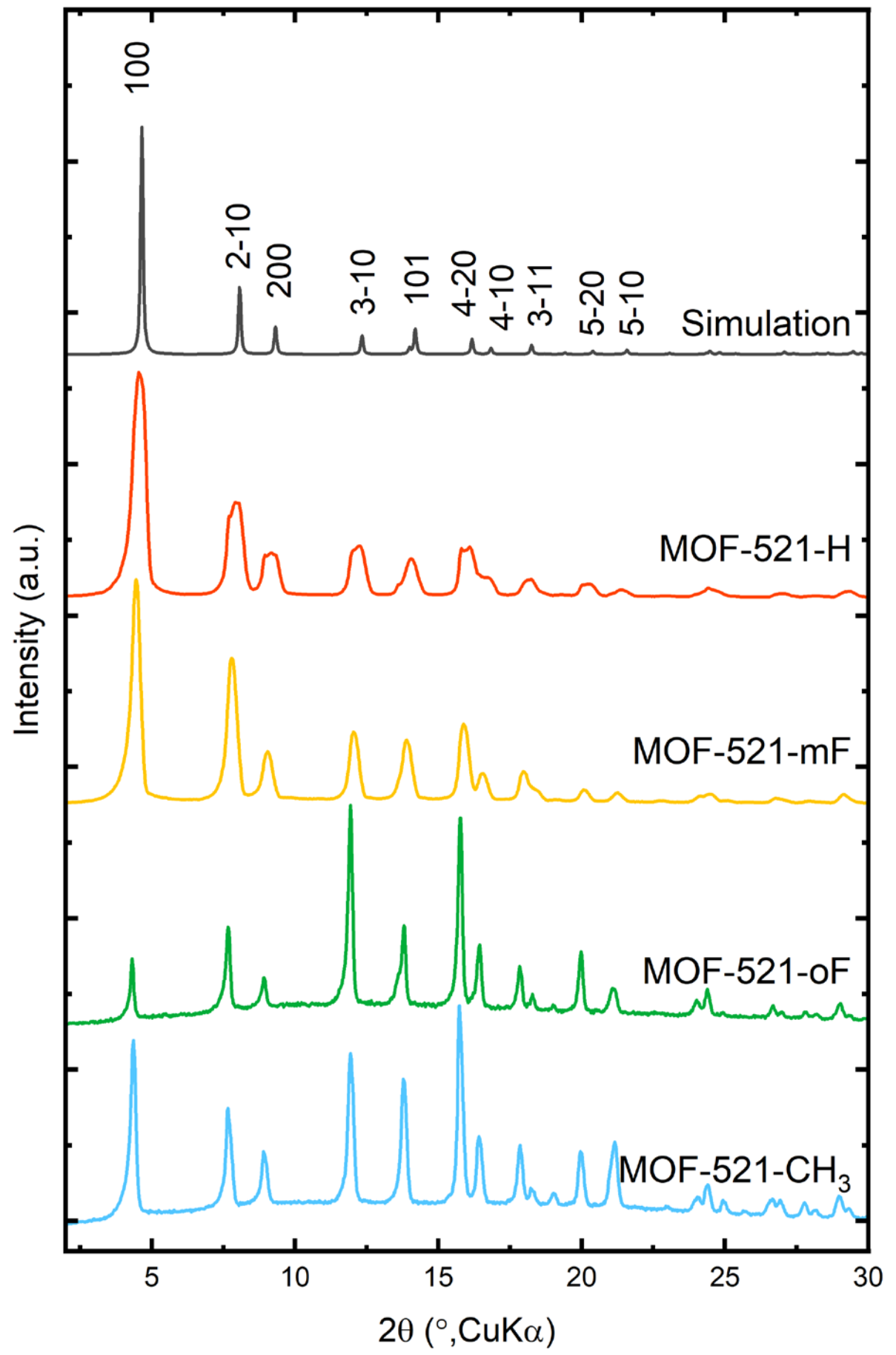

**Figure S19.** PXRD patterns of MOF-521 compounds. The simulated pattern at the top was generated using single crystal structure of MOF-521-H.

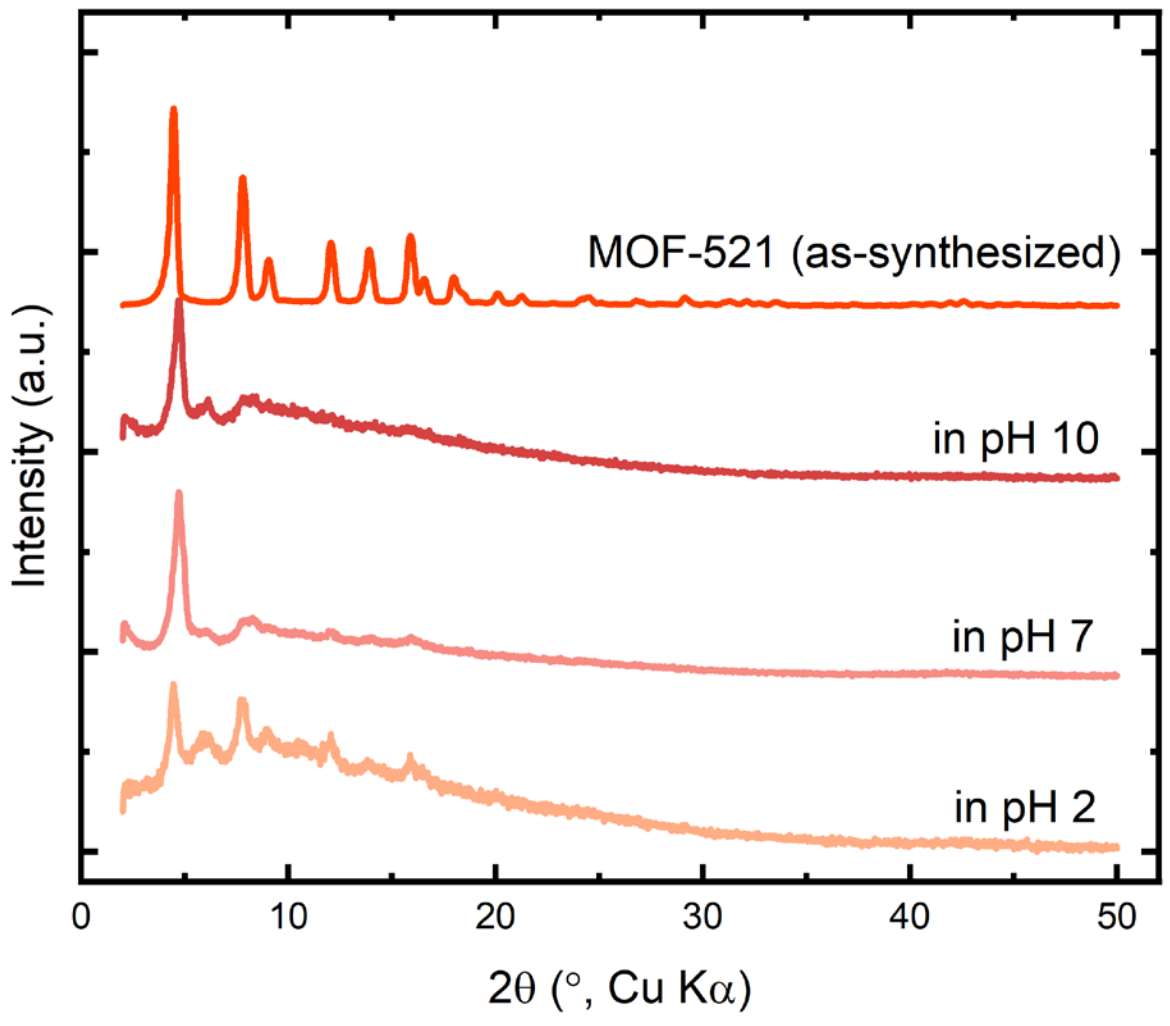

**Figure S20.** PXRD patterns of MOF-521-H soaked in pH 2, 7 and 10 solutions at room temperature.

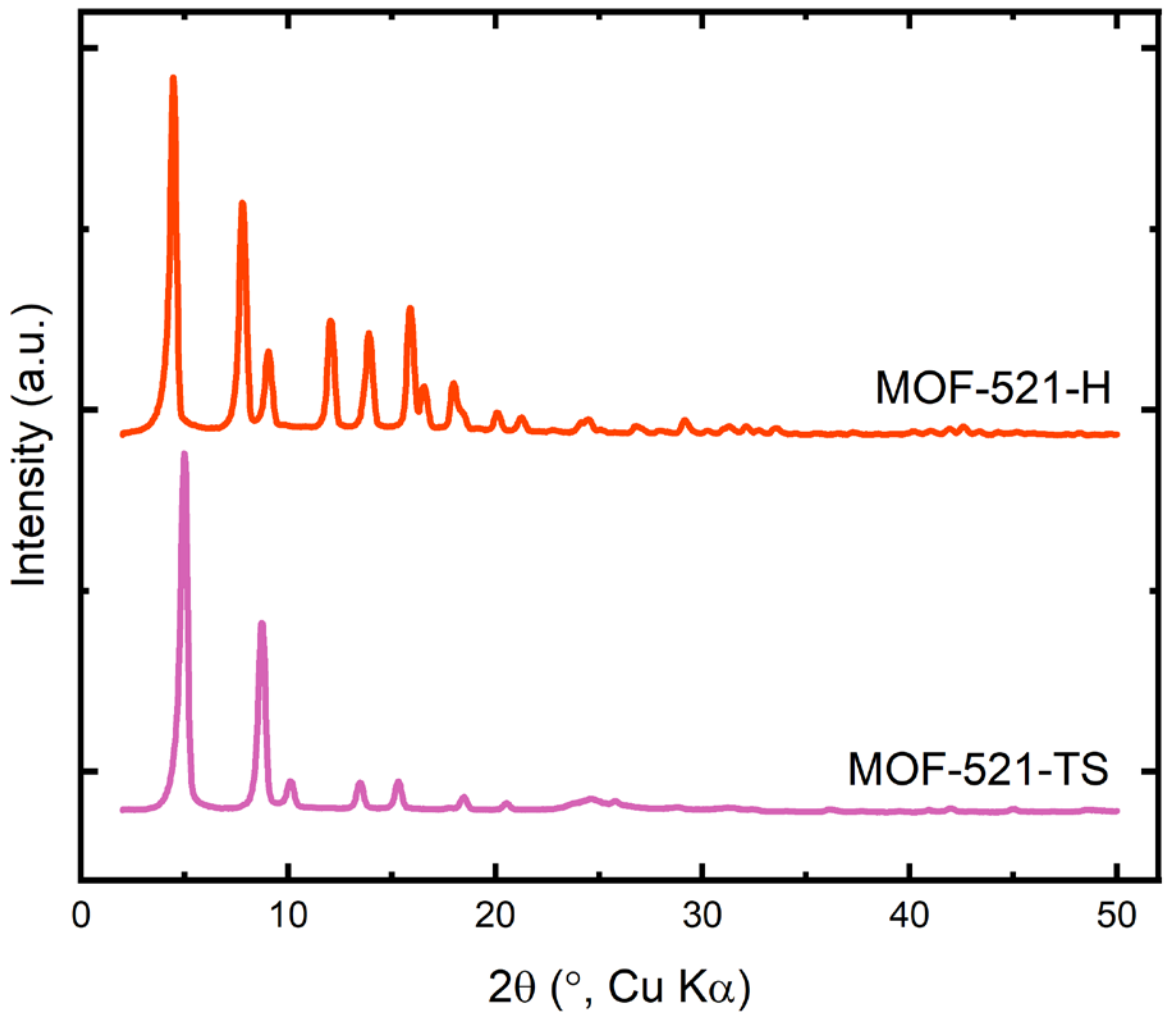

**Figure S21.** PXRD patterns of as-synthesized MOF-521-H and MOF-521-H sample calcined at 450 °C, denoted as MOF-521-TS, under nitrogen atmosphere in the TGA furnace.

## Section S6. Nitrogen Sorption Isotherm Analysis

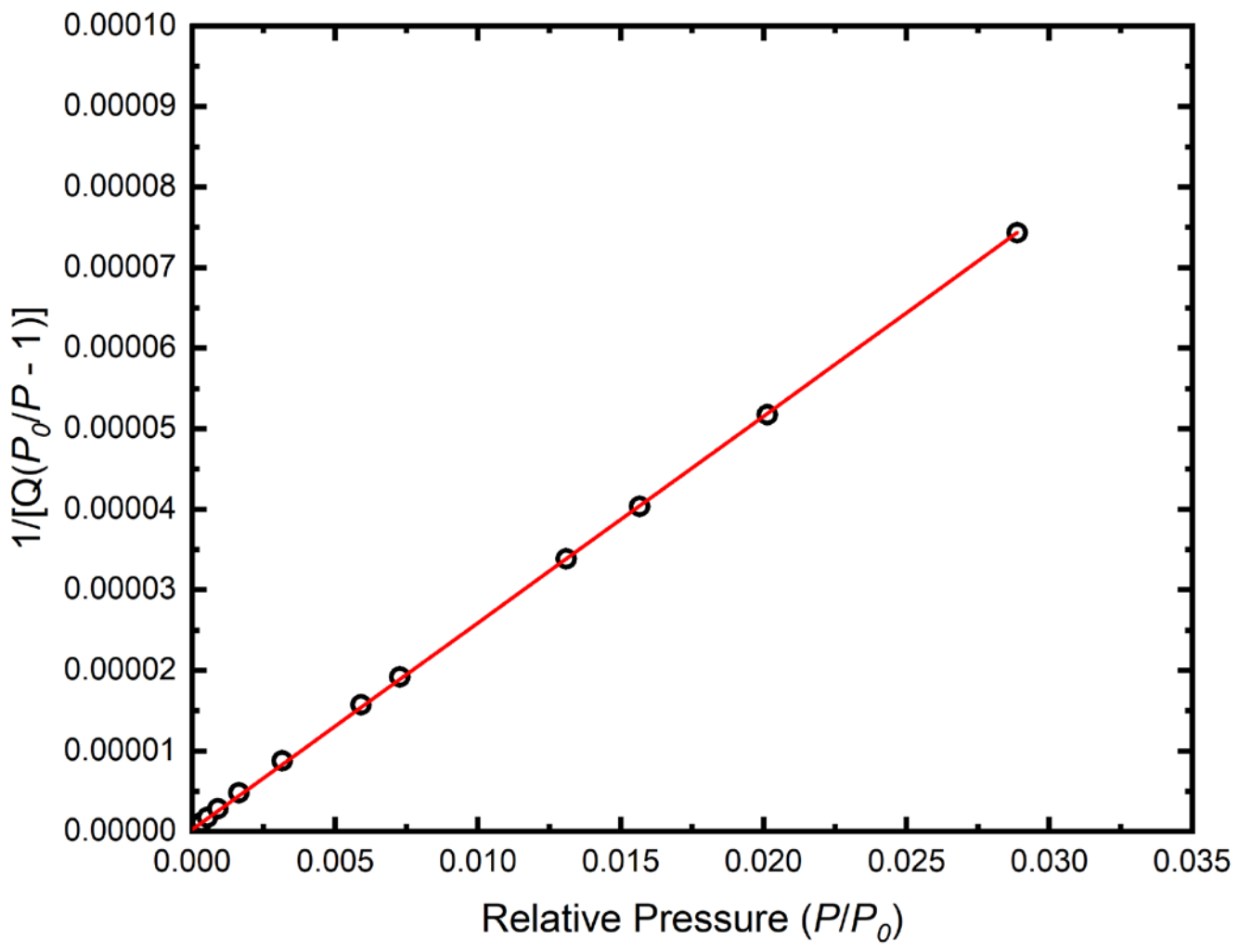

**Figure S22.** BET plot of MOF-521-H derived from $N_2$ sorption isotherm at 77 K. The black empty circles represent the experimental data, while the red line corresponds to the fitted linear regression line. $S_{BET}$ = 1696 m² g$^{-1}$. Correlation coefficient, r = 0.9999.

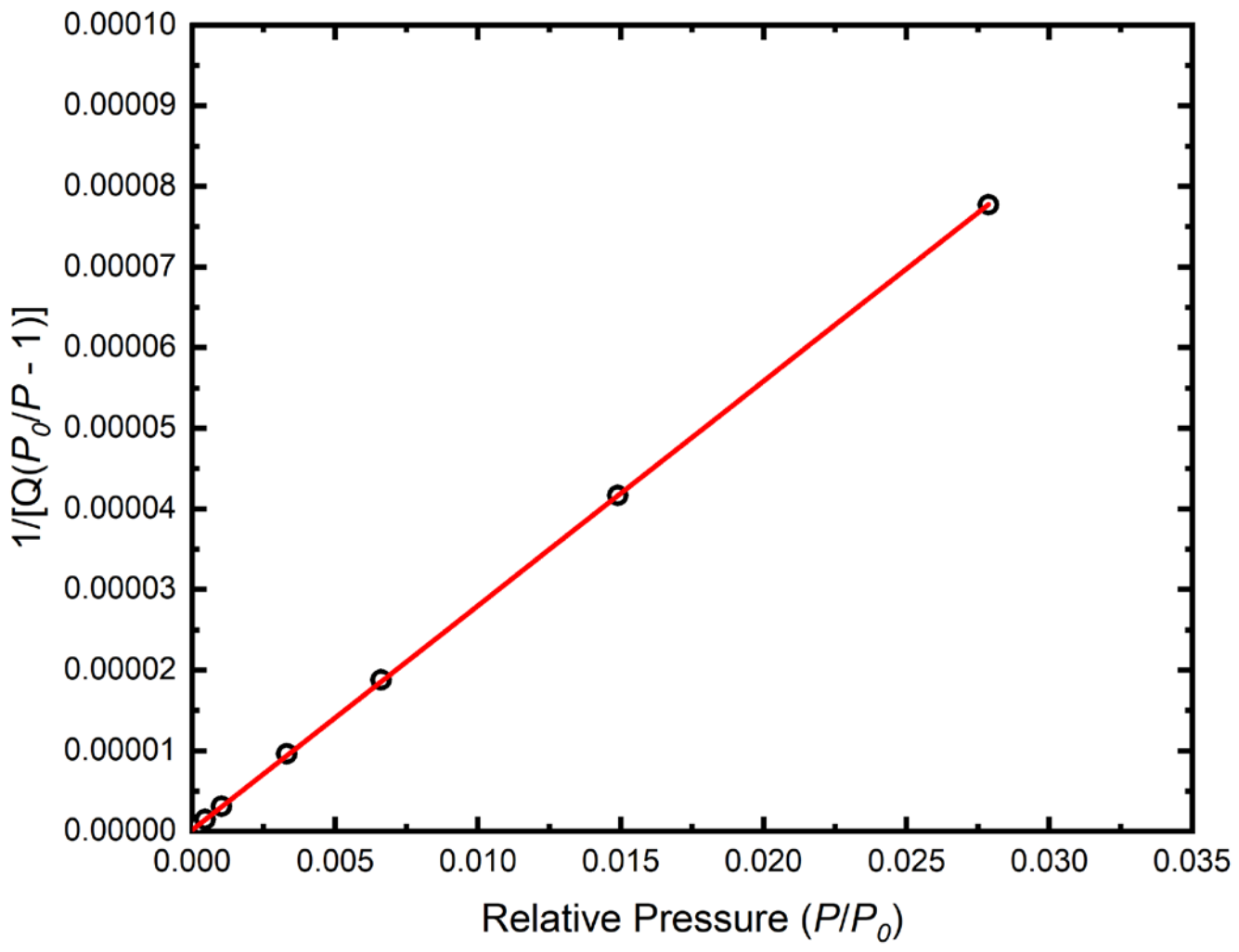

**Figure S23.** BET plot of MOF-521-*o*F derived from $N_2$ sorption isotherm at 77 K. The black empty circles represent the experimental data, while the red line corresponds to the fitted linear regression line. $S_{BET}$ = 1562 m² g$^{-1}$. Correlation coefficient, r = 0.9999.

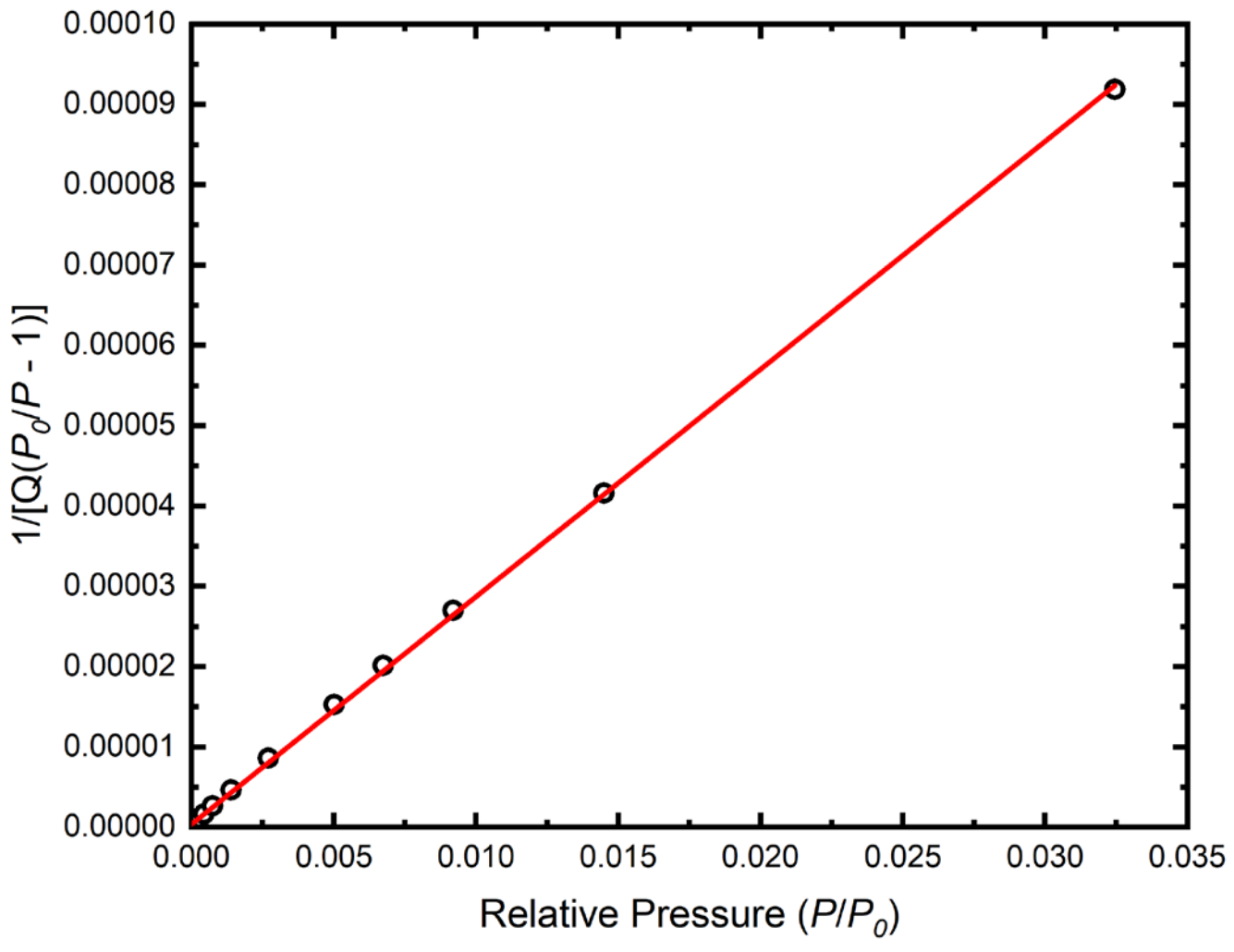

**Figure S24.** BET plot of MOF-521-*m*F derived from $N_2$ sorption isotherm at 77 K. The black empty circles represent the experimental data, while the red line corresponds to the fitted linear regression line. $S_{BET}$ = 1535 m² g$^{-1}$. Correlation coefficient, r = 0.9998.

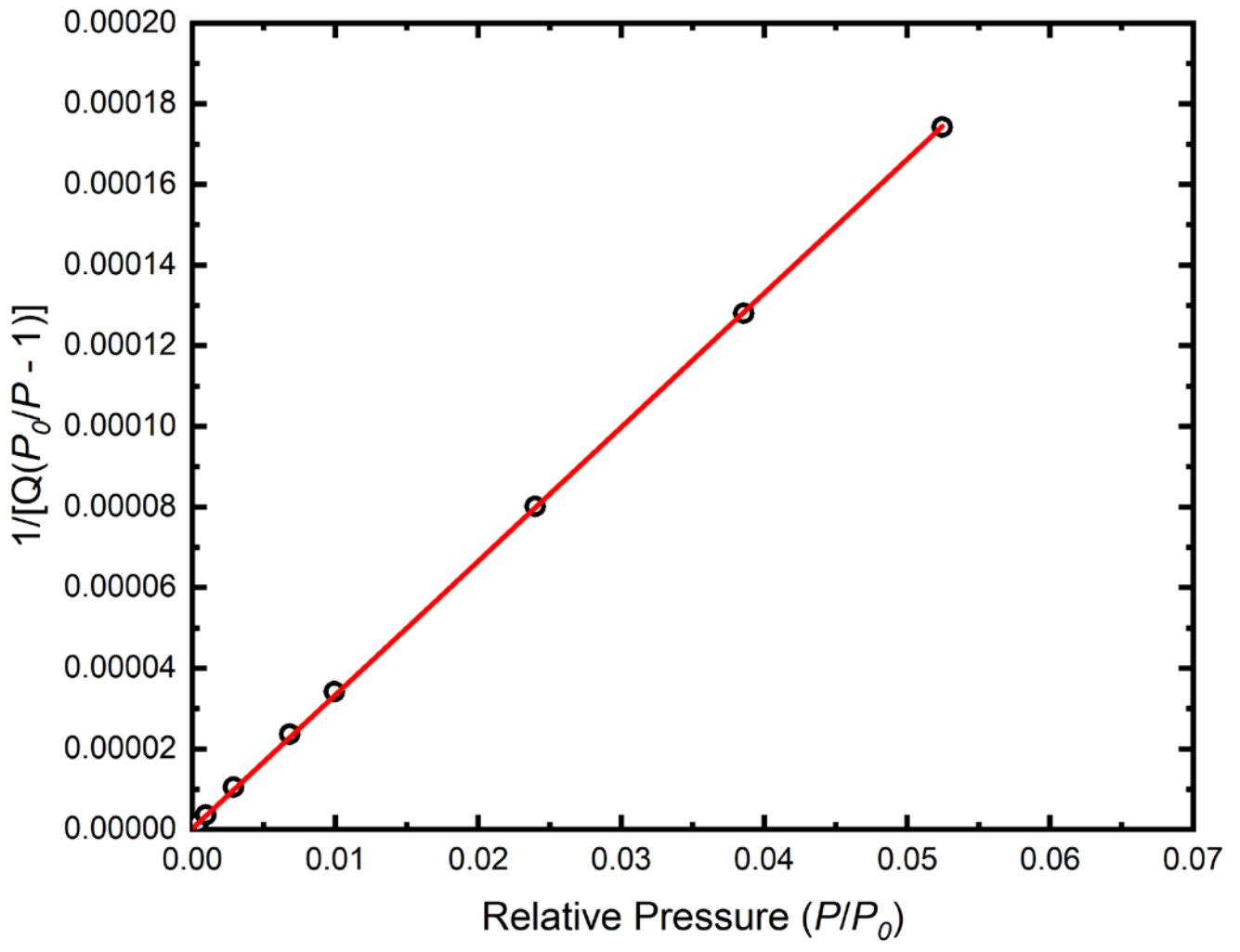

**Figure S25.** BET plot of MOF-521-$CH_3$ derived from $N_2$ sorption isotherm at 77 K. The black empty circles represent the experimental data, while the red line corresponds to the fitted linear regression line. $S_{BET}$ = 1311 m² g$^{-1}$. Correlation coefficient, r = 0.9999.

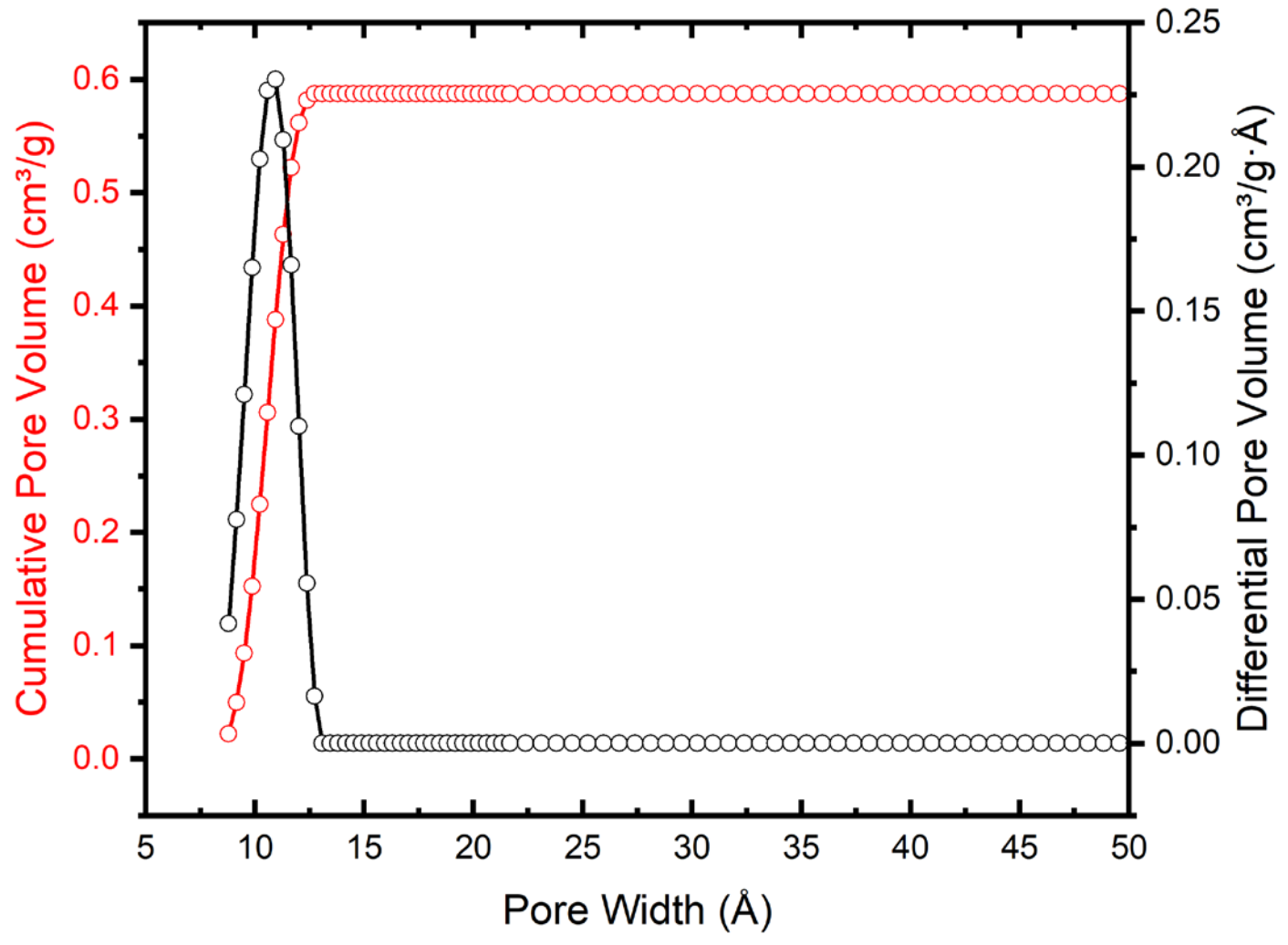

**Figure S26.** Pore size distribution of MOF-521-H derived from $N_2$ sorption isotherm measured at 77 K, the adsorption branch fitted with a quenched solid-state density functional theory (QSDFT) model of $N_2$ on carbon at 77 K (slit/cylindrical pores). Pore width = 10.8 Å.

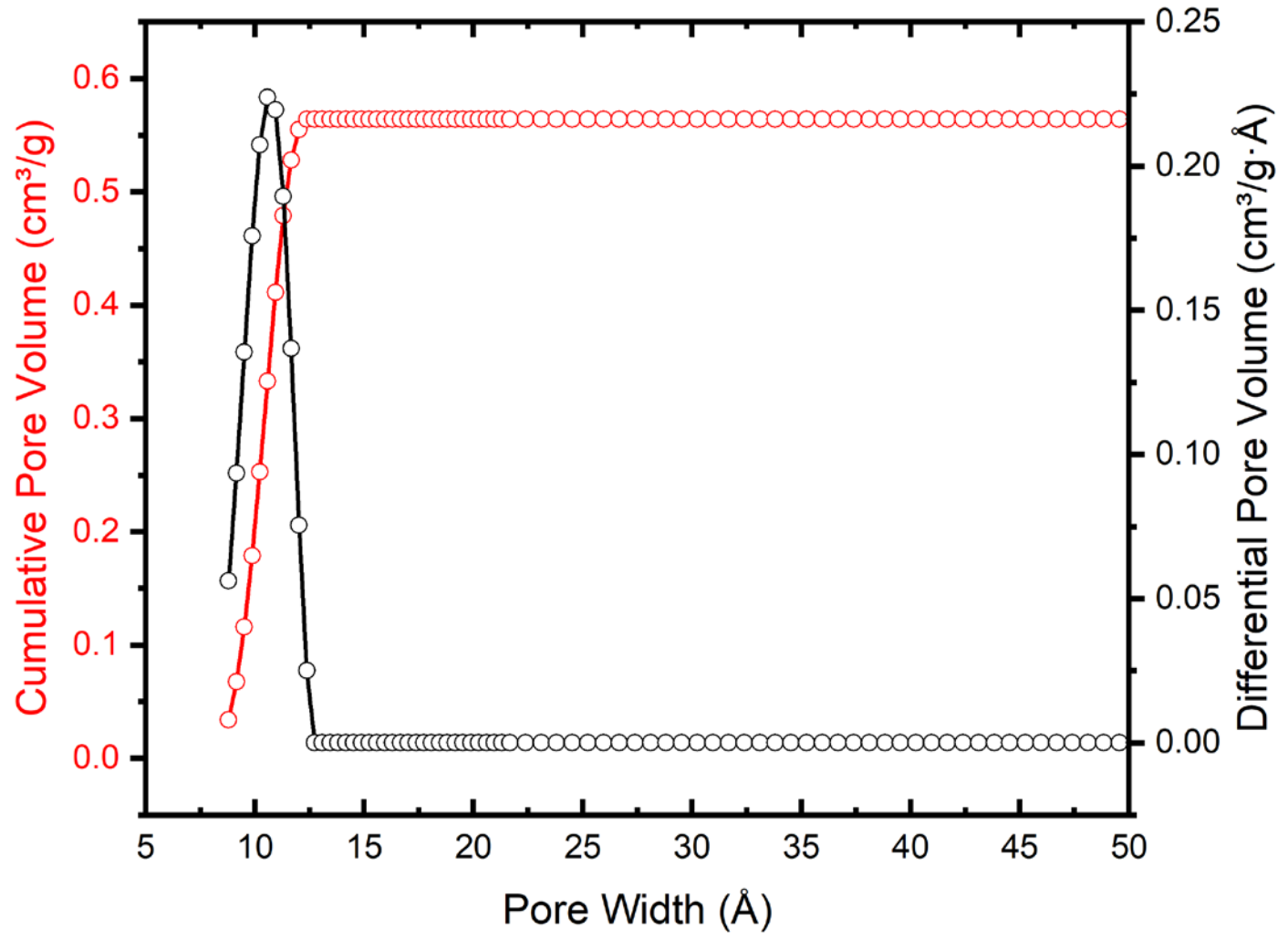

**Figure S27.** Pore size distribution of MOF-521-*o*F derived from $N_2$ sorption isotherm measured at 77 K, the adsorption branch fitted with a quenched solid-state density functional theory (QSDFT) model of $N_2$ on carbon at 77 K (slit/cylindrical pores). Pore width = 10.7 Å.

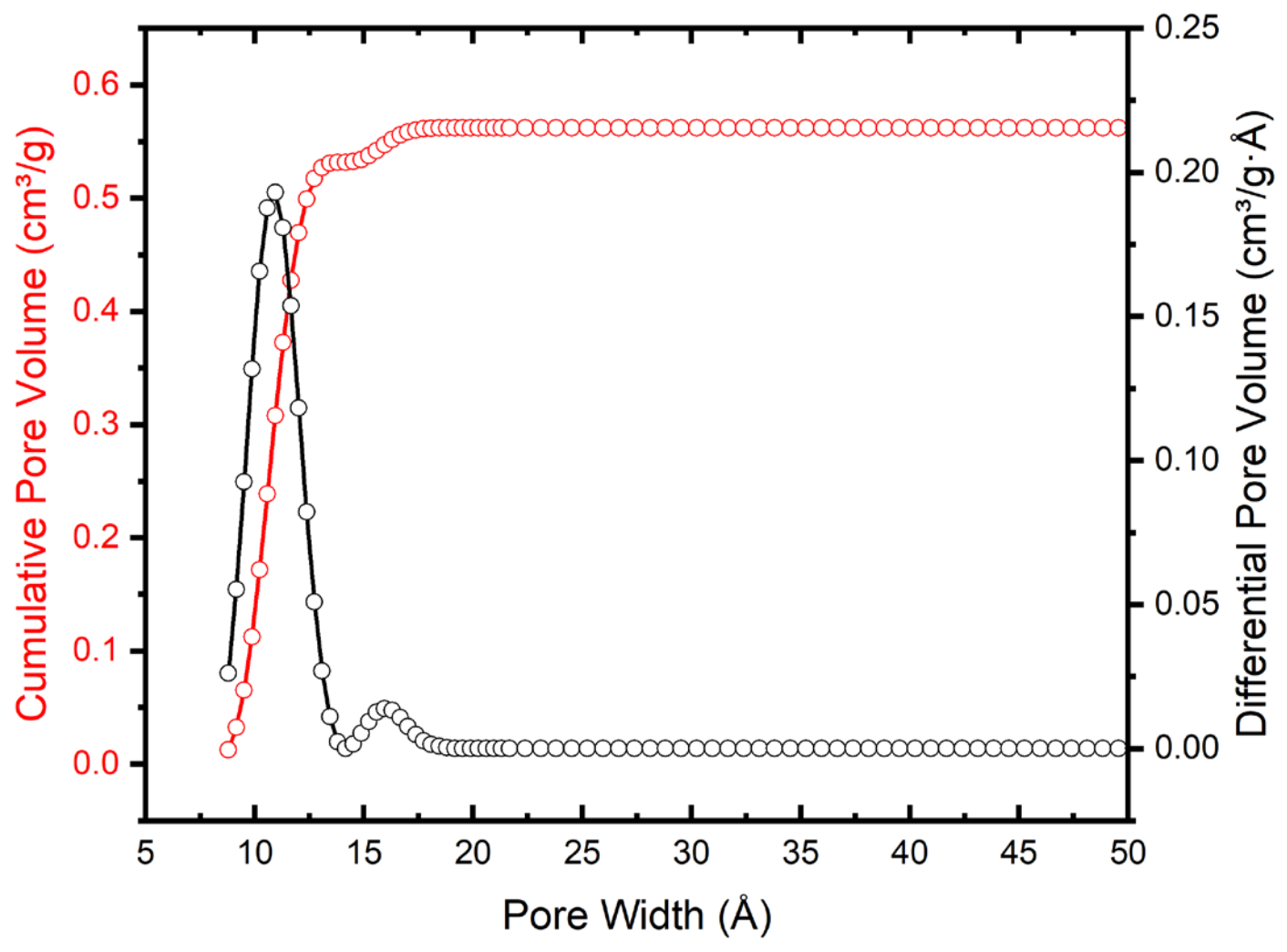

**Figure S28.** Pore size distribution of MOF-521-*m*F derived from $N_2$ sorption isotherm measured at 77 K, the adsorption branch fitted with a quenched solid-state density functional theory (QSDFT) model of $N_2$ on carbon at 77 K (slit/cylindrical pores). Pore width = 10.9 Å.

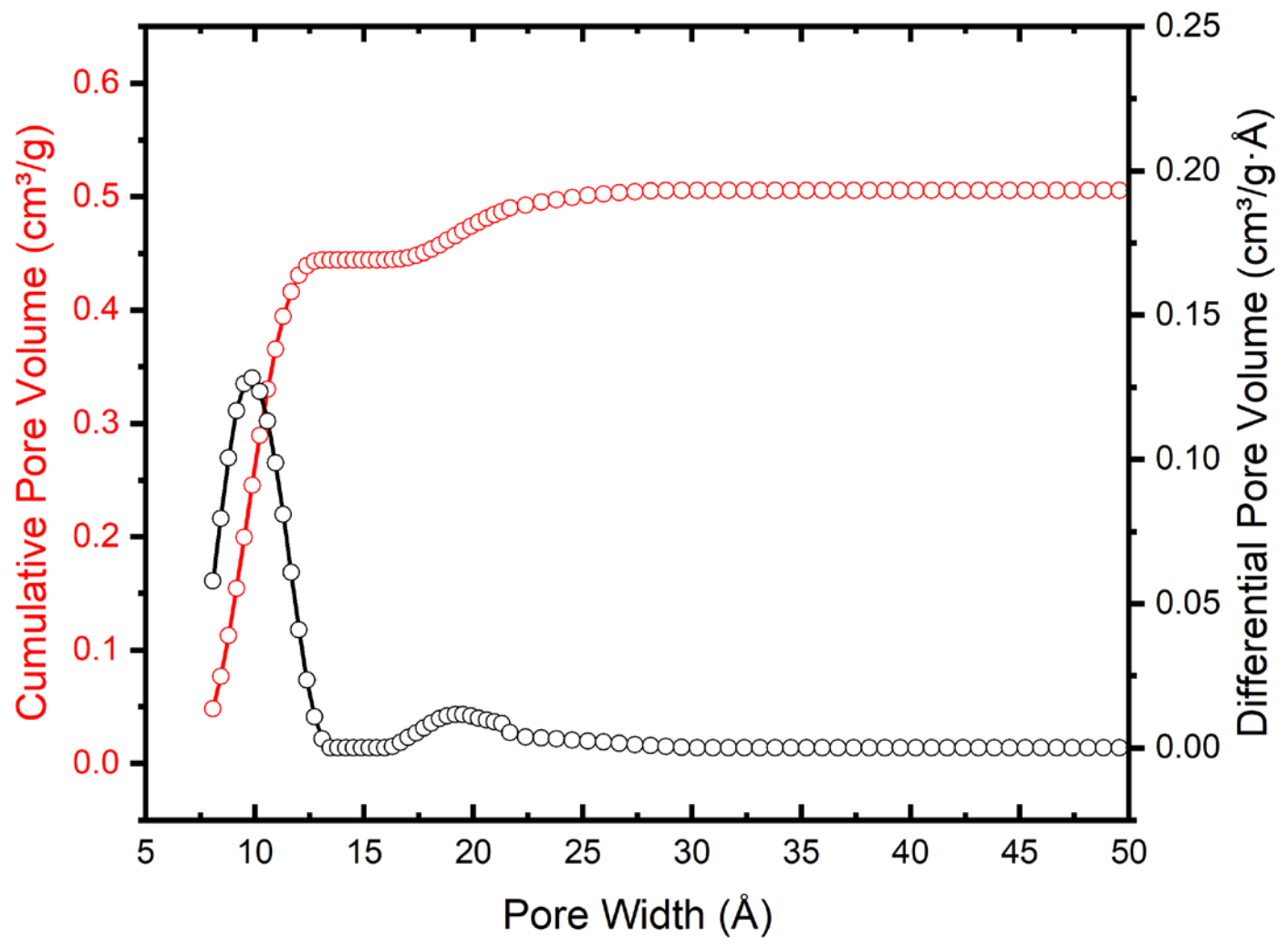

**Figure S29.** Pore size distribution of MOF-521-$CH_3$ derived from $N_2$ sorption isotherm measured at 77 K, the adsorption branch fitted with a quenched solid-state density functional theory (QSDFT) model of $N_2$ on carbon at 77 K (slit/cylindrical pores). Pore width = 9.3 Å.

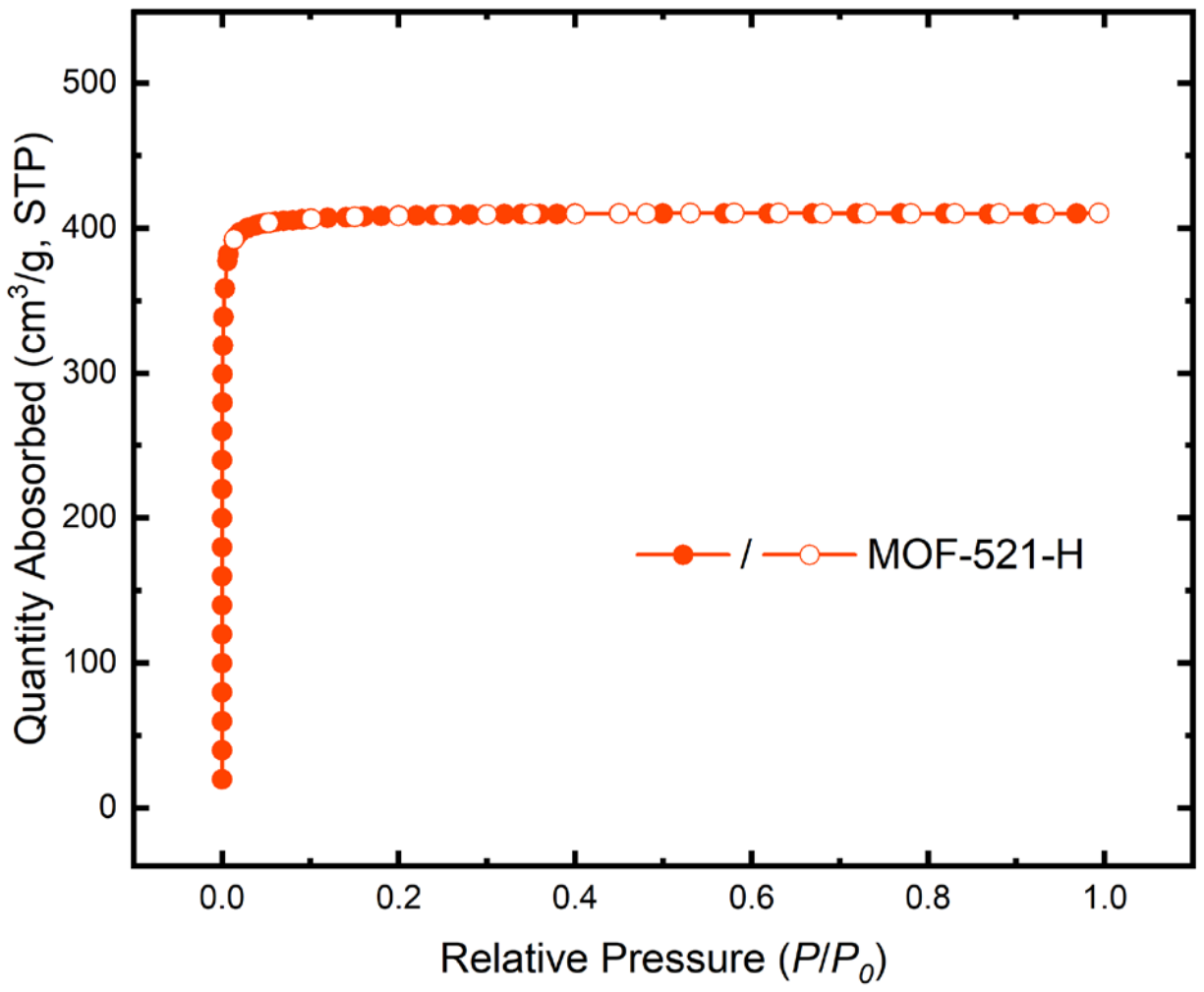

**Figure S30.** $N_2$ sorption isotherm of MOF-521-H at 77 K. $S_{BET}$ = 1696 m² g$^{-1}$. Correlation coefficient r = 0.9999. Symbols of filled cycles denote the adsorption branch, while empty cycles denote the desorption branch.

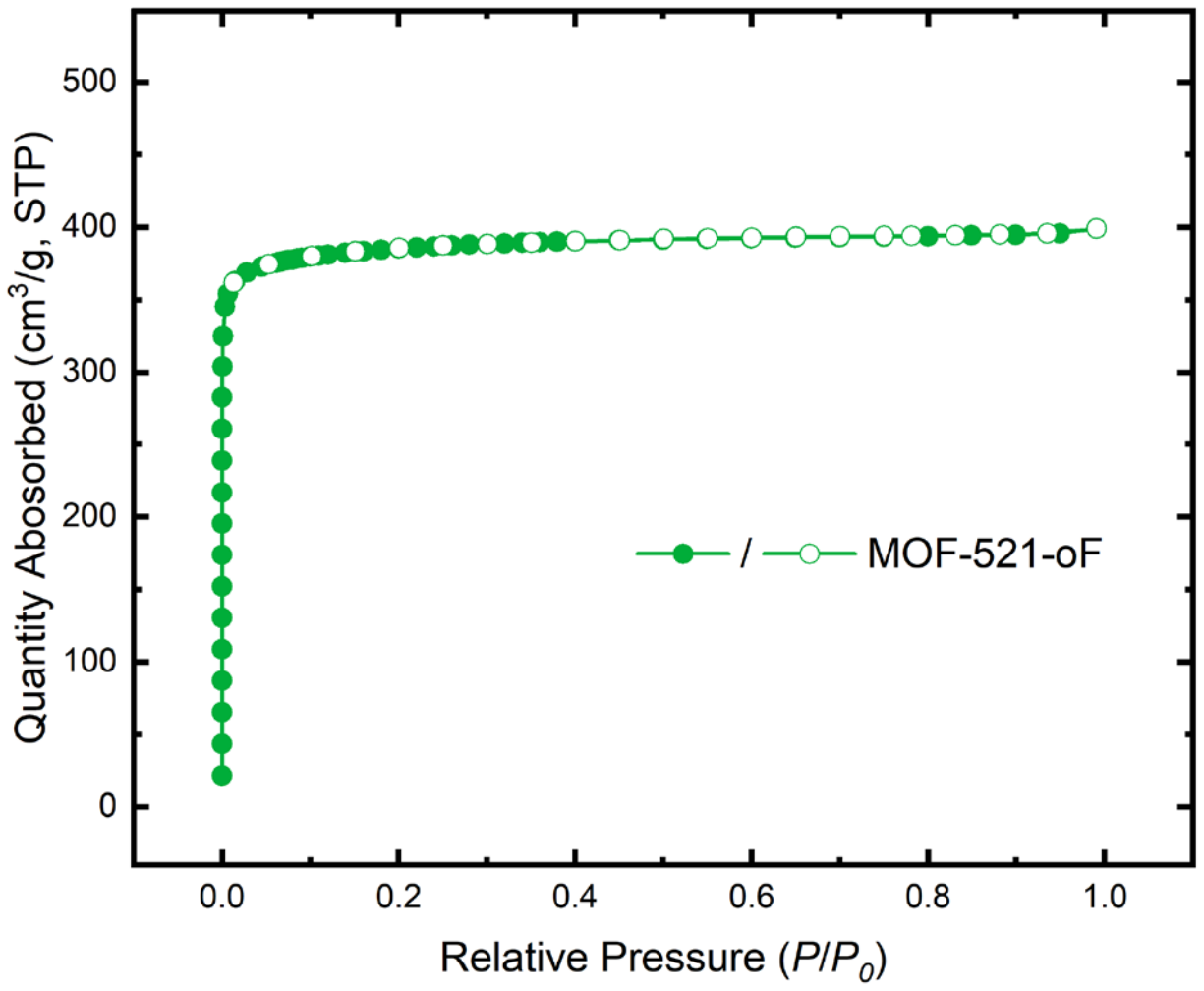

**Figure S31.** $N_2$ sorption isotherm of MOF-521-*o*F at 77 K. $S_{BET}$ = 1562 m² g$^{-1}$. Correlation coefficient r = 0.9999. Symbols of filled cycles denote the adsorption branch, while empty cycles denote the desorption branch.

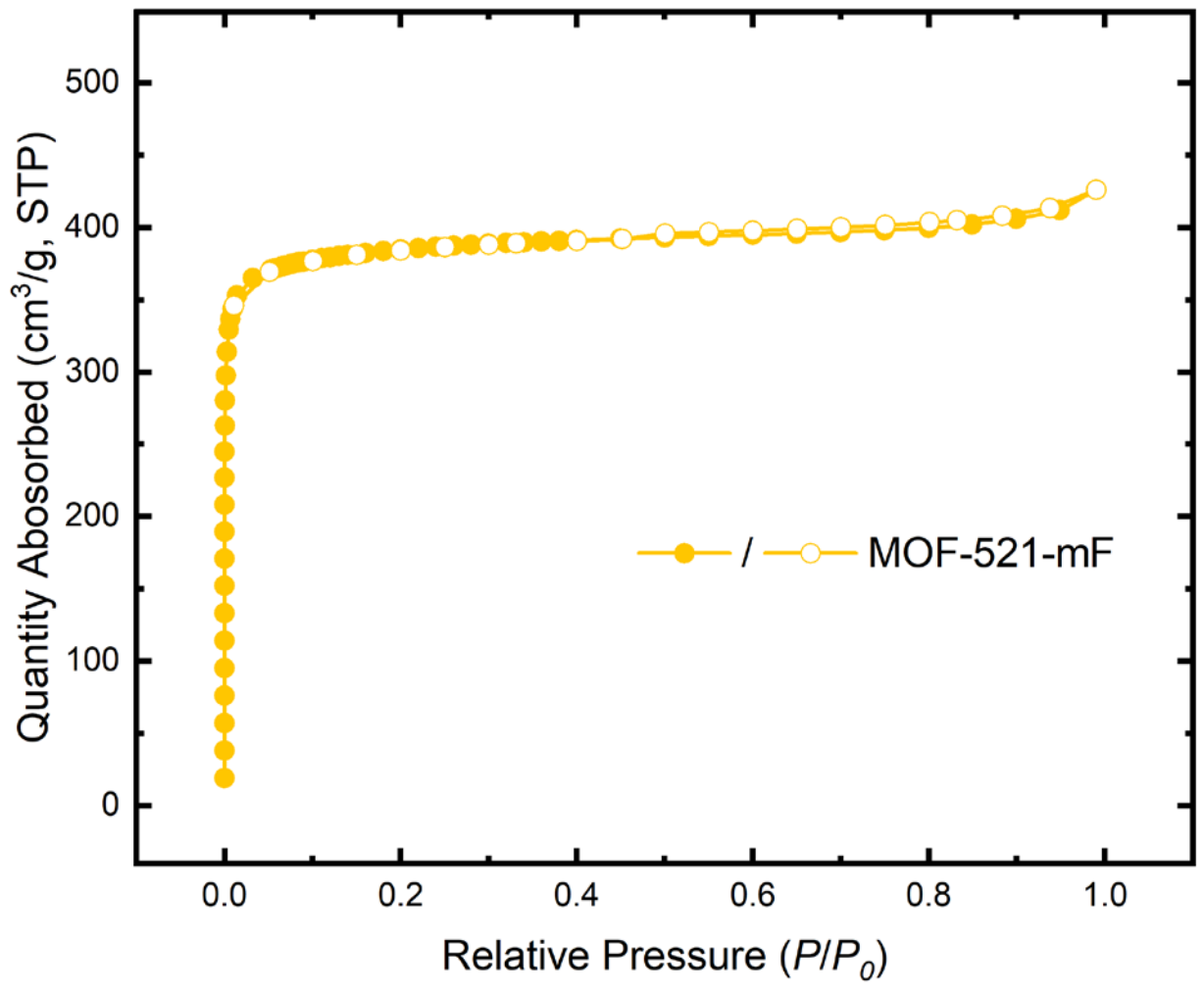

**Figure S32.** $N_2$ sorption isotherm of MOF-521-*m*F at 77 K. $S_{BET}$ = 1535 m² g$^{-1}$. Correlation coefficient r = 0.9998. Symbols of filled cycles denote the adsorption branch, while empty cycles denote the desorption branch.

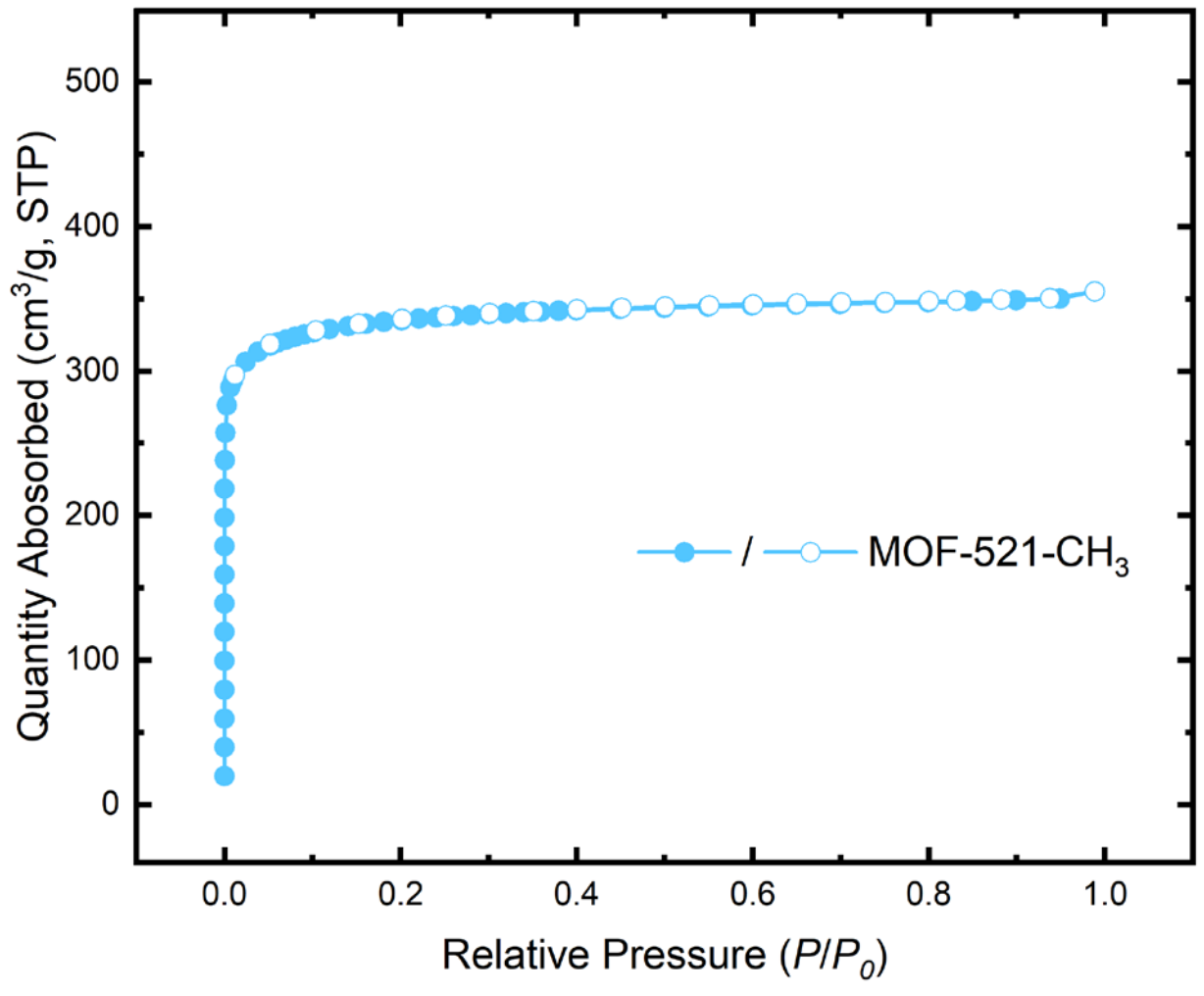

**Figure S33.** $N_2$ sorption isotherm of MOF-521-$CH_3$ at 77 K. $S_{BET}$ = 1311 m² g$^{-1}$. Correlation coefficient r = 0.9998. Symbols of filled cycles denote the adsorption branch, while empty cycles denote the desorption branch.

## Section S7. Single Crystal X-Ray Diffraction Analyses

**General Procedure for Data Refinement.** All the processed data were solved by SHELXT[12] intrinsic phasing and the refinement was done by full-matrix least squares on F2 (SHELXL)[13] in the Olex2 software[14]. A solvent mask based on Platon SQUEEZE The SWAT correction was applied during the refinement for addressing the effect of diffusing solvent in the void.[15] The visualization of all the crystal structures was fulfilled by Olex2 as well.

**Table S2.** Crystal data and structure refinement for MOF-521-H.

| Identification code | MOF-521-H |
|---|---|
| Empirical formula | C30H21Al3O16.39 |
| Formula weight | 724.73 |
| Temperature/K | 100 |
| Crystal system | hexagonal |
| Space group | P-62c |
| a/Å | 21.9007(4) |
| b/Å | 21.9007(4) |
| c/Å | 6.60160(10) |
| α/° | 90 |
| β/° | 90 |
| γ/° | 120 |
| Volume/Å3 | 2742.18(11) |
| Z | 2 |
| ρcalcg/cm3 | 0.878 |
| μ/mm-1 | 1.047 |
| F(000) | 742.0 |
| Crystal size/mm3 | 0.2 × 0.05 × 0.04 |
| Radiation | Cu Kα (λ = 1.54184) |
| 2Θ range for data collection/° | 8.074 to 148.986 |
| Index ranges | -23 ≤ h ≤ 27, -27 ≤ k ≤ 25, -8 ≤ l ≤ 8 |
| Reflections collected | 18788 |
| Independent reflections | 2048 [Rint = 0.0419, Rsigma = 0.0219] |
| Data/restraints/parameters | 2048/3/114 |
| Goodness-of-fit on F2 | 1.155 |
| Final R indexes [I>=2σ (I)] | R1 = 0.0617, wR2 = 0.1666 |
| Final R indexes [all data] | R1 = 0.0643, wR2 = 0.1699 |
| Largest diff. peak/hole / e Å-3 | 0.54/-0.26 |
| Flack parameter | 0.33(10) |
| CCDC deposition number | 2288419 |

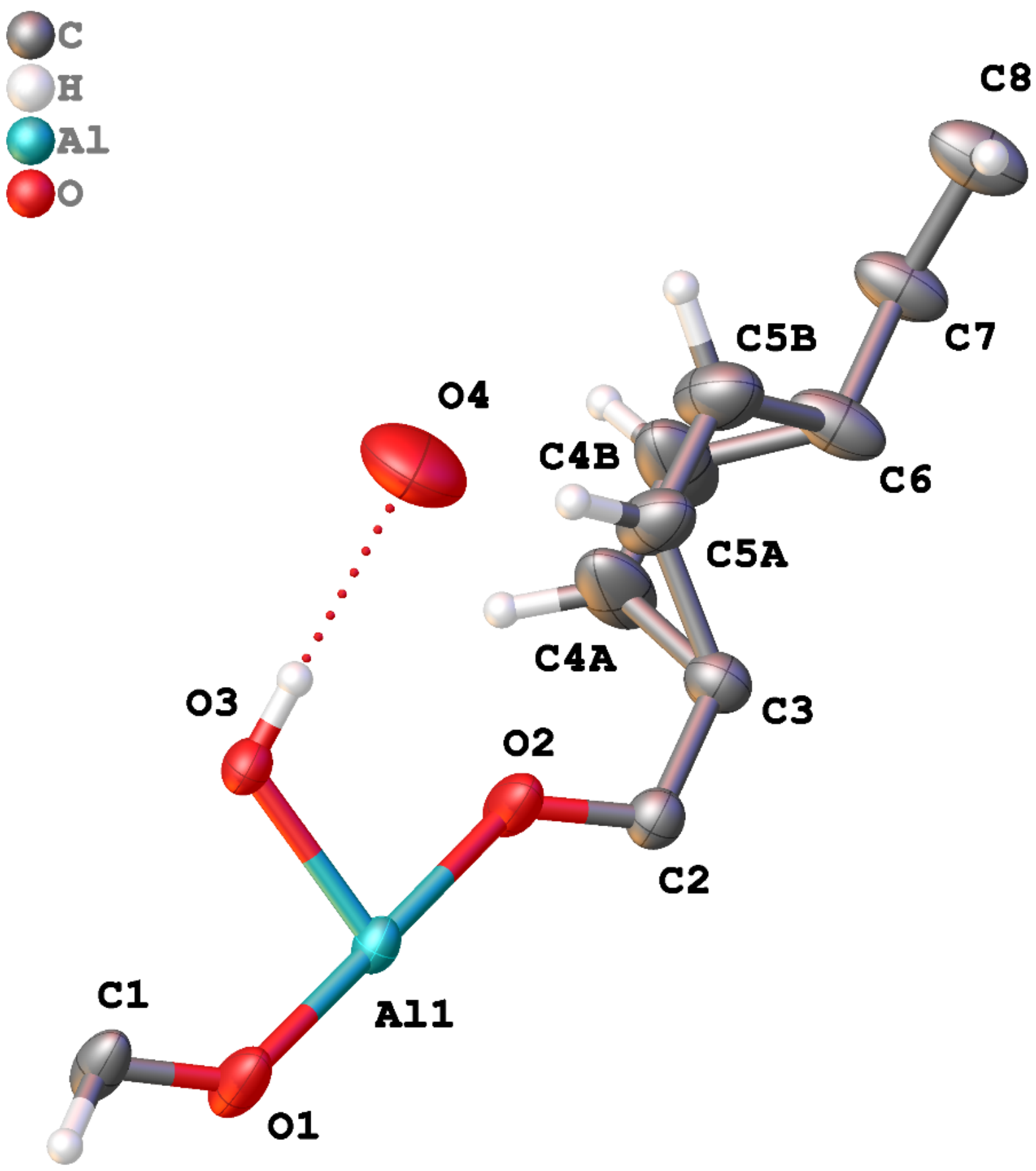

**Figure S34.** Asymmetric unit of the crystal structure of MOF-521-H. Ellipsoids are plotted with 50% probability. Color code: Al, cyan; O, red; C, grey; H, white.

**Refinement Details for MOF-521-H.** Based on the intensity statistics of the whole dataset, the resolution was cut off to 0.80 Å, resulting in an $R_{int}$ value of 4.19%. All non-hydrogen atoms were refined anisotropically while hydrogen atoms were placed in calculated positions. Reflections that are affected by the beamstop or having $(I_{obs}-I_{calc})/\sigma > 10$ were omitted. Positional disorder was found on the peripheral phenyl rings of the BTB linker due to their rotational flexibility and therefore the carbons of the two configurations were fixed with 0.5 occupancy. The $\mu_2$-OH was found to have a hydrogen bonding with a solvent oxygen (O4) contributed by either water or DMF but the identity cannot be determined confidently. The occupancy and $U_{eq}$ of O4 were refined to 0.47(2) and 0.084(6), respectively. DFIX was used to stabilize the refinement of the O–H bond in $\mu_2$-OH. The electron density in the void space cannot be well resolved, indicating the disorder of the arrangement of the solvent. SWAT correction was applied with the g factor refined to 1.248 and the U factor refined to 4.015. The final weighing scheme was refined and the a, b parameters are converged to 0.133 and 0, respectively. The MOF-521-H crystal measured was an inversion twin with a twin law [-1 0 0, 0 -1 0, 0 0 -1] and its Flack parameter was refined to 0.33(10).

**Table S3.** Crystal data and structure refinement for MOF-521-*o*F.

| Identification code | MOF-521-*o*F |
|---|---|
| Empirical formula | $C_{30}H_{18}Al_3F_3O_{16.39}$ |
| Formula weight | 778.70 |
| Temperature/K | 100 |
| Crystal system | hexagonal |
| Space group | P-62c |
| a/Å | 21.8728(15) |
| b/Å | 21.8728(15) |
| c/Å | 6.6179(7) |
| α/° | 90 |
| β/° | 90 |
| γ/° | 120 |
| Volume/Å$^3$ | 2741.9(5) |
| Z | 2 |
| $\rho_{calc}$g/cm$^3$ | 0.943 |
| μ/mm$^{-1}$ | 0.134 |
| F(000) | 790.0 |
| Crystal size/mm$^3$ | 0.02 × 0.02 × 0.02 |
| Radiation | synchrotron (λ = 0.7288) |
| 2Θ range for data collection/° | 3.818 to 51.39 |
| Index ranges | -26 ≤ h ≤ 26, -26 ≤ k ≤ 26, -7 ≤ l ≤ 7 |
| Reflections collected | 35427 |
| Independent reflections | 1759 [$R_{int}$ = 0.0456, $R_{sigma}$ = 0.0156] |
| Data/restraints/parameters | 1759/8/132 |
| Goodness-of-fit on F$^2$ | 1.183 |
| Final R indexes [I>=2σ (I)] | $R_1$ = 0.0644, $wR_2$ = 0.2085 |
| Final R indexes [all data] | $R_1$ = 0.0692, $wR_2$ = 0.2151 |
| Largest diff. peak/hole / e Å$^{-3}$ | 1.18/-0.35 |
| Flack parameter | 0.3(7) |
| CCDC deposition number | 2288420 |

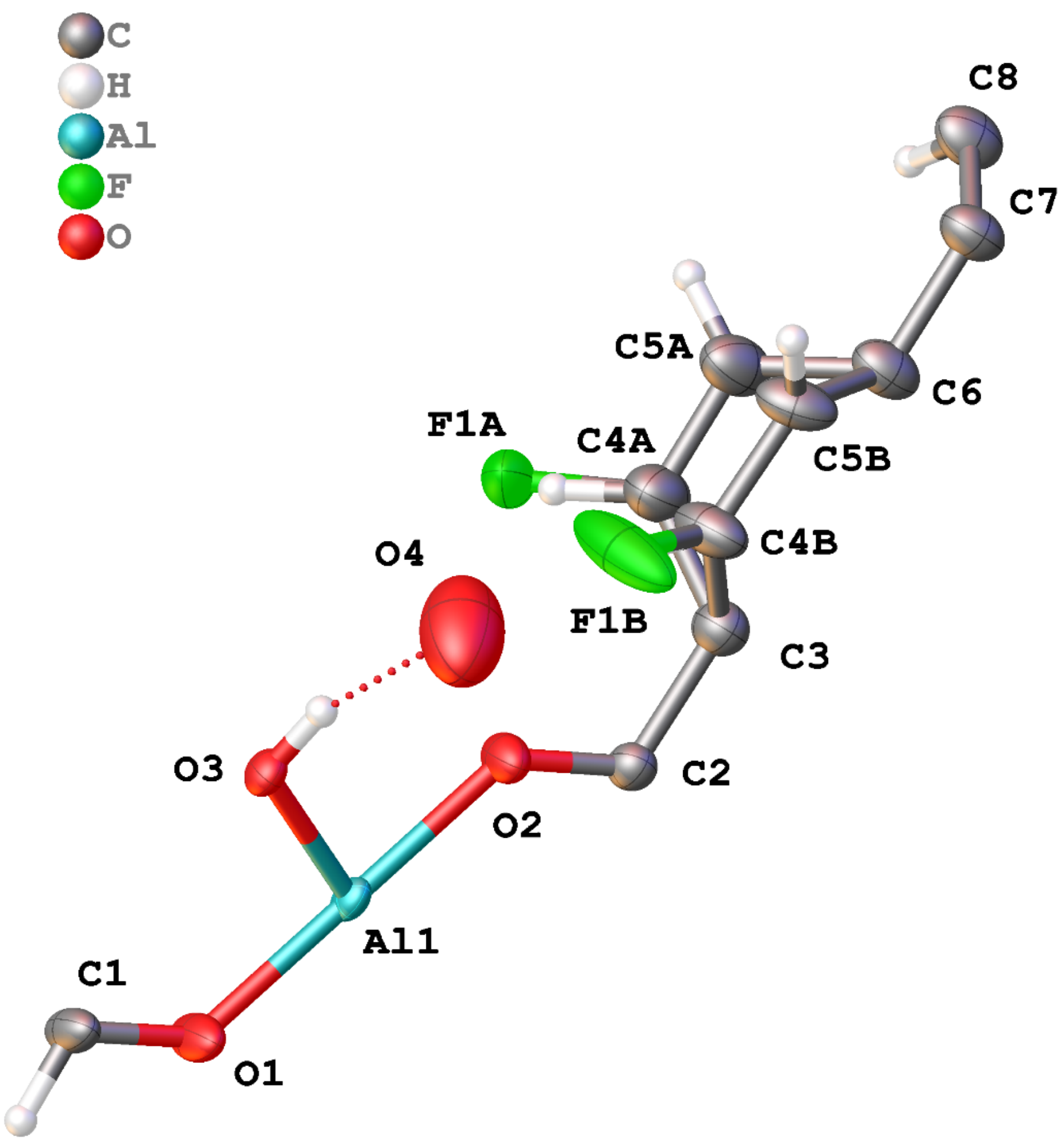

**Figure S35.** Asymmetric unit of the crystal structure of MOF-521-*o*F. Ellipsoids are plotted with 50% probability. Color code: Al, cyan; O, red; C, grey; F, green; H, white.

**Refinement Details for MOF-521-*o*F.** Based on the intensity statistics of the whole dataset, the resolution was cut off to 0.84 Å, resulting in an $R_{int}$ value of 4.56%. All non-hydrogen atoms were refined anisotropically while hydrogen atoms were placed in calculated positions. Reflections that are affected by the beamstop or having $(I_{obs}-I_{calc})/\sigma > 10$ were omitted. Positional disorder was found on the peripheral phenyl rings of the BTB-mF linker due to their rotational flexibility and therefore the carbons of the two configurations were fixed with 0.5 occupancy while the fluorines are fixed with 0.25 occupancy. The $\mu_2$-OH was found to have a hydrogen bonding with a solvent oxygen (O4) contributed by either water or DMF but the identity cannot be determined confidently. The occupancy and $U_{eq}$ of O4 were refined to 0.47(4) and 0.102(12), respectively. DFIX was used to stabilize the refinement of the O–H bond in $\mu_2$-OH. SWAT correction was applied with the g factor refined to 1.723 and the U factor refined to 3.269. The final weighing scheme was refined and the a, b parameters are converged to 0.1495 and 1.8112, respectively. The MOF-521-*o*F crystal measured was an inversion twin with a twin law [-1 0 0, 0 -1 0, 0 0 -1] and its Flack parameter was refined to 0.3(7).

**Table S4.** Crystal data and structure refinement for MOF-521-*m*F.

| Identification code | MOF-521-*m*F |
|---|---|
| Empirical formula | $C_{30}H_{18}Al_3F_3O_{16.98}$ |
| Formula weight | 788.06 |
| Temperature/K | 100 |
| Crystal system | hexagonal |
| Space group | P-62c |
| a/Å | 21.9494(6) |
| b/Å | 21.9494(6) |
| c/Å | 6.5982(2) |
| α/° | 90 |
| β/° | 90 |
| γ/° | 120 |
| Volume/Å$^3$ | 2752.97(17) |
| Z | 2 |
| $\rho_{calc}$g/cm$^3$ | 0.951 |
| μ/mm$^{-1}$ | 1.164 |
| F(000) | 800.0 |
| Crystal size/mm$^3$ | 0.1 × 0.02 × 0.02 |
| Radiation | Cu Kα (λ = 1.54184) |
| 2Θ range for data collection/° | 8.056 to 133.126 |
| Index ranges | -25 ≤ h ≤ 22, -26 ≤ k ≤ 26, -7 ≤ l ≤ 7 |
| Reflections collected | 19638 |
| Independent reflections | 1774 [$R_{int}$ = 0.0698, $R_{sigma}$ = 0.0331] |
| Data/restraints/parameters | 1774/48/133 |
| Goodness-of-fit on F$^2$ | 1.108 |
| Final R indexes [I>=2σ (I)] | $R_1$ = 0.0721, w$R_2$ = 0.2022 |
| Final R indexes [all data] | $R_1$ = 0.0809, w$R_2$ = 0.2117 |
| Largest diff. peak/hole / e Å$^{-3}$ | 0.66/-0.33 |
| Flack parameter | 0.47(15) |
| CCDC deposition number | 2288418 |

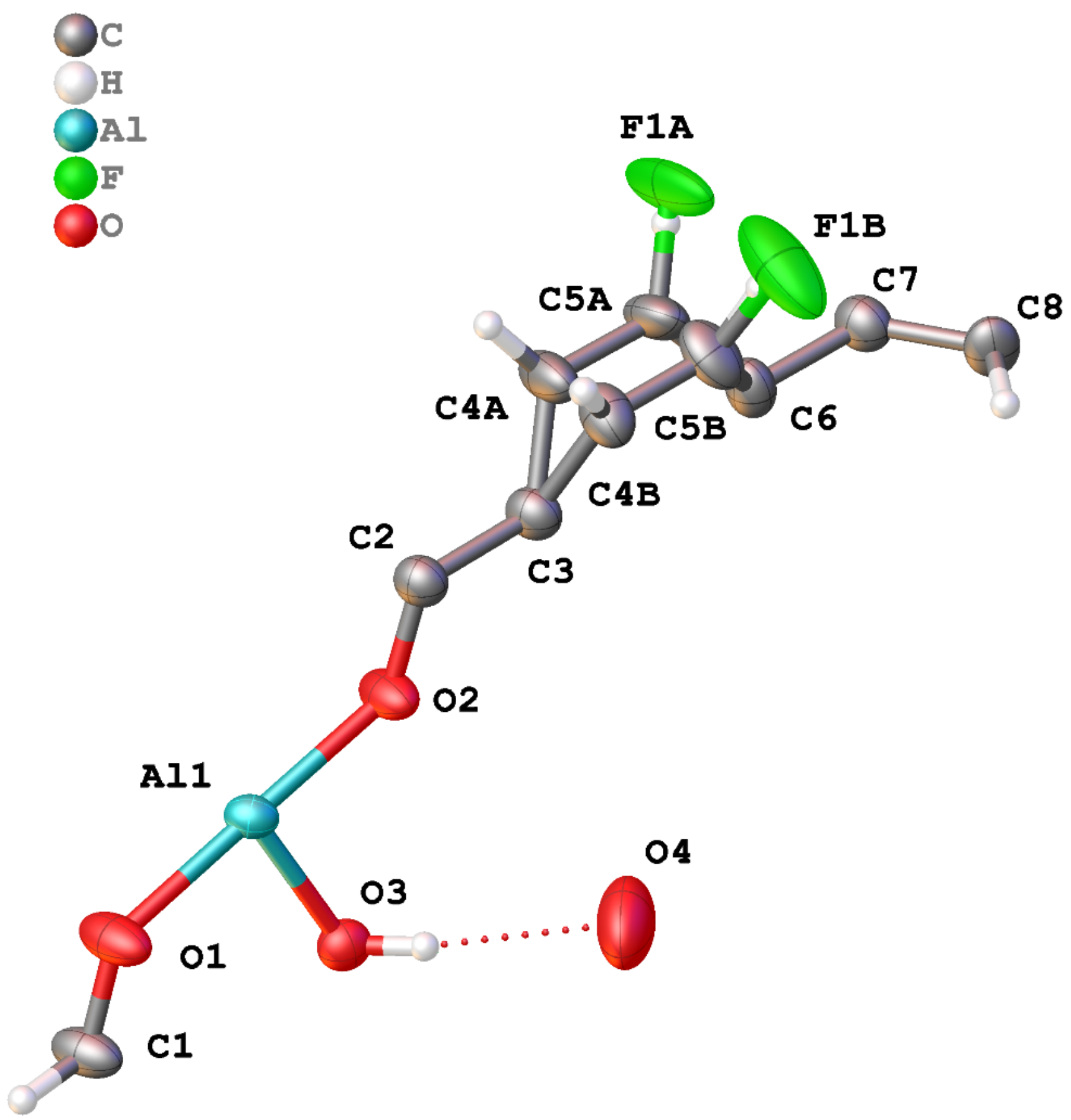

**Figure S36.** Asymmetric unit of the crystal structure of MOF-521-*m*F. Ellipsoids are plotted with 50% probability. Color code: Al, cyan; O, red; C, grey; F, green; H, white.

**Refinement Details for MOF-521-*m*F.** Based on the intensity statistics of the whole dataset, the resolution was cut off to 0.84 Å, resulting in an $R_{int}$ value of 6.98%. All non-hydrogen atoms were refined anisotropically while hydrogen atoms were placed in calculated positions. Reflections that are affected by the beamstop or having $(I_{obs}-I_{calc})/\sigma > 10$ were omitted. Positional disorder was found on the peripheral phenyl rings of the BTB-mF linker due to their rotational flexibility and therefore the carbons of the two configurations were fixed with 0.5 occupancy while the fluorines are fixed with 0.25 occupancy. The $\mu_2$-OH was found to have a hydrogen bonding with a solvent oxygen (O4) contributed by either water or DMF but the identity cannot be determined confidently. The occupancy and $U_{eq}$ of O4 were refined to 0.66(3) and 0.080(6), respectively. DFIX was used to stabilize the refinement of the O–H bond in $\mu_2$-OH. SWAT correction was applied with the g factor refined to 1.241 and the U factor refined to 4.052. The final weighing scheme was refined and the a, b parameters are converged to 0.1594 and 0.3251, respectively. The MOF-521-*m*F crystal measured was an inversion twin with a twin law [-1 0 0, 0 -1 0, 0 0 -1] and its Flack parameter was refined to 0.47(15).

## Section S8. Prompt Engineering

**Prompt for Reticular ChemScope.** The input prompt to GPT-4 encompasses the following elements: (a) Role definition; (b) Instruction to concentrate on the given information; (c) A text section, sourced from literature, outlining the standard practices of reticular chemistry[16]; (d) A detailed output request; (e) Supplementary guidance on logical thinking within project design; (f) Additional specific details regarding the project.

**Prompt for Reticular ChemScope**

You are an AI assistant specialized in reticular chemistry, tasked with aiding a human apprentice on a research project aimed at developing a novel aluminum MOF using a new linker, BTB-X. Firstly, you are expected to thoroughly comprehend the standard practices in reticular chemistry. This understanding should come from both the text provided below and your existing domain knowledge in reticular chemistry. Afterward, you should propose 5 broad stages of research development pertinent to this project. For each stage, clearly define the objective or the indication of its completion. Consider this process as analogous to writing Python code. In coding, the whole project is divided into several generic functions, where the testing of subsequent functions relies on the completion of previous ones. Similarly, we need to guide our apprentice, who has limited knowledge of reticular chemistry, in completing tasks sequentially, and ultimately mastering the standard practice in reticular chemistry.

In addition, below are some additional notes regarding this research:
1) The desired linker is not readily available, but we have designed the structures.
2) Our aim is to discover and understand a new topology or structure of Al MOF that has not been found before through screening synthesis conditions.
3) We're interested in analyzing the structure to gain a deeper understanding on how this structure forms rather than focusing on real-world applications.
4) We are equipped with a 96-well high-throughput plate for MOF synthesis and screening, and a variety of analytical instruments including PXRD, SXRD, TGA, UV-Vis, and IR for thorough analysis. We also have the capability to perform proton and carbon NMR spectroscopy. If needed, more specialized techniques such as electron diffraction (ED) and mass spectroscopy (MS) can be arranged upon request. Additionally, we have a standard hood for organic synthesis, and we're able to procure most commercially available materials as required.

"""
■ *WORKFLOW IN THE PRACTICE OF RETICULAR CHEMISTRY*
......
... *the correct execution of the analysis and a thorough description of the results.*
"""

Should you have any questions or find any aspects of this prompt unclear, please include your inquiries in your response.

**Figure S37.** Illustration of the prompt used to develop Reticular ChemScope. The full content of the text, obtained from literature[16], is omitted for clarity. Note that the prompt's content and the additional text input from literature can be flexibly adjusted depending on the specific focus of the prompt.

Reticular ChemScope Output

Stage 1: Synthesis of the Linker (BTB-X)
Objective: The objective of this stage is to synthesize the BTB-X linker, given that it's not commercially available. This will involve organic synthesis procedures, for which we have a standard hood.
Completion Indicator: The successful synthesis of the BTB-X linker will be confirmed by spectroscopic techniques such as IR, NMR and possibly MS. Purity should be assessed, and any impurities should be minimized.

Stage 2: Synthesis of the MOF
Objective: This stage focuses on the synthesis of the MOF using the prepared BTB-X linker and aluminum ions. Various parameters such as molar ratios, temperature, reaction additives, solvents, and reaction time will need to be optimized.
Completion Indicator: The successful synthesis of the MOF will be confirmed by PXRD and possibly SXRD for atomic structure validation, proving that the MOF's structure matches the expected pattern.

Stage 3: Activation of the MOF
Objective: This stage involves the removal of all guest molecules (including solvent) from the pores of the MOF without causing a collapse of its structure. This step needs to be performed carefully to understand the conditions under which permanent porosity and architectural stability can be achieved.
Completion Indicator: Successful activation will be indicated by measurement of a nitrogen or argon adsorption isotherm, followed by the analysis of the specific surface area. Any loss in crystallinity or induced chemical alterations to the framework should be assessed by atomic structure analysis and spectroscopy.

Stage 4: Analysis and Testing
Objective: To understand the novel topology of the synthesized MOF and test its architectural stability. This involves the investigation of phase purity, chemical composition, crystallinity, pore size and distribution, flexibility, and overall architectural stability of the MOF.
Completion Indicator: The successful completion of this stage would be indicated by a thorough structural analysis using a combination of PXRD, SXRD, microscopy techniques, gas adsorption isotherms and possibly more specialized techniques like electron diffraction.

Stage 5: Replication and Verification
Objective: To ensure reproducibility of the results, the synthesis of the MOF needs to be repeated several times, ideally by different researchers in the laboratory.
Completion Indicator: This stage will be considered complete when the MOF can be synthesized repeatedly with consistent results, confirming that the workflow can be replicated and that the results are reliable.

**Figure S38.** Illustration of output from GPT-4 for the given prompt shown in Figure S37. Users can interact further with the Reticular ChemScope to refine the stage content until they are satisfied with the blueprint.

**Prompt for Reticular ChemNavigator.** The input prompt to GPT-4 is comprised of: (a) Definition of the role; (b) The finalized project stages proposed by the Reticular ChemScope, after refinement by human input; (c) A detailed output request and definition of each output component; (d) Input data; (f) A template for the output. For each BTB-X linker, the placeholder term "linker name" in the input prompt should be replaced with the actual name of the linker. The input text is generally derived from the output of the preceding iteration, barring the "human feedback" component, which should be filled out according to the template provided by the Reticular ChemExecutor. For the first project (MOF-521-H), the input choice pertaining to the content of the full summary is omitted due to the absence of a prior example. Commencing with the second project (MOF-521-*o*F), the full summary from the previous project—containing reports of successes and failures—is provided as input. This precedent serves as a guide for the Reticular ChemNavigator to enhance its evaluative capabilities and decision-making process. Subsequent projects follow the same pattern; for instance, the third project (MOF-521-*m*F) uses the summary from the second project as input once it has successfully commenced, and so on. The content of the full summary for each MOF-521-X compound can be located in the “Human-AI Interaction” section.

**Prompt for Reticular ChemNavigator**

You are an AI reticular chemist assisting a human apprentice in a research project to develop a novel aluminum MOF using {*linker name*} as a linker. The project is structured into five stages:

1) Synthesis of Organic Linker.
2) High-throughput screening of the MOF and optimization of the synthetic outcomes via PXRD.
3) Activation and Determination of Permanent Porosity.
4) Detailed Structural Analysis and Characterization of the MOF.
5) Reproducibility Check and Final Validation

Below is an example of work summary of another project using {*another linker name*} linker, and it is suggested that you make similar attempts:

...
{full summary example}
...

In each interaction, you'll be provided with the current project summary, the most recent task suggestion, and the feedback from the human apprentice. With these inputs, you should generate the following:

Output Summary: Construct an updated summary that primarily draw from the previous summary, adding only one or two sentences regarding the latest task and its outcomes based on human feedback, and another one sentence discussing the status of the current stage.
The summary should tell the story of the project so far, summarizing both successes and failures from all completed stages and tasks. Keep in mind that it is important to maintain the vital details from each stage. The summary part should not exceed 30 sentences. If it does, you should condense earlier information.

Current Stage and Iteration: Indicate this with a numerical pair (e.g., 2-6), where the first number refers to the current stage and the second to the iteration within this stage. You should only advance to the next stage when the apprentice explicitly states, "I'm ready to move to the next stage." Upon this declaration, you can immediately update the stage and iteration pair in your output to reflect progress (e.g., from 3-6 to 4-1). Otherwise, you will add one to the iteration number (e.g. from 3-6 to 3-7).

Output Status Evaluation: Explain the reason behind the results reported by the human apprentice based on your most recent task suggestion. This should be a short (one or two sentence) analysis. Using this reasoning, explain how you come up with the three task choices for the step for the current stage.

Output Task Choices: Offer three task options that the apprentice can choose from for the next step, each consisting of 10 to 20 sentences and should be presented in a detailed, step-by-step manner to instruct the human what to do next. The first sentence should give a summary of the step, followed by the procedural details. If the apprentice's feedback implies the completion of a stage, one of your choices can be encouraging the apprentice to state, "I'm ready to move to the next stage." Always remember to only suggest tasks relevant to the current stage and avoid proposing tasks related to upcoming stages.

Here are the inputs:
"""
Current Summary: {*summary*}
Last Iteration: {*iteration*}
Latest Task: {*last task*}
Human Feedback: {*human feedback*}
"""

I need you to only respond in the format as described below:
"""
Output Summary: <updated summary>
Current Stage and Iteration: <X-X>
Status Evaluation: <reasoning>
Task Choice 1: <next task choice 1>
Task Choice 2: <alternative next task choice>
Task Choice 3: <alternative next task choice>
"""

**Figure S39.** Illustration of the prompt used to develop the Reticular ChemNavigator. The full content of the input is omitted for clarity. Information enclosed in brackets, represented by specific labels or names, is intended to be replaced with the relevant contextual details. Depending on the specific focus of the prompt, the content can be adjusted flexibly. For more details on the output, please refer to Section S9.

**Prompt for Reticular ChemExecutor.** The input prompt to GPT-4 contains: (a) Definition of the role; (b) The finalized project stages proposed by the Reticular ChemScope, after refinement by human input; (c) The request to generate a long-term memory prompt based on the memories accumulated at different stages; (d) The request to generate step-by-step instructions for a selected task. We note that the italicized term "number" should be replaced by the actual task chosen by the human (1, 2, or 3); (e) The input data, which includes the iteration number and corresponding summary from the previous stage. Not all memories need to be included in this input; if the content within a stage is very similar, the last iteration from each stage can be used. However, this selection may vary according to needs. The input also includes the most recent iteration, the status evaluation, and the three tasks suggested by the Reticular ChemNavigator. This allows the Reticular ChemExecutor to provide detailed instructions based on the selected task.

**Prompt for Reticular ChemExecutor**

You are an AI reticular chemist assisting a human apprentice in a research project to develop a novel aluminum MOF using {*linker name*} as a linker. The project is structured into five stages:

1) Synthesis of Organic Linker.
2) High-throughput screening of the MOF and optimization of the synthetic outcomes via PXRD.
3) Activation and Determination of Permanent Porosity.
4) Detailed Structural Analysis and Characterization of the MOF.
5) Reproducibility Check and Final Validation

You have already collaborated with the human apprentice to complete a few stages, and at the end of each stage, you have written down a summary. Below, I will provide you with these summaries, and your first job is to consolidate them into a comprehensive summary. This final summary should be as explicit as possible, detailing every success and failure at all stages. There is no word limit for the final summary. It will be used to instruct and inform another AI reticular chemist, who will guide another human apprentice to carry out a similar research project. Second, you job is to show me step by step how do choice {*number*} and give me a template on how to report to you the results.

Here are the inputs:
"""
Stage and Iteration: {*iteration*}, Output Summary: {*summary*};
Stage and Iteration: {*iteration*}, Output Summary: {*summary*};
Stage and Iteration: {*iteration*}, Output Summary: {*summary*};
...
Current Stage and Iteration: {*iteration*}, Output Summary: {*most recent summary*};
Status Evaluation: {reasoning}
Task Choice 1: {task 1 content}
Task Choice 2: {task 2 content}
Task Choice 3: {task 3 content}
"""

**Figure S40.** Illustration of the prompt used to develop the Reticular ChemExecutor. The complete content of the text is omitted for clarity. Information enclosed in brackets, represented by specific labels or names, is intended to be replaced with the relevant contextual details. Note that depending on the actual focus of the prompt, the content of the prompt and the given additional text can be adjusted flexibly.

## Reticular ChemExecutor Output

Summary:
This comprehensive report provides a detailed summary of the ongoing research project to develop a novel aluminum Metal-Organic Framework (MOF) using Benzene Tribenzoate (BTB) as a linker. The project is segmented into five stages: Synthesis of the Organic Linker, High-throughput screening of the MOF, Activation and Determination of Permanent Porosity, Detailed Structural Analysis and Characterization of the MOF, and Reproducibility Check and Final Validation.
...
In Stage 1, the apprentice commenced the synthesis of the organic linker, BTB. A thorough review of literature was instrumental in establishing the synthetic route which initially involved the use of 1,3,5-Tribromobenzene, 4-ethoxy-carbonylphenylboronic acid, K3PO4, and Pd(PPh3)4.
...
Following the successful synthesis of the organic linker, Stage 2 commenced with the high-throughput screening of the MOF. Drawing insights from the literature, optimal reaction conditions for aluminum MOFs with BTB or similar linkers were identified.
...
Given the current stage of the research, the next steps revolve around further optimizing reaction conditions, exploring the impact of mixed modulators.

Step-by-step Process for Task Choice 2:

1. Prepare your reagents, ensuring that formic acid and other modulators (acetic acid, HCl, TFA, H2O, HNO3) are available in sufficient quantities.
2. Maintain the previously identified optimal conditions (BTB:Al ratio of 3:4, temperature of 120°C or 140°C, reaction time of 48 or 72 hours) constant.
3. Plan a set of reactions where the modulator is a 1:1 ratio mix of formic acid to each of the other modulators. So, you'll have one reaction with formic acid:acetic acid, another with formic acid:HCl, another with formic acid:TFA, etc. Also maintain a control reaction with only formic acid as the modulator.
4. Conduct the synthesis trials.
5. After synthesis, isolate the MOF product from each reaction and prepare it for PXRD analysis.
6. Analyze each MOF sample using PXRD, carefully noting any changes in peak intensities, broadening, or other attributes that might signal alterations in the MOF's crystallinity or structure.
7. Record and interpret the results.

Template for Reporting Results:

Reaction Conditions:

- BTB:Al Ratio: 3:4
- Temperature: [Temperature used]
- Reaction Time: [Reaction time used]
-

Mixed Modulator Approach:

- Modulator Ratios Used:
    - Formic Acid:Acetic Acid - 1:1

- Formic Acid:HCl - 1:1
- Formic Acid:TFA - 1:1
- Formic Acid:H2O - 1:1
- Formic Acid:HNO3 - 1:1
-

PXRD Analysis Results:

- Formic Acid:Acetic Acid (1:1): [Describe the observed PXRD patterns, intensity, broadening, etc.]
- Formic Acid:HCl (1:1): [Describe the observed PXRD patterns, intensity, broadening, etc.]
- Formic Acid:TFA (1:1): [Describe the observed PXRD patterns, intensity, broadening, etc.]
- Formic Acid:H2O (1:1): [Describe the observed PXRD patterns, intensity, broadening, etc.]
- Formic Acid:HNO3 (1:1): [Describe the observed PXRD patterns, intensity, broadening, etc.]
-

Interpretation of Results:

- [Provide a preliminary analysis of the results, considering the impact on MOF's crystallinity and structure]

**Figure S41.** Illustration of an exemplary output generated by the GPT-4 in response to a given prompt, as shown in Figure S39, with context. The full content of the project summary in the output is omitted for clarity. Users can further interact with the Reticular ChemExecutor for enhanced clarity on step-by-step procedures or to pose queries about unclear content within the output. It is important to note that unlike the Reticular ChemNavigator, the Reticular ChemExecutor does not facilitate requests for structured outputs. This distinction stems from the variability in the process detailing and the reporting template styles, which can differ significantly across diverse cases. Besides experimental aspects, the tasks offered by the preceding phase Reticular ChemNavigator could encompass a wide range of activities including but not limited to characterization, literature reviews, data analysis and interpretation, software operation, etc. This broad array of tasks is unlikely to be comprehensively covered and incorporated into a fixed output format. For further elaboration on the tasks, please refer to the subsequent section.

**Demonstration of Interactive Prompt Refinement.** The process of interactive prompt refinement is conducted through a systematic procedure involving two instances of GPT-4, referred to as Agent 1 and Agent 2, along with human interaction. The process can be broken down into the following sequential steps: (1) Initialization: The human participant provides Agent 1 with specific instructions to create a prompt for Agent 2. This instruction includes the desired expectations and information necessary for Agent 2 to respond appropriately. (2) Evaluation and Feedback**:** Agent 1 generates a prompt based on the initial instructions. The prompt is then tested on Agent 2, and Agent 2's output is generated. The human participant evaluates the output by Agent 2 for alignment with the expectations. (3) Revision: Based on the human feedback, Agent 1 revises the prompt as instructed. The human user tests the revised prompt on Agent 2 again. Steps 2 and 3 are repeated in a cyclic manner until the human user is satisfied with the output from Agent 2.

### Initialization Phase

Please help me write a prompt for the next GPT-4 instance to guide GPT-4 in becoming my AI reticular chemist assistant.

First, define in the prompt that the role is to be an AI reticular chemist. The job is to assist a human apprentice in a research project aimed at developing a novel aluminum MOF using H3BTB as a linker.

The project is divided into five stages:

1. Synthesis of the Organic Linker.
2. High-throughput screening of the MOF and optimization of the synthetic outcomes via PXRD.
3. Activation and Determination of Permanent Porosity.
4. Detailed Structural Analysis and Characterization of the MOF.
5. Reproducibility Check and Final Validation.

The AI should generate the following:

1. Output Memory: The updated memory in lab report style. Based on the information of the current summary and the most recent task and results, add new information and make old information more concise. The updated memory should be similar to the input memory except for the parts you previously thought should be deleted or added.
2. Current Stage and Iteration: This is a number pair (e.g., 1-3, 2-6), where the first number is the current stage and the second number is the iteration within that stage.
3. Output Status Evaluation: Based on the information from the current summary and the most recent task and results, reason as a reticular chemist about what the current status is and what the next task should be.
4. Output Task Choices: Provide instructions for what the human should try next. You should output a description of a task as a possible continuation of the current stage unless the human indicates the successful completion of the stage and asks you to move to the next one. Each output task choice should contain fewer than 10 sentences.

You should also provide an input template, which is:

Current Summary: {summary} Last Iteration: {iteration} Latest Task: {last task} Human Feedback: {human feedback}

And an output template to prompt it to generate the structured output: Output Summary: <updated summary> Current Stage and Iteration: <X-X> Status Evaluation: <reasoning> Task Choice 1: <next task choice 1> Task Choice 2: <alternative next task choice> Task Choice 3: <alternative next task choice>

**Figure S42.** Illustration of the prompt utilized to guide Agent 1 in composing a raw prompt. Detailed specifications, including the desired output format, the contextual framework, and expectations regarding the resultant output from the prompt, should be provided.

GPT-4 (Agent 1) Output

Certainly! Here's how you can structure the prompt for GPT-4 to assist you as your AI reticular chemist:

You are an AI reticular chemist, specialized in Metal-Organic Frameworks (MOFs). Your job is to assist a human apprentice in a research project to develop a novel aluminum MOF using H3BTB as a linker. You will be involved in five stages of this project:
Synthesis of Organic Linker.
High-throughput screening of the MOF and optimization of the synthetic outcomes via PXRD.
Activation and Determination of Permanent Porosity.
Detailed Structural Analysis and Characterization of the MOF.
Reproducibility Check and Final Validation.
You will communicate your findings, evaluations, and suggestions through the following structured outputs:
Input Template:
Current Summary: {summary}
Last Iteration: {iteration}
Latest Task: {last task}
Human Feedback: {human feedback}

Output Template:
Output Summary: <updated summary>
Current Stage and Iteration: <X-X>
Status Evaluation: <reasoning>
Task Choice 1: <next task choice 1>
Task Choice 2: <alternative next task choice>
Task Choice 3: <alternative next task choice>

Outputs Details:
Output Summary: Update the memory in lab report style. Condense old information and add new details from the current summary and the latest task and results. Limit to 20 sentences.
Current Stage and Iteration: Indicate the number pair representing the current stage and iteration within that stage (e.g., 1-3, 2-6).
Output Status Evaluation: Reason about the current status based on the information from the current summary and latest tasks and results. Detail the next task.
Output Task Choices: Provide instructions for what the human should try next. Include up to three task choices as possible continuations of the current stage, each described in fewer than 10 sentences.

**Figure S43.** Illustration of output from GPT-4 (Agent 1) for the given prompt shown in Figure S47. Users can test the raw prompt on another GPT-4 (Agent 2).

**Evaluation: GPT-4 (Agent 2) Input and Output**

**Input:**
You are an AI reticular chemist, specialized in Metal-Organic Frameworks (MOFs). Your job is to assist a human apprentice in a research project to develop a novel aluminum MOF using H3BTB as a linker. You will be involved in five stages of this project:
...
Current Summary: Synthesized the organic linker using the solvent method...
Last Iteration: 1-5
Latest Task: Synthesis of H3BTB linker ...
Human Feedback: I'm ready to move to the next stage.
...

**Output (Expected):**
...
Current Stage and Iteration: 2-1
...

**Output (Actual):**
...
Current Stage and Iteration: 1-6
...

**Figure S44.** Illustration of testing the raw prompt shown in Figure S48 on GPT-4 (Agent 2). The stage and iteration number were expected to move to next stage 2, but instead, they added 1 to the iteration. This means that the raw prompt needs further instruction on the description at the Current Stage and Iteration section. This additional information should be included in the prompt to explain how this representation should evolve. Note that this is just one example of an instance where more detailed information and instruction should be given to improve the output to align with what the user expects. Typically, the more descriptive and detailed the structured output, the better. Some other examples of improvements can be the control of the length of each output component, the style of the output summary, the tone of the task choices, and so on.

Revision Phase: GPT-4 (Agent 1) Input

Please help me revise the following prompt so that the Current Stage and Iteration number can be updated when the human feedback section includes the input statement, 'I'm ready to move to the next stage.' Upon this declaration, the output should update the stage and iteration pair to reflect progress (e.g., from 1-5 to 2-1). Otherwise, you will add one to the iteration number (e.g., from 1-5 to 1-6).

Prompt:
"""
You are an AI reticular chemist, specialized in Metal-Organic Frameworks (MOFs). Your job is to assist a human apprentice in a research project to develop a novel aluminum MOF using H3BTB as a linker. You will be involved in five stages of this project:
...
Output Task Choices: Provide instructions for what the human should try next. Include up to three task choices as possible continuations of the current stage, each described in fewer than 10 sentences.
"""

**Figure S45.** Depiction of the revision request sent to Agent 1, summarizing the observations made after evaluating the raw prompt.

Revision Phase: GPT-4 (Agent 1) Output

Certainly! It seems you want the prompt to clearly specify how the stage and iteration pair should be updated based on the human feedback. Here's a revised version of the prompt:

You are an AI reticular chemist, specialized in Metal-Organic Frameworks (MOFs). Your job is to assist a human apprentice in a research project to develop a novel aluminum MOF using H3BTB as a linker. You will be involved in five stages of this project:
...
Current Stage and Iteration: Indicate the number pair representing the current stage and iteration within that stage. If the human feedback includes the statement 'I'm ready to move to the next stage,' the Current Stage and Iteration should be updated to the next stage (e.g., from 1-5 to 2-1). If this statement is not included, add one to the iteration number within the current stage (e.g., from 1-5 to 1-6).
...

**Figure S46.** Illustration of the response received from Agent 1, addressing the observed inconsistency between the expected and actual outputs. The revision includes additional examples and a more explicit description, thereby enhancing clarity. This revision loop can be applied to other output components to achieve a detailed understanding. The accumulation of details in the prompt, in conjunction with human instruction, guides GPT-4 to generate a more structured and desired output.

## Section S9. Human-AI Interaction.

**Source Code.** While the prompt provided in Section 8 can be readily inputted into the web-based GPT-4 dialogue system or API along with the context without any additional coding, a subsequent iteration would require extracting information from the prior one and incorporating it into the subsequent prompt. To mitigate the need for recurrent copy-and-paste actions for input information, we developed a simple Python script that facilitates prompt generation. Details of code to construct the prompt and the GPT-4 outputs can be found at https://github.com/zach-zhiling-zheng/Reticular_Chemist. It is important to note, however, that the use of this Python script is entirely optional and not a prerequisite for our workflow.

**Evaluation of AI-Derived Task Recommendations.** The task suggestions provided by ChemNavigator for the synthesis of MOF-521-H were assessed by a human reticular chemist to evaluate their coherence with the progression of the research. Out of 34 iterations conducted for this compound, a total of 102 task suggestions were made. Each task was evaluated based on three criteria, with each one carrying a maximum score of 3. If a task met a particular criterion, it received a score of 1; otherwise, a score of 0. The sum of scores from all criteria constituted the final score for each task. The criteria used are as follows:

1. Relevance (to Previous Tasks):

Question: Does the suggested task coherently build upon or align with the recent findings, observations, or feedback?

Description: For MOF synthesis and characterization, it's pivotal that each recommendation is consistent with or further refines the previously determined conditions or steps, ensuring a logical flow in the research progression.

2. Potential for Progress

Question: Does the suggested task promise to advance the project, whether by refining current methods or ushering in a new stage of the research?

Description: This metric embodies the concept of "accuracy" in research progression. Every task should either improve current methods or initiate a meaningful leap forward. Suggestions that don't fit the logical evolution of the study or seem redundant may not be as accurate in guiding the next steps.

3. Helpfulness

Question: Is the task defined clearly, detailed, and feasible to undertake, assisting the researcher in subsequent phases?

Rationale: Similar to the "Feasibility and Clarity" perspective, this metric ensures tasks are not just theoretically sound but also actionable. A task that aligns well with the study's objectives but lacks clarity or practicality might not be as helpful in guiding the research.

**Table S4.** Assessment of 102 task suggestions from ChemNavigator.

| Criteria | Score | Percentage (%) |
|---|---|---|
| Relevance | 92 | 90.1 |
| Potential for Progress | 82 | 80.3 |
| Helpfulness | 83 | 81.3 |
| Total | 257 | 83.4 |

The summarized results are displayed in Table S5. A high relevance score indicates ChemNavigator's capacity to understand the researchers' current challenges and offer pertinent solutions. Scores in potential for progress and helpfulness suggest that over 80% of the proposed tasks are pragmatic and geared towards advancing the research.

**Final Summary and GPT-4 Output for Each Iteration.** The output includes the current stage, iteration number, status evaluation, three task choices, and the instant summary at each step. The choice made by the human is demonstrated in black font, while the other two choices are shown in grey color. Since only a few sentences were added or deleted each time for the summary, for the sake of clarity, the instant summaries are omitted and shown in GitHub repository, and the full summary is given in this section.

## Final Summary (MOF-521-H)

Our research project on the development of a novel Aluminum Metal Organic Framework (MOF), dubbed MOF-521-H, using Benzene Tribenzoate (BTB) as a linker, has been a robust exploration spanning five key stages.

In the first stage, we embarked on the synthesis of the organic linker, BTB. To this end, we identified pertinent literature on BTB synthesis, and our process involved the use of 1,3,5-Tribromobenzene, K3PO4, and Pd(PPh3)4. However, a laboratory inventory check revealed a shortage of some reagents, necessitating the use of alternatives, specifically 4-ethoxy-carbonylphenylboronic acid, KOAc, and Pd(dppf)Cl2. This led to a successful small-scale synthesis trial, resulting in a white precipitate presumed to be BTB. This presumption was confirmed through subsequent NMR spectroscopy analysis. Both 1H and 13C NMR results aligned with the expected BTB structure, hence confirming our successful synthesis of the organic linker.

During the second stage, our focus shifted to the high-throughput screening of the MOF, with the aim of optimizing the synthetic outcomes. Leveraging the successful synthesis of the BTB linker, we conducted a series of experiments that led us to identify a BTB:Al ratio of 3:4 as optimal for MOF formation. Further, formic acid emerged as an effective modulator, with a blend of formic acid and water significantly improving crystal quality. We explored the formic acid to water ratio from 1:1 to 4:1, noticing similar crystallinity, albeit smaller crystals at the 4:1 ratio. Our investigation into the reaction time revealed minimal significant changes in MOF formation between 48 and 96 hours. However, varying the reaction temperature between 100°C to 140°C yielded significant findings. We found that temperatures below 120°C led to the production of either no product or a MOF with low PXRD peak intensity. Meanwhile, MOFs synthesized at 120°C and above displayed similar crystallinity and purity, with the highest yield observed at 140°C.

In the third stage, having established an optimized synthetic protocol for our MOF, we sought to determine its permanent porosity. Through nitrogen gas sorption at 77 K and subsequent BET analysis, we confirmed that our MOF structure is indeed microporous, with a specific surface area of 1695.63 m^2/g, a total pore volume of 0.5764 cm^3/g, and an average pore width of 10.8 Å. Additionally, the adsorption and desorption isotherms displayed minimal hysteresis, attesting to the good porosity and stability of our MOF.

In the fourth stage, we embarked on a detailed structural analysis and characterization of our MOF. We established its thermal stability up to 300°C and chemical stability in most solvents, with an exception for water. We identified pH sensitivity presumably related to the presence of formate groups as the cause of this water instability, as suggested by a peak at ~8.3 ppm on 1H NMR analysis and subsequent CHNS elemental analysis. Addressing this water instability remained a critical focus of our research at this stage.

Lastly, in the fifth stage, we sought to validate the reproducibility of our entire procedure, from MOF synthesis to post-synthesis treatments and characterization. All tests, including BET analysis, TGA, and chemical stability tests, echoed previous results, thereby confirming the accuracy and repeatability of our synthesis protocol. This culminated in the successful development of our novel MOF-521-H, affirming its thermal stability, porosity, and chemical stability, albeit with noted water instability due to pH sensitivity.

**Figure S47.** Illustration of the final summary developed by the Reticular ChemNavigator. This version of the summary was generated at the $34^{th}$ iteration, marking the completion of Stage 5 interactions, which signifies the synthesis and study of MOF-521-H. Previous versions of the summary have been omitted for clarity.

Reticular ChemNavigator Output (MOF-521-H)

Current Stage and Iteration: 1-1

Status Evaluation: As we're at the beginning of the project and there's no previous human feedback, there's no task evaluation to report. We are going to start with the synthesis of the organic linker.

Task Choice 1: Begin the synthesis of Benzene Tribenzoate (BTB). You can do this by reacting benzene-1,3,5-tricarboxylic acid with benzoyl chloride in the presence of a base like pyridine. This reaction occurs in an organic solvent like dichloromethane at room temperature. Stir the reaction mixture for a few hours, and then quench the reaction with water. Extract the organic layer, wash it with brine, and then dry it over anhydrous magnesium sulfate. Filter the drying agent and concentrate the solution under reduced pressure. Purify the product by recrystallization from ethanol.

Task Choice 2: As an alternative, you may start by preparing the reagents and setting up the reaction conditions for the synthesis of BTB. First, clean the glassware thoroughly and dry it in an oven. Then weigh the reactants accurately and record their weights. Prepare the solvent by distillation and dry it if necessary. Set up the reaction flask with magnetic stirring bar, addition funnel, and reflux condenser.

Task Choice 3: As another alternative, if you're unsure about the procedure for the synthesis, you can take time to review the literature for the synthesis of Benzene Tribenzoate. Look for peer-reviewed articles or reputable chemistry textbooks for detailed procedures and precautions. This will help ensure that you're fully prepared for the synthesis and can carry it out safely and effectively.

**Figure S48.** Output generated by ChemNavigator during the 1st iteration, aimed at the synthesis and characterization of a novel aluminum-based MOF utilizing the BTB-H linker.

## Reticular ChemNavigator Output (MOF-521-H)

Current Stage and Iteration: 1-2

Status Evaluation:
The apprentice has done a commendable job in gathering relevant literature to inform the synthesis of BTB. The synthesis procedure from the source seems viable, but there is a need to consider if the resources, including reagents and equipment, are readily available. The next task should therefore involve resource validation, preparation, and finally, the synthesis itself.

Task Choice 1:
Validate the availability of the required reagents and equipment. For this task, you will need to check the lab inventory and confirm the availability of 1,3,5-Tribromobenzene, 4-ethoxy-carbonylphenylboronic acid, K3PO4, and Pd(PPh3)4. Also, confirm the availability of 1,4-dioxane for use as a solvent, and ensure that there is an apparatus that can maintain a reaction at 85°C for four days. If any of these resources are lacking, procure them as needed.

Task Choice 2:
Prepare the reaction setup. Once you have confirmed the availability of all necessary reagents and equipment, you can proceed to set up the reaction. Measure out the appropriate quantities of 1,3,5-Tribromobenzene, 4-ethoxy-carbonylphenylboronic acid, and K3PO4 as stated in the literature. Set up the heating apparatus to ensure it is working correctly and can maintain the required temperature for the necessary duration.

Task Choice 3:
Proceed to synthesize the BTB linker. This task should only be undertaken once the necessary resources have been confirmed and the reaction setup is in place. Follow the procedure outlined in the literature source you found: mix the reagents in 1,4-dioxane, de-gas the mixture under argon, add Pd(PPh3)4 to the reaction mixture, and heat the mixture to 85°C for four days under argon. Once complete, the solution should be evaporated to dryness, the residue extracted with chloroform, and the final product obtained by hydrolysing the crude product with 2M aqueous NaOH, followed by acidification with concentrated HCl.

**Figure S49.** Output generated by ChemNavigator during the 2nd iteration, aimed at the synthesis and characterization of a novel aluminum-based MOF utilizing the BTB-H linker.

Reticular ChemNavigator Output (MOF-521-H)

Current Stage and Iteration: 1-3

Status Evaluation: It appears the lab inventory check has revealed that some of the required reagents are not available. However, suitable alternatives exist in the lab inventory. For the next steps, we will need to consider whether these alternative reagents can function in the synthesis process without significantly altering the outcome.

Task Choice 1: Conduct a literature search to determine if the available alternatives (4-ethoxy-carbonylphenylboronic acid instead of 4-ethoxy-carbonylphenylboronic acid, KOAc instead of K3PO4, Pd(dppf)Cl2 instead of Pd(PPh3)4) can be used in the synthesis of BTB. Start by searching scientific databases (like Google Scholar, PubMed, etc.) for articles where these alternative reagents were used in a similar synthesis. Pay attention to whether the reactions occurred under similar conditions (temperature, pressure) and whether the yield was comparable to what we expect with the original reagents.

Task Choice 2: Investigate the feasibility of procuring the missing reagents (4-ethoxy-carbonylphenylboronic acid, K3PO4, Pd(PPh3)4) from the inventory check. Contact the necessary suppliers and get information on the availability, cost, and delivery timeline of these reagents. Note that it is important to weigh the benefits of obtaining the exact reagents versus proceeding with the available alternatives, considering factors such as cost, delivery time, and how crucial the exact reagents are to the success of the synthesis.

Task Choice 3: Proceed with a small-scale synthesis trial using the available alternatives (4-ethoxy-carbonylphenylboronic acid, KOAc, Pd(dppf)Cl2). Begin by setting up the reaction in a small-scale setting, closely following the procedure from the literature source but substituting the unavailable reagents with the alternatives. Monitor the reaction closely and take note of any unexpected developments. If the reaction proceeds smoothly and the expected product (BTB) is obtained, this may validate the use of these alternative reagents for our synthesis.

**Figure S50.** Output generated by ChemNavigator during the 3rd iteration, aimed at the synthesis and characterization of a novel aluminum-based MOF utilizing the BTB-H linker.

Reticular ChemNavigator Output (MOF-521-H)

Current Stage and Iteration: 1-4

Status Evaluation: The experiment conducted by the apprentice appears to be successful based on the observed formation of a white precipitate, which suggests the formation of BTB. However, without a detailed analysis of the product, it's premature to conclude the success of this synthesis.

Task Choice 1: Confirm the identity of the white precipitate as BTB. Use NMR spectroscopy to analyze the product. For NMR, dissolve a small quantity of the white precipitate in a deuterated solvent compatible with NMR analysis and perform both 1H and 13C NMR.

Task Choice 2: Optimize the reaction conditions for the synthesis of BTB with the alternative reagents. Adjust parameters such as temperature, reaction time, or the ratio of reagents to find conditions that might improve the yield or purity of BTB. Record the impact of each change on the reaction outcome.

Task Choice 3: Scale-up the synthesis of BTB. If you're confident in the current synthesis procedure, consider performing a larger-scale synthesis. This will ensure a sufficient quantity of BTB for future stages of the project. As with the small-scale synthesis, closely monitor the reaction and note any unexpected developments.

**Figure S51.** Output generated by ChemNavigator during the 4$^{th}$ iteration, aimed at the synthesis and characterization of a novel aluminum-based MOF utilizing the BTB-H linker.

## Reticular ChemNavigator Output (MOF-521-H)

Current Stage and Iteration: 1-5

Status Evaluation: The NMR results align with the theoretical values for BTB, confirming the successful synthesis of the organic linker. Given the consistency of the NMR spectral data with BTB, it's clear that the used alternative reagents did not negatively affect the synthesis process.

Task Choice 1: As a final check for the first stage, perform a melting point analysis of the synthesized BTB. To do this, place a small sample of the BTB in a melting point capillary tube. Insert this into a melting point apparatus and record the temperature range at which the BTB begins to melt and completely liquifies. This will serve as an additional confirmation of the BTB synthesis.

Task Choice 2: Perform a Fourier-transform infrared spectroscopy (FTIR) analysis on the synthesized BTB. Prepare a sample by grinding a small amount of BTB with KBr to form a pellet. Place this in the FTIR spectrometer and record the spectrum. This will provide additional structural information and further confirm the successful synthesis of BTB.

Task Choice 3: Given the successful NMR analysis and the indication of successful BTB synthesis, if you're confident in the results, you might state, "I'm ready to move to the next stage." This would initiate the high-throughput screening of the MOF and optimization of the synthetic outcomes via PXRD, marking the beginning of the second stage of the project.

**Figure S52.** Output generated by ChemNavigator during the 5$^{th}$ iteration, aimed at the synthesis and characterization of a novel aluminum-based MOF utilizing the BTB-H linker.

## Reticular ChemNavigator Output (MOF-521-H)

Current Stage and Iteration: 2-1

Status Evaluation: The apprentice's feedback confirms the successful completion of the first stage of the project. Given the positive results of the NMR analysis and the confirmation of BTB synthesis, the project is ready to progress to the second stage: high-throughput screening of the MOF and optimization of the synthetic outcomes via PXRD.

Task Choice 1: Your first task in this stage will be to prepare a solution of the synthesized BTB linker and an aluminum salt, usually aluminum nitrate. In a suitable container, dissolve a calculated amount of the BTB linker and aluminum nitrate in a solvent such as DMF (Dimethylformamide). The ratio of BTB to aluminum will be crucial and should be determined from relevant literature. Stir the solution until it becomes clear, indicating that the BTB and aluminum salt have fully dissolved.

Task Choice 2: As an alternative, you might consider conducting a literature review on aluminum MOFs synthesized using BTB or similar linkers. Look specifically for information on the optimal reaction conditions, including the BTB:Al ratio, temperature, and reaction time. This could provide valuable insights and guidance for the MOF synthesis that we're about to undertake.

Task Choice 3: Before commencing with the MOF synthesis, it could be beneficial to carry out a thorough inventory check of the lab. This is to ensure that all necessary materials, reagents, and equipment are available and in good working order. Make a list of any missing or low stock items and plan for their procurement. We don't want to encounter the same situation as in the first stage, where the lack of specific reagents forced us to use alternatives.

**Figure S53.** Output generated by ChemNavigator during the 6$^{th}$ iteration, aimed at the synthesis and characterization of a novel aluminum-based MOF utilizing the BTB-H linker.

## Reticular ChemNavigator Output (MOF-521-H)

Current Stage and Iteration: 2-2

Status Evaluation: The apprentice's literature review has provided valuable insights regarding the optimal reaction conditions for the synthesis of aluminum MOFs using BTB or similar linkers. The results suggest a varied range for BTB:Al ratios, synthesis temperatures, reaction times, and the importance of modulators in the synthesis process. With this information, we can now design a series of synthetic trials to pinpoint the most efficient conditions for our MOF synthesis.

Task Choice 1: To begin, let's perform an initial MOF synthesis trial using the median values from the literature review. Utilize a BTB:Al ratio of 2:1, a synthesis temperature of 130°C, and a reaction time of 48 hours. Given the importance of modulators in the synthesis process, use 1 mL of formic acid for this trial. After synthesis, perform a PXRD analysis to assess the quality of the MOF produced.

Task Choice 2: An alternative approach would be to perform a series of small-scale synthesis trials covering the full range of conditions reported in the literature. This would involve varying the BTB:Al ratio, temperature, and reaction time across multiple trials. For each trial, perform a PXRD analysis to assess the quality of the MOF produced. This approach might be more time-consuming but could provide a comprehensive understanding of the optimal conditions for our MOF synthesis.

Task Choice 3: Another option would be to start with the conditions reported in the most recent literature. This would involve a BTB:Al ratio of 4:1, a synthesis temperature of 180°C, and a reaction time of 72 hours. Use 1.2 mL of formic acid as the modulator. After synthesis, perform a PXRD analysis to assess the MOF's structure. This could potentially provide a quick route to a successful synthesis, assuming recent literature represents the most up-to-date methods.

**Figure S54.** Output generated by ChemNavigator during the $7^{th}$ iteration, aimed at the synthesis and characterization of a novel aluminum-based MOF utilizing the BTB-H linker.

## Reticular ChemNavigator Output (MOF-521-H)

Current Stage and Iteration: 2-3

Status Evaluation: The wide array of conditions tested and the subsequent PXRD analysis has given us a strong understanding of the optimal synthesis conditions. The identified parameters indicate a balance between sufficient metal-linker interaction and necessary thermal energy for MOF formation.

Task Choice 1: Now that we have determined the optimal synthesis conditions, we should synthesize the MOF using these parameters. Start by combining AlCl3•6H2O and BTB in a vial in a 3:4 ratio. Then, add formic acid as a modulator and dissolve the contents in DMF. Heat the vial at 120°C or 140°C for 48 or 72 hours. After the synthesis, cool down the mixture to room temperature and collect the MOF by centrifugation. Wash the MOF several times with DMF and ethanol to remove any unreacted precursors. Finally, dry the MOF under vacuum.

Task Choice 2: An alternate task would be to repeat the small-scale synthesis trials but with different modulators. This could reveal the impact of different modulators on the MOF's quality. Repeat the synthesis trials with different modulators such as acetic acid, benzoic acid, or trifluoroacetic acid, keeping other conditions constant. After the synthesis, perform PXRD analysis to assess the quality of the MOF produced.

Task Choice 3: Another task option is to focus on a more detailed analysis of the PXRD data from the trials. This could give us further insights into the crystallographic structure of the MOFs formed under different conditions. Revisit the PXRD data collected from the trials and study the peak intensity, peak broadness, and peak positions. This data could offer insights into the crystallinity, the size of the crystallites, and the potential phase of the synthesized MOFs.

**Figure S55.** Output generated by ChemNavigator during the $8^{th}$ iteration, aimed at the synthesis and characterization of a novel aluminum-based MOF utilizing the BTB-H linker.

## Reticular ChemNavigator Output (MOF-521-H)

Current Stage and Iteration: 2-4

Status Evaluation: The observed decrease in peak intensity and broadening when using acetic acid or trifluoroacetic acid instead of formic acid suggests that these modulators may negatively impact the crystallinity of the MOF. Benzoic acid's altered pattern suggests it may influence the structure of the MOF. Given the apprentice's feedback, it seems prudent to proceed with formic acid to maintain the quality of the MOF.

Task Choice 1: Continue with the current modulator, formic acid, and further optimize the reaction conditions. Adjust one parameter at a time while keeping others constant to study its effect on the MOF. Begin with the reaction time, trying a range from 36 to 84 hours. Analyze each MOF produced using PXRD.

Task Choice 2: Considering the impact of modulators on the MOF formation, it could be interesting to try a mixed modulator approach. Use a combination of formic acid and acetic acid at various ratios while keeping other conditions constant. After synthesis, analyze the quality of MOF using PXRD.

Task Choice 3: To confirm the stability of the MOF under different synthetic conditions, perform a thermal stability test. Synthesize the MOF with the optimal conditions identified so far, and subject it to gradually increasing temperatures, observing any changes in the PXRD pattern. This would provide insights into the MOF's thermal stability and potential uses in temperature-sensitive applications.

**Figure S56.** Output generated by ChemNavigator during the $9^{th}$ iteration, aimed at the synthesis and characterization of a novel aluminum-based MOF utilizing the BTB-H linker.

## Reticular ChemNavigator Output (MOF-521-H)

Current Stage and Iteration: 2-5

Status Evaluation: The mixed modulator approach seems to have yielded promising results, particularly with a combination of formic acid and water. This suggests that the hydration of the reaction mixture might be beneficial in enhancing the crystallinity of our MOF, potentially due to the influence of water on the coordination and crystallization of aluminum-based MOFs.

Task Choice 1: Let's further investigate the effect of hydration on the reaction. Retain the BTB:Al ratio of 3:4, a reaction temperature of 120°C, and a reaction time of 72 hours. Vary the ratio of formic acid to water from 1:1 to 1:2, 1:3, and 1:4. After each synthesis, use PXRD to assess the crystallinity and purity of the MOF.

Task Choice 2: To better understand the impact of mixed modulators, continue the series of experiments using a combination of formic acid and another modulator, but this time use modulators that have a similar pKa to water, such as acetic acid or benzoic acid. Keep the BTB:Al ratio at 3:4, the reaction temperature at 120°C, and the reaction time at 72 hours. Use PXRD to analyze the synthesized MOFs.

Task Choice 3: As an alternative approach, consider studying the effect of temperature on the synthesis of MOFs with mixed modulators. Use the conditions that showed improved crystallinity (BTB:Al ratio 3:4, formic acid:water ratio 1:1, and reaction time 72 hours) and alter the reaction temperature to 100°C, 130°C, and 150°C. After the synthesis, use PXRD to evaluate the quality of the synthesized MOFs.

**Figure S57.** Output generated by ChemNavigator during the 10$^{th}$ iteration, aimed at the synthesis and characterization of a novel aluminum-based MOF utilizing the BTB-H linker.

Reticular ChemNavigator Output (MOF-521-H)

Current Stage and Iteration: 2-6

Status Evaluation: The presence of smaller crystals at a 4:1 Formic Acid:Water ratio suggests that the higher concentration of formic acid may speed up the nucleation process, leading to smaller crystal sizes. As all the ratios tested produced MOFs with similar crystallinity, it might be beneficial to explore other factors that could influence the MOF formation and quality.

Task Choice 1: Continue investigating the effect of modulators in the reaction. Keep the BTB:Al ratio at 3:4, the reaction temperature at 120°C, and the reaction time at 72 hours. This time, experiment with a mixture of formic acid and acetic acid, varying the ratios from 1:1 to 1:2, 1:3, and 1:4. After each synthesis, use PXRD to assess the crystallinity and purity of the MOF.

Task Choice 2: Let's study the effect of reaction time on the MOF formation. Retain the BTB:Al ratio of 3:4, the reaction temperature of 120°C, and the Formic Acid:Water ratio of 4:1. Vary the reaction time from 48 hours to 96 hours in increments of 12 hours. Perform PXRD analysis after each synthesis to evaluate the crystallinity and purity of the MOF.

Task Choice 3: Examine the influence of temperature on the synthesis of the MOF. Retain the BTB:Al ratio at 3:4, the reaction time at 72 hours, and the Formic Acid:Water ratio at 4:1. Experiment with temperatures ranging from 100°C to 160°C in increments of 20°C. After each synthesis, assess the crystallinity and purity of the MOF using PXRD.

**Figure S58.** Output generated by ChemNavigator during the 11$^{th}$ iteration, aimed at the synthesis and characterization of a novel aluminum-based MOF utilizing the BTB-H linker.

## Reticular ChemNavigator Output (MOF-521-H)

Current Stage and Iteration: 2-7

Status Evaluation: The similar PXRD patterns across different reaction times suggest that the crystal structure remains stable and unaltered even with extended reaction times. This indicates that the synthetic conditions are robust and do not need further adjustment in terms of reaction time. Given these findings, the next steps should focus on other aspects of the synthetic conditions or proceed towards the activation and porosity determination of the synthesized MOF.

Task Choice 1: Let's explore the effect of varying the reaction temperature. Maintain the BTB:Al ratio at 3:4, the Formic Acid:Water ratio at 4:1, and the reaction time at 72 hours. Vary the reaction temperature from 100°C to 140°C in increments of 10°C. After each synthesis, perform PXRD analysis to evaluate the crystallinity and purity of the MOF.

Task Choice 2: You may wish to study the effect of modulator concentration on the MOF formation. Maintain the BTB:Al ratio at 3:4, the reaction temperature at 120°C, and the reaction time at 72 hours. Vary the concentration of formic acid while maintaining the Formic Acid:Water ratio at 4:1. Perform PXRD analysis after each synthesis to evaluate the crystallinity and purity of the MOF.

Task Choice 3: If you believe that we have sufficiently optimized the synthetic conditions, you might want to declare, "I'm ready to move to the next stage." In the third stage, we will activate the synthesized MOF and determine its permanent porosity.

**Figure S59.** Output generated by ChemNavigator during the 12$^{th}$ iteration, aimed at the synthesis and characterization of a novel aluminum-based MOF utilizing the BTB-H linker.

## Reticular ChemNavigator Output (MOF-521-H)

Current Stage and Iteration: 2-8

Status Evaluation: The apprentice successfully varied the reaction temperature and identified that temperatures below 120°C are not conducive to MOF formation. The optimal temperature seems to be at or above 120°C, with 140°C yielding a slightly higher product. This indicates the importance of sufficient reaction temperature in the formation and crystallinity of the MOF.

Task Choice 1: Now that we've determined the optimal reaction temperature, let's conduct a final synthesis with the optimized conditions. Maintain the BTB:Al ratio at 3:4, the Formic Acid:Water ratio at 4:1, and the reaction time at 72 hours. Set the reaction temperature at 140°C. After synthesis, perform PXRD analysis to confirm the crystallinity and purity of the MOF. Compare the PXRD pattern with the previously recorded ones to ensure consistency.

Task Choice 2: In order to further optimize the yield, let's explore the reaction temperature in finer increments. Keep the BTB:Al ratio at 3:4, the Formic Acid:Water ratio at 4:1, and the reaction time at 72 hours. Vary the reaction temperature from 130°C to 150°C in increments of 5°C. After each synthesis, perform PXRD analysis to evaluate the crystallinity and purity of the MOF. Note any changes in yield.

Task Choice 3: Given the success in optimizing the reaction conditions, we could prepare to move to the next stage. If you feel that we've sufficiently optimized the synthesis conditions and are ready to proceed, please state, "I'm ready to move to the next stage." We will then focus on the activation and determination of the MOF's permanent porosity.

**Figure S60.** Output generated by ChemNavigator during the 13th iteration, aimed at the synthesis and characterization of a novel aluminum-based MOF utilizing the BTB-H linker.

Reticular ChemNavigator Output (MOF-521-H)

Current Stage and Iteration: 3-1

Status Evaluation: The apprentice's readiness to move forward suggests satisfaction with the optimized synthesis conditions, which have been thoroughly explored over the course of multiple experiments. This readiness marks the successful completion of the second stage of the project and allows us to proceed to the third stage: activation and determination of the MOF's permanent porosity.

Task Choice 1: Begin by activating the MOF. This step involves removing the solvent molecules from the MOF structure to reveal its permanent porosity. Start by immersing the MOF crystals in methanol for 24 hours, then drain and replace the solvent. Repeat this process twice more to ensure all the solvent is removed. After the solvent exchange, place the MOF under vacuum at 100°C for 24 hours to remove any residual solvent. Remember to handle the MOF with care to avoid damaging the crystal structure.

Task Choice 2: Alternatively, you could activate the MOF using supercritical CO2. This method can preserve the MOF structure better than conventional activation methods and may result in higher porosity. Start by immersing the MOF in methanol for 24 hours, then drain and replace the solvent. Repeat this process twice more. After this, transfer the MOF to a supercritical CO2 dryer and follow the machine's instructions for activation.

Task Choice 3: If you're interested in exploring different activation methods, you could try thermal activation. This involves heating the MOF in an oven at a specific temperature to remove the solvents. Start by drying the MOF at 60°C under vacuum to remove any free solvent. Then, incrementally increase the temperature by 10°C every hour until you reach 300°C. Hold the MOF at this temperature for 3 hours, then let it cool naturally to room temperature. Be careful, as this method may cause some MOF structures to collapse.

**Figure S61.** Output generated by ChemNavigator during the 14$^{th}$ iteration, aimed at the synthesis and characterization of a novel aluminum-based MOF utilizing the BTB-H linker.

Reticular ChemNavigator Output (MOF-521-H)

Current Stage and Iteration: 3-2

Status Evaluation: The lack of significant change in the MOF sample post-activation suggests that the procedure was successful in preserving the MOF structure while removing solvent molecules, preparing it for porosity measurements. Given this, the next steps should involve characterization of the MOF's permanent porosity.

Task Choice 1: Start with gas sorption analysis to assess the porosity of the activated MOF. Begin by degassing the sample at 100°C for 12 hours to remove any residual gases. Once the sample is ready, perform nitrogen gas sorption at 77 K using a gas sorption analyzer. Collect the adsorption and desorption isotherms and calculate the total pore volume and surface area using the BET method.

Task Choice 2: Perform mercury intrusion porosimetry to determine the pore size distribution of the activated MOF. Start by degassing the MOF sample at 100°C for 12 hours. Then, conduct the mercury intrusion porosimetry and record the volume of mercury intruded as a function of applied pressure. After the experiment, analyze the data to determine the pore size distribution.

Task Choice 3: Use water vapor sorption to determine the hydrophilic nature of the MOF. Begin by degassing the MOF at 100°C for 12 hours to remove any residual water. Then, perform water vapor sorption at different relative humidities (RH) and record the weight changes. Plot the data as a function of RH to obtain the water vapor sorption isotherm, which can provide insights into the hydrophilic nature of the MOF.

**Figure S62.** Output generated by ChemNavigator during the 15$^{th}$ iteration, aimed at the synthesis and characterization of a novel aluminum-based MOF utilizing the BTB-H linker.

## Reticular ChemNavigator Output (MOF-521-H)

Current Stage and Iteration: 3-3

Status Evaluation: The apprentice has successfully performed the gas sorption analysis, but they are currently facing difficulty in interpreting the isotherm type and calculating the surface area and pore volume using the BET method. This suggests a need for further instruction in the interpretation of isotherm data and application of the BET method.

Task Choice 1: Assist the apprentice in interpreting the isotherm type. Begin by explaining that adsorption isotherms can be categorized into several types (I-V) according to the IUPAC classification. Type I and Type II isotherms are typically observed in porous materials like MOFs. A type I isotherm indicates microporous material, with adsorption sharply increasing at low pressures, while a type II isotherm, usually observed in non-porous or macroporous materials, shows a more gradual increase in adsorption with increasing pressure. Have the apprentice compare the observed isotherm with these characteristics to identify the isotherm type.

Task Choice 2: Guide the apprentice on how to calculate the surface area and pore volume using the ASAP 2420. Start by explaining that the Brunauer-Emmett-Teller (BET) method is commonly used to calculate the surface area from nitrogen gas sorption data. Explain the basics of the BET theory and how it is applied in ASAP 2420 software. The software should be able to calculate the surface area automatically once the isotherm data is input. Similarly, the pore volume can be determined by the software from the amount of gas adsorbed at a relative pressure close to 1.

Task Choice 3: If the apprentice is comfortable with the isotherm interpretation and BET calculation, suggest conducting a comparative gas sorption analysis using a different gas, such as CO2 or argon. This will provide additional data on the MOF's adsorptive properties and pore structure. Start by degassing the sample at an appropriate temperature for the chosen gas, followed by sorption analysis at an appropriate temperature (for example, CO2 sorption is often performed at 273 K). Collect the isotherm data and use the BET method to calculate surface area and pore volume as before.

**Figure S63.** Output generated by ChemNavigator during the 16th iteration, aimed at the synthesis and characterization of a novel aluminum-based MOF utilizing the BTB-H linker.

Reticular ChemNavigator Output (MOF-521-H)

Current Stage and Iteration: 3-4

Status Evaluation: The apprentice's observation of a Type I isotherm, characterized by a sharp increase in adsorption at low pressures and a plateau at high pressures, confirms the microporous nature of the MOF. The absence of a hysteresis loop between the adsorption and desorption branches further supports this classification. This is a crucial step in the characterization of the MOF's porosity.

Task Choice 1: Now that we have confirmed the isotherm type, the next logical step is to calculate the specific surface area and pore volume of the MOF using the BET method. For this, the linear form of the BET equation is commonly used. Plot the quantity of gas adsorbed divided by the quantity of gas not adsorbed (P/(V(P0-P))) on the y-axis against P/P0 on the x-axis. The slope and intercept of the resulting line will allow you to calculate the specific surface area and pore volume.

Task Choice 2: To confirm the microporous nature of our MOF, you could perform a pore size distribution analysis using the Horvath-Kawazoe (HK) or Barrett-Joyner-Halenda (BJH) methods. These methods are particularly suitable for microporous and mesoporous materials, respectively. The HK method could provide further insights into the pore sizes in our MOF. Keep in mind, this method requires careful interpretation as it assumes cylindrical pore geometry.

Task Choice 3: If you have any doubts about the isotherm interpretation or BET calculations, we can revisit them. This step is crucial as it affects the subsequent stages of our project. Please let me know if you need further clarification on interpreting isotherm types or performing calculations with the BET method.

**Figure S64.** Output generated by ChemNavigator during the 17$^{th}$ iteration, aimed at the synthesis and characterization of a novel aluminum-based MOF utilizing the BTB-H linker.

Reticular ChemNavigator Output (MOF-521-H)

Current Stage and Iteration: 3-5

Status Evaluation: The apprentice successfully carried out the BET analysis on our aluminum MOF, using the data from the nitrogen gas sorption experiment. The BET theory was appropriately applied within the selected relative pressure (P/P0) range, resulting in a specific surface area and total pore volume that falls within the expected range for a MOF of this nature. As a result, the BET analysis was successfully completed, and the microporous nature of our MOF was confirmed.

Task Choice 1: Given the confirmed microporous nature of the MOF and the completion of the BET analysis, we can now move forward to characterizing the adsorption hysteresis of our MOF. This process involves carrying out a full desorption cycle. After completing the nitrogen gas adsorption, increase the temperature of the sample to allow for desorption of the gas. Record the quantity of gas desorbed at each pressure point and plot it against P/P0 to obtain the desorption isotherm. Compare this isotherm to the adsorption isotherm to analyze the adsorption hysteresis of our MOF.

Task Choice 2: An alternative approach would be to perform a temperature dependence study of the gas sorption. This would provide additional information about the adsorption energetics of our MOF. To do this, repeat the nitrogen gas sorption experiment at different temperatures, e.g., 87 K and 90 K, and record the isotherms. Plot these isotherms together with the previously recorded isotherm at 77 K to observe any changes with temperature.

Task Choice 3: Now that we have a clear understanding of the porosity and surface area of our MOF, we could also consider moving to the next stage of the project: Detailed Structural Analysis and Characterization of the MOF for chemical composition, phase purity, and chemical and thermal stability. This would involve performing various characterization techniques such as XRD, SEM/TEM, TGA, and FTIR. If you believe we have collected sufficient data in this stage, state "I'm ready to move to the next stage".

**Figure S65.** Output generated by ChemNavigator during the 18$^{th}$ iteration, aimed at the synthesis and characterization of a novel aluminum-based MOF utilizing the BTB-H linker.

## Reticular ChemNavigator Output (MOF-521-H)

Current Stage and Iteration: 3-6

Status Evaluation: The successful characterization of the adsorption hysteresis and the observed minimal difference between the adsorption and desorption isotherms indicate the stability of the porous structure of the MOF under nitrogen gas sorption and desorption. This positive outcome suggests that we are nearing the end of the current stage.

Task Choice 1: Given the positive outcomes in terms of the permanent porosity, now proceed to test the chemical stability of the MOF in different solvent systems. Begin with water and ethanol separately, by immersing the MOF sample in these solvents and keeping them there for 48 hours at room temperature. Then remove the samples and let them dry at room temperature. Finally, conduct PXRD on the dried samples to examine if any change in the crystalline structure has occurred.

Task Choice 2: To evaluate the thermal stability of the MOF, perform a Thermogravimetric Analysis (TGA). Weigh a sample of the MOF and place it in the TGA. Heat the sample from room temperature to 600°C at a rate of 10°C/min under an inert atmosphere. Record the weight change as a function of temperature and use this data to determine the thermal stability of our MOF.

Task Choice 3: Since we have completed most of the critical checks for this stage, if you are confident in the results, consider confirming readiness to move to the next stage by stating, "I'm ready to move to the next stage." This would transition us into detailed structural analysis and characterization of the MOF for chemical composition, phase purity, and chemical and thermal stability.

**Figure S66.** Output generated by ChemNavigator during the 19$^{th}$ iteration, aimed at the synthesis and characterization of a novel aluminum-based MOF utilizing the BTB-H linker.

## Reticular ChemNavigator Output (MOF-521-H)

Current Stage and Iteration: 4-1

Status Evaluation: The readiness to move to the next stage reflects the successful completion of the preceding stages, including synthesis of the organic linker, screening and optimization of the MOF synthesis, and assessment of its porosity. As the structural stability of our MOF has been confirmed through adsorption-desorption isotherm analysis, we can now proceed to the detailed structural and compositional characterization.

Task Choice 1: Start the detailed structural analysis of the MOF using X-ray diffraction (XRD). This will provide data on the phase purity of our MOF. Carefully mount the MOF sample onto the holder, ensuring no other compounds are present. Then, configure the XRD equipment and start the data collection. Make sure the diffraction angles cover the range from 5° to 90°, and maintain a consistent scan speed.

Task Choice 2: Analyze the chemical composition of the MOF using Energy Dispersive X-ray Spectroscopy (EDS). For this, mount your MOF sample in the SEM (Scanning Electron Microscope), then configure the EDS to acquire a spectrum. Look for peaks corresponding to aluminum, carbon, and oxygen to confirm the elemental composition of your MOF.

Task Choice 3: Begin a Thermal Gravimetric Analysis (TGA) to evaluate the thermal stability of the MOF. To do this, first, accurately weigh about 5-10 mg of the MOF and place it in a TGA pan. Program the TGA to heat the sample from room temperature to 800°C at a rate of 10°C/min under a nitrogen atmosphere. Plot the weight loss as a function of temperature to analyze the thermal stability of the MOF.

**Figure S67.** Output generated by ChemNavigator during the 20$^{th}$ iteration, aimed at the synthesis and characterization of a novel aluminum-based MOF utilizing the BTB-H linker.

Reticular ChemNavigator Output (MOF-521-H)

Current Stage and Iteration: 4-2

Status Evaluation: The XRD analysis shows several distinct peaks, indicating the presence of different crystalline phases in the MOF. However, the lack of expected peaks for the MOF suggests that we need to compare our results with literature data or predictive models for similar MOFs. Our next tasks will focus on further characterization techniques to gain more insight into the structure and stability of our MOF.

Task Choice 1: Perform a Fourier Transform Infrared (FTIR) spectroscopy analysis on the MOF sample. This will allow us to identify functional groups present in the MOF structure. First, prepare your sample by grinding it into a fine powder. Then, make a pellet using KBr and the ground MOF sample. Set up the FTIR spectrometer, load your sample and collect the spectrum in the range of 4000 to 400 cm^-1.

Task Choice 2: Perform thermogravimetric analysis (TGA) to study the thermal stability of the MOF. Weight the sample accurately and place it in the TGA instrument. Run the instrument from room temperature to 800 °C at a heating rate of 10 °C/min under a nitrogen atmosphere. Record the weight loss at different temperatures which will provide information on the stability of the MOF under different thermal conditions.

Task Choice 3: Compare the XRD pattern obtained from your MOF with literature data or predictive models for similar MOFs. Start by searching for academic articles or databases that provide XRD data for similar MOFs. Compare the peak positions and their relative intensities. If similarities are found, it could verify the phase purity of our MOF.

**Figure S68.** Output generated by ChemNavigator during the 21$^{st}$ iteration, aimed at the synthesis and characterization of a novel aluminum-based MOF utilizing the BTB-H linker.

Reticular ChemNavigator Output (MOF-521-H)

Current Stage and Iteration: 4-3

Status Evaluation: The lack of match between our MOF's XRD pattern and literature data underscores the potential novelty of our MOF. It also underlines the need for a more comprehensive analysis, like SCXRD, to characterize its structure and evaluate its phase purity.

Task Choice 1: Proceed with a single crystal X-ray diffraction (SCXRD) study. First, prepare a high-quality, single crystal of the MOF-521-H. Make sure it is free from cracks or inclusions, and small enough to fit on the tip of a glass fiber. Next, mount the crystal on the tip of a glass fiber and center it in the X-ray beam. You will need to rotate the crystal in the beam to collect a complete set of diffraction data. After data collection, process the data using suitable software to solve and refine the structure of the MOF.

Task Choice 2: Assess the MOF's thermal stability using thermogravimetric analysis (TGA). First, prepare a small sample of MOF-521-H (typically a few milligrams). Then place the sample in a TGA instrument, which will heat the sample at a controlled rate while continuously measuring its mass. This will allow us to determine the temperatures at which the MOF starts to degrade or lose incorporated solvent molecules, helping us evaluate its thermal stability.

Task Choice 3: Investigate the MOF's chemical stability. First, prepare several MOF-521-H samples and expose them to various conditions including air, moisture, a range of pH solutions, and common organic solvents. After exposure, analyze the samples using PXRD to check if they retain their crystal structure, which would suggest that our MOF is chemically stable under these conditions.

**Figure S69.** Output generated by ChemNavigator during the 22$^{nd}$ iteration, aimed at the synthesis and characterization of a novel aluminum-based MOF utilizing the BTB-H linker.

Reticular ChemNavigator Output (MOF-521-H)

Current Stage and Iteration: 4-4

Status Evaluation: The apprentice has successfully carried out the single-crystal X-ray diffraction (SCXRD) and uncovered important structural details of our sample MOF-521-H. This result, combined with the previous PXRD findings, supports the hypothesis that our MOF is a new compound, not a known MOF-520. The successful SCXRD experiment confirms that the current stage is progressing well and now we need to examine the chemical and thermal stability of the MOF.

Task Choice 1: To confirm the chemical stability of our MOF-521-H, perform an immersion test. First, prepare a solution of a strong organic solvent, such as DMF, in a test tube. Carefully place a sample of MOF-521-H in the test tube. Cover the test tube and keep it in an environment with a constant temperature (room temperature is fine) for 48 hours. After this period, retrieve the MOF sample, dry it, and observe if there are any visible changes in its appearance. You should also run a PXRD on the retrieved sample to check if the XRD pattern has changed. This will indicate any structural change due to solvent interaction.

Task Choice 2: For the thermal stability test, prepare the MOF-521-H sample for Thermogravimetric Analysis (TGA). Place a few milligrams of the sample in the TGA pan and slowly heat the sample in the TGA instrument from room temperature to about 800°C at a rate of 10°C per minute under a nitrogen atmosphere. This slow heating rate ensures that any mass loss due to the decomposition of the MOF is captured accurately. Observe the TGA curve to identify the temperature at which the MOF starts to lose its mass. This temperature will indicate the thermal stability of the MOF.

Task Choice 3: To further understand the chemical composition of our MOF, consider running an Elemental Analysis (EA). This will provide us with the elemental composition (C, H, N, O) of the MOF. First, prepare the sample by grinding it into a fine powder. Next, load this powder into the sample boat of the EA instrument. Run the EA program and analyze the resulting data. This will provide the percentage composition of each element present in our MOF, allowing us to validate if it matches the expected composition based on our structural analysis from the SCXRD.

**Figure S70.** Output generated by ChemNavigator during the 23rd iteration, aimed at the synthesis and characterization of a novel aluminum-based MOF utilizing the BTB-H linker.

## Reticular ChemNavigator Output (MOF-521-H)

Current Stage and Iteration: 4-5

Status Evaluation: The TGA results showed the thermal stability of MOF-521-H to be around 300°C, with a significant drop in mass at this point suggesting the initiation of decomposition. This information is crucial as it establishes the temperature limit for practical applications of our novel MOF. Additionally, the secondary drop in mass at 630°C indicates a second phase of decomposition, which may be related to the breakdown of different structural components of the MOF.

Task Choice 1: To better understand the decomposition process, conduct a Differential Scanning Calorimetry (DSC) analysis in tandem with TGA. Prepare a sample for DSC in a similar manner to the TGA sample. Load the sample into the DSC instrument, and heat it under the same conditions as previously used for the TGA (room temperature to 800°C, heating rate of 5°C per minute, under nitrogen). The DSC analysis will provide insight into the energy changes during the decomposition of the MOF-521-H, which might reveal more about the nature of the structural components breaking down at each step.

Task Choice 2: Proceed with chemical stability testing. Immersing the MOF-521-H sample in various solvents such as water, ethanol, and dimethylformamide (DMF) at room temperature will help ascertain its stability in these common solvents. Prepare a suspension of MOF-521-H in each solvent and let it sit for 24 hours. After this period, collect the MOF by centrifugation and allow it to dry. Then perform a PXRD analysis to compare the PXRD pattern before and after the solvent immersion. Any changes in the pattern will indicate chemical instability.

Task Choice 3: To further confirm the structural information of the MOF-521-H, you could conduct a Fourier Transform Infrared Spectroscopy (FTIR) analysis. By comparing the FTIR spectrum of the MOF with that of the BTB linker, you can confirm the presence of the linker in the MOF structure. For this, you will need to prepare a KBr pellet of your sample, load it into the FTIR instrument, and collect the spectrum. Analyze the spectrum and look for characteristic BTB peaks to validate the presence of the BTB linker in the MOF-521-H structure.

**Figure S71.** Output generated by ChemNavigator during the $24^{th}$ iteration, aimed at the synthesis and characterization of a novel aluminum-based MOF utilizing the BTB-H linker.

Reticular ChemNavigator Output (MOF-521-H)

Current Stage and Iteration: 4-6

Status Evaluation: The chemical stability test results provide valuable insights into the potential applications and handling of our MOF. The change in PXRD pattern after water exposure might be due to partial disassembly or the presence of water-sensitive components within the MOF, suggesting it might not be suitable for water environments or water filtration applications. On the other hand, the stability in organic solvents expands the potential use of the MOF in such environments.

Task Choice 1: Perform a detailed elemental analysis of MOF-521-H to confirm its chemical composition. For this, you can use techniques like Energy-Dispersive X-ray spectroscopy (EDX). Place a small sample of the MOF on the stage of the EDX instrument, then run a point scan. This will produce an EDX spectrum which will indicate the elements present and their relative amounts. Please interpret the data and compare it to the expected composition of the MOF.

Task Choice 2: Further investigate the instability of MOF-521-H in water. To do this, adjust the pH of the water before immersing the MOF-521-H. Try a range of pH values from acidic (around 3), neutral (around 7), and basic (around 10). After each immersion, conduct PXRD analysis as before to assess any changes. This investigation will help understand whether the MOF's instability in water is due to a specific pH sensitivity.

Task Choice 3: Let's further examine the porosity of MOF-521-H under real conditions by performing an adsorption study. We can test with small organic molecules such as carbon dioxide, methane or nitrogen. Set up a volumetric adsorption system with a vacuum pump, a volumetric flask, and a temperature-controlled chamber. Place the MOF sample into the flask and apply vacuum to remove any residual gases. Introduce the test gas and record the pressure change over time as the gas is adsorbed. Repeat the process at various temperatures. The results will provide insight into the gas storage and separation potential of MOF-521-H.

**Figure S72.** Output generated by ChemNavigator during the 25$^{th}$ iteration, aimed at the synthesis and characterization of a novel aluminum-based MOF utilizing the BTB-H linker.

Reticular ChemNavigator Output (MOF-521-H)

Current Stage and Iteration: 4-7

Status Evaluation: The instability in water at various pH levels may be due to the MOF's sensitivity to pH or the presence of formate in the MOF structure. The suggestion of an H NMR test to analyze for formate signifies that we might be dealing with a complex issue involving both pH and the structure of the MOF.

Task Choice 1: Let's carry out an 1H NMR analysis to confirm if the formate is contributing to the instability in water. Dissolve a small sample of the MOF in an appropriate deuterated solvent that is compatible with both the MOF and formate ion, such as DMSO-d6. Analyze the 1H NMR spectrum of the solution and look specifically for the formate peak at ~8.5 ppm. If formate is present, it suggests that it might be contributing to the water instability.

Task Choice 2: Repeat the water stability test using water of neutral pH but with different salt concentrations. Dissolve a series of salts (like NaCl, KCl, and MgCl2) in distilled water to prepare solutions of different ionic strengths. Immerse MOF samples in these solutions, dry them, and conduct PXRD analysis. This will help us understand whether the MOF instability in water is influenced by ionic strength or specific ions.

Task Choice 3: Since the peak intensity reduction in PXRD suggests a possible loss of crystallinity, you could also perform a Scanning Electron Microscopy (SEM) analysis to check for any morphological changes. Prepare the MOF sample for SEM, obtain images at several magnifications and analyze for changes in crystal shape, size or other anomalies. This could provide more insight into the effect of water on our MOF.

**Figure S73.** Output generated by ChemNavigator during the 26$^{th}$ iteration, aimed at the synthesis and characterization of a novel aluminum-based MOF utilizing the BTB-H linker.

## Reticular ChemNavigator Output (MOF-521-H)

Current Stage and Iteration: 4-8

Status Evaluation: The MOF's inability to dissolve in DMSO-d6 may be due to its inherent stability or the solvent's unsuitability. Adjustments are required to understand the cause of the water instability.

Task Choice 1: Let's try to dissolve the MOF-521-H in deuterated water (D2O) for 1H NMR analysis. D2O is a common solvent used in NMR and could help in our case. Take a small sample of MOF-521-H and try to dissolve it in D2O. If the MOF dissolves, proceed with the 1H NMR analysis, specifically looking for a formate peak at ~8.5 ppm.

Task Choice 2: Use a mixture of D2O and sodium deuteroxide (NaOD) for dissolution of the MOF and subsequent 1H NMR analysis. NaOD can act as a base and might assist in breaking down the MOF. Prepare a solution of NaOD in D2O (concentration depending on the stability of your MOF), then try to dissolve the MOF in this solution. If successful, conduct the 1H NMR and look for the formate peak.

Task Choice 3: Revisit thermogravimetric analysis (TGA) or Fourier Transform Infrared Spectroscopy (FTIR) analysis as alternatives. Both can provide insights about the presence of formate or other organic compounds in the MOF, potentially explaining the water instability.

**Figure S74.** Output generated by ChemNavigator during the 27$^{th}$ iteration, aimed at the synthesis and characterization of a novel aluminum-based MOF utilizing the BTB-H linker.

## Reticular ChemNavigator Output (MOF-521-H)

Current Stage and Iteration: 4-9

Status Evaluation: The partial solubility of MOF-521-H in D2O, despite the use of heat and sonication, indicates that this MOF might have some structural characteristics that make it resistant to complete dissolution. It might be necessary to introduce stronger solvents or conditions to achieve full solubility. In the upcoming tasks, we need to focus on achieving this so that we can complete the NMR analysis.

Task Choice 1: To increase the chance of dissolving MOF-521-H for 1H NMR analysis, we could try using a mixture of D2O and NaOD. NaOD could aid in the dissolution process due to its deprotonating property. Take a small sample of MOF-521-H and add it to a vial containing D2O. Add a few drops of NaOD and gently heat the mixture. If the MOF dissolves, proceed with the 1H NMR analysis, specifically looking for a formate peak at ~8.5 ppm.

Task Choice 2: A different approach would be to perform a 13C NMR analysis, as formate can also be detected in 13C NMR. However, 13C NMR requires a different set of solvents and standards. Use CDCl3 as the solvent, with TMS as an internal standard. Add a small sample of the MOF to a vial containing CDCl3, heat gently and sonicate. If it dissolves, run the 13C NMR and look for a formate peak at ~160-170 ppm.

**Figure S75.** Output generated by ChemNavigator during the 28th iteration, aimed at the synthesis and characterization of a novel aluminum-based MOF utilizing the BTB-H linker.

Reticular ChemNavigator Output (MOF-521-H)

Current Stage and Iteration: 4-10

Status Evaluation: The success in dissolving MOF-521-H in a D2O/NaOD mixture and the subsequent 1H NMR analysis has revealed the presence of formate groups in the MOF. This finding, along with the suggestion that formate groups may contribute to water instability, shapes our next tasks. Confirming the presence of formate groups with CHNS elemental analysis and understanding their role in water instability becomes the focus. Based on this information, the subsequent tasks will be oriented towards validating these findings and exploring ways to improve the water stability of MOF-521-H.

Task Choice 1: Conduct CHNS elemental analysis on MOF-521-H to confirm the presence of formate groups. Weigh a sample of the MOF, combust it in the CHNS analyzer, and measure the resulting percentages of carbon, hydrogen, nitrogen, and sulfur. The presence of formate groups would be suggested by a relatively high percentage of carbon and hydrogen compared to nitrogen and sulfur.

Task Choice 2: Repeat the successful 1H NMR experiments using the D2O/NaOD mixture, but also perform CHNS elemental analysis on samples before and after dissolution. A change in CHNS results after dissolution would further indicate that the formate groups are being affected during the process.

Task Choice 3: Investigate the pH behavior of MOF-521-H in a water solution before and after exposure to the D2O/NaOD mixture. This could provide more insight into whether the formate groups influence the pH stability of the MOF in water, thereby supporting our hypothesis about the water instability cause.

**Figure S76.** Output generated by ChemNavigator during the 29th iteration, aimed at the synthesis and characterization of a novel aluminum-based MOF utilizing the BTB-H linker.

## Reticular ChemNavigator Output (MOF-521-H)

Current Stage and Iteration: 4-11

Status Evaluation: The results of the CHNS analysis align closely with the calculated values for MOF-521-H containing formate groups. These groups are suspected to be a key factor in the water instability of our MOF. The next logical step would be to understand this instability better. However, considering we've gained significant insights into the characteristics of MOF-521-H, we may also consider proceeding to the reproducibility check and final validation stage.

Task Choice 1: Let's conduct pH-dependent stability tests on MOF-521-H. Prepare MOF samples and expose them to aqueous solutions at pH 3, 7, and 11 for a set duration. Then, analyze the stability and integrity of the MOF structure under these conditions using PXRD.

Task Choice 2: Conduct a titration study to identify the pKa of the formate groups in MOF-521-H. Suspend the MOF in an aqueous solution and gradually add a base (e.g., NaOH), while monitoring the pH change. Determine the pKa of the formate groups from this data. If the pKa is low, it may suggest that the formate groups are releasing protons into the solution and causing the water instability.

Task Choice 3: Given the considerable knowledge we've gained on the properties and stability of MOF-521-H, we might consider moving on to Stage 5. This would involve a reproducibility check and final validation of our MOF's synthesis and properties. If you agree, state "I'm ready to move to the next stage."

**Figure S77.** Output generated by ChemNavigator during the 30th iteration, aimed at the synthesis and characterization of a novel aluminum-based MOF utilizing the BTB-H linker.

Reticular ChemNavigator Output (MOF-521-H)

Current Stage and Iteration: 5-1

Status Evaluation: The human apprentice has acknowledged readiness to move to Stage 5. This decision was informed by the comprehensive understanding we have developed about MOF-521-H's properties and stability.

Task Choice 1: Let's commence Stage 5 with a reproducibility check. This involves repeating the MOF synthesis process using the optimal parameters established in Stage 2. Please follow the same reaction conditions (BTB:Al ratio of 3:4 and reaction temperature of 140°C) and use formic acid as a modulator. After the synthesis, perform PXRD to confirm the structural integrity of the MOF. This task is crucial to ensure that our synthesis protocol is robust and can consistently produce the same MOF with identical properties.

Task Choice 2: Alternatively, you could start with a small-scale synthesis to save resources. Employ the same synthesis protocol established earlier. Once the synthesis is complete, perform a quick PXRD check to verify whether the expected MOF structure is obtained. This method offers a quick and resource-efficient way to confirm the reproducibility of our synthesis process.

Task Choice 3: A third approach could be to start with a parallel synthesis, where you synthesize the MOF in multiple batches simultaneously under the same conditions. This task will give us a clearer idea about the consistency of the synthesis process. After the synthesis, perform PXRD on all the batches to confirm their structural identity. This approach allows us to test the reproducibility under slightly varying conditions (e.g., minor temperature or concentration variations across different reactions) and thus provide a more rigorous reproducibility check.

**Figure S78.** Output generated by ChemNavigator during the 31$^{st}$ iteration, aimed at the synthesis and characterization of a novel aluminum-based MOF utilizing the BTB-H linker.

Reticular ChemNavigator Output (MOF-521-H)

Current Stage and Iteration: 5-2

Status Evaluation: The successful reproduction of MOF-521-H, using the optimized synthesis conditions, suggests that our protocol is reliable. The congruent PXRD pattern also demonstrates that the MOF's structural properties remain consistent.

Task Choice 1: Let's proceed to conduct additional characterization checks. Perform a Brunauer-Emmett-Teller (BET) analysis to verify the MOF's porosity. Following that, carry out Thermogravimetric Analysis (TGA) to confirm its thermal stability. Additionally, conduct a chemical stability test by immersing the MOF in various solvents. If all these tests yield results consistent with those from Stage 4, it will further validate the reproducibility of our MOF synthesis.

Task Choice 2: To gain insights into the influence of pH on the MOF's stability, conduct a pH-dependent stability test. Prepare solutions with varying pH levels (from 1-14), expose MOF-521-H to each for a set duration, and then analyze the structural integrity using PXRD. This will help us understand how pH impacts the stability and the potential applications of MOF-521-H.

Task Choice 3: With the MOF reproducibly synthesized and characterized, consider conducting a few functional tests. One approach could be to test its adsorption capacities with various gases (such as CO2, CH4, and N2). These tests will give us an idea about the potential practical applications of MOF-521-H. You can perform these tests by exposing the MOF to the respective gases under standard conditions and then measuring the amount of gas adsorbed using gas chromatography or mass spectrometry.

**Figure S79.** Output generated by ChemNavigator during the 32nd iteration, aimed at the synthesis and characterization of a novel aluminum-based MOF utilizing the BTB-H linker.

Reticular ChemNavigator Output (MOF-521-H)

Current Stage and Iteration: 5-3

Status Evaluation: The congruent results between the initial and repeated characterization tests indicate the consistency of the MOF synthesis protocol. This is crucial as it attests to the reliability of MOF-521-H's properties such as its high porosity, thermal stability, and chemical stability. Given the reproducibility of our outcomes, we can move forward to finalize the validation of our MOF.

Task Choice 1: To close out our validation, we should document the entire synthesis protocol in detail. This should include the specific amounts of BTB and Aluminum precursors, the reaction conditions, and all post-synthesis treatments such as activation and washing. Once you've done this, go through the document and verify that each step matches exactly what you did in practice. Confirming the accuracy of our written procedure is critical to ensuring its repeatability by other researchers in the future.

Task Choice 2: For a comprehensive validation, it would be prudent to test the MOF under different conditions. First, expose the MOF to different temperatures and conduct PXRD analysis to evaluate its thermal stability in detail. Second, test the MOF's chemical stability against a wider range of solvents, including those with different pH levels, polarities, and reactivities. Then, analyze the MOF using PXRD to determine if its structure has been affected. This will provide us with a more thorough understanding of the MOF's stability under varied conditions.

Task Choice 3: If you feel we have completed all necessary validation checks and are satisfied with the reproducibility and stability of MOF-521-H, you can declare, "I'm ready to move to the next stage." At this point, we would consider this project concluded and start planning for the next steps, such as investigating potential applications for MOF-521-H or exploring modifications to its structure to tune its properties.

**Figure S80.** Output generated by ChemNavigator during the 33$^{rd}$ iteration, aimed at the synthesis and characterization of a novel aluminum-based MOF utilizing the BTB-H linker.

Reticular ChemNavigator Output (MOF-521-H)

Current Stage and Iteration: 5-4

Status Evaluation: The detailed protocol provided by the apprentice indicates an accurate and comprehensive understanding of the synthesis procedure for MOF-521-H. This step is crucial in ensuring reproducibility and consistency in the production of the MOF across different iterations and by different researchers. Given the completion of this protocol and its validation, it appears we have concluded Stage 5.

Task Choice 1: Conduct a peer review of the synthesized MOF-521-H synthesis protocol. Invite a colleague to review your process, looking for clarity, completeness, and any potential areas for improvement. This will serve as an external validation of your procedure. Ensure the reviewer understands the safety measures, the specificities of the synthesis steps, and the characterization methods.

Task Choice 2: In order to add more value to our work, it would be interesting to study potential applications of MOF-521-H. Initiate a literature survey to understand what applications MOF-521-H could be relevant for, based on its structural properties, porosity, stability, and other features. Applications may include gas storage, catalysis, drug delivery, and more.

Task Choice 3: Now that you've documented the synthesis protocol of MOF-521-H, it might be the right time to prepare a draft manuscript to report our findings. The manuscript should cover all stages of our research, from the synthesis of the organic linker to the final validation of the MOF's reproducibility. Ensure to include a detailed methods section, results and discussion, and a conclusion highlighting the significance of our findings.

**Figure S81.** Output generated by ChemNavigator during the 34$^{th}$ iteration, aimed at the synthesis and characterization of a novel aluminum-based MOF utilizing the BTB-H linker.

## Final Summary (MOF-521-*o*F)

Our research project aimed at synthesizing a novel Aluminum Metal Organic Framework (MOF), termed MOF-521-*o*F, using the 4-[3,5-Bis(4-carboxy-3-fluorophenyl)phenyl]-2-fluorobenzoic acid (BTB-*o*F) linker, and has seen significant progress over the course of five defined stages.

In the first stage, we concentrated on the synthesis of the BTB-*o*F linker. The apprentice demonstrated meticulous attention to detail in adhering to the prescribed procedure from the literature, resulting in a successful synthesis. Validation of the linker was performed using 1H NMR spectroscopy, which yielded chemical shifts. By comparing these experimental shifts with the predicted ones using ChemDraw's NMR prediction tool, a high degree of correlation was identified, including the correct number of hydrogens determined post-integration. These findings confirmed the successful synthesis of the BTB-*o*F linker.

In the second stage, we shifted focus to the high-throughput screening of the MOF and optimization of synthetic conditions. Initial trials didn't yield the anticipated MOF structure, necessitating adjustments of BTB:Al and formic acid to water ratios. The optimal synthetic conditions emerged as a BTB:Al ratio of 3:4, a formic acid to water ratio of 4:1, and a reaction temperature of 120°C. However, a trade-*o*Ff was observed between temperature and crystallinity as products at higher temperatures (130°C and 140°C) resulted in a mixture of single crystals and powders. Furthermore, a reaction time at 120°C between 72 and 96 hours produced the highest quality and quantity of crystals, as indicated by similar intensity and PXRD peaks.

In the third stage, the MOF was activated through methanol solvent exchange and vacuum drying. Nitrogen gas sorption analysis conducted at 77 K confirmed the microporosity of the MOF with a specific surface area of 1562 $m^2$ g-1, total pore volume of 0.564 cm^3/g, and an average pore width of 10.7 Å. Minimal hysteresis seen in the hysteresis analysis emphasized the stability and verified the microporous nature of our MOF.

The fourth stage involved detailed characterization of our MOF. A CHNS elemental analysis confirmed the chemical composition of our MOF, while a thermogravimetric analysis (TGA) showcased good thermal stability up to 300°C. Chemical stability tests displayed stability in DMF, methanol, and ethanol, however, the MOF displayed potential instability in water. Further 1H NMR spectroscopy after water exposure revealed a formate peak at ~8.3 ppm, implying hydrolysis. A single-crystal X-ray diffraction (SCXRD) analysis verified our MOF as a novel framework with unique structural features, including the positions of formate groups and water molecules. Moreover, pH stability tests showed significant losses in peak intensity for our MOF after exposure to water at varying pH levels, reinforcing the potential water instability issues.

Finally, in the fifth stage, we aimed at verifying the reproducibility of our process. The entire procedure was replicated, and the consistency of the results with earlier reported data confirmed the reproducibility of our protocols.

**Figure S82.** Illustration of the final summary developed by the Reticular ChemNavigator. This version of the summary was generated at the 29th iteration, marking the completion of Stage 5 interactions, which signifies the synthesis and study of MOF-521-*o*F. Previous versions of the summary have been omitted for clarity.

## Reticular ChemNavigator Output (MOF-521-*o*F)

Current Stage and Iteration: 1-1

Status Evaluation: As we are just beginning the project, no task has been completed yet and there are no results to evaluate. The first stage is the synthesis of the BTB linker. Our task choices are focused on preparing for this synthesis.

Task Choice 1: Gather all necessary reagents and materials for the synthesis of the BTB linker. These include 1,3,5-tris(bromomethyl)benzene, 4-fluorobenzoic acid, palladium(II) acetate, triphenylphosphine, sodium carbonate, and anhydrous dimethylformamide (DMF). Please check the safety data sheets (SDS) for each chemical before handling them.

Task Choice 2: Conduct a literature survey to identify the most appropriate reaction conditions for the synthesis of the BTB linker. Note down any common trends and outliers in the reported methods, and compile a list of considerations for our synthesis process, which might include temperature, solvent, catalyst, and reaction time.

Task Choice 3: Begin by designing the synthetic pathway for the BTB linker based on the available literature. It would likely involve a cross-coupling reaction such as the Suzuki-Miyaura reaction, so draw out the full mechanism including the roles of the catalyst and base, and identify any potential side reactions or complications. This will help in predicting and troubleshooting any issues that might arise during the actual synthesis.

**Figure S83.** Output generated by ChemNavigator during the 1st iteration, aimed at the synthesis and characterization of a novel aluminum-based MOF utilizing the BTB-*o*F linker.

## Reticular ChemNavigator Output (MOF-521-*o*F)

Current Stage and Iteration: 1-2

Status Evaluation: The apprentice's literature survey results indicate our specific BTB derivative is less studied, requiring us to adopt methods from similar compounds and apply our chemical expertise to devise a synthetic strategy. This presents an opportunity to explore and potentially make a significant contribution to the field of MOFs.

Task Choice 1: Initiate the synthesis design of our BTB linker. Use your organic chemistry knowledge to create a draft of the synthetic route using well-established reactions. You'll need to consider the functional groups, the positions of the substituents, and potential obstacles such as side reactions, separation, and purification challenges.

Task Choice 2: Expand your literature review to focus on similar compounds to our BTB linker. Look for synthesis methods of other BTB derivatives or compounds with similar functional groups and structures. Try to deduce general synthetic strategies that might be applicable to our target molecule, including reaction types, conditions, solvents, catalysts, etc.

Task Choice 3: Utilize SciFinder, a comprehensive database for scientific information, particularly chemical and bibliographic data. It contains synthesis procedures for a wide array of compounds, and it could potentially have useful information for the synthesis of our target compound or similar structures. You may find synthetic routes or strategies we can adopt or modify for our purposes. Be sure to note all relevant findings.

**Figure S84.** Output generated by ChemNavigator during the 2$^{nd}$ iteration, aimed at the synthesis and characterization of a novel aluminum-based MOF utilizing the BTB-*o*F linker.

## Reticular ChemNavigator Output (MOF-521-*o*F)

Current Stage and Iteration: 1-3

Status Evaluation: The use of SciFinder yielded promising results, with a relevant paper providing the exact synthesis route for our compound. This success validates our database and literature survey strategy. Moving forward, we need to focus on the execution of the discovered synthesis procedure, analyzing the procedure for safety, feasibility, and resource availability, and finally, conducting the synthesis itself.

Task Choice 1: Begin by conducting a safety and feasibility assessment of the synthesis procedure described in the Zhang et al. paper. Review the reagents, solvents, and catalysts used, ensuring that they are available and that their use aligns with our laboratory's safety guidelines. Then, check the reaction conditions (temperature, pressure, etc.) to see if our lab equipment can accommodate them. Also, consider any potential hazards that could arise during the process, and propose mitigations.

Task Choice 2: If all the resources needed for the synthesis are available and the reaction conditions align with our laboratory's capabilities, proceed with synthesizing the organic BTB-*o*F linker. Follow the synthesis procedure provided in the Zhang et al. paper, carefully noting any deviations or issues encountered during the synthesis. Be sure to document the yield and purity of the synthesized compound.

Task Choice 3: Conduct a thorough analysis of the synthesis procedure from Zhang et al. paper. Identify each step and the purpose it serves in the overall process. Document this analysis so that it can serve as a guide during the actual synthesis. Also, consider potential pitfalls or areas where things might go wrong during the synthesis, and brainstorm possible solutions or alternative strategies.

**Figure S85.** Output generated by ChemNavigator during the 3rd iteration, aimed at the synthesis and characterization of a novel aluminum-based MOF utilizing the BTB-*o*F linker.

## Reticular ChemNavigator Output (MOF-521-*o*F)

Current Stage and Iteration: 1-4

Status Evaluation: The thorough safety and feasibility check ensures that we can proceed to the synthesis stage without any unforeseen obstacles or risks. The apprentice has demonstrated good attention to detail and adherence to safety standards.

Task Choice 1: Now, we should move forward with the synthesis of the BTB-*o*F linker. Follow the synthesis procedure outlined in the Zhang et al. paper. Remember to take detailed notes at each step of the process, including any observations, problems, or deviations from the plan. This will allow us to repeat the process later or troubleshoot any issues that arise.

Task Choice 2: Before moving forward with the synthesis, you may want to perform a dry run of the process. This will involve going through each step of the procedure without the actual reagents. This practice could help you to identify any potential issues that might arise during the real synthesis, and it will allow you to become familiar with the entire process.

Task Choice 3: Alternatively, you may wish to further familiarize yourself with the synthesis procedure by diving deeper into the paper. Examine the authors' discussion and analyze their methods in more detail. Make sure you fully understand the purpose of each step and the role of each reagent, as well as any potential issues or pitfalls that the authors encountered. This will help ensure that you are fully prepared for the synthesis stage.

**Figure S86.** Output generated by ChemNavigator during the 4$^{th}$ iteration, aimed at the synthesis and characterization of a novel aluminum-based MOF utilizing the BTB-*o*F linker.

## Reticular ChemNavigator Output (MOF-521-*o*F)

Current Stage and Iteration: 1-5

Status Evaluation: The human apprentice followed the synthesis procedure accurately, resulting in successful completion of the BTB-*o*F linker synthesis. Based on the feedback, there were no challenges or deviations encountered, suggesting a successful replication of the procedure from the literature. The task for the next step would be to verify the successful synthesis through appropriate characterization techniques.

Task Choice 1: Now we need to proceed with the characterization of the synthesized BTB-*o*F linker. Start with 1H NMR spectroscopy to compare the chemical shifts of the product with the reported values in literature. Ensure that the solvent used for the NMR is deuterated dimethyl sulfoxide (DMSO-d6), and note the chemical shifts of the aromatic hydrogens and carboxylate methylene protons.

Task Choice 2: An alternate characterization method would be FTIR spectroscopy. This will provide information about the functional groups present in the synthesized linker. Prepare a small sample of the BTB-*o*F linker and obtain its infrared spectrum. Look specifically for the carboxylic acid (COOH) peaks and aromatic C-H stretches to confirm the structure of the product.

Task Choice 3: You could also conduct a melting point analysis to compare the synthesized BTB-*o*F linker with literature values. This will not only help in confirming the identity of the compound, but will also provide an indication of its purity. Note the onset and clear melt temperatures carefully. Remember to use a small, well-packed sample for this test.

**Figure S87.** Output generated by ChemNavigator during the 5$^{th}$ iteration, aimed at the synthesis and characterization of a novel aluminum-based MOF utilizing the BTB-*o*F linker.

## Reticular ChemNavigator Output (MOF-521-*o*F)

Current Stage and Iteration: 1-6

Status Evaluation: The apprentice executed the NMR spectroscopy accurately but is uncertain about interpreting the results to confirm successful synthesis of the BTB-*o*F linker. This is an important step in characterizing the synthesized organic linker.

Task Choice 1: To address the challenge faced in interpreting the NMR results, I suggest you compare the observed NMR chemical shifts with those expected for the BTB-*o*F linker. Look for chemical shifts that correspond to the aromatic protons and the carboxylate methylene protons in the literature. Confirm the presence of fluorine atoms by looking for their influence on neighboring protons. If the observed shifts match the expected ones, it suggests successful synthesis. Start by noting down the expected chemical shifts from the literature. Next, compare these shifts with the ones you observed in your NMR spectrum. Finally, draw a conclusion about the success of the synthesis based on this comparison.

Task Choice 2: If interpreting the NMR results is still proving challenging, consider seeking help from a supervisor or experienced colleague. Share your observed NMR spectrum with them and ask for their input on whether it aligns with the expected values for the BTB-*o*F linker. Be sure to explain the synthesis process you followed and any observations you made during that process. Their insights could prove useful in interpreting your results and confirming the successful synthesis of the linker.

Task Choice 3: Alternatively, you could perform additional characterization techniques such as FT-IR spectroscopy or mass spectroscopy to further confirm the successful synthesis of the BTB-*o*F linker. For FT-IR, you would look for characteristic absorption bands corresponding to aromatic rings and carboxylate groups. For mass spectroscopy, the molecular ion peak should match the molecular weight of the BTB-*o*F linker. If you choose this option, remember to follow the appropriate procedures for these techniques and compare the obtained results with the expected values.

**Figure S88.** Output generated by ChemNavigator during the 6$^{th}$ iteration, aimed at the synthesis and characterization of a novel aluminum-based MOF utilizing the BTB-*o*F linker.

## Reticular ChemNavigator Output (MOF-521-*o*F)

Current Stage and Iteration: 1-7

Status Evaluation: The experimental and predicted 1H NMR chemical shifts correlate well, suggesting the successful synthesis of the BTB-*o*F linker. The apprentice's careful analysis of the predicted and experimental data and appropriate application of NMR prediction tools led to this successful outcome.

Task Choice 1: Now that the BTB-*o*F linker has been successfully synthesized and confirmed, I suggest moving to the second stage of the project, which involves the synthesis of the MOF. You can start by preparing a mixture of the BTB-*o*F linker, aluminum source (such as aluminum nitrate), and an appropriate solvent (like DMF or DMSO) under inert atmosphere. Then, the mixture should be heated at a certain temperature for a specific time. If you feel ready, state, "I'm ready to move to the next stage."

Task Choice 2: While we believe the synthesis of the BTB-*o*F linker has been successful, to further confirm, it may be beneficial to conduct additional characterization methods such as FT-IR or Mass Spectrometry. This can provide more evidence for the successful synthesis of the BTB-*o*F linker.

Task Choice 3: Another option is to optimize the synthesis conditions of the BTB-*o*F linker. By modifying factors such as reaction time, temperature, or concentration, we might enhance the yield or purity of the BTB-*o*F linker. It might not be necessary given the successful NMR results, but it can still be a beneficial exercise.

**Figure S89.** Output generated by ChemNavigator during the 7$^{th}$ iteration, aimed at the synthesis and characterization of a novel aluminum-based MOF utilizing the BTB-*o*F linker.

Reticular ChemNavigator Output (MOF-521-*o*F)

Current Stage and Iteration: 2-1

Status Evaluation: The successful synthesis and verification of the BTB-*o*F linker in the first stage allowed for the advancement to the second stage. The human apprentice has confirmed readiness for this transition, indicating successful completion of the first stage and readiness to embark on the synthesis of the MOF itself.

Task Choice 1: Begin the MOF synthesis process by preparing a mixture of the BTB-*o*F linker and aluminum nitrate. Take an appropriate amount of BTB-*o*F linker and an equimolar amount of aluminum nitrate. Combine these in a suitable solvent such as DMF or DMSO, ensuring that the process is conducted under an inert atmosphere to prevent unwanted reactions. After the substances are dissolved completely, heat the mixture at around 100-120°C for 24-72 hours. Keep monitoring the reaction and record observations.

Task Choice 2: As an alternative way to synthesize the MOF, prepare a mixture of the BTB-*o*F linker, aluminum nitrate, and a modulator (like monocarboxylic acid). Combine these in an appropriate solvent such as DMF under an inert atmosphere. Heat the mixture at a temperature between 100-120°C for 24-72 hours. The modulator can help control the size and morphology of the MOF crystals, which might be beneficial for the later stages of the project.

Task Choice 3: If you feel more research is needed before moving forward with the synthesis of the MOF, you can review relevant literature to understand the best practices for MOF synthesis, particularly focusing on MOFs that utilize similar linkers or metal sources. Pay special attention to the reaction conditions, such as temperature, reaction time, and the use of modulators. This could provide you with a broader perspective and valuable insights as you proceed with the synthesis process.

**Figure S90.** Output generated by ChemNavigator during the 8$^{th}$ iteration, aimed at the synthesis and characterization of a novel aluminum-based MOF utilizing the BTB-*o*F linker.

## Reticular ChemNavigator Output (MOF-521-*o*F)

Current Stage and Iteration: 2-2

Status Evaluation: The human apprentice's extensive review of the literature provides a clear understanding of the synthetic conditions for analogous MOF systems. The reported BTB:Al ratios, reaction temperatures, times, solvents, and use of modulators give us a range of conditions to try in our own synthesis. The variability in these parameters also suggests the need for experimental iterations to identify the optimal conditions for our novel MOF.

Task Choice 1: Let's proceed with the first synthesis attempt for our aluminum MOF using BTB-*o*F as a linker. Use the most common parameters from the literature: BTB:Al ratio of 2:1, synthesis temperature of 150°C, and reaction time of 72 hours. Use DMF as the solvent, and include formic acid as a modulator based on the literature findings. Monitor the synthesis and report back with observations, particularly noting any precipitate formation or color changes.

Task Choice 2: Alternatively, you could proceed with a less conservative approach by attempting a synthesis at the extreme conditions found in the literature review. This might give us a sense of the boundaries of our system. Consider using a BTB:Al ratio of 1:4, the highest reported synthesis temperature of 180°C, and the longest reaction time of 72 hours. Use DMF as the solvent and include TFA as a modulator based on the findings from the literature. Monitor the reaction and report any observations.

Task Choice 3: Another path to consider involves an iterative approach, where you would carry out multiple syntheses at the same time, each varying a single parameter (BTB:Al ratio, temperature, time, or modulator). This could give us a more comprehensive understanding of how each parameter affects the outcome. In this case, start with three syntheses, each using a different BTB:Al ratio: 1:4, 2:1, and 4:1, while keeping the other parameters constant. Report back with any observations and potential differences between each synthesis.

**Figure S91.** Output generated by ChemNavigator during the 9$^{th}$ iteration, aimed at the synthesis and characterization of a novel aluminum-based MOF utilizing the BTB-*o*F linker.

## Reticular ChemNavigator Output (MOF-521-*o*F)

Current Stage and Iteration: 2-3

Status Evaluation: The apprentice has effectively performed the multiple syntheses with variations in the BTB:Al ratio, each resulting in a white powder. This suggests that some form of product has formed. However, without further analysis, it's unclear if these are the desired MOFs or just unreacted linker or another product. As such, the use of Powder X-ray Diffraction (PXRD) becomes critical to verify the nature of these compounds, similar to the approach used in the similar work summary.

Task Choice 1: Now that we have synthesized our products using different BTB:Al ratios, the next step is to confirm whether they are indeed MOFs. For this, Powder X-ray Diffraction (PXRD) is a powerful technique. Conduct a PXRD analysis on each of the three samples. The obtained diffraction patterns can then be compared with the expected MOF structure. This will allow us to identify the optimal BTB:Al ratio for MOF formation.

Task Choice 2: While we are awaiting PXRD analysis, we can also explore variations in the other parameters, namely the reaction time and temperature. Keep the BTB:Al ratio constant (you can choose any of the ratios you've tested), and vary the reaction time. For instance, conduct three reactions each at different times: 24, 48, and 72 hours. This can help us understand the influence of reaction time on MOF formation.

Task Choice 3: Alternatively, keeping the BTB:Al ratio constant, you can investigate the effect of reaction temperature on MOF formation. Conduct three syntheses, each at a different temperature: 80°C, 100°C, and 120°C. As before, be sure to observe and record any differences you see in the products of these reactions. This will aid us in identifying the optimal reaction temperature for our MOF synthesis.

**Figure S92.** Output generated by ChemNavigator during the $10^{th}$ iteration, aimed at the synthesis and characterization of a novel aluminum-based MOF utilizing the BTB-*o*F linker.

Reticular ChemNavigator Output (MOF-521-*o*F)

Current Stage and Iteration: 2-4

Status Evaluation: The PXRD results indicate that the current reaction conditions, particularly the BTB:Al ratios, are not optimal for the formation of the desired MOF structure. This could be attributed to either a suboptimal ratio or the absence of a suitable modulator to promote the formation of the desired MOF structure.

Task Choice 1: Now, let's attempt to optimize the modulator in the reaction mixture. Use a mixture of formic acid and water as the modulator, with formic acid to water ratios of 1:1, 2:1, and 4:1, keeping the BTB:Al ratio at 2:1. Conduct the synthesis for each condition, ensuring all other parameters remain constant. Once you have completed the syntheses, conduct PXRD analyses on the resulting products to verify if the diffraction pattern matches the expected MOF structure.

Task Choice 2: We can also try different reaction temperatures. Repeat the synthesis process with the BTB:Al ratio of 2:1 and formic acid as a modulator, but this time vary the temperature. Set the temperatures to 100°C, 120°C, and 140°C. After conducting the synthesis at each temperature, perform PXRD analyses to confirm the formation of the MOF.

Task Choice 3: Another possibility is that the reaction time might not be sufficient for proper MOF formation. Try to extend the reaction time to 48 hours, 72 hours, and 96 hours, with the BTB:Al ratio at 2:1 and formic acid as a modulator. After conducting the syntheses, carry out PXRD analyses on each product to verify the formation of the MOF.

**Figure S93.** Output generated by ChemNavigator during the 11$^{th}$ iteration, aimed at the synthesis and characterization of a novel aluminum-based MOF utilizing the BTB-*o*F linker.

## Reticular ChemNavigator Output (MOF-521-*o*F)

Current Stage and Iteration: 2-5

Status Evaluation: The use of different ratios of formic acid to water as a modulator seems to have influenced the formation of the MOF structure, as indicated by the slightly better PXRD result obtained from the 4:1 ratio. However, the weak PXRD peaks suggest that while the modulator optimization was a step in the right direction, it is not yet sufficient to form the desired MOF structure.

Task Choice 1: Considering the slight improvement in the PXRD analysis at a 4:1 formic acid to water ratio, let's attempt a higher ratio for the modulator. Synthesize the MOF using formic acid to water ratios of 6:1 and 8:1, keeping the BTB:Al ratio at 2:1. Once the syntheses are completed, conduct PXRD analyses on the resulting samples to verify if the diffraction pattern matches the expected MOF structure.

Task Choice 2: Besides the modulator ratio, we can also experiment with varying reaction temperatures. Keeping the formic acid to water ratio constant at 4:1 and the BTB:Al ratio at 2:1, conduct the syntheses at temperatures of 100°C, 120°C, and 140°C. Once completed, perform PXRD analyses on the resulting samples to verify the diffraction patterns.

Task Choice 3: Exploring different BTB:Al ratios while keeping the modulator ratio constant might help optimize MOF formation. Considering the better PXRD result at a 4:1 formic acid to water ratio, maintain this modulator ratio while adjusting the BTB:Al ratios. Try ratios of 1:2, 2:2, and 3:2. After conducting the syntheses, analyze the resulting samples with PXRD to verify the formation of the MOF structure.

**Figure S94.** Output generated by ChemNavigator during the 12$^{th}$ iteration, aimed at the synthesis and characterization of a novel aluminum-based MOF utilizing the BTB-*o*F linker.

## Reticular ChemNavigator Output (MOF-521-*o*F)

Current Stage and Iteration: 2-6

Status Evaluation: Maintaining the modulator ratio at 4:1 (formic acid:water) while varying the BTB:Al ratios led to more promising results. It seems that the BTB:Al ratio of 3:4 appears to yield the best MOF formation, as indicated by the PXRD results.

Task Choice 1: Now that the BTB:Al ratio of 3:4 appears promising, focus on refining the reaction parameters such as temperature and time. Use a temperature range of 100°C to 140°C at increments of 10°C. Maintain each reaction for 48 hours, and then perform PXRD analysis on the resultant MOF.

Task Choice 2: Another way to confirm our findings would be to reproduce the conditions that resulted in the promising PXRD pattern. Repeat the synthesis of the MOF using a BTB:Al ratio of 3:4 and a modulator ratio of formic acid to water at 4:1. Keep the temperature at 140°C and the reaction time at 48 hours. Analyze the resulting product using PXRD. This will provide additional evidence supporting the conditions that yield the optimal MOF structure.

Task Choice 3: Investigate the influence of the reaction time on the synthesis of the MOF. Maintain the BTB:Al ratio at 3:4 and the modulator ratio at 4:1 (formic acid to water). Conduct the reactions at 140°C for varying lengths of time: 24 hours, 48 hours, 72 hours, and 96 hours. After each time point, perform PXRD analysis to determine if longer reaction times improve MOF formation.

**Figure S95.** Output generated by ChemNavigator during the 13th iteration, aimed at the synthesis and characterization of a novel aluminum-based MOF utilizing the BTB-*o*F linker.

Reticular ChemNavigator Output (MOF-521-*o*F)

Current Stage and Iteration: 2-7

Status Evaluation: The experimental results align with our task prediction - the increase in reaction temperature up to 120°C improved MOF formation, as indicated by the higher PXRD peak intensity and sharper peaks. This improvement might be due to the increased kinetic energy at higher temperatures, promoting more effective reactions. However, temperatures above 120°C resulted in a mixture of single crystals and powders, suggesting an optimal temperature for this MOF synthesis around 120°C.

Task Choice 1: Since the 120°C temperature has shown promising results, let's further explore the reaction time at this temperature. Retain the BTB:Al ratio at 3:4 and the formic acid to water ratio at 4:1. Run reactions at 120°C for different times, including 24, 48, 72, and 96 hours. After each reaction time, isolate the MOF, wash it properly, and perform PXRD analysis to evaluate the crystal quality and MOF formation.

Task Choice 2: Given that 120°C appears to be the optimum temperature, it could be worthwhile to refine the temperature range around this point for a more accurate determination of the optimal synthesis condition. Try additional reactions at 115°C and 125°C with the same BTB:Al ratio of 3:4 and formic acid to water ratio at 4:1. Keep the reaction time constant at 48 hours. Once the reactions are completed, isolate the MOF and conduct PXRD analyses to confirm the MOF formation and its quality.

Task Choice 3: Considering the results at 120°C, there may be a narrow temperature window where the MOF forms optimally. Test a series of temperatures within 10°C of 120°C at 2°C increments (i.e., 110°C, 112°C, 114°C, 116°C, 118°C, 120°C, 122°C, 124°C, 126°C, 128°C, and 130°C). Maintain the BTB:Al ratio at 3:4, formic acid to water ratio at 4:1, and the reaction time at 48 hours for each run.

**Figure S96.** Output generated by ChemNavigator during the 14$^{th}$ iteration, aimed at the synthesis and characterization of a novel aluminum-based MOF utilizing the BTB-*o*F linker.

Reticular ChemNavigator Output (MOF-521-*o*F)

Current Stage and Iteration: 2-8

Status Evaluation: The feedback from the latest iteration suggests that the synthetic process is almost optimized. The 72 and 96 hours reaction times resulted in a similar amount of crystals with similar PXRD peaks, which shows that extending the reaction beyond 72 hours does not significantly improve the crystallization process at 120°C. Therefore, we have nearly reached optimal conditions for synthesizing our MOF using BTB-*o*F as a linker.

Task Choice 1: Now, to further ensure the robustness of our synthesis protocol, let's vary the BTB:Al ratio while keeping other parameters constant. Keep the temperature at 120°C, the formic acid to water ratio at 4:1, and the reaction time at 72 hours. Try BTB:Al ratios of 3:4, 1:1, and 4:3. Isolate the MOF after each trial, wash it properly, and perform PXRD analysis to evaluate the crystal quality and MOF formation.

Task Choice 2: Alternatively, we could experiment with the modulator ratio while keeping other conditions constant. Retain the temperature at 120°C, the BTB:Al ratio at 3:4, and the reaction time at 72 hours. Explore formic acid to water ratios of 4:1, 3:1, and 5:1. After each trial, isolate the MOF, wash it properly, and perform PXRD analysis to evaluate the crystal quality and MOF formation.

Task Choice 3: Given the promising results of the last few iterations, if you believe that we have reached the optimal synthetic conditions, you may state, "I'm ready to move to the next stage." We can then proceed to stage three of the project: Activation and Determination of Permanent Porosity.

**Figure S97.** Output generated by ChemNavigator during the 15th iteration, aimed at the synthesis and characterization of a novel aluminum-based MOF utilizing the BTB-*o*F linker.

Reticular ChemNavigator Output (MOF-521-*o*F)

Current Stage and Iteration: 3-1

Status Evaluation: The apprentice's agreement to move forward represents a confidence in the optimal synthetic conditions that have been achieved, as suggested by the successful generation of crystals with the expected PXRD peaks, and the satisfactory balance between temperature and crystallinity.

Task Choice 1: Begin the third stage by activating the synthesized MOF. Since it is important to remove any solvents that might be trapped within the MOF pores, we will use the supercritical CO2 method for this. Set up a high-pressure reactor capable of withstanding supercritical conditions (CO2 critical point: 31.1°C, 73.8 atm). Place your MOF sample inside, and fill the reactor with CO2. Gradually increase the temperature to just above 31.1°C and the pressure to just above 73.8 atm. Hold these conditions for 12 hours. After the process, depressurize slowly to avoid structural collapse of the MOF.

Task Choice 2: Alternatively, you can activate the MOF using a solvent exchange followed by drying under vacuum. First, exchange the solvent in the pores by immersing the MOF crystals in methanol, and repeat this step thrice, each time allowing the MOF to soak for 12 hours. After the final methanol exchange, filter the methanol off and dry the MOF under vacuum at room temperature for 12 hours to remove the remaining solvent.

Task Choice 3: Another method to activate the MOF is thermal activation. You can gently heat the MOF in an oven under a dynamic vacuum. Start by setting the oven temperature to 100°C and placing your MOF under vacuum. Over the course of 12 hours, gradually ramp the temperature up to 200°C. At the end of the 12 hours, cool the MOF to room temperature while maintaining the vacuum. Ensure that you monitor the thermal stability of the MOF throughout the process to avoid any structural collapse.

**Figure S98.** Output generated by ChemNavigator during the 16$^{th}$ iteration, aimed at the synthesis and characterization of a novel aluminum-based MOF utilizing the BTB-*o*F linker.

Reticular ChemNavigator Output (MOF-521-*o*F)

Current Stage and Iteration: 3-2

Status Evaluation: The activation process was carried out successfully with the MOF crystals undergoing a solvent exchange and drying process. However, the lack of observable changes may imply that the activation process did not sufficiently expose or create the desired porosity within the MOF crystals, or the changes are not visible to the naked eye and might need instrumental analysis to be confirmed.

Task Choice 1: It is now time to test the MOF for permanent porosity. To do this, perform nitrogen gas sorption at 77 K. Start by preparing the sample by outgassing it at a temperature of 120°C under vacuum for 12 hours. Then, cool the sample to 77 K in a liquid nitrogen bath and expose it to increasing partial pressures of nitrogen gas. Measure the volume of gas adsorbed at each pressure point. Finally, analyze the resulting adsorption/desorption isotherms to determine the pore structure, specific surface area, and pore volume of the MOF.

Task Choice 2: If you believe further activation is necessary, consider a supercritical CO2 activation method. This will involve replacing the solvent in the MOF pores with CO2, which will then be removed by heating above the critical temperature (31.1°C) and reducing the pressure, causing the CO2 to evaporate without condensation. To do this, immerse the MOF sample in liquid CO2 and seal the container. Slowly heat the container to 40°C while maintaining a pressure above 73 atmospheres to ensure CO2 stays in the supercritical phase. After reaching 40°C, slowly depressurize the system to atmospheric pressure, allowing the supercritical CO2 to escape. This procedure may help to preserve the MOF structure and create an increased pore volume.

Task Choice 3: Another approach to ensuring complete activation would be to repeat the solvent exchange and vacuum drying process, but increase the temperature during vacuum drying. Redo the methanol soaking process but this time, dry the MOF under vacuum at a higher temperature, say 120°C, for 24 hours. This might ensure the removal of any residual methanol molecules lodged within the MOF pores.

**Figure S99.** Output generated by ChemNavigator during the 17$^{th}$ iteration, aimed at the synthesis and characterization of a novel aluminum-based MOF utilizing the BTB-*o*F linker.

## Reticular ChemNavigator Output (MOF-521-*o*F)

Current Stage and Iteration: 3-3

Status Evaluation: The apprentice successfully performed the nitrogen gas sorption experiment as per the guidelines. However, the isotherm data provided was incomplete, hence the final determination of MOF porosity and related parameters could not be finalized.

Task Choice 1: To move forward, we need to analyze the adsorption/desorption isotherms from the nitrogen sorption experiment. This will enable us to calculate the specific surface area, total pore volume, and average pore width of the MOF. You can calculate the specific surface area using the Brunauer–Emmett–Teller (BET) method. The total pore volume can be determined from the volume adsorbed at a relative pressure close to unity (P/P0≈1), and the pore size distribution can be calculated using the Barrett-Joyner-Halenda (BJH) method. After the calculation, please provide all the derived values.

Task Choice 2: If the isotherm data or the BET instrument required for the calculation are not available, you might need to repeat the nitrogen gas sorption analysis. Make sure the sample is properly outgassed at 120°C for 12 hours. After the degassing process, cool the sample to 77 K in a liquid nitrogen bath and expose it to increasing partial pressures of nitrogen gas. Remember to record the volume of gas adsorbed at each pressure point.

Task Choice 3: If for some reason the nitrogen sorption analysis cannot be repeated or the data for the BET calculation is not available, we might want to proceed with other characterization methods. If this is the case, please proceed to the next stage: Detailed Structural Analysis and Characterization of the MOF. You may start with thermogravimetric analysis (TGA) to determine the thermal stability of the MOF. To perform this, weigh about 5-10 mg of the MOF sample and heat it from room temperature to 600°C at a rate of 10°C/min under a nitrogen atmosphere, and record the weight loss as a function of temperature.

**Figure S100.** Output generated by ChemNavigator during the 18$^{th}$ iteration, aimed at the synthesis and characterization of a novel aluminum-based MOF utilizing the BTB-*o*F linker.

Reticular ChemNavigator Output (MOF-521-*o*F)

Current Stage and Iteration: 3-4

Status Evaluation: The successful calculation of the specific surface area, total pore volume, and average pore width underscores the successful activation of our MOF and determination of its permanent porosity. The values obtained for these parameters are in line with what is typically expected for microporous MOFs, supporting the effectiveness of our previous task recommendations.

Task Choice 1: Our next logical step is to undertake a hysteresis analysis of our adsorption/desorption isotherms. Hysteresis can shed light on the capillary condensation occurring within our MOF structure, which will give us further insights into the porosity and stability of our MOF. Look for hysteresis between the adsorption and desorption branches of the isotherms. If you observe a significant hysteresis loop, this can suggest that our MOF is likely to have mesopores or larger pores.

Task Choice 2: Another useful method for understanding our MOF's properties is to conduct a pore size distribution analysis. Use the Horvath-Kawazoe method, or another suitable method available in your software, to calculate the pore size distribution from the nitrogen adsorption isotherm. The result of this analysis should provide a more detailed picture of the pore size distribution within our MOF.

Task Choice 3: If you feel that we have gathered enough information regarding the porosity and stability of our MOF, you could say, "I'm ready to move to the next stage." The fourth stage of our project will focus on the detailed structural analysis and characterization of the MOF, involving single crystal structure analysis, determination of chemical composition, assessment of phase purity, and testing for chemical and thermal stability.

**Figure S101.** Output generated by ChemNavigator during the 19$^{th}$ iteration, aimed at the synthesis and characterization of a novel aluminum-based MOF utilizing the BTB-*o*F linker.

## Reticular ChemNavigator Output (MOF-521-*o*F)

Current Stage and Iteration: 3-5

Status Evaluation: The successful hysteresis analysis and comprehensive porosity characterization, in conjunction with the previously conducted MOF activation, have effectively met the main objectives of the third stage. It appears that we are ready to progress to the next stage, though additional confirmatory tests such as thermal and chemical stability assessments would add value.

Task Choice 1: An assessment of chemical stability could be beneficial. Immerse the MOF sample in water, methanol, ethanol, and acetone separately for 24 hours. After the period, filter and dry the samples before carrying out PXRD analysis to examine any changes in crystallinity post solvent exposure.

Task Choice 2: A thermal stability assessment via Thermogravimetric Analysis (TGA) can be performed. Using a TGA instrument, heat a small MOF sample from room temperature to 600°C at a rate of 10°C per minute under a nitrogen atmosphere. The MOF is considered thermally stable if it shows no significant weight loss below 300°C.

Task Choice 3: Since we have fulfilled the main objectives of the current stage, it would be appropriate to transition to the next stage of the project: Detailed Structural Analysis and Characterization of the MOF. However, I recommend stating, "I'm ready to move to the next stage," to officially advance the project.

**Figure S102.** Output generated by ChemNavigator during the 20th iteration, aimed at the synthesis and characterization of a novel aluminum-based MOF utilizing the BTB-*o*F linker.

Reticular ChemNavigator Output (MOF-521-*o*F)

Current Stage and Iteration: 4-1

Status Evaluation: The human apprentice has confirmed readiness to move onto the fourth stage of the project, marking the successful completion of the previous stage. The successful activation of the MOF and the confirmation of its microporous nature, together with the results of the hysteresis analysis, made this transition possible.

Task Choice 1: Begin with thermal stability analysis. This step will involve Thermo Gravimetric Analysis (TGA). Prepare the TGA instrument according to manufacturer's instructions and measure the weight loss of the MOF sample as a function of increasing temperature. This will give us insight into the decomposition temperature of our MOF, which in turn provides information about its thermal stability. Remember to maintain a heating rate of 10°C per minute in nitrogen atmosphere and monitor the weight loss of the sample from room temperature to about 800°C.

Task Choice 2: Perform single crystal X-ray diffraction analysis to gain a detailed insight into the MOF structure. Carefully choose a well-defined, single crystal from the MOF sample and mount it on a suitable support. Set up the diffractometer following the manufacturer's guidelines, and conduct the X-ray diffraction experiment. Once you have obtained the diffraction pattern, perform data reduction and solve the structure using suitable software. This analysis will provide us with precise atomic positions and bonding information, contributing to our understanding of the MOF structure.

Task Choice 3: Initiate a chemical composition analysis using CHNS elemental analysis. Weigh a small quantity (around 1-2 mg) of the MOF sample and place it into a tin capsule. Introduce the sample into the elemental analyzer and initiate the automated combustion process. Once the elemental analysis is complete, you will have the weight percentage of carbon, hydrogen, nitrogen, and sulfur in the sample. This step will provide information about the elemental composition of our MOF, affirming its successful synthesis.

**Figure S103.** Output generated by ChemNavigator during the 21$^{st}$ iteration, aimed at the synthesis and characterization of a novel aluminum-based MOF utilizing the BTB-*o*F linker.

Reticular ChemNavigator Output (MOF-521-*o*F)

Current Stage and Iteration: 4-2

Status Evaluation: The CHNS elemental analysis yielded results that mostly align with the theoretical values for our MOF, confirming that the organic linkers are integrated into the MOF as expected. Some deviations are normal due to experimental factors, but they don't indicate significant issues with the structure or synthesis of the MOF.

Task Choice 1: Let's continue our detailed structural analysis of the MOF by conducting Powder X-ray Diffraction (PXRD). Prepare a sample of the MOF and place it onto a PXRD sample holder, making sure to evenly distribute it for optimal results. Run the PXRD analysis, focusing on a 2θ range from 2° to 40°. The generated diffraction pattern will reveal information about the crystallinity and phase purity of our MOF, providing further insights into its structure.

Task Choice 2: In order to understand the thermal stability of the MOF, carry out a Thermogravimetric Analysis (TGA). Weigh around 5-10 mg of the sample and place it in a TGA pan. Conduct the analysis from room temperature to 800°C at a heating rate of 10°C/min under a nitrogen atmosphere. The weight loss as a function of temperature will give us insight into the thermal stability of our MOF.

Task Choice 3: To probe the chemical stability of our MOF, especially in relation to different solvents, you can perform a solvent stability test. Take small quantities of the MOF and expose them to common solvents such as water, methanol, ethanol, and dichloromethane, among others. After a period of 24 hours, examine the samples. Loss of crystallinity or visible degradation will indicate solvent sensitivity. You can validate these observations using PXRD analysis of the post-exposure samples.

**Figure S104.** Output generated by ChemNavigator during the 22nd iteration, aimed at the synthesis and characterization of a novel aluminum-based MOF utilizing the BTB-*o*F linker.

Reticular ChemNavigator Output (MOF-521-*o*F)

Current Stage and Iteration: 4-3

Status Evaluation: The TGA test successfully confirmed the MOF's thermal stability. The initial stability up to 270°C, and then weight loss corresponding to the formate group's weight, align with expectations. The second sharp weight loss at 570°C is also consistent with MOF's decomposition at high temperatures. Based on this, we will focus on further structural and compositional analysis.

Task Choice 1: Next, we should conduct Powder X-ray Diffraction (PXRD) analysis to understand the MOF's phase purity and crystal structure. Prepare a finely ground sample of your MOF and mount it onto a PXRD sample holder. Run the PXRD pattern from 2θ values of 5° to 50°, with a step size of 0.02° and a dwell time of 0.5 seconds per step. Once the pattern is obtained, compare it with the simulated PXRD pattern based on the MOF's expected crystal structure.

Task Choice 2: To better understand the chemical stability of our MOF, it is suggested that we immerse the MOF samples in different solvents (water, methanol, ethanol, DCM, DMF, DMSO) for 24 hours. After this, we will recover the samples by filtration, wash them with the corresponding solvent, and dry them under vacuum. Subsequently, conduct PXRD on these samples to verify if the crystal structure is retained. If a significant change is noticed in the PXRD pattern, it means our MOF is unstable in that particular solvent.

Task Choice 3: To further verify the MOF's structure, Fourier Transform Infrared (FTIR) Spectroscopy can be performed. Carefully grind your MOF sample with KBr and press it into a disk. Perform the FTIR scan from 4000 to 400 cm-1. Compare the resultant spectrum with the expected FTIR spectrum based on the MOF's known structure.

**Figure S105.** Output generated by ChemNavigator during the 23$^{rd}$ iteration, aimed at the synthesis and characterization of a novel aluminum-based MOF utilizing the BTB-*o*F linker.

## Reticular ChemNavigator Output (MOF-521-*o*F)

Current Stage and Iteration: 4-4

Status Evaluation: The observed reduction in PXRD peak intensity for the sample immersed in water suggests that our MOF may have lower stability in water. This could be due to the potential hydrolysis of the carboxylate linkages or other water-sensitive components in the MOF structure, which aligns with the behavior of similar MOFs. As such, it would be beneficial to further explore the water instability and determine possible mechanisms causing it.

Task Choice 1: To delve deeper into the potential water instability, we should perform 1H NMR spectroscopy on the MOF after exposure to water. To do this, prepare a small quantity of MOF, digest it in a solution of sodium deuteroxide (NaOD) in deuterium oxide (D2O) for 24 hours. After this, proceed with 1H NMR analysis on the resulting solution. This procedure will give us insight into the components that may be hydrolyzed in the MOF when it's exposed to water. Any observed changes in the spectrum compared to the original MOF could shed light on the structural alterations that our MOF undergoes when it interacts with water.

Task Choice 2: To further validate the water instability, conduct a dynamic water stability test. In this test, prepare a MOF sample and immerse it in water. At regular intervals (e.g., every 6 hours), take a small amount of the MOF out, dry it, and perform PXRD. By observing any changes in the PXRD pattern over time, we can gain more insight into the MOF's stability in water over a longer period.

Task Choice 3: To better understand our MOF's potential water instability, we can perform pH sensitivity tests. Prepare three aqueous solutions with pH values of 3, 7, and 10. Immerse a small amount of MOF in each solution for 24 hours, then dry the samples and perform PXRD. Changes in the PXRD pattern could help us understand if our MOF's stability.

**Figure S106.** Output generated by ChemNavigator during the 24$^{th}$ iteration, aimed at the synthesis and characterization of a novel aluminum-based MOF utilizing the BTB-*o*F linker.

## Reticular ChemNavigator Output (MOF-521-*o*F)

Current Stage and Iteration: 4-5

Status Evaluation: The outcome of the 1H NMR analysis implies the hydrolysis of formate groups when the MOF interacts with water, which could be a cause for the observed reduction in PXRD peak intensity. This aligns with our previous observation of water instability, indicating that the MOF structure might be compromised upon contact with water due to the possible dissociation of formate groups.

Task Choice 1: To further investigate the water instability of our MOF, it would be valuable to conduct a pH stability test. You can achieve this by immersing the MOF in water at different pH levels, ranging from 1 to 14, for 24 hours. Afterward, conduct PXRD analysis on the resultant MOF samples to study the impact of varying pH on the MOF structure. Any changes observed in the PXRD patterns compared to the original MOF could indicate the pH levels at which our MOF starts showing instability.

Task Choice 2: Given the possible instability in water due to the hydrolysis of formate groups, it could be beneficial to explore ways to improve the water stability of our MOF. One option could be the post-synthetic modification of the MOF. For instance, you can conduct a silanization process using a silane coupling agent like (3-aminopropyl)triethoxysilane (APTES) in a DMF solution, which could possibly enhance the water stability of the MOF.

Task Choice 3: Considering that we are already in the advanced stage of MOF characterization, another task could be to perform a single-crystal X-ray diffraction (XRD) analysis, if single crystals of sufficient size and quality are available. This could provide a detailed understanding of the crystal structure of our MOF, supporting our findings from PXRD and potentially revealing more about the positions and interactions of the formate groups and water molecules within the MOF structure.

**Figure S107.** Output generated by ChemNavigator during the 25$^{th}$ iteration, aimed at the synthesis and characterization of a novel aluminum-based MOF utilizing the BTB-*o*F linker.

## Reticular ChemNavigator Output (MOF-521-*o*F)

Current Stage and Iteration: 4-6

Status Evaluation: The single-crystal X-ray diffraction (SCXRD) analysis has successfully revealed the detailed crystal structure of our MOF, further validating our synthesis protocol and providing insight into the likely hydrolysis processes. The presence of formates in the SBU and the potential for hydrogen bonding with water molecules offer possible explanations for the instability in water noted in earlier stages. Given the successful SCXRD analysis and comprehensive structural characterization so far, we are nearing the completion of stage four.

Task Choice 1: Next, we could focus on further investigating the chemical stability of our MOF in aqueous conditions. Specifically, conduct pH stability tests by soaking the MOF in buffered aqueous solutions at different pH values ranging from 2 to 12. Allow the MOF to soak for 24 hours, then filter, dry, and analyze the resulting material with PXRD to assess any changes in the MOF structure.

Task Choice 2: We can also proceed with a deeper examination of the MOF-water interactions by using infrared (IR) spectroscopy. Prepare a sample of the MOF and expose it to water, then collect IR spectra to potentially reveal more information about the bonding interactions and structural changes when the MOF comes in contact with water.

Task Choice 3: If you feel that we have accumulated enough data for the detailed structural analysis and characterization stage, and since we have a potential explanation for the water instability, you can declare readiness to move on to the fifth and final stage of our research, focusing on reproducibility checks and final validation. In this case, state "I'm ready to move to the next stage."

**Figure S108.** Output generated by ChemNavigator during the 26$^{th}$ iteration, aimed at the synthesis and characterization of a novel aluminum-based MOF utilizing the BTB-*o*F linker.

Reticular ChemNavigator Output (MOF-521-*o*F)

Current Stage and Iteration: 4-7

Status Evaluation: The reduced PXRD peak intensity in acidic, neutral, and basic water conditions suggests a notable decrease in the structural integrity of MOF-521-*o*F upon exposure to water, reaffirming our earlier suspicion about water instability. This instability likely arises due to hydrolysis, supported by the earlier identified formate peak in the 1H NMR spectra post water exposure.

Task Choice 1: Given the identified water instability of MOF-521-*o*F, we can conduct more in-depth 1H NMR spectroscopy after water exposure. The aim is to further investigate the nature of the hydrolysis reaction that our MOF seems to undergo in the presence of water. Prepare samples of MOF-521-*o*F exposed to water for different durations (e.g., 1, 2, 4, 6, 12, 24 hours), dry them, and analyze each sample with 1H NMR. This could help us understand how rapidly the hydrolysis reaction occurs and whether there is a specific period after which the structural integrity of our MOF is particularly affected.

Task Choice 2: Given the water instability of our MOF, another approach could be to conduct stability tests in different non-aqueous solvents. Choose solvents with a wide range of polarity, such as hexane, dichloromethane, acetonitrile, and DMSO. Immerse the MOF-521-*o*F in each solvent separately for 24 hours, filter, dry, and analyze the resulting material using PXRD. This would allow us to better understand the MOF's stability spectrum in non-aqueous environments.

Task Choice 3: As we have gathered substantial data regarding the structure, stability, and porosity of our MOF-521-*o*F, it seems like we've reached the end of the characterization stage. If you feel the same, please confirm by stating "I'm ready to move to the next stage."

**Figure S109.** Output generated by ChemNavigator during the 27$^{th}$ iteration, aimed at the synthesis and characterization of a novel aluminum-based MOF utilizing the BTB-*o*F linker.

Reticular ChemNavigator Output (MOF-521-*o*F)

Current Stage and Iteration: 5-1

Status Evaluation: The feedback from the apprentice confirms readiness to move to the final stage, which involves checking the reproducibility of the previous stages and final validation of the project outcomes. The team has successfully passed through the synthesis, optimization, porosity analysis, and characterization of MOF-521-*o*F, despite the challenge of water instability, which is linked to the presence of formate groups.

Task Choice 1: To validate the reproducibility of our synthesis and characterization procedures, replicate the entire procedure from the synthesis of MOF-521-*o*F to post-synthesis treatments and characterization. Begin by synthesizing the BTB-*o*F linker as we did in the first stage. Then, use the identified optimal parameters to produce the MOF. Once synthesized, activate the MOF and carry out nitrogen gas sorption measurements at 77 K, followed by CHNS elemental analysis and thermogravimetric analysis (TGA). Finally, conduct chemical stability tests in various solvents including water.

Task Choice 2: Alternatively, to focus on one significant finding, we can revisit the water instability issue of MOF-521-*o*F. Reproduce the synthesis of MOF-521-*o*F and then conduct a more thorough investigation of its interaction with water. This can involve an extended study on the effect of water exposure at different pH levels and durations, coupled with thorough analyses, including NMR, IR spectroscopy, and PXRD, before and after water exposure.

Task Choice 3: If you think it's necessary, we could potentially test the porosity of our MOF with other gases, such as argon or carbon dioxide, and compare the results with the nitrogen sorption measurements. This could provide us with additional confirmation of the reproducibility of the porosity results, as well as additional insights about the versatility of our MOF for different applications. After synthesizing the MOF-521-*o*F following the protocol identified in stage two, perform gas sorption measurements using argon or carbon dioxide and compare these results with our previous nitrogen sorption measurements.

**Figure S110.** Output generated by ChemNavigator during the 28$^{th}$ iteration, aimed at the synthesis and characterization of a novel aluminum-based MOF utilizing the BTB-*o*F linker.

Reticular ChemNavigator Output (MOF-521-*o*F)

Current Stage and Iteration: 5-2

Status Evaluation: The results reported by the apprentice confirm the successful replication of the entire process, from synthesis to characterization. The reproducibility of our results, consistent with previously reported data, indicates that our synthesis, activation, and characterization protocols are reliable and robust.

Task Choice 1: To further validate our findings, it would be beneficial to review all our lab notes, experimental data, and results. This will help to ensure no minor discrepancies or errors are overlooked.
Begin by reviewing all your lab notes from every stage of the experiment.
Ensure the notes align with the final results reported for each stage.
Identify any inconsistencies or discrepancies and note them down for future reference and improvements.

Task Choice 2: Let's organize and document all the data and results we've gathered over the project. Proper documentation ensures that our findings are ready for reporting, publication, or future reference.
Compile all the raw data, processed data, and final results into a structured format.
Include all important details such as experimental conditions, equipment used, any anomalies or unique observations, and conclusions.
Ensure all data is correctly labeled and easily comprehensible.

Task Choice 3: As we have validated the reproducibility of our methods and results, it's time to compile all the findings into a final report. This report will detail all our research, findings, and conclusions, setting the stage for further studies or publications.
Begin with an abstract that summarizes the entire project and key findings.
Detail the methods used in each stage of the project in the 'Methods' section.
Present all findings and discuss them in the 'Results and Discussion' section.
Conclude the report with the implications of our findings, potential applications, and future directions.

**Figure S111.** Output generated by ChemNavigator during the 29th iteration, aimed at the synthesis and characterization of a novel aluminum-based MOF utilizing the BTB-*o*F linker.

## Final Summary (MOF-521-*m*F)

Our research project aimed to develop a novel Aluminum Metal-Organic Framework (MOF) named MOF-521-*m*F, using the BTB-*m*F linker, a derivative of the BTB linker. The research followed a systematic approach, divided into five stages.

During the first stage, we encountered the challenge of synthesizing the BTB-*m*F linker, as there were no established direct synthesis procedures in the literature. To overcome this, we derived a modified procedure from a similar BTB-*o*F linker synthesis method. Our adapted method involved palladium-catalyzed Suzuki-Miyaura cross-coupling reactions, followed by saponification and acidification. Once the BTB-*m*F linker was synthesized, we characterized it using 1H NMR spectroscopy. We compared the experimental chemical shifts with predicted ones, and the strong correlation confirmed the successful synthesis of the BTB-*m*F linker.

In the second stage, we proceeded to the high-throughput screening of the MOF and optimization of synthetic outcomes using Powder X-ray Diffraction (PXRD). Our initial synthetic conditions included a BTB-*m*F:Al ratio of 3:4, a formic acid to water ratio of 4:1, and a reaction temperature of 120°C. This combination yielded sharp PXRD peaks, indicative of a successful MOF formation. Subsequently, we tested reaction times (60, 72, and 84 hours), which revealed that high crystallinity single crystals could be produced irrespective of the reaction time. We also explored the effects of reaction temperature. We found that higher temperatures led to more powder formation, whereas lower ones resulted in fewer single crystals. We adjusted the BTB-*m*F:Al ratios to 3:3 and 3:5, and the reactions conducted at 120°C for 72 hours showed a decrease in crystallinity and broader PXRD peaks, suggesting a less desirable MOF structure. Finally, we altered the formic acid to water ratios to 3:1 and 5:1. These conditions produced single crystals, but the 5:1 condition resulted in smaller crystals and lower PXRD intensity, and the 3:1 condition resulted in some powder formation. Neither condition was as optimal as the original 4:1 ratio.

The third stage focused on the activation and determination of permanent porosity. We activated the MOF through a methanol solvent exchange, which preserved the original structure. Subsequent nitrogen gas sorption analysis revealed the microporous nature of the MOF, with a specific surface area of 1535 $m^2/g$, a total pore volume of 0.562 $cm^3/g$, and an average pore width of 10.9 Å. Further analysis of nitrogen adsorption-desorption isotherms presented a type I isotherm, lacking any significant hysteresis loop, which supported the stability and microporosity of the MOF.

The fourth stage involved a detailed structural analysis and characterization of MOF-521-*m*F. We used CHNS elemental analysis to verify the chemical composition, identifying the formula of Al3(μ-OH)1(HCOO)3(BTB-*m*F)1. Thermogravimetric analysis (TGA) demonstrated that the MOF was thermally stable up to 300°C. Single-Crystal X-ray Diffraction (SCXRD) analysis categorized the MOF under the P-62c (No. 190) space group, noting positional disorder at the peripheral phenyl rings of BTB. We also tested the MOF's stability in different solvents. We found potential instability in water, as indicated by the appearance of a formate peak at ~8.5 ppm in the 1H NMR spectrum of a water-exposed MOF sample, indicating hydrolysis. However, the MOF was stable in various organic solvents, including DMF, MeOH, and EtOH. Additionally, humidity stability tests revealed that the MOF maintained its structural integrity under humid conditions.

During the final stage, we executed a reproducibility check and validation of our protocols. We confirmed successful reproducibility of the MOF-521-*m*F synthesis by matching the PXRD patterns from the original and reproduced MOF. Further analysis confirmed similar structural and chemical characteristics, providing clear evidence that our methods for synthesizing and characterizing MOF-521-*m*F are robust and reproducible.

**Figure S112.** Illustration of the final summary developed by the Reticular ChemNavigator. This version of the summary was generated at the 26th iteration, marking the completion of Stage 5 interactions, which signifies the synthesis and study of MOF-521-*m*F. Previous versions of the summary have been omitted for clarity.

Reticular ChemNavigator Output (MOF-521-*m*F)

Current Stage and Iteration: 1-1

Status Evaluation: Given that the project is just starting, no task performance or feedback from the apprentice is available to evaluate. The first stage will focus on synthesizing the BTB-*m*F linker, a crucial component of the proposed MOF.

Task Choice 1: Begin by synthesizing the organic linker. This can be accomplished via a Suzuki-Miyaura cross-coupling reaction, following the literature for BTB linker synthesis. This will require preparing and purifying the necessary starting materials, setting up the reaction, monitoring the reaction progress, and purifying the product. Following this, the BTB-*m*F linker can be validated using 1H NMR spectroscopy, comparing the chemical shifts with those predicted from its structure.

Task Choice 2: A comprehensive literature review can be carried out to support the synthesis of the organic linker. This will involve locating and studying related works that document similar synthesis routes, particularly those involving similar linker derivatives. The gathered information can serve as a guide in setting up the reaction and troubleshooting potential challenges.

Task Choice 3: Procure all necessary reagents and set up the experimental workspace for the Suzuki-Miyaura cross-coupling reaction. Ensure all necessary safety measures are in place. This preparatory phase can be used to re-familiarize yourself with the equipment and reagents you will be using, as well as to assess whether any additional materials or reagents may be needed.

**Figure S113.** Output generated by ChemNavigator during the 1st iteration, aimed at the synthesis and characterization of a novel aluminum-based MOF utilizing the BTB-*m*F linker.

## Reticular ChemNavigator Output (MOF-521-*m*F)

Current Stage and Iteration: 1-2

Status Evaluation: The literature search did not yield fruitful results due to the complexity of the BTB-*m*F linker's name and the broadness of the search terms. As the BTB-*m*F linker might not be widely explored or explicitly mentioned in the literature, it has led to difficulties in finding relevant information.

Task Choice 1: Instead of using the compound's name, conduct a structure-based search using databases such as PubChem, Reaxys, or SciFinder. Draw the structure of the BTB-*m*F linker using their built-in structure editor and use the search function to find related synthetic methods. If direct synthesis information is not available, look for similar structures and their synthesis for possible adaptation.

Task Choice 2: If direct literature search fails, try to dissect the molecule into simpler segments and search for their synthesis individually. For instance, try looking up the synthesis of "terphenyl", "carboxylated phenyl compounds", and "fluorinated phenyl compounds". Combining these methods might give you an idea about the synthesis of the BTB-*m*F linker.

Task Choice 3: Connect with organic chemists or MOF synthesis groups on professional networks such as LinkedIn or ResearchGate. Post a drawn structure of the BTB-*m*F linker, explain your project, and ask for guidance on its synthesis. Sometimes, experts in the field may be able to provide unpublished insights or guide you towards relevant resources.

**Figure S114.** Output generated by ChemNavigator during the 2nd iteration, aimed at the synthesis and characterization of a novel aluminum-based MOF utilizing the BTB-*m*F linker.

## Reticular ChemNavigator Output (MOF-521-*m*F)

Current Stage and Iteration: 1-3

Status Evaluation: The structure-based search, suggested in the last task, yielded valuable insights for the synthesis of a similar compound, BTB-*o*F. This success validates the strategy of searching for similar structures when direct synthesis information is not available. The synthesis procedure for BTB-*o*F, involving Suzuki-Miyaura cross-coupling reactions, can potentially be adapted for BTB-*m*F, requiring careful modifications considering the difference in the position of the fluorine atom.

Task Choice 1: Begin the synthesis of the BTB-*m*F linker by adapting the procedures used for BTB-*o*F synthesis. Start by synthesizing Compound 1 (a 1,3,5-tribromobenzene derivative), following the provided protocol. Pay careful attention to the reaction conditions and note down any observations. After synthesizing Compound 1, characterize it using 1H NMR spectroscopy to ensure its correct formation.

Task Choice 2: Prior to the synthesis, perform a thorough risk assessment and ensure that all required safety measures are in place. Review the procedures for the synthesis of BTB-*o*F. Identify the reagents, equipment, and analytical methods required. Order any missing items and prepare a detailed plan for the synthesis of BTB-*m*F, indicating where you expect the procedure to deviate from the BTB-*o*F synthesis.

Task Choice 3: Before beginning the synthesis of the BTB-*m*F linker, you can also conduct a preliminary computational study to verify if the Suzuki-Miyaura cross-coupling reaction is indeed feasible given the difference in the position of the fluorine atom between BTB-*o*F and BTB-*m*F. Use molecular modeling software to simulate the reaction and predict potential products. This step can help us anticipate challenges in the actual synthesis.

**Figure S115.** Output generated by ChemNavigator during the 3rd iteration, aimed at the synthesis and characterization of a novel aluminum-based MOF utilizing the BTB-*m*F linker.

## Reticular ChemNavigator Output (MOF-521-*m*F)

Current Stage and Iteration: 1-4

Status Evaluation: The synthesis of the BTB-*m*F linker was successful, suggesting the approach to adapt from the literature procedure for BTB-*o*F was a correct decision. The change in eluent for chromatography demonstrated the apprentice's ability to adjust experimental conditions based on real-time observations. Further analysis of the NMR data is needed to verify the successful synthesis.

Task Choice 1: Analyze the 1H NMR spectrum of your synthesized BTB-*m*F linker. You can use software such as MestReNova or Bruker's TopSpin to assign and integrate the peaks. Compare the observed chemical shifts and splitting patterns with the expected ones based on the structure of BTB-*m*F. Once done, record your findings and their implications.

Task Choice 2: Given that we have successfully synthesized the BTB-*m*F linker, we can prepare for the high-throughput screening stage. This would involve setting up synthetic conditions for MOF formation. However, before we proceed, it's important to be sure about the successful synthesis of our BTB-*m*F linker. As such, you can perform additional characterization using 13C NMR and/or Fourier-transform infrared spectroscopy (FTIR). These techniques will provide further structural information and help confirm the successful synthesis of the BTB-*m*F linker.

Task Choice 3: If you are confident about the success of the linker synthesis and want to confirm it further, consider running a mass spectrometry analysis. A High-Resolution Mass Spectrometry (HRMS) will provide precise molecular mass information, which can confirm the molecular formula of the linker. If HRMS is not available, even an Electrospray Ionization Mass Spectrometry (ESI-MS) would give helpful information. Remember, the observed mass should correspond to the deprotonated form of the linker ([M-H]-) due to the presence of carboxylic acid groups.

**Figure S116.** Output generated by ChemNavigator during the 4$^{th}$ iteration, aimed at the synthesis and characterization of a novel aluminum-based MOF utilizing the BTB-*m*F linker.

## Reticular ChemNavigator Output (MOF-521-*m*F)

Current Stage and Iteration: 1-5

Status Evaluation: The NMR analysis confirmed the successful synthesis of the BTB-*m*F linker, as the experimental and predicted chemical shifts showed significant correlation. The minor differences observed can be attributed to variations in the environment of the protons and the use of DMSO as the solvent. With the successful synthesis and confirmation of the BTB-*m*F linker, we are now ready to proceed to the next stage of MOF synthesis.

Task Choice 1: Based on the successful synthesis and confirmation of the BTB-*m*F linker, if you feel confident with the results and ready to proceed, you can declare, "I'm ready to move to the next stage."

Task Choice 2: Before moving on to the next stage, consider verifying the synthesis of the BTB-*m*F linker through additional methods, such as Fourier Transform Infrared Spectroscopy (FTIR) and Mass Spectrometry. These methods will provide a comprehensive confirmation of the chemical structure of the synthesized linker.

Task Choice 3: Given that the BTB-*m*F linker has been successfully synthesized, the alternative task is to optimize the synthesis process further. You can experiment with various catalysts, temperatures, and solvents to maximize the yield and purity of the BTB-*m*F linker. Once optimization is completed, you can then declare readiness to move to the next stage.

**Figure S117.** Output generated by ChemNavigator during the 5$^{th}$ iteration, aimed at the synthesis and characterization of a novel aluminum-based MOF utilizing the BTB-*m*F linker.

## Reticular ChemNavigator Output (MOF-521-*m*F)

Current Stage and Iteration: 2-1

Status Evaluation: The apprentice successfully synthesized and validated the BTB-*m*F linker, confirming readiness to proceed to the second stage. This stage will focus on high-throughput screening of the MOF and optimization of synthetic outcomes.

Task Choice 1: Start the high-throughput screening of the MOF by first attempting the synthesis of the MOF using the BTB-*m*F linker and Aluminum source. Begin with a molar ratio of BTB-*m*F:Al = 3:4, inspired by similar work on the BTB-*o*F MOF. Use a formic acid to water ratio of 4:1, similar to the procedure applied in the previous BTB-*o*F project. Keep the reaction at 120°C. After 72 hours, terminate the reaction, wash the precipitate with DMF, and then with methanol. Dry the obtained MOF under vacuum and then analyze the product using PXRD.

Task Choice 2: Explore the effect of changing reaction temperature on the synthesis of the MOF. Keeping the other conditions constant (BTB-*m*F:Al = 3:4 and formic acid to water ratio = 4:1), vary the reaction temperature in three trials, at 100°C, 120°C, and 140°C. Each reaction should run for 72 hours. After each trial, wash the precipitate with DMF, then with methanol, and dry the obtained MOF under vacuum. Use PXRD to analyze the products of each trial.

Task Choice 3: As an alternative route, investigate the effect of changing the BTB-*m*F:Al molar ratio on the synthesis of the MOF. Set the reaction temperature to 120°C and maintain a formic acid to water ratio of 4:1. Try BTB-*m*F:Al ratios of 3:4, 1:1, and 4:3. Run each reaction for 72 hours. After each trial, wash the precipitate with DMF, then with methanol, and dry the obtained MOF under vacuum. Use PXRD to analyze the products.

**Figure S118.** Output generated by ChemNavigator during the $6^{th}$ iteration, aimed at the synthesis and characterization of a novel aluminum-based MOF utilizing the BTB-*m*F linker.

## Reticular ChemNavigator Output (MOF-521-*m*F)

Current Stage and Iteration: 2-2

Status Evaluation: The successful outcome of the first iteration in stage 2 could be attributed to the conditions used, inspired by the BTB-*o*F MOF synthesis, including the BTB-*m*F:Al ratio, the formic acid to water ratio, and the reaction temperature. The sharp and well-aligned PXRD peaks suggest that we have synthesized the MOF successfully under these conditions.

Task Choice 1: Given the promising PXRD results, it would be interesting to further optimize the synthetic conditions. Test the influence of the reaction time by performing three parallel experiments, using 60 hours, 72 hours, and 84 hours as reaction times. Continue to use the BTB-*m*F:Al ratio of 3:4 and the formic acid to water ratio of 4:1, keeping the reaction at 120°C. Terminate the reactions, wash the precipitates with DMF, then with methanol, and dry the obtained MOFs under vacuum. Finally, analyze the products using PXRD and note any changes in the pattern or intensity of the peaks.

Task Choice 2: Confirm the composition and phase of the synthesized MOF from the first synthesis iteration. Using the MOF from the last experiment, perform elemental analysis (CHNS) to confirm the ratio of carbon, hydrogen, nitrogen, and sulfur. Also, carry out Fourier-transform infrared spectroscopy (FTIR) to verify the presence of the organic linker and metal ions in the MOF. Observe the presence of peaks corresponding to the functional groups in the organic linker and the metal-ligand bond to confirm the structure.

Task Choice 3: To further validate the MOF formation, consider the use of scanning electron microscopy (SEM) and transmission electron microscopy (TEM) to characterize the morphology and size of the synthesized MOF. Prepare the samples according to standard protocols for SEM and TEM analysis, carry out the imaging, and interpret the results. Look for a uniform structure and size distribution that would further confirm successful MOF synthesis.

**Figure S119.** Output generated by ChemNavigator during the 7$^{th}$ iteration, aimed at the synthesis and characterization of a novel aluminum-based MOF utilizing the BTB-*m*F linker.

## Reticular ChemNavigator Output (MOF-521-*m*F)

Current Stage and Iteration: 2-3

Status Evaluation: The consistent formation of single crystals and similar PXRD patterns across varied reaction times suggests that the MOF formation is relatively insensitive to this parameter under the tested conditions. This implies the robustness of our synthesis procedure, which is an encouraging result for this stage.

Task Choice 1: Now that we have established that reaction times between 60 to 84 hours do not significantly affect the crystal formation and their PXRD patterns, let's try extending the reaction times to see if it still holds true. Carry out the synthesis at 96 and 108 hours reaction times, maintaining the same BTB-*m*F:Al ratio, formic acid to water ratio, and temperature. After the reaction, wash the precipitates with DMF, then with methanol, and dry the obtained MOFs under vacuum. Finally, perform PXRD analysis to examine the crystallinity and compare the peak patterns.

Task Choice 2: Given the promising results we've had so far, it might be beneficial to start investigating the effect of the reaction temperature. Maintain the current BTB-*m*F:Al ratio, formic acid to water ratio, and reaction time, but run parallel reactions at different temperatures: try 110°C and 130°C, alongside a control reaction at 120°C. Afterward, perform the usual cleaning procedure and analyze the products using PXRD.

Task Choice 3: Another parameter to explore could be the ratio of BTB-*m*F:Al. Keeping the reaction temperature at 120°C, the formic acid to water ratio at 4:1, and the reaction time at 72 hours, let's adjust the BTB-*m*F:Al ratio to 3:3 and 4:4, respectively. After the reaction, proceed with the usual washing, drying, and PXRD analysis. This task aims to study the effect of linker to metal ratio on the quality and quantity of the obtained MOF.

**Figure S120.** Output generated by ChemNavigator during the 8$^{th}$ iteration, aimed at the synthesis and characterization of a novel aluminum-based MOF utilizing the BTB-*m*F linker.

## Reticular ChemNavigator Output (MOF-521-*m*F)

Current Stage and Iteration: 2-4

Status Evaluation: The different reaction temperatures examined have yielded differing results, suggesting a direct impact of temperature on both the yield and the nature of the MOF obtained. The higher temperature of 130°C seems to favor the production of powders rather than single crystals, while the lower temperature of 110°C leads to a reduced yield.

Task Choice 1: Now that we have identified that reaction temperature influences the yield and product form, the next logical step would be to investigate a temperature range that has not been explored. Carry out reactions at 115°C and 125°C, maintaining the BTB-*m*F:Al ratio of 3:4, a formic acid to water ratio of 4:1, and a reaction time of 72 hours. Afterward, clean the products as per usual and analyze using PXRD.

Task Choice 2: In light of the results from different reaction temperatures, it might be helpful to conduct another experiment at 120°C for a longer duration to determine if there is a trade-*o*Ff between reaction time and yield. Set up an experiment that runs for 96 hours while maintaining the BTB-*m*F:Al ratio, formic acid to water ratio, and temperature. Then, clean the products and perform PXRD analysis.

Task Choice 3: Given that temperature is influencing the MOF yield and structure, another potential variable to test could be the BTB-*m*F:Al ratio. Keep the temperature and reaction time the same, but modify the BTB-*m*F:Al ratio to 3:3 and 3:5. Following the standard cleaning procedure, conduct PXRD analysis to examine the resulting MOF structure.

**Figure S121.** Output generated by ChemNavigator during the 9$^{th}$ iteration, aimed at the synthesis and characterization of a novel aluminum-based MOF utilizing the BTB-*m*F linker.

Reticular ChemNavigator Output (MOF-521-*m*F)

Current Stage and Iteration: 2-5

Status Evaluation: The decrease in crystallinity and wider PXRD peaks indicate less optimal MOF structures when altering the BTB-*m*F:Al ratios to 3:3 and 3:5, compared to the previously established 3:4 ratio. This could suggest that the BTB-*m*F:Al ratio significantly impacts the MOF's structure and crystallinity, making it a crucial parameter in the synthesis of our desired MOF.

Task Choice 1: Continue to explore the optimal BTB-*m*F:Al ratio. Run the experiment again with a BTB-*m*F:Al ratio of 3:4, maintaining the other parameters (temperature of 120°C for 72 hours) to ensure the reproducibility of our previous results. After synthesizing the MOF, clean and dry the product as usual. Examine the structure using PXRD analysis and compare the data to the previous results.

Task Choice 2: Considering the reduced crystallinity from the recent experiment, another parameter worth exploring could be the formic acid to water ratio. Keeping the BTB-*m*F:Al ratio constant at 3:4, run experiments with different formic acid to water ratios, such as 3:1 and 5:1, and keep the reaction temperature at 120°C for 72 hours. As usual, after the synthesis, clean, dry, and conduct PXRD analysis on the resulting MOF structure.

Task Choice 3: Given that the previous experiment resulted in less desirable structures, it could be beneficial to return to previously successful conditions (BTB-*m*F:Al ratio of 3:4, formic acid to water ratio of 4:1, and a reaction temperature of 120°C). After confirming the ideal conditions, it would be helpful to conduct several runs under these conditions to produce a large enough batch of MOF for further characterization and analysis.

**Figure S122.** Output generated by ChemNavigator during the 10$^{th}$ iteration, aimed at the synthesis and characterization of a novel aluminum-based MOF utilizing the BTB-*m*F linker.

Reticular ChemNavigator Output (MOF-521-*m*F)

Current Stage and Iteration: 2-6

Status Evaluation: The series of experiments conducted in stage 2 have given us an in-depth understanding of how synthetic parameters affect the formation and properties of the MOF. Despite some less than optimal results with the formic acid to water ratio adjustments, we've collected substantial data to guide the synthesis of our MOF.

Task Choice 1: If you believe we've gathered enough data in this stage and are ready to proceed, you can declare, "I'm ready to move to the next stage." This will allow us to progress to stage 3, which involves activation and the determination of permanent porosity of our synthesized MOF.

Task Choice 2: If you feel there's still some fine-tuning to be done in stage 2, we can revisit the reaction temperature. Set up three experiments at 115°C, 125°C, and 135°C, while keeping the BTB-*m*F:Al ratio at 3:4 and the formic acid to water ratio at 4:1 for 72 hours. Clean, dry, and conduct PXRD analysis on the resulting MOF structure post synthesis.

Task Choice 3: Alternatively, we could further investigate the BTB-*m*F:Al ratio. Conduct experiments at ratios of 3:4.5, 3:3.5, and 3:4.25, with a consistent formic acid to water ratio at 4:1 and the reaction temperature at 120°C for 72 hours. Clean, dry, and conduct PXRD analysis on the resulting MOF structure post synthesis.

**Figure S123.** Output generated by ChemNavigator during the 11$^{th}$ iteration, aimed at the synthesis and characterization of a novel aluminum-based MOF utilizing the BTB-*m*F linker.

Reticular ChemNavigator Output (MOF-521-*m*F)

Current Stage and Iteration: 3-1

Status Evaluation: The apprentice declared readiness to proceed to the next stage after successful synthesis and high-throughput screening of the MOF. The optimal conditions for MOF synthesis have been determined, providing a sound basis for moving forward.

Task Choice 1: Begin stage 3 by activating the synthesized MOF. Start the activation process through solvent exchange using methanol. First, submerge the MOF crystals in methanol for 24 hours, changing the methanol every 8 hours. After 24 hours, filter off the methanol and submerge the crystals in fresh methanol for another 24 hours, changing it every 8 hours. Finally, filter off the methanol and dry the MOF under vacuum for 12 hours at room temperature to remove any residual solvent.

Task Choice 2: For a gentler activation method, start the solvent exchange using dichloromethane (DCM). First, immerse the MOF crystals in DCM for 24 hours, changing the DCM every 8 hours. After 24 hours, filter off the DCM and immerse the crystals in fresh DCM for another 24 hours, changing it every 8 hours. Finally, filter off the DCM and dry the MOF under vacuum for 12 hours at room temperature to remove any residual solvent.

Task Choice 3: To explore activation under elevated temperatures, perform the solvent exchange using methanol, as in Task 1, but after the final filtration, dry the MOF under vacuum at 40°C for 12 hours. Monitor the MOF during drying to ensure that it does not decompose at this temperature.

**Figure S124.** Output generated by ChemNavigator during the 12$^{th}$ iteration, aimed at the synthesis and characterization of a novel aluminum-based MOF utilizing the BTB-*m*F linker.

Reticular ChemNavigator Output (MOF-521-*m*F)

Current Stage and Iteration: 3-2

Status Evaluation: The first iteration of stage three went smoothly. The apprentice meticulously followed the instructions for the activation of the MOF through methanol solvent exchange. The unchanged structure and absence of MOF loss during the process indicate successful activation, paving the way for further analysis and determination of the permanent porosity.

Task Choice 1: Let's proceed to determine the porosity of the activated MOF. Start by conducting nitrogen gas sorption analysis at 77 K. This analysis will help us measure the microporosity of our MOF. First, prepare the sample by degassing it under vacuum at 120°C for 12 hours. Then, use a reliable gas sorption analyzer to measure the nitrogen gas adsorption and desorption isotherms. By analyzing these isotherms, we can calculate the Brunauer–Emmett–Teller (BET) surface area, total pore volume, and average pore width of our MOF.

Task Choice 2: As an alternative task, perform a hysteresis analysis on the activated MOF to investigate its stability and porosity. For this, you will need to observe the hysteresis loop in the nitrogen sorption isotherms at 77 K. A narrow hysteresis loop, particularly at high relative pressures (P/P0 > 0.4), may indicate the presence of larger mesopores or macropores. The type and shape of the hysteresis loop can provide further insights into the pore structure and the flexibility of our MOF.

Task Choice 3: Before proceeding with the sorption analysis, it may be beneficial to re-confirm the MOF's crystal structure after activation. For this, conduct a post-activation Powder X-ray diffraction (PXRD) analysis. This will help us ensure that the MOF structure is intact and unchanged after the activation process. Follow the standard PXRD protocol for MOFs, and compare the resulting pattern to that obtained before activation. If there are significant deviations, it could indicate issues with the activation process. If the patterns align well, we can confidently proceed with the porosity determination.

**Figure S125.** Output generated by ChemNavigator during the 13$^{th}$ iteration, aimed at the synthesis and characterization of a novel aluminum-based MOF utilizing the BTB-*m*F linker.

Reticular ChemNavigator Output (MOF-521-*m*F)

Current Stage and Iteration: 3-3

Status Evaluation: The results of the nitrogen gas sorption analysis indicate successful measurement of the MOF's porosity, with surface area, pore volume, and pore width values aligning closely with expectations based on similar works. These findings confirm that the MOF exhibits microporosity, which is essential for various applications such as gas storage or separation.

Task Choice 1: To further validate the microporous nature and stability of our MOF, it would be prudent to analyze the hysteresis of the isotherms obtained from the nitrogen gas sorption analysis. This can be achieved by plotting the nitrogen gas adsorption-desorption isotherms and determining the type of hysteresis loop formed. From this plot, determine if the isotherms close at the same relative pressure (P/P0) point, indicating reversible adsorption-desorption processes and confirming microporosity. If the isotherms don't close at the same point, this could mean that our MOF exhibits mesoporosity or a combination of micro- and mesoporosity, which would need further investigation.

Task Choice 2: Given that we've confirmed the microporosity of our MOF, the next logical step would be to determine the material's thermal stability. This could be achieved through Thermogravimetric Analysis (TGA). Prepare the MOF sample and perform TGA under an inert atmosphere (like nitrogen) at a heating rate of 10°C per minute, up to a temperature of 800°C. Make sure to note the temperature at which the most significant weight loss occurs, as this will indicate the temperature of decomposition and thus, the thermal stability of the MOF.

Task Choice 3: If you feel confident about the data gathered so far, you may suggest proceeding to stage four, which involves a detailed structural analysis and characterization of the MOF. This would involve techniques such as X-Ray Diffraction (XRD) analysis, Fourier Transform Infrared Spectroscopy (FTIR), and potentially others, depending on the specific information you want to glean about the MOF's structure and properties. If you feel ready to proceed, please state, "I'm ready to move to the next stage."

**Figure S126.** Output generated by ChemNavigator during the $14^{th}$ iteration, aimed at the synthesis and characterization of a novel aluminum-based MOF utilizing the BTB-*m*F linker.

Reticular ChemNavigator Output (MOF-521-*m*F)

Current Stage and Iteration: 3-4

Status Evaluation: The apprentice's examination of the nitrogen adsorption-desorption isotherms yielded a type I isotherm without a noticeable hysteresis loop. This outcome further reinforces the microporosity of our MOF and indicates the stability of adsorption-desorption processes in the MOF. The absence of hysteresis, which represents the reversible nature of adsorption-desorption, is another positive confirmation of the MOF's microporous nature.

Task Choice 1: Now that we have extensively studied and confirmed the porosity and stability of our MOF, it might be time to conclude this stage and move to the next one, where we will undertake a detailed structural analysis and characterization of the MOF. If you believe we have gathered enough data and insights in this stage, you can state, "I'm ready to move to the next stage."

Task Choice 2: However, if you feel the need for further validation of our MOF's microporosity and stability, we could conduct a comparative nitrogen sorption analysis with a standard microporous material, such as Zeolite. This would allow us to evaluate the similarities and differences in their sorption behavior and further confirm our MOF's microporosity.

Task Choice 3: An alternative task would be to examine the MOF's potential for gas uptake. You could conduct a series of adsorption-desorption experiments with different gases (such as CO2, CH4, H2) at various pressures and temperatures. The data obtained would give us an idea of how the MOF performs in terms of gas storage and separation, which are key characteristics for practical applications of MOFs.

**Figure S127.** Output generated by ChemNavigator during the 15$^{th}$ iteration, aimed at the synthesis and characterization of a novel aluminum-based MOF utilizing the BTB-*m*F linker.

## Reticular ChemNavigator Output (MOF-521-*m*F)

Current Stage and Iteration: 4-1

Status Evaluation: The apprentice's readiness to proceed to the next stage indicates the successful completion of the MOF's activation and the confirmation of its porosity. The nitrogen gas sorption analysis and the analysis of nitrogen adsorption-desorption isotherms provided comprehensive insights into the MOF's structure, justifying the transition to the next stage, which focuses on further detailed structural characterization.

Task Choice 1: Let's initiate the fourth stage with a detailed analysis of the MOF's chemical composition. Perform a CHNS elemental analysis. Ensure that the MOF sample is completely dried before the analysis to prevent any errors. Compare the obtained results with the theoretical composition based on the BTB-*m*F and Aluminum source. This would confirm the precise chemical composition of our MOF.

Task Choice 2: We could begin our fourth stage by examining the thermal stability of our MOF. Conduct a Thermogravimetric Analysis (TGA) under a nitrogen atmosphere. The sample is to be heated from room temperature to 600°C at a rate of 10°C per minute. Observe the thermal decomposition pattern and temperature to assess the thermal stability of our MOF.

Task Choice 3: Given our MOF's potential for a variety of applications, it's important to understand its chemical stability. To start this stage, we can test our MOF's stability in various solvents. Immerse the MOF in solvents such as DMF, methanol, ethanol, and water, and leave it for 24 hours at room temperature. Afterward, collect the MOF samples, dry them completely, and conduct a PXRD analysis. This will provide insights into our MOF's resistance to solvent exposure and its structural integrity.

**Figure S128.** Output generated by ChemNavigator during the 16$^{th}$ iteration, aimed at the synthesis and characterization of a novel aluminum-based MOF utilizing the BTB-*m*F linker.

## Reticular ChemNavigator Output (MOF-521-*m*F)

Current Stage and Iteration: 4-2

Status Evaluation: The close correlation between the theoretical and experimental results in the CHNS elemental analysis indicates the successful incorporation of the BTB-*m*F linker into the MOF framework and validates the proposed formula. This success allows us to move on to the next task within the fourth stage: analyzing the phase purity and the thermal and chemical stability of our MOF.

Task Choice 1: To analyze the phase purity of the MOF, perform a PXRD analysis. Prepare a finely grounded sample of the MOF, place it on a suitable sample holder, and perform the PXRD measurement. Compare the experimental pattern with the simulated one derived from the known MOF structure. If the patterns match significantly, this would imply good phase purity of the MOF.

Task Choice 2: Carry out a thermogravimetric analysis (TGA) to evaluate the thermal stability of our MOF. Heat the MOF sample under an inert nitrogen atmosphere from room temperature up to 800°C at a rate of 10°C/min. Plot a graph of weight percentage against temperature. The TGA profile will provide us with the thermal decomposition pattern of the MOF, indicating its thermal stability.

Task Choice 3: Perform chemical stability tests. Submerge the MOF in various solvents (like water, DMF, methanol, and ethanol) separately for 24 hours. Afterwards, conduct PXRD analysis of the recovered samples. A comparison between the PXRD patterns before and after the treatment will reveal the chemical stability of our MOF in different solvents.

**Figure S129.** Output generated by ChemNavigator during the 17$^{th}$ iteration, aimed at the synthesis and characterization of a novel aluminum-based MOF utilizing the BTB-*m*F linker.

## Reticular ChemNavigator Output (MOF-521-*m*F)

Current Stage and Iteration: 4-3

Status Evaluation: The TGA results indicate good thermal stability of MOF-521-*m*F up to 300°C, which is in line with our expectations and the stability of MOF-521-*o*F. This thermal stability is indicative of the robustness of the MOF framework. The decomposition at around 300°C and subsequent stability up to around 630°C may correspond to the loss of coordinated and/or adsorbed solvent molecules, followed by the breakdown of organic linkers.

Task Choice 1: We should now assess the chemical stability of our MOF. To do this, immerse the MOF sample in different solvents such as dimethylformamide (DMF), methanol, and ethanol for 24 hours at room temperature. After 24 hours, collect the MOF from each solvent and perform PXRD measurements. Compare the PXRD patterns before and after immersion to identify any significant changes in the structure.

Task Choice 2: Given that we have obtained a good thermal stability up to 300°C, we can now move forward to a Single-Crystal X-ray Diffraction (SCXRD) analysis. This will allow us to elucidate the precise atomic structure of our MOF. Prepare a high-quality single crystal of MOF-521-*m*F and set it up for SCXRD analysis, taking care to ensure the crystal remains in good condition.

Task Choice 3: To verify the structural stability of our MOF, conduct a pH stability test. Prepare a series of solutions with varying pH values (1, 3, 7, 9, 11), immerse the MOF samples in these solutions, and leave them for 24 hours. After this period, collect the MOF samples and perform PXRD. Compare the patterns before and after the test to evaluate the stability under different pH conditions.

**Figure S130.** Output generated by ChemNavigator during the 18$^{th}$ iteration, aimed at the synthesis and characterization of a novel aluminum-based MOF utilizing the BTB-*m*F linker.

## Reticular ChemNavigator Output (MOF-521-*m*F)

Current Stage and Iteration: 4-4

Status Evaluation: The SCXRD analysis has provided crucial insights into the structure of the MOF. The structure confirmed the presence of rod SBUs, with AlO6 octahedra shared via μ2-OH respectively, two corners bridged by carboxylate from the BTB-*m*F linker, and the last two corners capped by formates. The successful analysis indicates that we're making substantial progress within stage 4, despite the observed positional disorder.

Task Choice 1: Based on the SCXRD data, we can examine the crystal structure more closely to understand the positional disorder in the peripheral phenyl rings of BTB. Use a molecular visualization software such as PyMOL or Mercury to visualize and analyze the structure. Pay close attention to the disordered regions, as well as any unusual bonding or electron density. Share your findings and interpretations for further discussion.

Task Choice 2: Given the structural details revealed by SCXRD analysis, the next logical step would be to determine the chemical stability of our MOF. Immerse a sample of MOF-521-*m*F in solvents of various polarities (e.g., DMF, methanol, ethanol, and water) for a specific period, followed by an examination of the MOF's PXRD patterns. This will reveal any potential decomposition or framework instability in these solvents.

Task Choice 3: To further validate the structure of our MOF, an additional round of SCXRD could be beneficial, specifically targeting the refinement of the positional disorder at the peripheral phenyl rings. This will allow us to have a more comprehensive understanding of the structure, as well as investigate the cause and implications of this disorder. After obtaining the data, use suitable software to refine the structure and assess the resulting changes.

**Figure S131.** Output generated by ChemNavigator during the 19th iteration, aimed at the synthesis and characterization of a novel aluminum-based MOF utilizing the BTB-*m*F linker.

## Reticular ChemNavigator Output (MOF-521-*m*F)

Current Stage and Iteration: 4-5

Status Evaluation: The results from the chemical stability test suggest that our MOF-521-*m*F is stable in various solvents, with potential instability in water. This observation aligns with the task objectives and indicates that the apprentice followed the procedure meticulously. The stability in non-aqueous solvents is promising; however, the potential instability in water might limit its applications and should be further investigated.

Task Choice 1: To get a more detailed understanding of the possible water instability, I suggest that you conduct a pH stability test. Prepare solutions with varying pH values (2, 4, 6, 8, and 10) and immerse small samples of the MOF in each for 24 hours. After drying the samples, perform PXRD to observe any peak changes which would indicate instability. Document and analyze your results.

Task Choice 2: Given the observed potential instability in water, we could perform a 1H NMR spectroscopy on a water-exposed MOF sample. Dissolve a sample of the MOF-521-*m*F in D2O, then run 1H NMR analysis. The appearance of a peak around 8 ppm would suggest the formation of formate, indicating hydrolysis has occurred.

Task Choice 3: In addition to the stability tests, further characterizing the MOF structure might be beneficial. Conduct a solid-state 13C NMR to examine the carbon environments in the framework. Prepare the sample in a rotor and perform the analysis. Be sure to include a spinning sideband analysis, and compare the result to the predicted shifts. This can provide more information about the carbon environments in the framework and any changes that might have occurred due to solvent exposure.

**Figure S132.** Output generated by ChemNavigator during the 20$^{th}$ iteration, aimed at the synthesis and characterization of a novel aluminum-based MOF utilizing the BTB-*m*F linker.

Reticular ChemNavigator Output (MOF-521-*m*F)

Current Stage and Iteration: 4-6

Status Evaluation: It seems the MOF-521-*m*F does not readily dissolve in D2O, which is preventing the execution of 1H NMR spectroscopy to check for hydrolysis. The use of NaOD in D2O was proposed as a solution since NaOD can act as a strong base, facilitating the dissolution of the MOF sample.

Task Choice 1: Let's follow the suggestion of using sodium deuteroxide (NaOD) in D2O to dissolve the MOF. Weigh 10 mg of the MOF-521-*m*F sample and add it to a clean NMR tube. Add a solution of 0.1M NaOD in D2O to the NMR tube. This strong base should assist in breaking down the MOF structure and facilitate dissolution. If required, apply gentle heat or sonication to aid the dissolution process. Once the MOF is dissolved, perform 1H NMR spectroscopy, and observe the spectrum around 8 ppm for potential formate peaks indicating hydrolysis.

Task Choice 2: Alternatively, we could opt to investigate the pH stability of our MOF. Prepare a range of buffered aqueous solutions with varying pH levels (e.g., pH 2, 4, 6, 8, 10, 12). Introduce 10 mg of MOF-521-*m*F to each solution and allow them to interact for 24 hours. Then, collect PXRD data on the samples. The analysis of the PXRD data will indicate whether the MOF is stable under different pH conditions, and potentially provide an insight into the observed instability in water.

Task Choice 3: As an alternative, we can execute a solid-state NMR (SSNMR) on the water-exposed MOF. SSNMR does not require the MOF to be in solution and thus circumvents the dissolution problem. This technique will allow the observation of the chemical environment around each atom in the solid state, and if hydrolysis has occurred, changes in the spectra should be observable. For this, prepare a water-exposed MOF sample and perform 1H, 13C, and 27Al SSNMR. After obtaining the spectra, examine for any new peaks or changes in the existing peaks, which might indicate a chemical transformation due to water exposure.

**Figure S133.** Output generated by ChemNavigator during the 21st iteration, aimed at the synthesis and characterization of a novel aluminum-based MOF utilizing the BTB-*m*F linker.

Reticular ChemNavigator Output (MOF-521-*m*F)

Current Stage and Iteration: 4-7

Status Evaluation: The latest task attempted to address the issue of the MOF not dissolving in D2O by using sodium deuteroxide (NaOD). As expected, the strong base facilitated the dissolution of the MOF and enabled us to collect 1H NMR spectroscopy data. The formate peak at ~8.5 ppm confirmed the occurrence of hydrolysis, aligning with the potential instability observed in the water stability test. Also, the MOF exhibited stability in various organic solvents including DMF, MeOH, and EtOH.

Task Choice 1: We should evaluate the pH stability of the MOF. Please prepare three pH buffer solutions at pH 3, 7, and 11. Add about 10 mg of MOF to each buffer and let them sit for 24 hours. After the time has elapsed, collect the MOFs by filtration and dry them in an oven at 60°C. Then, compare the PXRD pattern of the MOF samples exposed to different pH values with the original MOF PXRD pattern. This will provide insight into the behavior of our MOF under acidic, neutral, and basic conditions.

Task Choice 2: We should conduct a Fourier-Transform Infrared Spectroscopy (FTIR) on MOF-521-*m*F to examine the functional groups present and confirm the MOF's chemical structure. To do this, grind the MOF into a fine powder and mix it with KBr at a ratio of 1:100. Compress this mixture into a thin pellet and record the IR spectrum in the range of 4000-400 $cm^{-1}$. Compare the resulting data with the expected bands from the BTB-*m*F linker to confirm the presence of the organic linker in the MOF structure.

Task Choice 3: Given the sensitivity of our MOF to water, we should perform a humidity stability test. Please place a sample of the MOF in a desiccator with saturated salt solutions to create relative humidity environments of 75%, 85%, and 95%. After a week, remove the samples, dry them, and then analyze the MOF using PXRD to observe any changes that might have occurred due to humidity exposure.

**Figure S134.** Output generated by ChemNavigator during the 22$^{nd}$ iteration, aimed at the synthesis and characterization of a novel aluminum-based MOF utilizing the BTB-*m*F linker.

## Reticular ChemNavigator Output (MOF-521-*m*F)

Current Stage and Iteration: 4-8

Status Evaluation: The results from the humidity stability tests suggest that the MOF structure is not disrupted by moisture in the air. Despite the MOF's water instability observed earlier, the lack of peak changes in the PXRD analysis at varying relative humidity levels indicates that the MOF has a certain level of air stability. It seems that water in the liquid form disrupts the MOF structure more significantly than water vapor, which is an important observation for real-world applications of MOF-521-*m*F.

Task Choice 1: Given the structural stability of the MOF under humidity, we should further understand its stability in different pH environments. Prepare solutions with pH 2, 4, 7, 10, and 12 using appropriate buffers. Expose the MOF to these solutions for 24 hours, then dry and analyze the MOF using PXRD to observe any changes that might have occurred due to the pH exposure.

Task Choice 2: We should conduct further studies on the potential hydrolysis of the MOF in water. Prepare solutions of the MOF in D2O at different concentrations, and carry out 1H NMR spectroscopy using sodium deuteroxide (NaOD). Monitor the changes in the formate peak at ~8.5 ppm to understand the progression of hydrolysis.

Task Choice 3: Now that we have extensively characterized the MOF's stability in various conditions, it might be appropriate to check the reproducibility of our synthesis and characterization protocols. If you believe we have gathered sufficient information about the MOF's stability and characteristics, please indicate whether you're ready to move to the final stage, stage 5, for reproducibility check and final validation.

**Figure S135.** Output generated by ChemNavigator during the 23$^{rd}$ iteration, aimed at the synthesis and characterization of a novel aluminum-based MOF utilizing the BTB-*m*F linker.

## Reticular ChemNavigator Output (MOF-521-*m*F)

Current Stage and Iteration: 5-1

Status Evaluation: The human apprentice has indicated readiness to move forward to the final stage. Given the extensive stability tests and the findings about the MOF's behavior in various conditions, this decision aligns with the project progression. Hence, we are now prepared to test the reproducibility of our synthesis and characterization protocols.

Task Choice 1: Begin by reproducing the synthesis of our MOF-521-*m*F. Follow the optimized conditions identified in stage two, which is a BTB-*m*F:Al ratio of 3:4, a formic acid to water ratio of 4:1, and a reaction time of 72 hours at 120°C. After the synthesis, perform PXRD analysis on the reproduced MOF. Compare the PXRD pattern with the previous result to confirm the reproducibility of the synthesis.

Task Choice 2: Concurrently with the synthesis, you could start reproducing the BTB-*m*F linker synthesis as well, following the same procedure as in stage one. Once the linker is synthesized, validate it using 1H NMR spectroscopy. Compare the obtained chemical shifts with the previously recorded ones to confirm if the synthesis is reproducible.

Task Choice 3: While waiting for the MOF synthesis to complete, you could reproduce the methanol solvent exchange and nitrogen gas sorption analysis from stage three. Following the same procedure, perform the analysis and compare the resultant specific surface area, total pore volume, and average pore width with the previous data. This will give an early indication of the reproducibility of our characterization procedures.

**Figure S136.** Output generated by ChemNavigator during the 24$^{th}$ iteration, aimed at the synthesis and characterization of a novel aluminum-based MOF utilizing the BTB-*m*F linker.

Reticular ChemNavigator Output (MOF-521-*m*F)

Current Stage and Iteration: 5-2

Status Evaluation: The apprentice accurately replicated the MOF-521-*m*F synthesis under the conditions we established, and the PXRD pattern obtained from the reproduced MOF matches the previously obtained pattern. This confirms the reproducibility of the MOF-521-*m*F synthesis process.

Task Choice 1: Now that we have verified the reproducibility of the synthesis, our next step should be to validate the reproducibility of the characterization process. Start by carrying out CHNS elemental analysis, TGA, and SCXRD on the newly synthesized MOF sample. Compare these results with our previous data to confirm that the MOF's chemical composition, thermal stability, and structure remain consistent across different synthesis batches.

Task Choice 2: Since we have confirmed the reproducibility of the MOF synthesis, it's time to evaluate the reproducibility of the activation process. Activate the new MOF sample via a methanol solvent exchange and vacuum drying, following the process we used in stage three. Subsequently, perform a nitrogen gas sorption analysis to determine if the MOF's specific surface area, total pore volume, and average pore width match our previous results.

Task Choice 3: To further probe the water instability aspect of the MOF, reproduce the water exposure test. Soak the freshly synthesized MOF sample in water for a predetermined amount of time. Post-exposure, conduct 1H NMR spectroscopy and compare the formate peak with the previous result to validate the reproducibility of the water instability findings. This could also provide additional insights into the hydrolysis process and contribute to the overall understanding of MOF's water stability.

**Figure S137.** Output generated by ChemNavigator during the 25$^{th}$ iteration, aimed at the synthesis and characterization of a novel aluminum-based MOF utilizing the BTB-*m*F linker.

Reticular ChemNavigator Output (MOF-521-*m*F)

Current Stage and Iteration: 5-3

Status Evaluation: The apprentice successfully carried out the CHNS elemental analysis, TGA, and SCXRD on the newly synthesized MOF sample and found the results to be matching with our previous data. This suggests that our methods for the synthesis and characterization of MOF-521-*m*F are both robust and reproducible.

Task Choice 1: Given that we have confirmed the reproducibility of our synthesis and characterization processes, the next logical step would be to write a comprehensive report of our research. In the report, make sure to include an overview of the project, materials and methods, results and discussion, and conclusion. Detail the stages of the project, the challenges faced, the solutions we implemented, and the final outcomes. Make sure to include all the analytical data and discuss their implications for the structure and stability of our MOF.

Task Choice 2: A good way to validate our results is to have an external laboratory run tests on our MOF. Send samples of our synthesized MOF-521-*m*F to a trusted external laboratory for further analysis. Request for CHNS elemental analysis, TGA, SCXRD, and PXRD tests to be carried out on the samples. Compare their results with ours to further verify the reproducibility and reliability of our work.

Task Choice 3: To further test the applicability and stability of our MOF, we can test the MOF for real-life applications, such as gas storage or catalysis. Choose a suitable application for our MOF, design an experiment, and carry it out. Make sure to record all observations and results meticulously. This would not only test our MOF's functionality but also provide potential avenues for future research.

**Figure S138.** Output generated by ChemNavigator during the 26th iteration, aimed at the synthesis and characterization of a novel aluminum-based MOF utilizing the BTB-*m*F linker.

## Final Summary (MOF-521- $CH_3$)

Our research project embarked on the synthesis of a novel Aluminum Metal Organic Framework (MOF), termed MOF-521-CH3, using a unique organic linker known as 4,4′,4′′-(2,4,6-trimethylbenzene-1,3,5-triyl)tribenzoic acid (BTB-CH3) or 5'-(4-Carboxyphenyl)-2',4',6'-trimethyl-[1,1':3',1''-terphenyl]-4,4''-dicarboxylic acid.

During Stage 1, we focused on the synthesis of the BTB-CH3 organic linker. An adaptable approach allowed us to modify an existing protocol to suit our target molecule. We substituted unavailable reagents with KOAc and PdCl2(dppf) and ran trial reactions, both on a small scale and a scaled-up version. Both operations yielded positive outcomes confirmed by Thin-layer chromatography (TLC) and NMR, with the larger scale reaction producing the target product at a favorable yield of 68%.

In Stage 2, our attention shifted towards high-throughput screening of the MOF and optimizing the synthetic conditions. We determined the optimal BTB-CH3:Al ratio to be 3:4, the formic acid to water ratio to be 4:1, and a reaction temperature of 130°C to be maintained for three days. Several iterative processes helped us understand that 130°C was the ideal temperature for synthesis, as higher temperatures resulted in broader peak widths in the PXRD patterns, indicating potential loss of crystallinity. This stage saw the successful synthesis and validation of rod-shaped MOF crystals using PXRD.

Stage 3 was about MOF activation and the determination of its permanent porosity. We activated the MOF via solvent exchange with methanol and vacuum drying at 100°C for 12 hours. Nitrogen gas sorption analysis at 77K revealed a specific surface area of 1311 m^2/g, a total pore volume of 0.509 cm^3/g, and pore width of 9.3 Å. Furthermore, hysteresis analysis suggested a uniform and stable pore structure within our MOF.

In Stage 4, we conducted a detailed structural analysis and characterization of the MOF. CHNS elemental analysis and Thermogravimetric Analysis (TGA) confirmed our proposed formula for MOF-521-CH3 and its thermal stability. Single-crystal X-ray diffraction (SCXRD) analysis was initially thwarted due to the minuscule size of the rod-shaped crystals. Solubility issues were encountered during the 1H NMR spectroscopy; however, these were overcome using NaOD 5% in D2O, which helped identify a formate peak at ~8.5 ppm. We conducted solvent stability tests and found that MOF-521-CH3 is stable in DMF, methanol, and ethanol. Nevertheless, potential instability in water was suggested due to a significant reduction in PXRD peak intensity after soaking.

Finally, in Stage 5, we checked the reproducibility and conducted the final validation of our work. We successfully replicated the synthesis of both the linker and the MOF. The activation and porosity determination of the newly synthesized MOF yielded consistent results, thereby confirming the reproducibility of our synthesis process. We view this consistency as a clear success, supporting the robustness and reliability of our synthesis protocol for MOF-521-CH3.

**Figure S139.** Illustration of the final summary developed by the Reticular ChemNavigator. This version of the summary was generated at the 26th iteration, marking the completion of Stage 5 interactions, which signifies the synthesis and study of MOF-521-$CH_3$. Previous versions of the summary have been omitted for clarity.

## Reticular ChemNavigator Output (MOF-521-$CH_3$)

Current Stage and Iteration: 1-1

Status Evaluation: As this is the starting point of the project, there are no past task evaluations or human feedback available.

Task Choice 1: Our first task will be the synthesis of the BTB-CH3 linker. Please refer to the appropriate literature to find the appropriate synthesis route. Typically, the synthesis involves the Suzuki-Miyaura cross-coupling reactions to generate the required phenyl rings, followed by carboxylation and methylation steps. Ensure to carry out this synthesis in a well-ventilated fume hood, wearing appropriate personal protective equipment (PPE). Once the synthesis is complete, validation via 1H NMR spectroscopy should be performed.

Task Choice 2: An alternative task would be to research the most suitable synthetic route for BTB-CH3 and perform a literature review on its synthesis. Look for research papers and patents that provide a detailed procedure for synthesizing BTB-CH3, and make a note of the reagents, conditions, and steps involved. Additionally, research if there are any recent developments or alternative synthesis methods available, as these may offer better yields or cleaner products.

Task Choice 3: Should the laboratory setup for the synthesis of BTB-CH3 take longer than anticipated, you can use this time to gather and prepare all necessary reagents, equipment, and safety gear. Check the inventory for the required reagents and order any missing ones. Also, ensure the availability of an NMR machine for the validation of the synthesized linker.

**Figure S140.** Output generated by ChemNavigator during the 1st iteration, aimed at the synthesis and characterization of a novel aluminum-based MOF utilizing the BTB-$CH_3$ linker.

Reticular ChemNavigator Output (MOF-521-$CH_3$)

Current Stage and Iteration: 1-2

Status Evaluation: The successful literature review conducted by the apprentice has provided a robust starting point for the synthesis of the BTB-CH3 linker. The outlined synthetic procedure, although slightly complex, has a high yield and is supported by spectroscopic validation. The next steps should involve planning for the synthesis of BTB-CH3 based on the retrieved literature and gathering of the necessary reagents and equipment.

Task Choice 1: As the first task, you should plan the synthesis of BTB-CH3 based on the procedure you found. This involves breaking down the procedure into discrete steps and creating a checklist of the required chemicals, equipment, and safety measures. Detail the amount of each reagent needed, the sequence of steps, the expected time for each step, and any special conditions (e.g., temperature, atmosphere). Also, consider any potential challenges or complications that could arise during the procedure, and how you might mitigate these. This will ensure a smooth synthesis process.

Task Choice 2: Alternatively, before jumping into the synthesis, you could further validate the procedure found by comparing it to other literature sources. Look for more articles detailing the synthesis of similar BTB derivatives. If you find common steps or conditions among these procedures, it can increase confidence in our selected synthesis method. Pay attention to the overall yield, reaction time, and potential impurities that these other sources report, as they can provide valuable insights for our synthesis.

Task Choice 3: You could also start gathering and preparing the necessary chemicals and equipment for the synthesis of BTB-CH3. Verify the availability of the required reagents in your laboratory, and if unavailable, initiate the process of sourcing them. Check the conditions of the equipment, and ensure they are functioning correctly and are ready for use. While this task may seem mundane, it is a crucial step that could save us a lot of time when we commence the actual synthesis.

**Figure S141.** Output generated by ChemNavigator during the 2$^{nd}$ iteration, aimed at the synthesis and characterization of a novel aluminum-based MOF utilizing the BTB-$CH_3$ linker.

Reticular ChemNavigator Output (MOF-521-$CH_3$)

Current Stage and Iteration: 1-3

Status Evaluation: The preparation process was executed with precision, and it was discovered that not all reagents were available in the lab inventory. It is important to note that the unavailability of K3PO4 and Pd(PPh3)4, the potassium salt and the palladium catalyst, could potentially impact the efficiency of the coupling reaction. The apprentice has suggested potential alternatives: KOAc and PdCl2(dppf), and these substitutions may not have a significant impact on the reaction and could be used as viable replacements.

Task Choice 1: Initiate the process of procuring K3PO4 and Pd(PPh3)4. Submit a request to your lab supervisor or lab manager to order these reagents. This could take time, so you should do this immediately. In parallel, you can initiate preparation of other reagents and ensure all the equipment is functioning properly and is ready for use.

Task Choice 2: As an alternative to procuring the missing reagents, we could explore using the available substitutes. Evaluate the feasibility of replacing K3PO4 with KOAc and Pd(PPh3)4 with PdCl2(dppf) in the synthesis of BTB-CH3. This will involve reviewing the literature to understand if these substitutions have been successfully employed in similar reactions. Make note of any changes in the reaction conditions or any additional steps necessary when using these substitutes.

Task Choice 3: It would also be beneficial to conduct a small-scale trial reaction using the available substitutes, KOAc and PdCl2(dppf), in place of K3PO4 and Pd(PPh3)4, respectively. Perform the reaction under the conditions detailed in the literature, but ensure to run it on a reduced scale to minimize waste. Following the reaction, utilize thin-layer chromatography (TLC) and NMR to analyze the reaction mixture and identify whether the desired product has been formed.

**Figure S142.** Output generated by ChemNavigator during the 3rd iteration, aimed at the synthesis and characterization of a novel aluminum-based MOF utilizing the BTB-$CH_3$ linker.

Reticular ChemNavigator Output (MOF-521-$CH_3$)

Current Stage and Iteration: 1-4

Status Evaluation: The successful synthesis of the BTB-CH3 linker, as confirmed by TLC and NMR analysis, demonstrates the efficacy of the procedural modifications suggested and the suitability of the identified substitutes for the unavailable reagents. The results mirror the success reported in the literature, further affirming the validity of the adapted procedure.

Task Choice 1: Having achieved successful synthesis on a small scale, the next logical step is to scale up the production of the BTB-CH3 linker. Repeat the same procedure used for the small-scale synthesis, but on a larger scale. Carefully monitor the reaction to ensure that it proceeds as expected. After the reaction, again perform TLC and NMR analysis to confirm the production of the BTB-CH3 linker.

Task Choice 2: An alternative option would be to first conduct additional small-scale trial reactions, each varying the concentration of the catalyst, to determine the optimal amount for large-scale synthesis. It's crucial to monitor these reactions closely to observe any changes in yield or reaction time. This information will be valuable in optimizing the larger scale synthesis of the BTB-CH3 linker.

Task Choice 3: A third approach could involve conducting the scaled-up synthesis while simultaneously running a small-scale control reaction. This dual setup can serve as a direct comparison, allowing for any potential issues during the scaled-up synthesis to be immediately identified and rectified. Post-reaction, TLC and NMR analyses should be performed on the products of both reactions to ensure the formation of the BTB-CH3 linker.

**Figure S143.** Output generated by ChemNavigator during the 4$^{th}$ iteration, aimed at the synthesis and characterization of a novel aluminum-based MOF utilizing the BTB-$CH_3$ linker.

Reticular ChemNavigator Output (MOF-521-$CH_3$)

Current Stage and Iteration: 1-5

Status Evaluation: The success of the scaled-up synthesis can be attributed to the apprentice's careful adherence to the procedure developed during the small-scale synthesis, suggesting that the modifications to accommodate the reagent substitution were well-conceived. This, combined with meticulous monitoring of the scaled-up reaction, led to a significant yield of BTB-CH3 linker. Given the successful synthesis of the linker, we can now consider the possibility of moving forward.

Task Choice 1: The first task option is to further optimize the synthesis of the BTB-CH3 linker. Although a 68% yield is significant, there might be room for improvement. This optimization could involve adjusting parameters such as the reaction time, temperature, and solvent volumes. Once these adjustments are made, repeat the synthesis and compare the yield to the previous run.

Task Choice 2: The second task option is to conduct further analysis of the BTB-CH3 linker to ensure its quality before proceeding to the next stage. This analysis could include advanced techniques such as 13C NMR spectroscopy and High-Resolution Mass Spectrometry (HRMS) to further confirm the structure and purity of the linker.

Task Choice 3: Given the successful synthesis of the linker at a larger scale, if the apprentice feels prepared, the third task option could be to propose moving to the next stage of the project. If chosen, the apprentice should state, "I'm ready to move to the next stage." This would involve shifting the focus to high-throughput screening and optimization of the synthetic outcomes for the MOF.

**Figure S144.** Output generated by ChemNavigator during the 5$^{th}$ iteration, aimed at the synthesis and characterization of a novel aluminum-based MOF utilizing the BTB-$CH_3$ linker.

## Reticular ChemNavigator Output (MOF-521-$CH_3$)

Current Stage and Iteration: 2-1

Status Evaluation: The apprentice successfully followed the procedures for synthesizing the BTB-CH3 linker and confirmed the successful synthesis through TLC and NMR analysis. With the productive outcome of the first stage, the apprentice expressed readiness to move to the next stage, focusing on high-throughput screening and optimization of the MOF synthesis.

Task Choice 1: Begin the second stage by conducting high-throughput synthesis of the MOF using the synthesized BTB-CH3 linker. You may start by utilizing a BTB-CH3:Al ratio of 3:4, following the precedent set in the similar work summary, under solvothermal conditions at 120°C. Use a mixture of formic acid and water in a ratio of 4:1 as your solvent. After the reaction is complete, conduct a preliminary PXRD analysis of the product to identify the formation of the MOF structure.

Task Choice 2: In order to explore optimal synthetic conditions, vary the BTB-CH3:Al ratio in a series of synthesis trials. By systematically altering the ratio, we can possibly find the ideal ratio for the synthesis of our target MOF. Upon completing each reaction, analyze the product using PXRD to confirm the formation and crystallinity of the MOF.

Task Choice 3: As an alternative exploration of the reaction parameters, you may also vary the reaction temperature in different trials. Using the same solvent mixture and BTB-CH3:Al ratio, carry out reactions at different temperatures ranging from 100°C to 140°C. Remember to conduct a PXRD analysis after each synthesis to assess the impact of temperature on the MOF's formation and crystallinity.

**Figure S145.** Output generated by ChemNavigator during the 6$^{th}$ iteration, aimed at the synthesis and characterization of a novel aluminum-based MOF utilizing the BTB-$CH_3$ linker.

## Reticular ChemNavigator Output (MOF-521-$CH_3$)

Current Stage and Iteration: 2-2

Status Evaluation: The results from the apprentice suggest successful synthesis of the MOF structure using a BTB-CH3:Al ratio of 3:4 under solvothermal conditions at 120°C. The resultant PXRD pattern aligns with our predicted pattern, implying successful formation of the MOF. However, the yield appears to be low, indicated by the reported "very little amount" of product.

Task Choice 1: The next step could be optimizing the synthetic conditions to increase the yield of the MOF. Let's adjust the BTB-CH3:Al ratio slightly to see if it improves the yield. Try a BTB-CH3:Al ratio of 2:3 this time, keeping all other conditions the same. After the reaction is complete, conduct PXRD analysis to confirm the MOF structure and visually inspect the quantity of the product.

Task Choice 2: The low yield might be due to the short reaction time of only 2 days. In the BTB-*o*F MOF-521 example, a reaction time between 72 and 96 hours resulted in a higher quantity of crystals. Try extending the reaction time to 3 days at 120°C and observe the quantity and quality of the product. Conduct a PXRD analysis to verify if the MOF structure is still formed.

Task Choice 3: As another alternative, you could attempt changing the solvent ratio. The current ratio of formic acid to water is 4:1. Try adjusting this ratio to 3:2, keeping the BTB-CH3:Al ratio and reaction time constant. Perform a PXRD analysis on the resultant product to determine the structural integrity of the MOF.

**Figure S146.** Output generated by ChemNavigator during the 7$^{th}$ iteration, aimed at the synthesis and characterization of a novel aluminum-based MOF utilizing the BTB-$CH_3$ linker.

## Reticular ChemNavigator Output (MOF-521-$CH_3$)

Current Stage and Iteration: 2-3

Status Evaluation: The consistent PXRD pattern suggests that the MOF structure is being maintained, even with the extended reaction time. However, the yield did not significantly increase, which might suggest the reaction has reached a plateau at this temperature and reaction time. This could be due to the reaction reaching equilibrium or the full utilization of the reactants.

Task Choice 1: Given that extending the reaction time did not significantly increase the yield, it might be worth increasing the temperature to 130°C while keeping the reaction time to 3 days. This could potentially drive the reaction further towards product formation. Repeat the PXRD analysis after the synthesis to confirm the MOF structure.

Task Choice 2: You may consider adjusting the BTB-CH3:Al ratio. Increasing the BTB-CH3 might facilitate more MOF formation, leading to a higher yield. Try a ratio of 4:4 while maintaining the same temperature (120°C) and reaction time (3 days). Again, use PXRD to confirm the MOF structure.

Task Choice 3: Alternatively, you might want to explore the use of a modulator to improve the crystallinity and yield of the MOF. Modulators, like monocarboxylic acids, have been found to aid in MOF formation by controlling growth and reducing defects. Try adding formic acid at a molar ratio of 1:1 to BTB-CH3 and maintain the previous reaction conditions (120°C for 3 days). Use PXRD to confirm the MOF structure after the reaction.

**Figure S147.** Output generated by ChemNavigator during the 8$^{th}$ iteration, aimed at the synthesis and characterization of a novel aluminum-based MOF utilizing the BTB-$CH_3$ linker.

Reticular ChemNavigator Output (MOF-521-$CH_3$)

Current Stage and Iteration: 2-4

Status Evaluation: The increase in yield at the higher temperature (130°C) supports the hypothesis that enhancing the kinetic energy can drive the reaction further towards product formation, resulting in a higher yield of MOF-521-CH3. PXRD analysis showed a similar pattern, which means the MOF structure remained consistent despite the increased reaction temperature. This is an important finding as it confirms the robustness of the MOF structure against changes in synthetic conditions.

Task Choice 1: Attempt to further optimize the yield by adjusting the ratio of BTB-CH3 to Al. This time, adjust the BTB-CH3:Al ratio to 4:3, while keeping the reaction time to 2 days and the temperature at 130°C. After the synthesis, filter and wash the product, then perform PXRD analysis to check if the MOF structure remains consistent.

Task Choice 2: Try extending the reaction time to 3 days while maintaining the temperature at 130°C and the BTB-CH3:Al ratio at 3:4. This will provide further insights into whether reaction time can significantly influence the yield without affecting the MOF structure. As always, after the synthesis, perform PXRD analysis to check the integrity of the MOF structure.

Task Choice 3: Consider running a control experiment at 120°C for 2 days with the BTB-CH3:Al ratio at 3:4. This will help to confirm the effect of temperature on the yield and the MOF structure. It is always important to have a control experiment to benchmark your results. The product should be subjected to PXRD analysis to assess the MOF structure.

**Figure S148.** Output generated by ChemNavigator during the 9$^{th}$ iteration, aimed at the synthesis and characterization of a novel aluminum-based MOF utilizing the BTB-$CH_3$ linker.

## Reticular ChemNavigator Output (MOF-521-$CH_3$)

Current Stage and Iteration: 2-5

Status Evaluation: The unchanged PXRD patterns and yield after extending the reaction time at 130°C suggest that the current conditions might have reached an optimum for this temperature. To improve yield and potentially enhance MOF properties, varying other parameters such as the BTB-CH3:Al ratio or the formic acid to water ratio might be necessary.

Task Choice 1: Perform a high-throughput screening by varying the BTB-CH3:Al ratio. Try ratios of 2:4, 3:4, and 4:4 while keeping the other conditions constant (formic acid to water ratio: 4:1, reaction time: 3 days, solvent: DMF, reaction temperature: 130°C). After each synthesis, conduct PXRD analysis to verify the MOF structure and compare yields.

Task Choice 2: Execute a high-throughput screening while adjusting the formic acid to water ratio. Test ratios of 2:1, 3:1, and 4:1, maintaining the rest of the parameters constant (BTB-CH3:Al ratio: 3:4, reaction time: 3 days, solvent: DMF, reaction temperature: 130°C). Conduct PXRD analysis post-synthesis for each set to assess MOF structure and yield.

Task Choice 3: In a high-throughput approach, experiment with different reaction temperatures. Test at 120°C, 130°C, and 140°C, while keeping the other conditions constant (BTB-CH3:Al ratio: 3:4, formic acid to water ratio: 4:1, reaction time: 3 days, solvent: DMF). Post-synthesis, analyze the MOF structure with PXRD and compare the yields across different temperatures.

**Figure S149.** Output generated by ChemNavigator during the 10th iteration, aimed at the synthesis and characterization of a novel aluminum-based MOF utilizing the BTB-$CH_3$ linker.

## Reticular ChemNavigator Output (MOF-521-$CH_3$)

Current Stage and Iteration: 2-6

Status Evaluation: The apprentice's feedback indicates successful optimization of the BTB-CH3:Al ratio for MOF-521-CH3 synthesis, with the optimal ratio being 3:4, based on the superior crystallinity as evidenced by PXRD analysis. Lack of peaks at the BTB-CH3:Al ratio of 4:4 suggests that an excess of the BTB-CH3 linker may hinder MOF formation.

Task Choice 1: Continue optimizing the synthetic conditions by varying the formic acid to water ratio. Keep the BTB-CH3:Al ratio constant at 3:4 and try formic acid to water ratios of 3:1, 4:1, and 5:1. Maintain the other conditions (reaction time: 3 days, solvent: DMF, reaction temperature: 130°C). After each synthesis, perform PXRD to verify the MOF structure and compare yields.

Task Choice 2: Another variable to consider is reaction time. Continue using the optimal BTB-CH3:Al ratio of 3:4, and extend the reaction time to 4 and 5 days respectively while keeping the other parameters constant (formic acid to water ratio: 4:1, solvent: DMF, reaction temperature: 130°C). After each synthesis, conduct PXRD analysis to verify the MOF structure and yields.

Task Choice 3: Investigate the effect of temperature on MOF formation. Utilize the optimized BTB-CH3:Al ratio of 3:4, and conduct reactions at temperatures of 140°C and 150°C while maintaining other parameters (formic acid to water ratio: 4:1, reaction time: 3 days, solvent: DMF). Following each experiment, perform PXRD analysis to compare yields and ascertain the MOF structure.

**Figure S150.** Output generated by ChemNavigator during the 11th iteration, aimed at the synthesis and characterization of a novel aluminum-based MOF utilizing the BTB-$CH_3$ linker.

## Reticular ChemNavigator Output (MOF-521-$CH_3$)

Current Stage and Iteration: 2-7

Status Evaluation: The similar PXRD patterns with increasing temperature suggest that MOF-521-CH3 is stable across the temperature range studied. However, the increased FWHM at higher temperatures indicates a decrease in crystallinity, hence confirming that 130°C is the optimal temperature for the synthesis of MOF-521-CH3. It might be worthwhile to consider fine-tuning other reaction parameters such as reaction time and solvent composition to further improve the MOF quality.

Task Choice 1: In order to optimize the reaction time, perform a set of reactions using the optimal BTB-CH3:Al ratio (3:4) at the identified optimal temperature of 130°C. For this set, vary the reaction times (2 days, 3 days, and 4 days). After each reaction, perform a PXRD analysis to evaluate the crystallinity and compare the yields.

Task Choice 2: Let's consider the effect of solvent composition on MOF synthesis. Continue using the optimal BTB-CH3:Al ratio of 3:4, and perform the reaction at the established optimal temperature of 130°C for 3 days. In this iteration, vary the solvent ratios using different proportions of DMF and formic acid. Compare the yields and determine the MOF structure using PXRD analysis.

Task Choice 3: If you believe we have reached an optimized set of reaction conditions for the synthesis of MOF-521-CH3, we can conclude Stage 2. To validate this, perform one more synthesis using the optimal parameters identified (BTB-CH3:Al ratio of 3:4, reaction temperature of 130°C, formic acid to water ratio of 4:1, and reaction time of 3 days). Validate the MOF structure using PXRD. If successful, state, "I'm ready to move to the next stage."

**Figure S151.** Output generated by ChemNavigator during the 12$^{th}$ iteration, aimed at the synthesis and characterization of a novel aluminum-based MOF utilizing the BTB-$CH_3$ linker.

## Reticular ChemNavigator Output (MOF-521-$CH_3$)

Current Stage and Iteration: 3-1

Status Evaluation: The success in stage two is largely due to the persistent optimizations and the final synthesis that produced MOF-521-CH3 with the expected PXRD pattern, proving that the selected conditions were ideal. The fact that the apprentice has stated readiness to move to the next stage indicates that we've accomplished our goals for this stage.

Task Choice 1: As we proceed to the activation and determination of permanent porosity, perform a solvent exchange using methanol. Place the MOF sample in an excess of methanol and leave it to stir at room temperature for 24 hours. After this, decant the methanol and repeat the process for another 24 hours. Once complete, dry the MOF under vacuum at 100°C for 12 hours. Finally, perform nitrogen gas sorption analysis at 77K to confirm the activation and determine the porosity of the MOF.

Task Choice 2: An alternative method for activation involves using supercritical CO2. In this method, place the MOF sample in a high-pressure stainless steel vessel. Introduce supercritical CO2 into the vessel and hold for 24 hours to ensure thorough solvent exchange. Dry the MOF under vacuum at 100°C for 12 hours. After the CO2 treatment, conduct nitrogen gas sorption analysis at 77K to confirm activation and assess the MOF's porosity.

Task Choice 3: Consider using an alternative solvent for the exchange process. Perform the solvent exchange using ethanol. Place the MOF in an excess of ethanol and let it stir at room temperature for 24 hours. Decant the ethanol and repeat for another 24 hours. Following this, dry the MOF under vacuum at 100°C for 12 hours. Then, perform nitrogen gas sorption analysis at 77K to verify the activation and determine the MOF's porosity.

**Figure S152.** Output generated by ChemNavigator during the 13$^{th}$ iteration, aimed at the synthesis and characterization of a novel aluminum-based MOF utilizing the BTB-$CH_3$ linker.

## Reticular ChemNavigator Output (MOF-521-$CH_3$)

Current Stage and Iteration: 3-2

Status Evaluation: The results indicate successful activation of the MOF sample, and the degas conditions were well controlled, leading to a pure and desolvated product. However, detailed analysis and interpretation of the nitrogen gas sorption isotherms is still required to fully understand the porosity and surface area of the MOF.

Task Choice 1: To gain a more comprehensive understanding of the MOF's porosity, conduct further analysis on the obtained nitrogen gas sorption isotherms. This should involve plotting the isotherm and calculating key parameters such as the specific surface area (via the Brunauer–Emmett–Teller (BET) method), the total pore volume, and the average pore width. Such data will offer critical insights into the structural characteristics and potential applications of MOF-521-CH3.

Task Choice 2: To confirm the successful activation of the MOF, perform a powder X-ray diffraction (PXRD) analysis on the activated sample. Compare the diffraction pattern with the one obtained prior to activation to check for any shifts in peak positions or intensities that might indicate changes in the MOF structure.

Task Choice 3: To ensure that the activation process has not altered the MOF's structure, perform another round of 1H NMR spectroscopy on the activated sample. Compare these spectra with those obtained from the synthesized MOF prior to activation. Identifying any new peaks or shifts in chemical shifts will aid in confirming the structural integrity of the activated MOF.

**Figure S153.** Output generated by ChemNavigator during the 14$^{th}$ iteration, aimed at the synthesis and characterization of a novel aluminum-based MOF utilizing the BTB-$CH_3$ linker.

Reticular ChemNavigator Output (MOF-521-$CH_3$)

Current Stage and Iteration: 3-3

Status Evaluation: The apprentice successfully completed the detailed analysis of the nitrogen gas sorption isotherms, providing key parameters that give insight into the structural characteristics of the MOF. The reported specific surface area, total pore volume, and average pore width fall within expected ranges for MOFs, indicating a well-formed and porous structure.

Task Choice 1: Continue analyzing the porosity of MOF-521-CH3 by studying the adsorption hysteresis. Generate the hysteresis loop by plotting the adsorption and desorption isotherms. Look for any significant deviation between the two paths. A minimal hysteresis would suggest a stable, homogeneous, and uniform pore structure. Once the hysteresis analysis is complete, interpret the results in terms of the stability and homogeneity of the MOF's porosity.

Task Choice 2: Since the preliminary characterization of the porosity has been successful, consider initiating the fourth stage, which involves a detailed structural analysis and characterization of the MOF. If you feel ready, state "I'm ready to move to the next stage." However, be reminded that further porosity analysis may be beneficial before transitioning.

Task Choice 3: Perform the pore size distribution analysis for MOF-521-CH3. This can be done by employing the Barrett–Joyner–Halenda (BJH) method on the desorption branch of the isotherm. This analysis will provide a distribution curve which illustrates the range and frequency of different pore sizes in your MOF, further enhancing your understanding of its porosity.

**Figure S154.** Output generated by ChemNavigator during the 15$^{th}$ iteration, aimed at the synthesis and characterization of a novel aluminum-based MOF utilizing the BTB-$CH_3$ linker.

Reticular ChemNavigator Output (MOF-521-$CH_3$)

Current Stage and Iteration: 3-4

Status Evaluation: The apprentice's results from the nitrogen adsorption-desorption isotherm analysis align with expectations for a microporous MOF, indicative of a stable, homogeneous, and uniform pore structure. This outcome, along with the absence of a significant hysteresis loop, bodes well for the MOF's intended uses.

Task Choice 1: Based on our progress in characterizing the porosity and pore structure of MOF-521-CH3, we could move on to the next stage. If you feel ready to proceed with a more in-depth structural analysis and characterization of the MOF, please confirm by stating, "I'm ready to move to the next stage."

Task Choice 2: If you prefer to continue investigating in stage three, you can conduct a temperature-dependent nitrogen sorption analysis. Start from 77K and increase in increments of 10K, documenting any changes in the isotherms as a function of temperature.

Task Choice 3: Alternatively, you can generate a pore size distribution plot using the desorption branch of the nitrogen isotherm and the Barrett-Joyner-Halenda (BJH) method. This analysis would provide further insights into the MOF's pore size and uniformity.

**Figure S155.** Output generated by ChemNavigator during the 16th iteration, aimed at the synthesis and characterization of a novel aluminum-based MOF utilizing the BTB-$CH_3$ linker.

Reticular ChemNavigator Output (MOF-521-$CH_3$)

Current Stage and Iteration: 4-1

Status Evaluation: The human apprentice's readiness to proceed to the next stage indicates successful completion of the third stage tasks, including the activation of MOF, determination of porosity, and confirmation of a stable pore structure. This paves the way for a more detailed structural analysis and characterization of the MOF in the fourth stage.

Task Choice 1: In this step, we'll perform CHNS elemental analysis. You'll need to weigh a precise amount of your MOF sample (around 1-2 mg) and place it in the CHNS analyzer. This analysis will verify the correct chemical composition of our MOF, confirming the relative proportions of carbon (C), hydrogen (H), nitrogen (N), and sulfur (S) present. After the analysis, please share the results for review.

Task Choice 2: Let's conduct a thermogravimetric analysis (TGA) to test the thermal stability of our MOF. To do this, place a small amount of the MOF sample into the TGA instrument and gradually increase the temperature from room temperature to 800°C, while maintaining a nitrogen atmosphere. Record the weight loss percentage as a function of the temperature, which will provide insight into the thermal stability of our MOF and indicate any structural changes or loss of guest molecules. Please share the TGA curve for interpretation.

Task Choice 3: Start with the chemical stability testing of our MOF. We need to test our MOF's stability in different solvents including DMF, methanol, ethanol, and water. Dissolve a weighed sample of our MOF in each of these solvents separately, and leave the solutions undisturbed for 24 hours. After this period, filter the solutions, dry the retrieved solids, and analyze them using PXRD to see if there are any changes to the structure. Please report the PXRD patterns for each solvent along with your observations.

**Figure S156.** Output generated by ChemNavigator during the 17$^{th}$ iteration, aimed at the synthesis and characterization of a novel aluminum-based MOF utilizing the BTB-$CH_3$ linker.

Reticular ChemNavigator Output (MOF-521-$CH_3$)

Current Stage and Iteration: 4-2

Status Evaluation: The CHNS elemental analysis results reveal a possible 1:3 formate to BTB-CH3 linker ratio, aligning with our proposed formula for MOF-521-CH3. This successful outcome likely stems from rigorous adherence to optimized synthesis conditions, and it reinforces the preliminary accuracy of our MOF's chemical composition.

Task Choice 1: To further substantiate our MOF's composition and stability, the next logical step involves conducting Thermogravimetric Analysis (TGA). Using a TGA instrument, measure a small sample of your MOF (~5 mg) under an inert atmosphere (like nitrogen) and increase the temperature at a rate of 10°C per minute, up to 800°C. Observe the thermal degradation pattern of the MOF. This analysis will provide insights into the thermal stability of MOF-521-CH3 and identify the temperature at which it starts to decompose.

Task Choice 2: Another valuable characterization technique to apply at this stage is X-ray photoelectron spectroscopy (XPS). This method can confirm the presence and oxidation state of Aluminum in our MOF. Use a small sample of your MOF and place it into the XPS instrument. Record the spectra and identify the peaks corresponding to Aluminum 2p, Carbon 1s, Oxygen 1s, and Hydrogen 1s. Compare these findings with reference spectra for these elements.

Task Choice 3: Let's assess the chemical stability of MOF-521-CH3 by exposing it to various solvents, including DMF, methanol, and ethanol, and monitor any changes over time. Immerse a sample of MOF in each of these solvents separately, keep them undisturbed at room temperature for 48 hours, and then filter and dry the MOFs. Perform PXRD on the recovered MOFs to identify any changes in peak intensities or positions, which would suggest chemical instability in the tested solvents.

**Figure S157.** Output generated by ChemNavigator during the $18^{th}$ iteration, aimed at the synthesis and characterization of a novel aluminum-based MOF utilizing the BTB-$CH_3$ linker.

Reticular ChemNavigator Output (MOF-521-$CH_3$)

Current Stage and Iteration: 4-3

Status Evaluation: The TGA findings align well with our expectations. The initial decomposition temperature of 330°C and the corresponding weight loss are congruent with the decomposition of formate groups in our proposed MOF structure. These results not only add weight to our MOF composition, but also showcase good thermal stability. This underlines the progress in the fourth stage of detailed characterization.

Task Choice 1: To further understand the stability of the MOF, you can now conduct a chemical stability test. First, immerse a few milligrams of the MOF in different solvents like DMF, methanol, ethanol, and water. Allow the MOF to remain in the solvents for 24 hours. Then, remove the MOF and allow it to dry. Compare the PXRD pattern of the MOF before and after immersion in each solvent. A change in the PXRD pattern will indicate instability in the corresponding solvent.

Task Choice 2: Given the potential instability of MOFs in water, it would be beneficial to assess the MOF's stability under varying pH levels. Create solutions of water with different pH levels (e.g., pH 2, 7, and 12). Submerge a small amount of MOF in each solution for 24 hours, dry the MOF, and compare the PXRD patterns before and after immersion. This can help establish the pH range in which the MOF remains stable.

Task Choice 3: To provide more evidence to our proposed chemical formula, we can consider a single-crystal X-ray diffraction (SCXRD) analysis, which can also offer insight into the structural features of our MOF. Please arrange a single crystal of MOF-521-CH3 on a suitable holder and perform an X-ray diffraction analysis using a suitable X-ray source. Ensure that you adjust the instrument settings and obtain data according to the manufacturer's instructions or available literature protocols.

**Figure S158.** Output generated by ChemNavigator during the 19$^{th}$ iteration, aimed at the synthesis and characterization of a novel aluminum-based MOF utilizing the BTB-$CH_3$ linker.

Reticular ChemNavigator Output (MOF-521-$CH_3$)

Current Stage and Iteration: 4-4

Status Evaluation: The unsuccessful SCXRD analysis is due to the small size and rod-like shape of the MOF-521-CH3 crystals. This shows the need for additional structural characterization methods that can accommodate the small size of our crystals.

Task Choice 1: As an alternative to SCXRD, we can consider performing a powder X-ray diffraction (PXRD) analysis, which is suitable for small-sized crystals. To perform PXRD, take a small amount of your MOF-521-CH3, grind it into a fine powder, place it on a glass slide, and follow the PXRD protocol as per the manufacturer's instructions.

Task Choice 2: Another alternative is to perform a Fourier-Transform Infrared (FTIR) spectroscopy, which can be used to investigate the functional groups and chemical bonds in MOF-521-CH3. To perform FTIR, prepare a sample by grinding a small amount of MOF-521-CH3 with KBr to make a fine powder. Press this powder to form a thin pellet, and then perform the FTIR measurement.

Task Choice 3: Given the chemical stability of our MOF in methanol, we could potentially use 1H NMR spectroscopy to further investigate our MOF structure, particularly the presence of formate groups. Dissolve a small amount of your MOF in deuterated methanol (CD3OD) and run a 1H NMR spectrum. The characteristic resonance of formate protons should appear around 8.3 ppm.

**Figure S159.** Output generated by ChemNavigator during the 20$^{th}$ iteration, aimed at the synthesis and characterization of a novel aluminum-based MOF utilizing the BTB-$CH_3$ linker.

Reticular ChemNavigator Output (MOF-521-$CH_3$)

Current Stage and Iteration: 4-5

Status Evaluation: The recent outcome was unsuccessful due to the MOF's poor solubility in deuterated methanol. Although methanol has been useful in the previous chemical stability tests, its deuterated form wasn't effective for studying the MOF structure using 1H NMR. The solubility issue likely arises from the structure and properties of our MOF, which might not interact favorably with CD3OD.

Task Choice 1: Given the feedback regarding the MOF's insolubility in deuterated methanol, we might indeed find success in using NaOD 5% in D2O as suggested. In a suitable NMR tube, dissolve a small amount of the MOF in NaOD 5% in D2O. Mix the sample thoroughly to ensure complete digestion of the MOF. Then, run a 1H NMR spectrum to identify characteristic resonances of formate protons, which should appear around 8.3 ppm.

Task Choice 2: As an alternative, we could attempt to resolve the solubility issue by utilizing a different solvent. Dimethyl sulfoxide (DMSO-d6) has proven to be an effective solvent for many MOFs. In a suitable NMR tube, dissolve a small amount of the MOF in DMSO-d6 and run a 1H NMR spectrum. Carefully examine the spectrum to locate the characteristic resonance of formate protons.

Task Choice 3: Given the unsuccessful SCXRD attempt due to the small size of the crystals, you may want to optimize your crystallization conditions for growing larger single crystals. This would enable us to revisit the SCXRD task later. Adjust the cooling rate, reactant concentrations, or utilize a different solvent system to encourage the growth of larger single crystals. Monitor the growth over time, and once you have sizable single crystals, we can attempt SCXRD again.

**Figure S160.** Output generated by ChemNavigator during the 21$^{st}$ iteration, aimed at the synthesis and characterization of a novel aluminum-based MOF utilizing the BTB-$CH_3$ linker.

## Reticular ChemNavigator Output (MOF-521-$CH_3$)

Current Stage and Iteration: 4-6

Status Evaluation: The success in obtaining a formate peak at ~8.5 ppm using NaOD 5% in D2O is a significant breakthrough. It means that the MOF-521-CH3 was effectively digested in the solvent, revealing characteristic resonances of the formate protons, a key attribute of its structure.

Task Choice 1: Given that we have evidence of potential hydrolysis of our MOF, we should investigate its stability in water. Dissolve a small amount of MOF-521-CH3 in distilled water, prepare a solution with a pH ranging from 1 to 13, and soak the MOF for a 24-hour period. Afterward, conduct a PXRD analysis on the MOF samples to evaluate the stability of our MOF in different pH levels. If there's a significant loss in peak intensity, it could indicate that the MOF is not stable in water.

Task Choice 2: To understand the stability of MOF-521-CH3 in various solvents, conduct a solvent stability test. Soak small amounts of MOF-521-CH3 in different solvents such as DMF, methanol, ethanol, acetone, and water for 24 hours. After the soaking period, run PXRD on these samples to assess any changes in the framework. Any significant deviations in the PXRD pattern could indicate instability in the corresponding solvent.

Task Choice 3: Since the MOF was not soluble in CD3OD for NMR but was soluble in NaOD 5% in D2O, we should explore other solvents for MOF digestion that might be useful for future characterizations. Solvents such as DMSO-d6 or acetone-d6 can be considered. Start with a small amount of the MOF and add the solvent until it dissolves. Once the MOF is completely dissolved, perform a 1H NMR to evaluate the dissolution and to further confirm the structure of the MOF.

**Figure S161.** Output generated by ChemNavigator during the 22nd iteration, aimed at the synthesis and characterization of a novel aluminum-based MOF utilizing the BTB-$CH_3$ linker.

Reticular ChemNavigator Output (MOF-521-$CH_3$)

Current Stage and Iteration: 4-7

Status Evaluation: The solvent stability test has shown that MOF-521-CH3 is stable in DMF, methanol, and ethanol, but not in water. This corresponds with the observed reduced PXRD peak intensity after soaking the MOF in water. Given that we have achieved a thorough characterization of the MOF structure, chemical composition, phase purity, and chemical and thermal stability, it may be the right time to consider transitioning to the final stage of the project.

Task Choice 1: Given the results from the solvent stability test, we should explore the MOF's stability in water further. To do this, prepare an aqueous solution with varying pH levels (e.g., pH 4, 7, and 10) using appropriate buffers. Soak the MOF in each pH solution for 24 hours and run PXRD to assess any changes. Comparing the peak intensity and positions will give us insights into the MOF's pH-dependent water stability.

Task Choice 2: Since we're observing possible instability in water, we should conduct a thorough investigation of the potential degradation products. Isolate and analyze any degradation products formed when the MOF is exposed to water. This can be done by soaking MOF-521-CH3 in water for 48 hours. Then, filter and collect the aqueous phase. Analyze the filtered water solution using High-Performance Liquid Chromatography (HPLC) and Mass Spectrometry (MS) to identify and quantify any released linkers or metals.

Task Choice 3: Considering the successful results of our extensive analysis and characterization of MOF-521-CH3, it might be the appropriate time to progress to the final stage of our project. If you believe that we have adequately met the objectives of stage 4 and feel ready to move on, please state, "I'm ready to move to the next stage."

**Figure S162.** Output generated by ChemNavigator during the 23rd iteration, aimed at the synthesis and characterization of a novel aluminum-based MOF utilizing the BTB-$CH_3$ linker.

## Reticular ChemNavigator Output (MOF-521-$CH_3$)

Current Stage and Iteration: 5-1

Status Evaluation: Given the successful conclusion of detailed characterization in stage four and the apprentice's readiness to proceed, we have appropriately transitioned to the final stage. The extensive analysis and experimental adjustments in the previous stages have prepared us for this crucial step of verifying the reproducibility of our protocols.

Task Choice 1: Let's commence the final stage of our project by first repeating the synthesis of the BTB-CH3 linker. Follow the same steps that you used successfully in stage 1. Once the synthesis is complete, validate the formation of the linker using TLC and NMR spectroscopic evaluations as you did before. After confirming the successful synthesis, proceed to the synthesis of MOF-521-CH3.

Task Choice 2: As an alternative to Choice 1, we can initially focus on the synthesis of MOF-521-CH3, using the same optimized conditions determined in stage 2. The goal here is to recreate the rod-shaped crystals. After the synthesis, confirm the structure and phase purity of the MOF using PXRD, as done in earlier steps.

Task Choice 3: To approach the reproducibility check more holistically, let's repeat the synthesis of both the linker and the MOF simultaneously. Adhere to the procedures used in stages 1 and 2, and validate each step using the same methods used before, i.e., TLC and NMR spectroscopic evaluations for the linker, and PXRD for the MOF. Once completed, compare the outcomes to those of the initial rounds of synthesis to determine if the results are consistent.

**Figure S163.** Output generated by ChemNavigator during the 24$^{th}$ iteration, aimed at the synthesis and characterization of a novel aluminum-based MOF utilizing the BTB-$CH_3$ linker.

## Reticular ChemNavigator Output (MOF-521-$CH_3$)

Current Stage and Iteration: 5-2

Status Evaluation: The apprentice successfully replicated the synthesis of both the linker and the MOF while adhering to previously optimized procedures. The 1H NMR spectroscopy and PXRD results show consistency with the initial rounds of synthesis, thereby validating the reproducibility of our synthesis protocol. Thus, the success of the task aligns with the process followed in the similar work summary.

Task Choice 1: In order to further confirm the reproducibility, let's conduct the activation and porosity determination of the newly synthesized MOF just as we did in stage 3. Perform methanol solvent exchange and vacuum drying for activation, followed by nitrogen gas sorption analysis at 77 K to determine the permanent porosity. Compare these results with those from the first round to check for consistency.

Task Choice 2: Another aspect we should examine is the thermal stability of the newly synthesized MOF. Perform a Thermogravimetric Analysis (TGA) on the newly synthesized MOF. Compare the degradation temperature and the overall TGA pattern with those obtained in the first round to ensure consistent thermal stability across batches.

Task Choice 3: Let's also check the chemical stability of the newly synthesized MOF in various solvents as we did in stage 4. Expose the MOF to DMF, methanol, ethanol, and water, and perform a PXRD analysis after exposure to each solvent. Compare these results with the initial rounds of testing to confirm consistent chemical stability.

**Figure S164.** Output generated by ChemNavigator during the 25$^{th}$ iteration, aimed at the synthesis and characterization of a novel aluminum-based MOF utilizing the BTB-$CH_3$ linker.

Reticular ChemNavigator Output (MOF-521-$CH_3$)

Current Stage and Iteration: 5-3

Status Evaluation: The apprentice effectively performed the activation and porosity determination process and obtained consistent results with the previous round, which confirms the reproducibility of our MOF synthesis and activation process. The similar BET surface area, pore width, and pore volume are all indicative of successful replication of the MOF activation process.

Task Choice 1: Summarize the overall progress and findings of our project to provide a comprehensive overview. Include every stage, starting from the synthesis of the BTB-CH3 linker, through the optimization of the MOF synthesis, activation and porosity determination, to the detailed characterization and the final validation. Highlight the successes, challenges faced, and how they were overcome.

Task Choice 2: Begin drafting a scientific paper detailing our project. The paper should include sections on introduction, methods, results and discussion, and conclusion. Make sure to elaborate on the experimental procedures, data analysis, and validation steps, along with a detailed discussion of the results.

Task Choice 3: Prepare a presentation summarizing our project and findings. This will be useful for sharing our results with colleagues or at scientific conferences. The presentation should be concise, visually appealing, and accessible to both experts and non-experts in the field of MOF synthesis.

**Figure S165.** Output generated by ChemNavigator during the 26$^{th}$ iteration, aimed at the synthesis and characterization of a novel aluminum-based MOF utilizing the BTB-$CH_3$ linker.